Matti Pietikäinen and Olli Silvén

# CHALLENGES OF ARTIFICIAL INTELLIGENCE:

## FROM MACHINE LEARNING AND COMPUTER VISION TO EMOTIONAL INTELLIGENCE

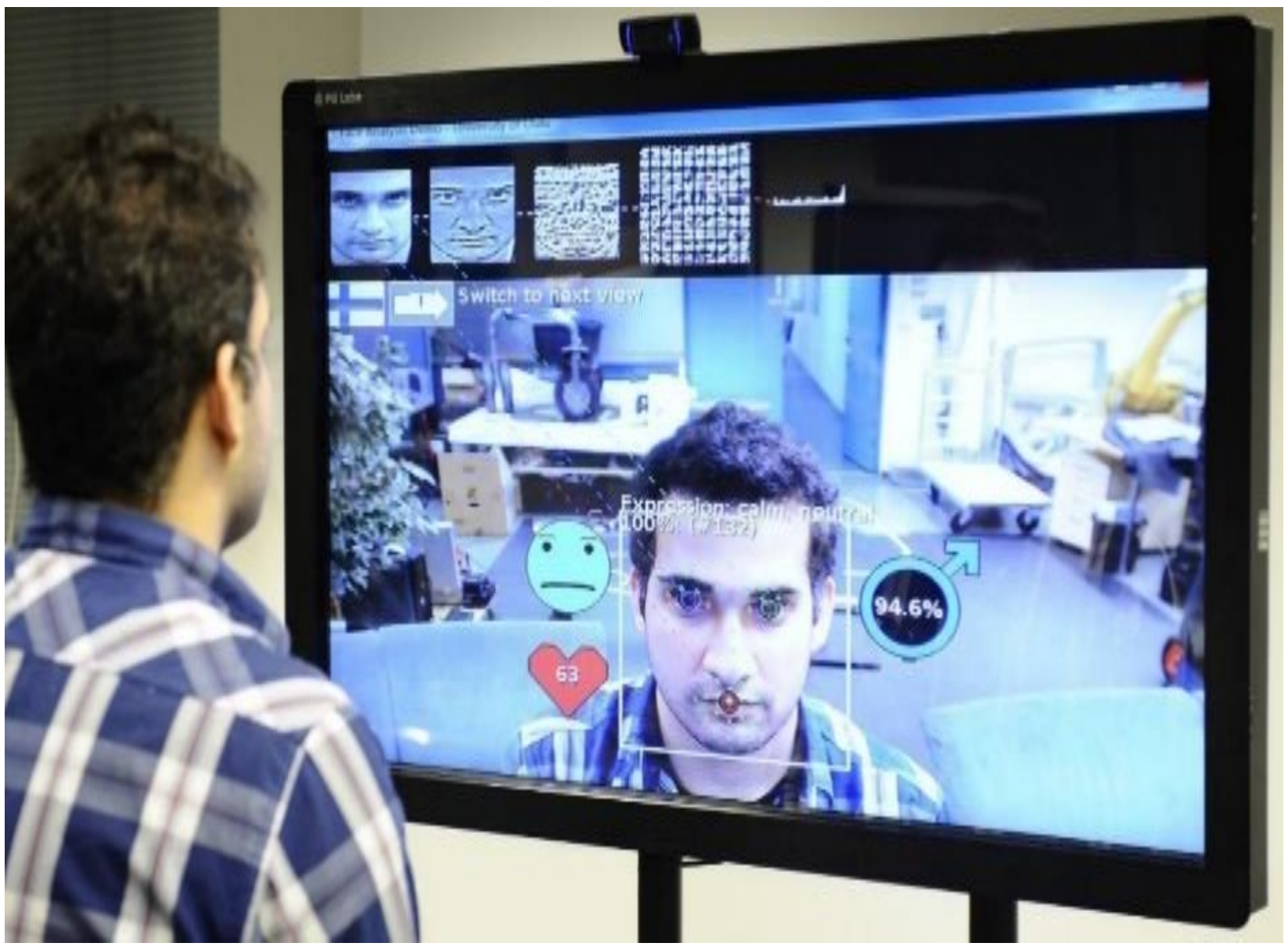

Center for Machine Vision and Signal Analysis (CMVS)
University of Oulu
https://www.oulu.fi/cmvs/

**CHALLENGES OF ARTIFICIAL INTELLIGENCE - FROM MACHINE LEARNING AND COMPUTER VISION TO EMOTIONAL INTELLIGENCE**

Authors of the book:

Matti Pietikäinen is Professor Emeritus at the University of Oulu's Center for Machine Vision and Signal Analysis (CMVS). He is a pioneer in Finnish machine vision research - investigating and teaching artificial intelligence since the early 1980s. He is one of Finland's most cited researchers in technology and computing, and has received several major international awards and honors, most recently King-Sun Fu Prize (2018) and Highly Cited Researcher (2018).

Biography: Matti Pietikäinen

Olli Silvén is Professor of Signal Processing Engineering at the Center for Machine Vision and Signal Analysis. He has been studying machine vision and artificial intelligence since 1981. In particular, he has focused on technologies and implementations of real-time energy-efficient machine vision systems and industrial applications. Since 2018, he has lectured on the highly popular course Introduction to Artificial Intelligence at the University of Oulu.

Biography: Olli Silvén

# Foreword

Artificial intelligence has become a part of everyday conversation and our lives. It is considered as the new electricity that is revolutionizing the world. Artificial intelligence is heavily invested in both industry and academy. Artificial intelligence technology leaders like Google, Facebook and Amazon are growing and their dominance is already causing concern. The United States has been a leading country in both research and application of artificial intelligence, but China, which is investing heavily in the topic, is expected to surpass it in the next few years. Artificial intelligence is feared to take a lot of jobs, and even in the next few decades, it is believed to become super-intelligent - and to take power from people.

However, there is also a lot of hype in the current artificial intelligence debate. There is often talk of artificial intelligence even in a context that does not actually represent actual artificial intelligence, but rather the regular evolution of digital technology towards more advanced functionalities. The combination of different technologies enables the development of completely new types of applications that mimic intelligent operations, humanoid robots being futuristic examples. Artificial intelligence based on so-called deep learning has achieved impressive results in many problems, but its limits are already visible. Artificial intelligence has been under research since the 1940s, and the industry has seen many ups and downs due to over-expectations and related disappointments that have followed.

The purpose of this book is to give a realistic picture of artificial intelligence, its history, its potential and limitations. We believe that artificial intelligence is a helper - not a ruler of humans. We begin by describing what artificial intelligence in the true sense of the word is and how it has evolved over the decades. After fundamentals, we explain the importance of massive data for the current mainstream of artificial intelligence. The most common representations for artificial intelligence, methods, and machine learning are covered. In addition, the main application areas are introduced.

Computer vision has long been central to the development of artificial intelligence. The book provides a general introduction to computer vision, and includes an exposure to the results and applications of our own research. Emotions are central to human intelligence, but little use has been made in artificial intelligence.

In this book we present the basics of emotional intelligence and our own research on the topic.

We discuss super-intelligence that transcends human understanding, explaining why such achievement seems impossible on the basis of present knowledge, and how artificial intelligence could be improved. Finally, a summary is made of the current state of artificial intelligence and what to do in the future.

In most countries, the growing importance of artificial intelligence skills has been noted over the last few years. In the appendix of the book, we look at the development of artificial intelligence education, especially from the perspective of contents at our own university.

This book is based on the authors' 40 years of experience in cutting-edge research, teaching and application of artificial intelligence and machine vision to a variety of problems. The book is intended for a wide range of readers: high school students, higher-education students, researchers, professionals working in industry or elsewhere, anyone interested in artificial intelligence – including the decision-makers.

Much of the content of the book is based on the work of our research group over the past decades, as evidenced by literary references and artwork. The first edition in Finnish was published in November 2019. We received valuable content suggestions for it from Satu Räsänen, Janne Heikkilä, Guoying Zhao and Tapio Seppänen, Li Liu, Tuomas Holmberg and Jie Chen assisted with the illustration. We also got useful comments from Pirkko Ekdahl and Tuukka Bogdanoff. We are grateful to all those who supported us in the completion of the book! For this English edition Google Translate based on artificial intelligence was used to produce the first translated versions of the text for most of the chapters, and then edited by the authors. Additional material written for the second Finnish edition to be published in early 2022 is also included.

Matti Pietikäinen Olli Silvén

# Table of Contents

# 1 Introduction

## 1.1 About Artificial Intelligence

Did you notice when Netflix recommended you a TV show you might like? Did you notice that the customer service agent Chris was really a chatbot? Or did you know that in industry the quality of products has already long been monitored with artificial intelligence technologies? Artificial intelligence has become a part of our daily lives, and its impacts extend beyond everything (Pietikäinen et al., 2017).

We are in the midst of artificial intelligence hype. Undeniably, artificial intelligence and machine learning are prerequisites for a new industrial revolution in which machines are helping or even replacing humans in advanced functions. Modern artificial intelligence systems learn through massive training data and become capable of identifying objects and making accurate predictions. Examples of recent developments include the ability of devices to recognize speech, predictive text input, better language translation programs, advanced Internet search engines that show you tailored ads and recommendations, intelligent robots, computer vision based recognition of faces and other objects, as well as the development of driver's assistance systems and autonomous vehicles.

Artificial intelligence creates opportunities to improve the energy efficiency of societies. It can promote people's learning, safety and health by enabling machines and the environment to adapt their actions to emotional states and behaviors.

On the other hand, artificial intelligence can be used to influence people, infamously demonstrated by Cambridge Analytica in the 2016 US presidential election, which polarized voters' attitudes by targeting news and fake news based on people's media behavior. This has contributed to the regulatory requirements of technology (Hern, 2018).

Artificial intelligence can be defined in many ways. Regardless of the definition, artificial intelligence research has two main objectives: to make computers more usable, and to understand the principles that make intelligence possible. During the recent artificial intelligence boom it has been suggested that the term "machine learning" should be used instead of artificial intelligence. However, learning is only one, albeit essential, part of intelligent activity and artificial intelligence, so such a definition is neither correct nor complete. For example, a human being has nearly 100 billion neurons in the central nervous system, but the rest of the body has much more for sensory perception, motor functions, and information transfer.

Since the early days of artificial intelligence research, the subject has been approached in two different ways: through symbolic data processing and data-driven connectionist neural networks.

Symbolic artificial intelligence derives its name from the aim to represent properties of a phenomenon or object by symbols that are abstractions of the observations. Symbolic artificial intelligence builds databases and rules for how the world associated with that problem is working.

Symbolic processing is particularly well suited for reasoning and using abstractions in problem solving. The artificial intelligence hype of the 1980s was based on rule-based expert systems, using symbolic artificial intelligence for applications such as medical diagnoses or legal advice. The big problem was that creating and teaching the systems had to be done largely manually. However, the systems could justify how they reached the conclusions.

Artificial neural networks are based on data-driven connectionist computing. They consist of simple computing elements that are connected to each other, resembling human nerve tissue, and can be taught. Recent developments in artificial intelligence are based on deep learning neural networks that are able to automatically classify objects in the image, recognize speech, or, for example, make predictions of market developments based on examples. The main problem is that huge amounts of examples are required for training, and the systems can’t justify their decisions. While even a common housefly appears to learn from a single example, deep learning, at least in its present form, does not seem to be a future-proof machine learning model.

## 1.2 Artificial Intelligence Hype in Finland

In the 1980s the first artificial intelligence enthusiasm spread to Finland. In Spring 1984, Tekes, the National Technology Agency, launched the FINPRINT IT Development Program. The aim was to achieve and maintain international competitiveness, especially in the IT hardware and software industry (Tekes: Tekoäly 1985-89). Of its five subprograms, two were related to artificial intelligence: 1) Artificial intelligence (Linnainmaa, 1989) and 2) Pattern recognition applications (Rämö, 1988).

Now watching the FINPRINT-motivated A-Studio TV program from 1988, the optimism and the applications considered at that time were very similar to the current ones: military intelligence, intelligent weapons and Star Wars, disability technology, nuclear power plant monitoring, Finnish language database interpreter, restoration of works of art, and planning cancer treatment.

The main technology was mostly rule-based expert systems, to which human experts' knowledge had to be brought largely manually. However, learning neural network systems were already seen as a future trend. Even then, ethical issues related to strong artificial intelligence resembling human intelligence were discussed. The Encyclopedia of Artificial Intelligence, which presents the different areas of artificial intelligence in Finnish, was published in 1993 (Hyvönen et al., 1993).

Tekes and the Academy of Finland have also had other programs in the field of artificial intelligence. Particularly important for machine vision applications and our own research was the Tekes Machine Vision in Industrial Automation Program 1992-1996 (Pau & Savisalo, 1996).

The Pattern Recognition Society of Finland was founded in 1977 and the Finnish Artificial Intelligence Society (FAIS) in 1986. The author of this book, Matti Pietikäinen, has been a key member in both societies. Teaching of artificial intelligence and digital image processing was started at the University of Oulu in the academic year 1982-83. The first AI course, based on Patrick Henry Winston's classic book Artificial Intelligence, took place in the Spring of 1983 (Winston, 1977).

At that time symbolic computing and knowledge engineering were at the center of artificial intelligence. In addition to a few other research units, our group received a Symbolics computer for the research and teaching from the Ministry of Education. The computer supported symbolic processing and LISP programming language particularly well.

## 1.3 Data and Machine Learning in Central Roles

In the late 1980s, artificial intelligence established itself as a scientific discipline. The new hypotheses had to be comprehensively tested, the results analyzed, and they had to be compared with previous results. The use of statistical models based on probability estimation increased. We moved from the "toy world" problems to real world applications. The computing speed and capacity of computers increased, allowing for more sophisticated methods and larger amounts of data.

With advanced tools, deep learning neural networks can easily produce impressive results in many problems. However, the performance may still fall far from the real application needs and may require much more computational resources than traditional methods. For this reason, developers should also explore alternative schemes that may reach the performance requirements.

The importance of knowing alternative approaches can be seen from a recent study by Ferrari Dacrema et al. (2019) on neural approaches to recommendation systems. These are used e.g. to predict the "rating" or "preference" a user would give to an item, such as in playlist generators for video and music services, product recommenders for services, or content recommenders for social media platforms (Wiki-Recommender). Their study indicated that reproducing recent results published in top forums was very challenging and that in this application area, most of the proposed modern machine learning based solutions could be outperformed with much simpler heuristic methods such as nearest-neighbor or graph-based approaches.

Understanding the fundamentals, differences, and uses of both symbolic and data driven artificial intelligence is important, as well as understanding the different ways of representation in AI. Data and representation of knowledge related to the application problem, search methods, feature representations describing the properties of the objects, and methods for reducing the dimensions of the feature representation play a central role. The roles of data and knowledge are viewed crucial (Figure 1.1).

Machine learning can be done in a variety of ways: using supervised learning, unsupervised learning, or the recently much-discussed reinforcement learning. A major breakthrough in machine learning (Kritzhevsky et al., 2012) is based on the massive increase in data available over the Internet, the rapid development of computer hardware that enables to use it, and method inventions since the 1980s. Such method innovations include multilayer neural network and the convolutional neural network (CNN) using the backpropagation algorithm and their subsequent improvements. It was shown that in image recognition using multilayer learning neural network and a massive number of training samples improved the results significantly. Since then, many new applications have been developed that were previously considered impossible.

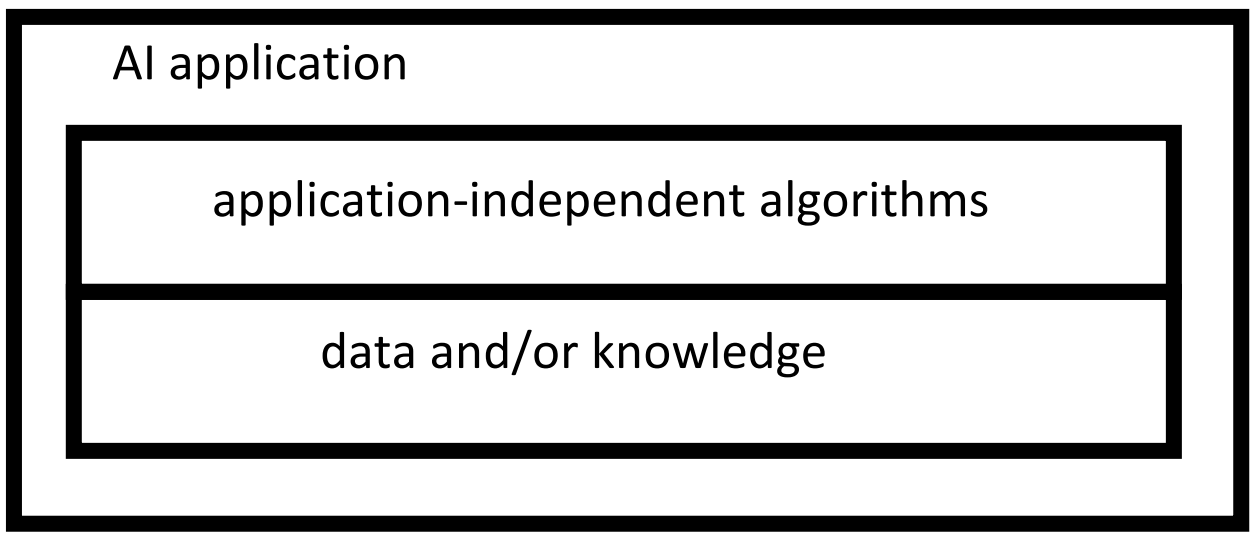

Figure 1.1. The roles of data and knowledge in artificial intelligence.

In the current AI boom, major technology companies such as Google, Amazon, Facebook, Microsoft, Apple, IBM, Nvidia, Uber, and Chinese internet giants Alibaba, Baidu and Tencent are investing heavily in various artificial intelligence applications and their development tools. The technology platform industry formed by such companies and the massive amount of data they own seem to dominate the industry, and their power is constantly growing. It is usually necessary for other players to use the platforms and computing capabilities provided by such companies to keep up with the development, which can also be a risk to privacy.

Andrew Ng, a leading expert in deep learning said in his famous assessment that "artificial intelligence is new electricity", meaning that it can revolutionize industry in the same way that electricity once did. Technology investor Steve Jurvetson made a more questionable assessment in Helsinki at Slush 2016: "Deep learning is mankind's greatest invention since the scientific method." “It can be used to teach a machine to do anything and it will revolutionize our world faster than anyone believes” (Kotilainen, 2017).

However, development is significantly hampered by the increasing energy consumption of the latest artificial intelligence applications - doubling every 3-4 months according to Open AI (Tuomi, 2019). For example, training the best current natural language interpretation methods have been estimated to produce approximately the same amount of carbon dioxide as five passenger cars over their lifetime.

In the foreseeable future, artificial intelligence will, based on our understanding, be integrated with other technologies, improving the functionality and usability of applications and adaptability to new situations. The current hype will be replaced by regular digital technology development, where a natural step is to introduce smarter features for machines and other systems.

In general, people want the latest technology for their smart phones, home appliances, and cars. The companies that can deliver will grow and survive. For example, the fate of Nokia smartphones was largely due to its inability to respond to Apple's touchscreen user interface. Properly applied artificial intelligence can also increase security, enhance operations, save energy, etc. More generally, only companies, communities and countries that are able to exploit the opportunities provided by new technologies, including artificial intelligence can succeed in the international competition.

By integrating various technologies related to artificial intelligence, such as mass data, machine learning, speech recognition, machine vision and robotics, a whole new range of applications can be created - a futuristic example is humanoid robots. The most challenging applications consist of many integrated components. For example, fully autonomous vehicles would require reliable 2-D and 3-D vision and other sensor information to work in changing conditions, map data, satellite navigation, extremely fast and reliable response to human, animal and other traffic actions, and continuous redesign of the driving route - and machine learning from all that. How, then, can we trust that such complex systems will work reliably in all practical circumstances?

## 1.4 On the Impact of and Investing in AI

Artificial intelligence and automation will have a significant impact on many jobs. For example, robots are already replacing many jobs in the industry today, and automated cash registers are replacing human operated ones in stores. Amazon has recently introduced a fully automated store where customers sign up with their smartphone (Leinonen, 2018). Versatile camera and sensor systems monitor what each customer purchases, and the off-site system automatically bills purchases from the customer's account. Similar solutions requiring high sensor densities will become more common, e.g., enabled by the future wireless communication technologies provided by the University of Oulu's 6G flagship program (www.oulu.fi/6gflagship/).

Many jobs will disappear or be partially replaced by technology as digitalization is spreading across society. The workforce to implement and maintain the new solutions needs to be educated, possessing the ability to orientate themselves to new tasks. Consequently, the future artificial-intelligence reliant society will need data-literate citizens capable of reading, using, interpreting and communicating data (Executive Office, 2016).

In the field of artificial intelligence, major national projects have been or are being launched. The United States has been a leader in artificial intelligence research and development, although under President Trump’s administration, the national efforts were not increased in line with President Obama's plans. Now major technology companies such as Amazon, Apple, Google, Facebook and Microsoft are investing in the industry. In February-2019, President Trump signed a decree on artificial intelligence and its regulation in response to the threat posed by China in this area. Apparently, the provision does not apply to research and development. In Spring 2021 President Biden promised to increase significantly funding of strategically central research areas, including AI.

Professors Geoffrey Hinton and Yoshua Bengio, who began their careers in Canada, have been pioneers in developing deep learning methods with Professor Yann LeCun. (These three scientists won The Turing Award 2018, which is often called the Nobel Prize in computing.) The Canadian government has recently launched an extensive program to stay ahead and support the development of the artificial intelligence industry in the country.

China has made artificial intelligence a strategic priority and invests huge amounts in AI research. Thus, it can be expected to reach the US level, and by the end of the 2020s it could be the leading developer and exploiter of artificial intelligence. Many other countries, such as Japan, the United Kingdom, Germany and the United Arab Emirates, have also prepared and launched their own national artificial intelligence programs.

Finland, too, has responded to the major societal effects of artificial intelligence, and has begun to invest in research and applications in the field. On 7 February 2017, the then Prime Minister

Juha Sipilä opened a discussion session at Säätytalo (the House of Estates) entitled "How we accept artificial intelligence and its potential as a society". After that, a steering group was set up to find out how Finland could be promoted as the world's best provider of artificial intelligence. Since then, several actions have been planned. A summary of these actions is presented in (Ailisto et al., 2019).

Education plays a key role in how we can respond to major challenges. There has been talk of retraining up to a million Finns, but this should also mean upgrading current skills to increase the data literacy mentioned above. Based on our own long-term experience, in Appendix L1 we present what should be taught to a wider audience, and what to future specialists in artificial intelligence.

## 1.5 Machine Vision Research at the University of Oulu

Matti Pietikäinen, the first author of this book, began research on computer vision and artificial intelligence in the Fall of 1980, when he went on to do his PhD research at the University of Maryland for 14 months, supervised by Professor Azriel Rosenfeld, a pioneer of digital image analysis and machine vision. The study was on the classification of textures that tell about the surface structure of objects in the images. Inspired by excellence in guidance and a highly international research environment, he defended his dissertation in Oulu in May 1982. After returning to Oulu from Maryland in late 1981, he subsequently founded the Machine Vision Research group, which later became a highly respected research unit in its field internationally.

Olli Silvén, the co-author of the book, was recruited as the first graduate student in the new group in December 1981. After earning his doctoral degree, he has played a key role in many of the research group's projects, particularly in those focusing on machine vision applications. In Spring 2018, he held the first Introduction to Artificial Intelligence course for University of Oulu students representing all backgrounds. The contents of the course form the basis of chapters on the fundamentals of artificial intelligence (Chapters 3 and 4) in this book.

Computer vision has been the key engine of artificial intelligence research since it was linked to the deep learning breakthrough in 2012. In this book, we generally look from the computer vision point of view what has been achieved and what is going on. In addition, we introduce the main findings of the machine vision research at the University of Oulu. The most significant international breakthrough has been the so-called Local Binary Pattern (LBP) method and its application to face recognition (Pietikäinen et al., 2011). LBP-related publications have been among the most cited Finnish publications in the field of artificial intelligence and computer science in the 2000s.

The major application areas of our group's research include industrial visual quality control, biometric identification, remote heart rate measurement, speech recognition from mouth movements, 3-D machine vision and its use in augmented reality, intelligent robots and human-machine interaction, and medical applications. This book gives the reader a good idea of what computer vision is, what applications it can be used for, and what challenges it faces.

Emotions and empathy play a central role in human-to-human interaction. Much of the communication between people is nonverbal: facial expressions, gestures, body movements, and variations in speech tone.

When talking to an intelligent robot or chatbot, a person feels "machine-like", insensitive speech most unpleasant. The use of emotions, or related affective computing, is expected to be the next major artificial intelligence breakthrough. Many companies, such as Affectiva, Microsoft, Apple and Google, have started investing into this domain. However, there are still many unresolved research challenges ahead of any large-scale deployment of technology.

This book discusses emotional intelligence, its applications and challenges from the perspective of machine vision. For instance, recognizing micro-expressions associated with hidden emotions is considered (Chapter 9).

## 1.6 The Risks and Challenges of AI

The risks of artificial intelligence have recently been the subject of discussion due to the threat scenarios presented, for example, by the late physicist Stephen Hawking and entrepreneur Elon Musk. However, we believe that artificial intelligence will be a helper for humans, not a ruler. Given the significant capability limitations of current artificial intelligence and machine learning technologies over human intelligence, we do not believe that any "super-intelligence" can be achieved, at least not in the foreseeable future.

In any case, the introduction of artificial intelligence involves many ethical issues that need to be discussed and necessary measures taken, for example, through appropriate regulations. However, over-control, e.g., in terms of data availability, can become a barrier to development and thus prevent Europe (and Finland) from being among the leading artificial intelligence providers.

On the other hand, Finland is a society based on trust, which can support the use of advanced solutions still in their infancy. At the same time, efforts must be made to locally process human information, avoiding excessive data collection into global cloud computing centers. Promoting such developments requires long-term research efforts, where the strategic focus is on the good of individual citizen.

The machine learns from the training material provided, whether good or bad. Consequently, machine learning methods have been found to be vulnerable to adversarial attacks (Heaven, 2019). By teaching almost completely true-like but falsified data, the systems can be made to produce false results. This can be a major problem, for example, in medical applications, self-driving vehicles, or weapons systems (Finlayson et al., 2019).

There are also high risks associated with applications where machine learning and artificial intelligence may not play a key role. These include combining vast amounts of information from multiple sources and creating a accurate profiles of individuals. The results can be used for both good and bad. The sources of information range from social media behavior, Internet searches, shopping data, health services, and banking. In these cases, decentralization provides at least moderate protections against security disasters.

Privacy also needs to be taken care of, for example, when using concealed face recognition or other identity disclosure technology. Information security is best ensured by solutions where machine learning and human-centered analytics are done with resources embedded solely in a vehicle, a room, or even wearable technology.

Such a potential privacy-driven development would clearly favor a different kind of earning logic than the current one of the Internet giants. They are determined to protect their centralized model, which also appears to be in the interests of many superpowers.

Finland's special strengths in the development of distributed artificial intelligence applications are strongly linked to the expertise in embedded information and communications technology required by the Internet of Things (IoT). However, utilizing this requires strategic choices and targeted research funding.

## 1.7 Realism is Needed instead of Hype in AI

Artificial intelligence as discipline has experienced many hype seasons and winters throughout its history. The current artificial intelligence boom is based almost exclusively on deep neural network technology, the potential of which could have been overlooked without an increase in the supply of mass data. That is why Google, Amazon, Facebook, Apple and other big players who have obtained data from the Internet have been at the forefront of development. Applications with no or limited mass data have received less attention.

This has clearly not been understood, but artificial intelligence researchers have largely focused on experimenting with and tuning deep learning schemes. Even the pioneer of deep learning, Professor Geoffrey Hinton, has recently admitted, like Max Planck, a physicist, that "science progresses from one funeral at a time" (LeVine, 2017).

Figure 1.2 shows the peaks of AI development over the decades. These will be addressed in later chapters of the book. We will also look at how artificial intelligence should be developed to be better, and more like human intelligence, to achieve the next breakthrough.

Experts know the limitations of current artificial intelligence technology, to which it can be applied well and poorly. They also know that it will at some point be replaced or supplemented with better solutions. Without an effective link between top expertise in research and application developers, we are easily in the same situation as with expert systems based on processing symbolic information in the 1980s. Everyone tried to apply them, but in the end, the biggest winners were tool and platform developers.

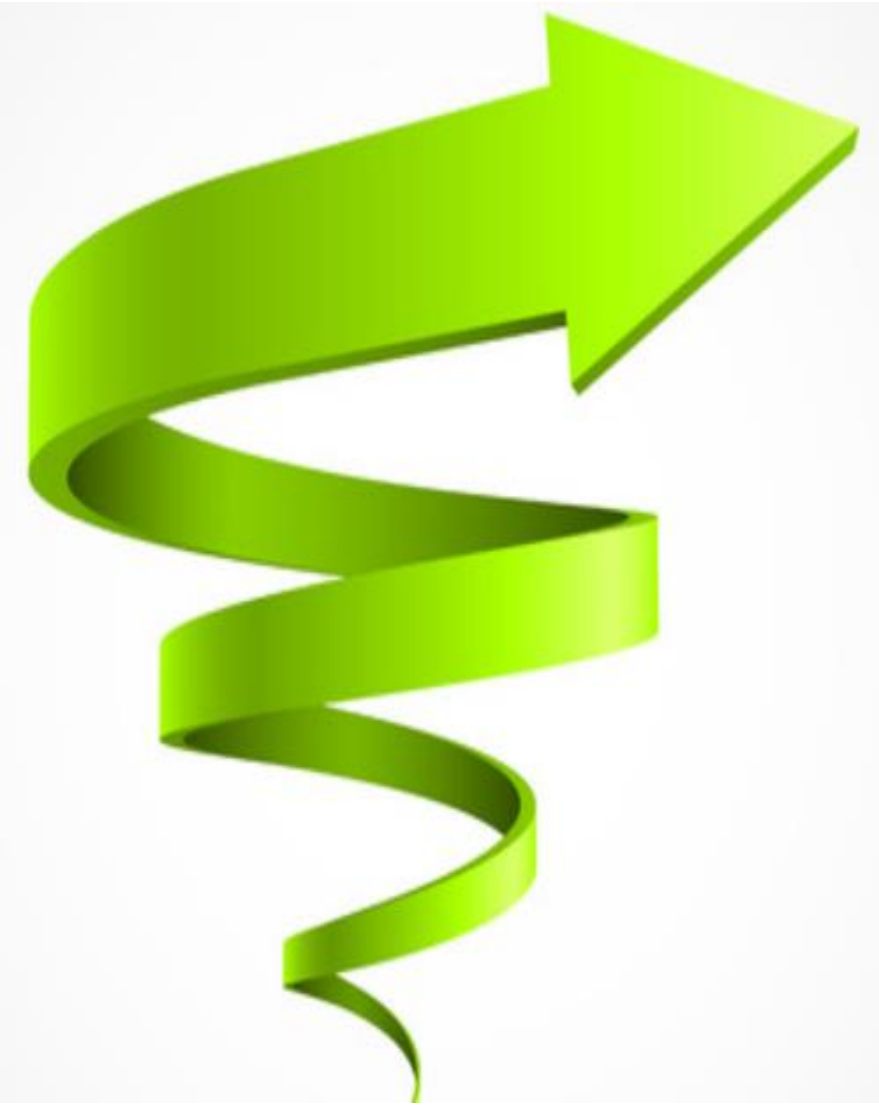

Figure 1.2. AI hype cycles over time. (© 123RF)

For example, the authors of this book, living in northern conditions, do not believe in the introduction of vehicles that are completely autonomous in changing traffic and weather. There are enormous expectations, but also complex technical challenges and ethical issues that appear unsurmountable. It would be better for us to focus on limited, secure application environments or tasks that facilitate human time. For instance, in January 2020 a Canadian automotive components manufacturer decided to concentrate on driver's assistance systems instead of self-driving (Bickis, 2020).

Instead of artificial intelligence a better term could be machine intelligence. Distinctly technological, it could better withstand highs and lows. With the advancement of technology, applications and systems are gaining features that appear intelligent, and there is no need for actual comparison with real intelligence or human intelligence.

At the end of the book, we discuss what's true and hype about today's artificial intelligence. In recent years, machine learning and artificial intelligence applications have finally begun to come to fruition. Development has been accelerated by advances in computing technology, particularly the Internet, which has provided the ability to acquire massive data sets and thereby supported the development of AI methodology. We have witnessed the departure from experiments in limited laboratory environments to global applications used by much of the humanity.

The objective of this book is to provide a first-hand view of the development, present and future of artificial intelligence, machine vision and emotional intelligence. Because of the over-expectations of artificial intelligence, the book aims to give a realistic picture of the situation today and expected developments over the next few years. In addition, the book introduces the basics that anyone studying AI needs to know to understand its potential for a variety of problems.

Our approach is largely data, methodology, technology and application oriented. We leave the social implications and ethical issues to experts in their respective fields. We think that in Finnish these are well addressed in Maija-Riitta Ollila's book, "Tekoälyn etiikkaa" (Ethics of Artificial Intelligence), published in 2019 (Ollila, 2019). In English, Paula Boddington in her book "Towards a code of ethics for artificial intelligence" (Boddington, 2017) considers how to produce realistic and workable ethical codes or regulations for AI.

## 1.8 References

Ailisto H (ed.), Neuvonen A, Nyman H, Halen M & Seppälä S (2019) Tekoälyn kokonaiskuva ja kansallinen osaamiskartoitus – loppuraportti (A general view to AI and a survey on national know-how – Final report). Valtioneuvoston kanslia, 15.1.2019, 84 p.

Bickis I (2020) Magna scaling back Lyft partnership as fully self-driving systems look further off. Toronto Star, January 16, 2020.

Boddington P (2017) Towards a Code of Ethics for Artificial Intelligence. Cham, Switzerland, Springer, 124 p.

Executive Office (2016) Preparing for the Future of Artificial Intelligence. President Obama's National Science and Technology Council Committee on Technology, USA, 2016.

Ferrari Dacrema M, Cremonesi P & Jannach D (2019) Are we really making much progress? A worrying analysis of recent neural recommendation approaches. Proc. 13th ACM Conference on Recommender Systems, 101-109.

Finlayson SG, Bowers JD, Ito J, Zittrain JL, Beam AL & Kohane IS (2019) Adversarial attacks on medical machine learning. Science 363 (6433):1287-1289.

Heaven D (2019) Deep trouble for deep learning. Nature 574:163-166, 10.10.2019.

Hern A (2018) Cambridge Analytica scandal "highlights need for AI regulation", Guardian 16.4.2018.

Hyvönen E, Karanta I & Syrjänen M (eds.) (1993) Tekoälyn ensyklopedia (Encyclopedia of Artificial Intelligence). Gaudeamus, 356 p.

Kotilainen S (2017) Tekoälyn vallankumous on alkanut – tätä kaikkea se tarkoittaa (The revolution of AI has begun – it means this all). Tivi 5/2017.

Krizhevsky A, Sutskever I &, Hinton GE (2012) Imagenet classification with deep convolutional neural networks. Advances in Neural Information Processing Systems, 1097-1105.

Leinonen T (2018) Verkkokaupan jätti perusti kivijalkakaupan (The giant of e-commerce established a ground floor shop). Kaleva 20.2.2018.

LeVine S (2017) Artificial intelligence pioneer says we need to start over. Communications of the ACM.

Linnainmaa S (ed.) (1989) Tekoäly 1985-1989 (Artificial Intelligence 1985-1989). IT Development Program, Tekes, 232 p.

Ollila M-R (2019) Tekoälyn etiikkaa (Ethics of Artificial Intelligence). Otava, 366 p.

Pau LF & Savisalo H (1996) Machine Vision Technology Programme 1992-1996 – Evaluation report. Tekes Technology Programme Report 16/96, 34 p.

Pietikäinen M, Hadid A, Zhao G & Ahonen T (2011) Computer Vision Using Local Binary Patterns. Springer, 207 p.

Pietikäinen M, Silvén O & Pirttikangas S (2017) Tekoälyn opetusta on lisättävä ja syvennettävä (We should have more and deeper AI teaching in Finland). Helsingin Sanomat 20.4.2017.

Rämö E (ed.) (1988) Hahmontunnistuksen sovelluksia (Applicatios of pattern recognition). IT Development Program, Tekes, 82 p.

Tuomi I (2019) Sähkönkulutus on tekoälyn kompastuskivi (Electricity consumption is an obstacle for AI). Helsingin Sanomat 31.8.2019.

Winston PH (1977) Artificial Intelligence. 1st Edition, Addison Wesley.

Wiki-Recommender: Recommender system

# 2 What is AI and How has it Progressed?

## 2.1 What is AI?

According to an early definition artificial intelligence is research on issues that are needed to make computers intelligent. This means that you should first define what intelligence means. In another definition, AI tackles problems for which solutions are not yet known, i.e., the border between an intelligent and ordinary machine is changing.

In the current AI boom it has also been suggested that instead of AI we should talk about machine learning. Machine learning is, however, only one, albeit a very important part of intelligent behavior and AI, which means that this definition is not correct.

In their widely used book Stuart Russell and Peter Norvig (2010) divide the definitions of AI into four different categories. These can be placed as in Table 2.1 into two dimensions.

Table 2.1 Grouping AI into different definitions

| | |
|---|---|
| Think like humans | Think rationally |
| Act like humans | Act rationally |

The definitions in the upper row are related to the processes of human thinking and decision making, whereas the lower row deals with human behavior. The approaches in the left column compare the success of AI to human performance. The right column is concerned with approaches that measure machine performance against an ideal rational measure. A system is rational when it makes right things based on the knowledge it has.

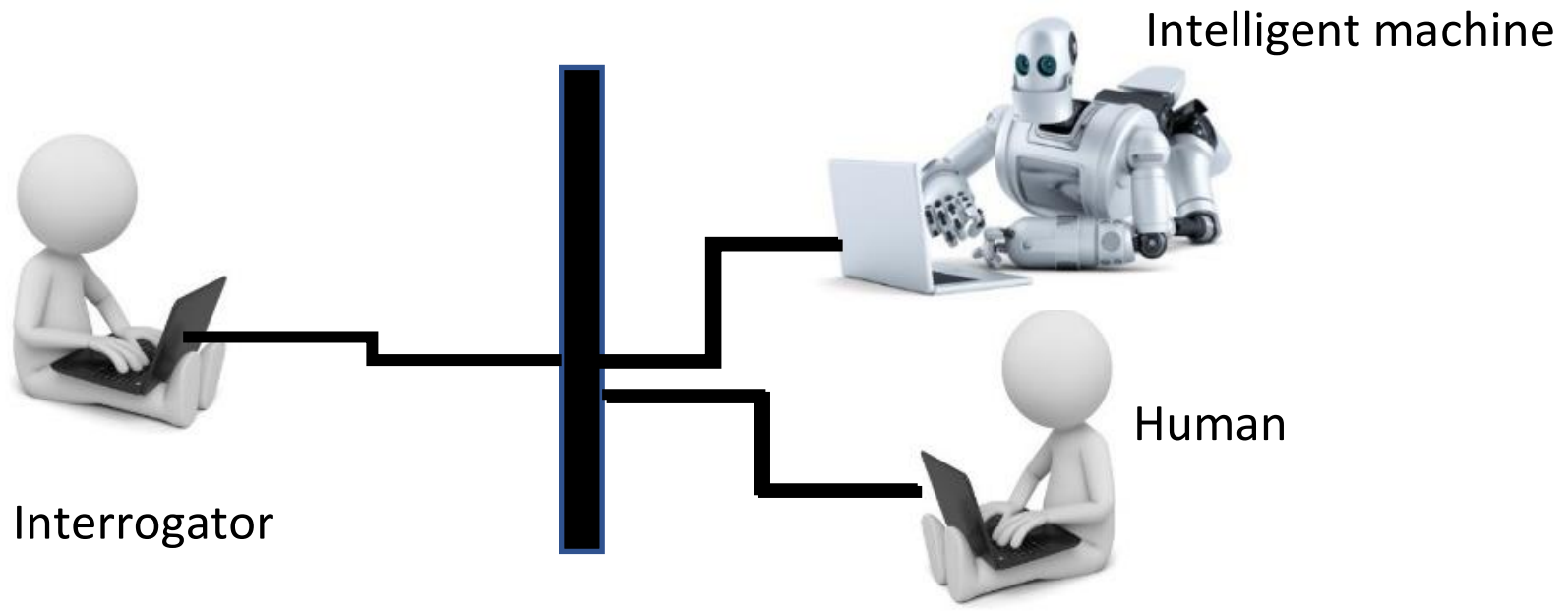

Figure 2.1. The Turing test. (© 123RF)

For **acting like humans**, Alan Turing presented the following widely referred method to test machine intelligence (Russell & Norvig, 2010). We assume that on one side of the wall is the evaluator of intelligence (human interrogator) and on the other side either a human or a machine (Figure 2.1).

The discussion is done, for example, with computer terminals. If the interrogator concludes that he or she has been discussing with a human, even though the party was a computer, according to Turing's definition the machine has intelligence.

To pass the test, the machine has to

1) be able for natural language processing (e.g., English or Finnish) needed for communication,
2) have knowledge representations in order to store what machine already knows or learns to know,
3) be able to automatic reasoning in order to use stored knowledge for answering questions and making new conclusions, and
4) use machine learning in order to adapt to new situations and be able to express and generalize models from its working environment.

The Turing test avoided direct human-machine interaction, assuming that intelligence can be measured based on written verbal communications. The variant called total Turing test includes two other necessary properties of an intelligent system: perceptual information obtained, e.g., with a video camera sensor and possibility to manipulate physical objects. Therefore, we need also

5) machine vision and other sensor data, such as speech, for perception of objects, and
6) robotics for manipulating objects and moving in the working environment.

These six areas in fact form the main parts of the scientific field called *Artificial Intelligence*.

**Thinking like humans** can be achieved in three different ways. 1) We can try to find the solution through our own thinking, 2) by using psychological tests to understand actions of another person, and 3) by observing brain activities with the help of technology. With these means we can try to find out a good enough theory about human mind, on the basis of which it would be possible to write computer programs, for example, to solve different types of problems. The multi-disciplinary *cognitive science* considers computational AI models together with experimental models and has evolved into its own field of science.

**Thinking rationally** (i.e., "the laws of thought") principle has its basis in decision processes used in philosophy. For example, "Socrates is a human" and "All humans are mortal", therefore "Socrates is mortal". These kinds of laws of thought called *logic* are considered to master the actions of human mind. The problem is that it is often very difficult to express problems to be solved with logic, especially in problems containing uncertainty. The computational complexity will also become far too high in real-world problems.

**Acting rationally** principle based on agents is utilized by Russell ja Norvig (2010). Figure 2.2 shows the basic structure of an agent. An agent is anything such that percepts its environment with sensors and acts in its environment with the help of actuators. Human senses include vision, hearing, touch, smell and taste. Hands, legs and mouth are actuators. As sensors a robot can have different types of cameras and microphones, and motors, legs or wheels, and loudspeakers and avatar displays as actuators. The agent can also just be a goal-oriented program, which reacts to the inputs in its working environment and autonomously performs some action. A very simple agent could, for example, compute the square root of a number given as input and return the result to the user or another program.

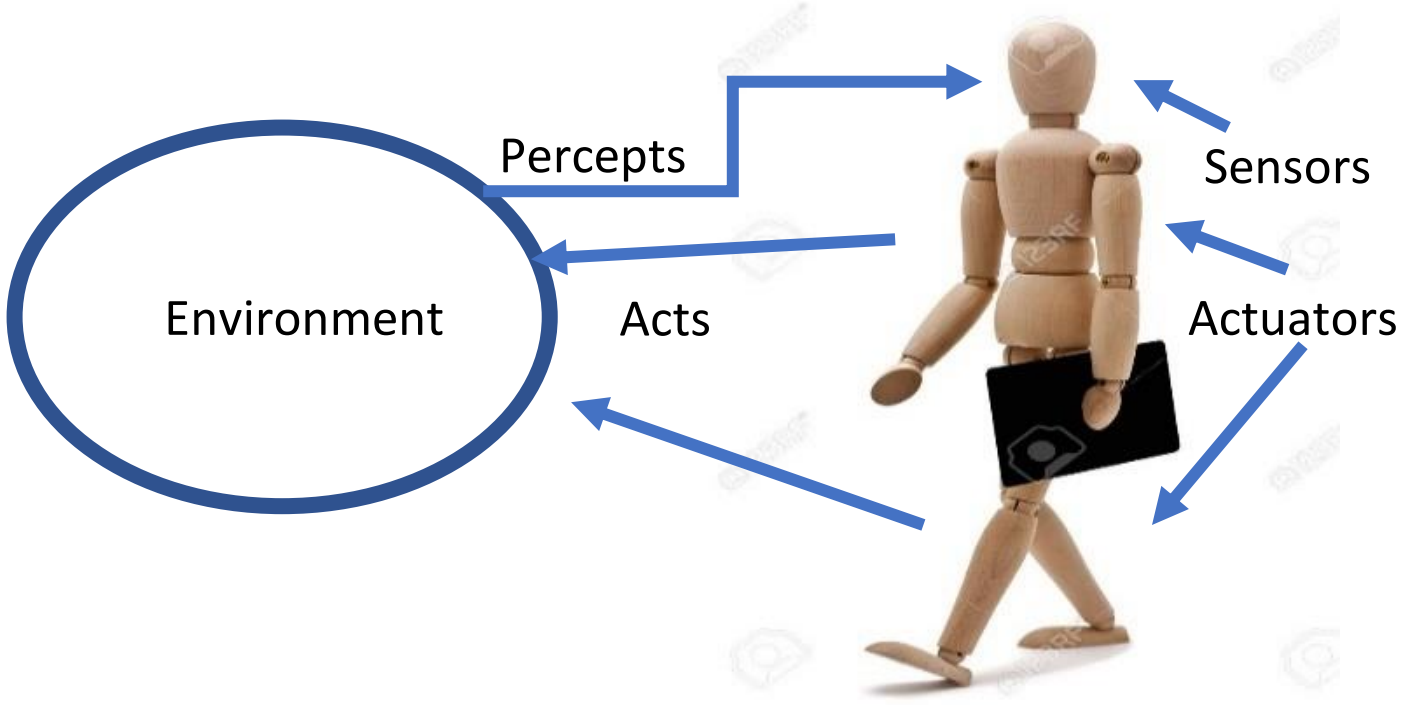

Figure 2.2. A rationally acting agent. (© 123RF)

An intelligent agent can act autonomously percept its environment, adapt to changes, defining and pursuing goals, and continuing its own existence. The rational agent tries to achieve the best possible result with the knowledge it has at each moment.

For example, an autonomous mobile robot could percept its environment using machine vision, avoid unexpected obstacles in front of it and due to this action change its original route, and finally find the destination where it was planned to go. The best result could be the shortest or safest possible route.

All the skills needed for passing the Turing test would also allow rational actions. With its knowledge representation and ability to reason the agent can make good decisions. It should be able to use a natural language to cope with complex environments based on instructions given by a human. Ability to learn is also needed for understanding new information and for efficient behavior.

The rational agent principle has clear advantages compared to the other definitions of AI mentioned above. In addition to the proper way of reasoning there are also other mechanisms for achieving rational actions. The agent approach is also much more useful for developing various applications than those based on human behavior or human thinking.

## 2.2 Symbolic AI vs. Connectionist Neural Networks

From the beginning of AI research the domain has been approached by two very different manners: symbolic data processing and connectionist computing based on neural networks.

The term **symbolic AI** is based on the assumption that a given part or property of a phenomenon or object can be represented with symbols. For example, the most well-known equation of Albert Einstein's relativity theory, $E = mc^2$ (where E = energy, m = mass and c = speed of light), is a symbolic representation. The value for energy can be calculated for different mass values without seeing or learning any specific values beforehand.

In symbolic AI, large knowledge bases and rules are built to describe how the world related to a given problem is working. For example, one of the rules related to recognizing animals in the zoo could be: ”If the animal is a mammal and it has a long neck, then it is a giraffe”.

Ontologies are often used in building systems. In computer science, the ontology means a description of concepts and their relations in a given application problem. By using rules and knowledge it is straightforward to build a system than can do reasoning.

Symbolic systems can also answer to questions. After getting a reply, it is easy to follow the chain of reasoning to find out why the system ended up to a given solution.

Symbolic processing fits very well to reasoning and using abstractions in problem solving. The AI boom in 1980s was mainly based on rule-based expert systems (Figure 2.3).

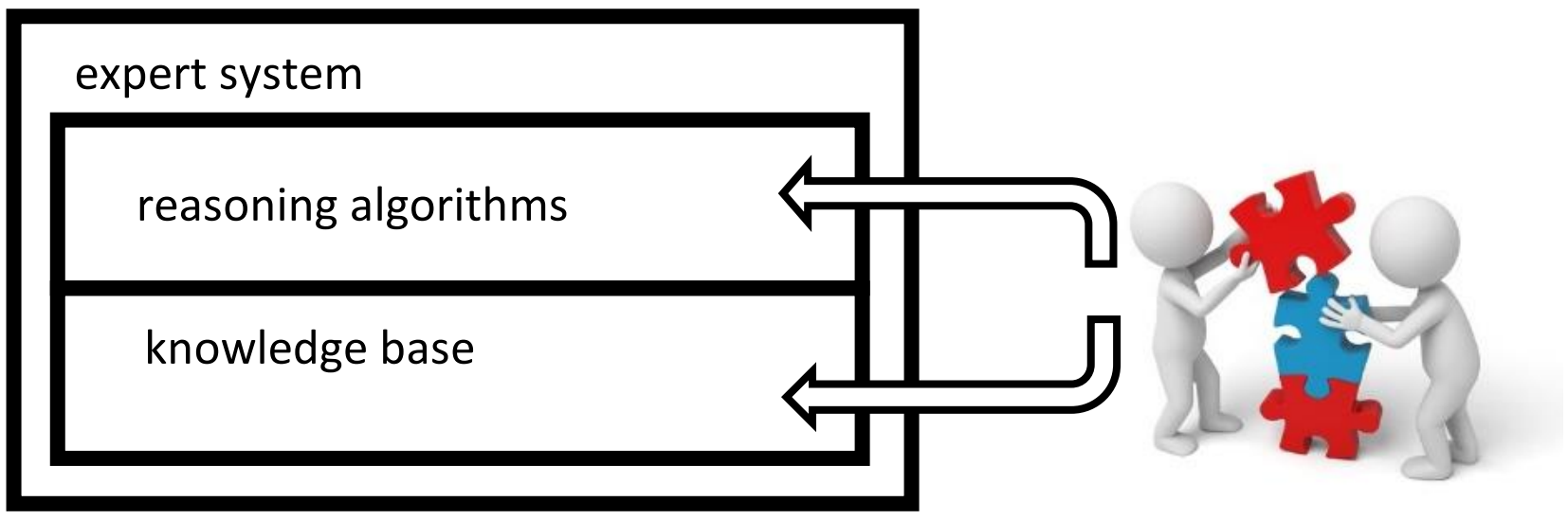

Figure 2.3. A rule-based system. (© 123RF)

Learning is the major problem, because the creation of rules and knowledge is very laborious. Often the rules and knowledge are built manually.

Artificial neural networks (or neural networks) are based on data-driven **connectionist computations**. They are composed of very simple computing elements that are interconnected in a way that roughly resembles human nervous tissue. The neural networks can be trained.

Neural networks suit particularly well to the problems of pattern recognition (Figure 2.4) and prediction with the help of numerous training samples. With vision humans can recognize quickly different types of objects, other people and their expressions and gestures. By hearing we can recognize speech and other sounds. For this kind of perception complex symbolic reasoning is not needed.

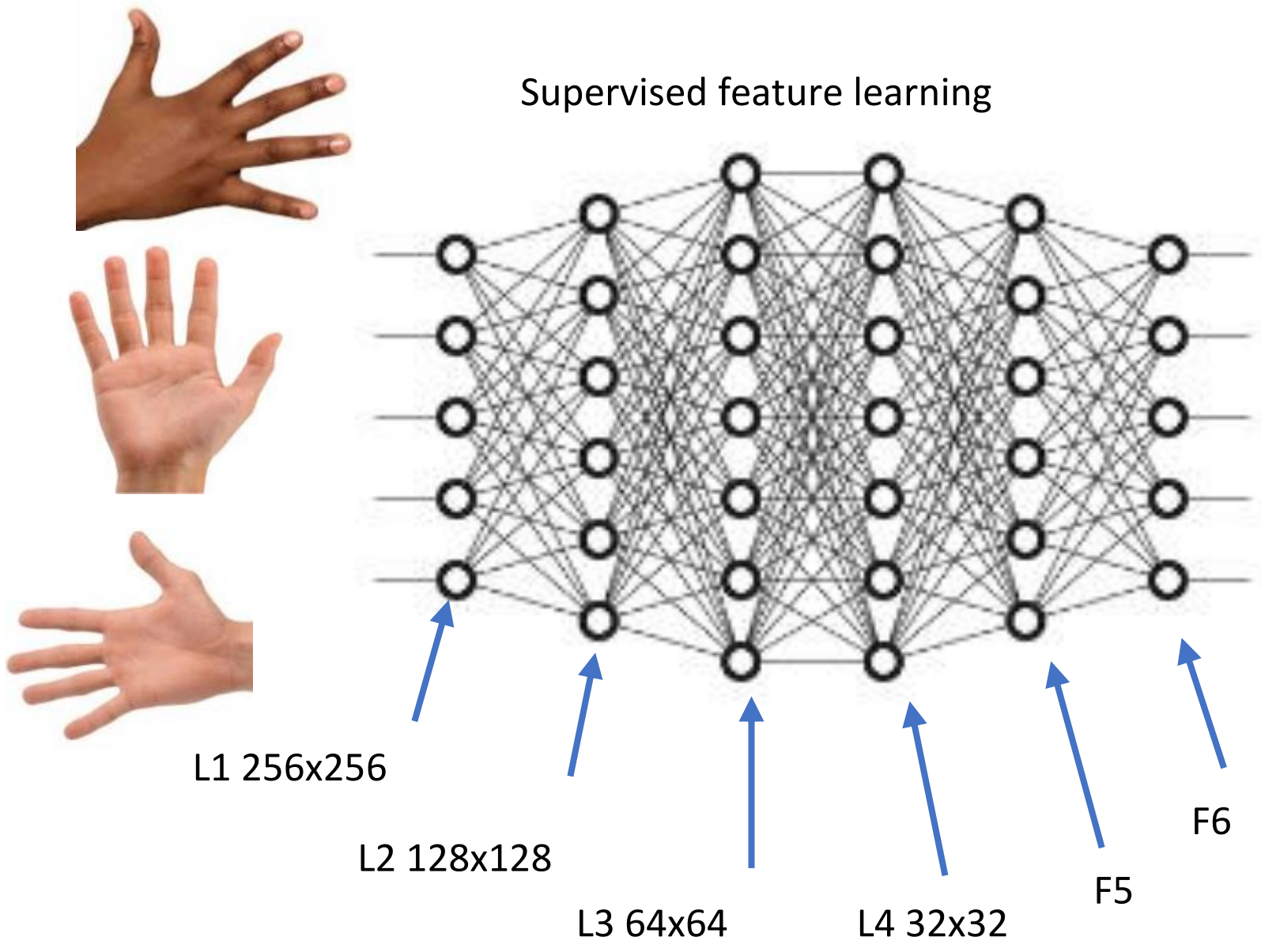

Figure 2.4. Pattern recognition with a multi-layer neural network. (© 123RF)

It is also possible to do pattern recognition, prediction and machine learning with methods of traditional statistical pattern recognition and data analysis (Duda et al., 2001), (Bishop, 2006). For this reason, instead of connectionist computing we use the term *data-driven artificial intelligence* in Chapter 3.

Traditionally, pattern recognition systems are trained with two optional approaches: supervised or unsupervised. In **supervised learning** training samples from known different categories are used, for example, representing numbers *0-9* for a number recognition problem. Based on these samples the system will learn classification criteria for the new unknown samples.

In **unsupervised learning** the system will learn classification criteria directly from the data. Usually this is called *clustering*. A major part of today's applications employ supervised learning. Together with the supervised and unsupervised learning, the third principle called **reinforcement learning** is also introduced in Chapter 4.

**Search methods** play a central role in symbolic AI. Many problems can be represented with simple directed graphs (Figure 2.5) or decision trees, which are composed of nodes that are connected to other nodes with links. An example search problem could be to find a route from a city to another by representing cities with nodes, connected with links. We could search for any route from Oulu to Helsinki or for the shortest route.

When the number of nodes in a network increases the search space grows, often making the search computationally heavy. There are various heuristic methods which can be used to reduce the search space (i.e. the number of nodes to be traversed) in a given application. However, the computational complexity is a central reason why it has not been possible in AI applications to go ahead from “toy-world” problems to realistic situations.

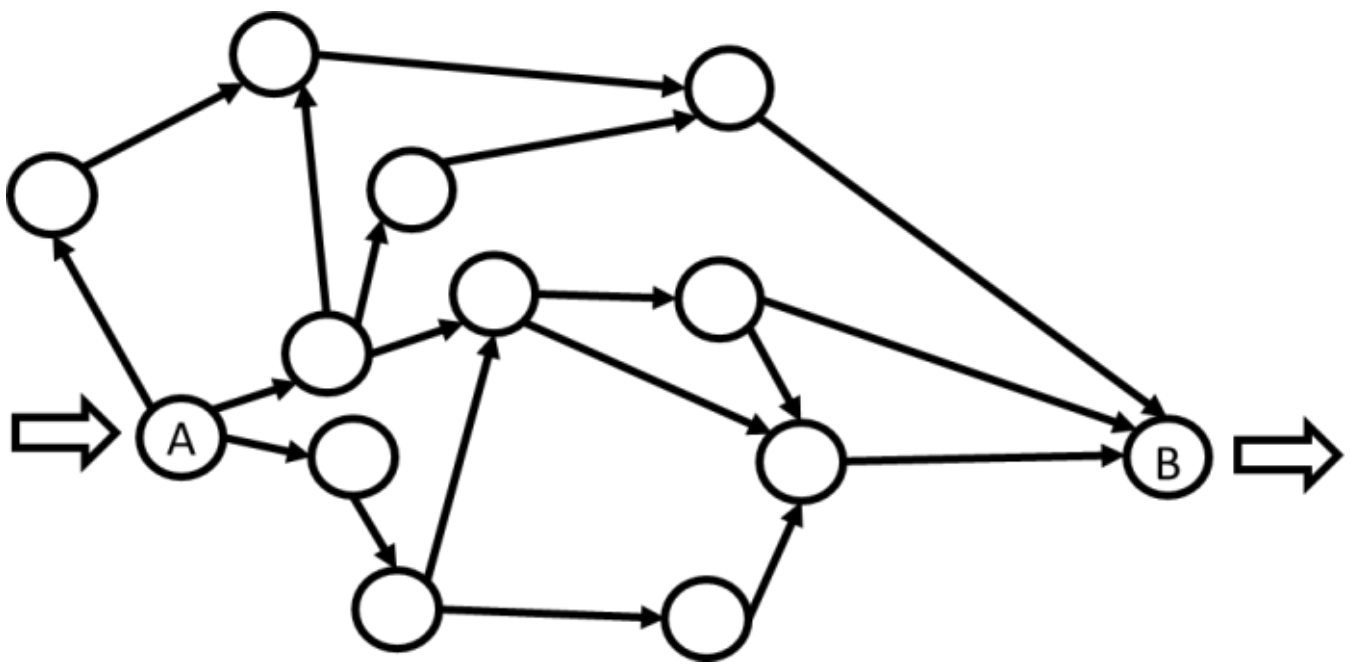

Figure 2.5. A directed search network.

## 2.3 Central Areas of AI

Figure 2.6 illustrates areas that are central to artificial intelligence. These are needed for implementing rational agents depicted in Figure 2.2.

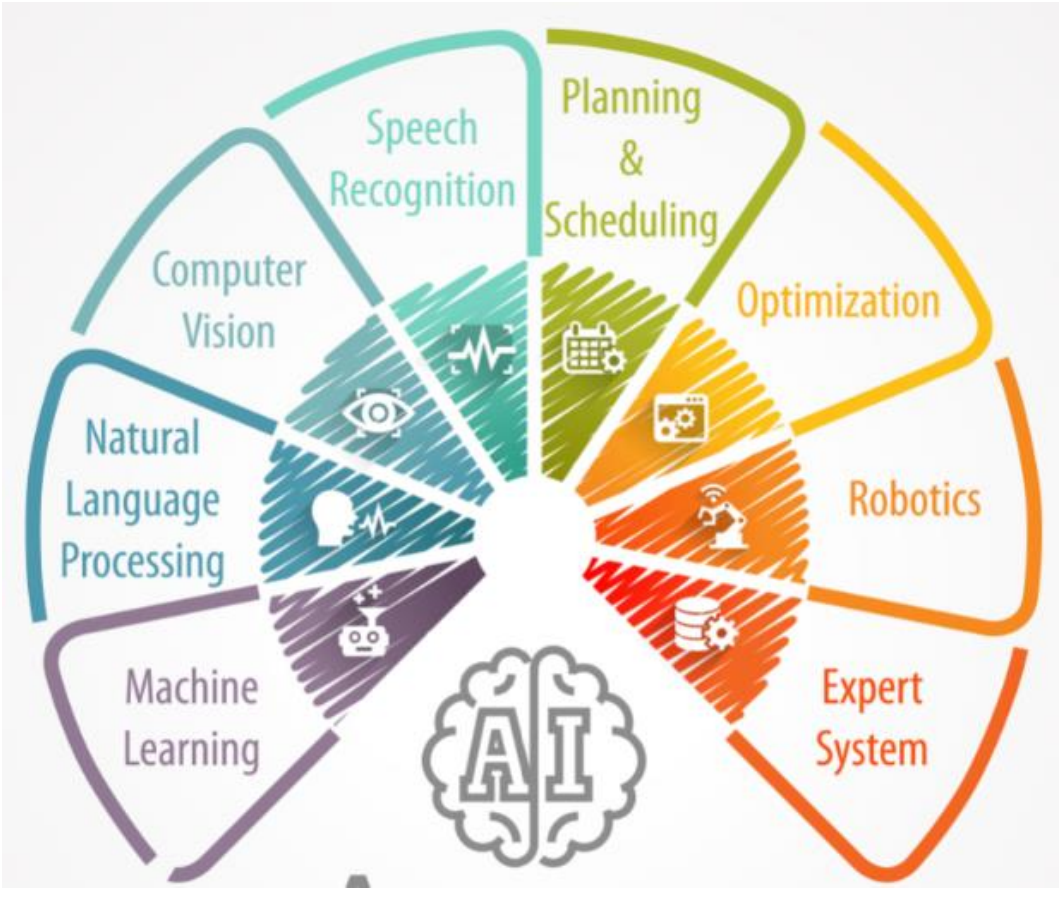

Figure 2.6. Central areas of AI. (© 123RF)

In addition to *machine learning* an intelligent system needs, as humans do, abilities for *natural language processing* and interpretation of sensory data using *computer vision* and *speech recognition*. The machine should also *plan and schedule* its actions. For achieving best possible results, the whole system and its different tasks should be *optimized*. With *robotics* the machine can execute different actions in the real world. There are also software robots in addition to mechanical robots. *Expert systems* (Figure 2.3), typically based on rule-based reasoning, are systems designed for solving some specific application problems.

## 2.4 Weak and Strong AI – or Super AI?

When discussing about artificial intelligence, it is often divided into three different levels of strength: weak, strong, or super AI.

All current applications of AI, including those presented in this book, represent *weak AI*. Only solutions for specific problems in chosen applications are possible. Human-like consciousness cannot be achieved. Examples of current weak AI applications include text recognition, speech recognition, language translation, recognition of humans, their faces, expressions and gestures from images and videos, playing chess and other games.

In *strong AI* the machine is approaching human intelligence, having at least some kind of consciousness about itself. It can use different types of background information in planning and making decisions. A fully autonomous car moving in a continuously changing environment would need already partially strong AI, at least for making decisions for handling conflict situations. Many lower level decisions, such as recognizing other cars, traffic signs or people, changing lane or avoiding crash, could be done with weak AI.

Strong AI is necessary, for example, for natural dialogue between humans and machines. The machine should be able to understand the contents of the dialogue, but this cannot be achieved with weak AI. Chapter 10 considers weaknesses of current artificial intelligence and how could we make progress towards strong AI.

A machine would have **super AI**, if its intelligence is higher than the sharpest and most talented human beings have. It could be born through “intelligence explosion” or “technological singularity”. Some scientists believe that super intelligence will be born soon after we have developed strong AI that resembles human intelligence with an ability to actively acquire new knowledge and experiences. The singularity is achieved when a strong AI can develop independently new even stronger solutions. Chapter 10 deals with super AI and discusses reasons why, according to our current knowledge, achieving it seems impossible.

## 2.5 History of Artificial Intelligence

In this section we will introduce a short history of AI, starting from some early related work in the 19th century. We focus on the progress since the birth of AI in the 1950s, and end in 2012 when the breakthrough in deep learning research took place. Information and original references for our presentation can be found, for example, from the history surveys of the textbooks by (Russell & Norvig, 2010) and (Negnevitsky, 2011). Additional information is given in the AI history article by Wikipedia (Wiki-History1), and in other articles presented in Internet, for example (Web-History2) and (Web-History3).

Some background for convolutional neural networks (CNN) was obtained from the ICPR 2018 keynote speech of Prof. Zhi-Hua Zhou titled ”An exploration to non-NN style deep learning” (Zhou, 2018). For the AI-related history of robotics and its applications we found pointers from an article in Finnish newspaper Helsingin Sanomat (Paukku, 2017).

In addition, some of the information for this survey has been borrowed from the professorship’s inaugural lecture given by the first author of this book (Pietikäinen, 1992). Interesting additional information on some early projects of AI can be found in (Knight, 2006).

The progress in many fields of science during the past few centuries had influence on the birth of artificial intelligence. Russell and Norvig (2010) examined this progress from the points of view of philosophy, mathematics, economics, neural sciences, computer engineering, control theory and cybernetics, and linguistics, respectively.

**19th century – 1956: Birth of AI**. There has been interest on intelligent machines since those days when the first computers were developed. British Charles Babbage, who designed the Analytical Engine based on mechanical calculations in the middle of 19th century, and his partner Ada Lovelace saw that this kind of programmable machine might be useful for intelligent operations like playing chess or composing music. Lord Byron’s daughter Ada Lovelace has been called as the world’s first computer programmer. Konrad Zuse from Germany, who in 1941 finished his Z-3 computer that is regarded as the word’s first programmable computer, applied Z-3 to the chess playing problem.

In 1872 Charles Darwin published the book ”The Expression of Emotions in Man and Animals”, in which he presented that facial expressions can be described as discrete categories related to emotions.

Already during the first years of computing history some scientists from different fields started to discuss about possibilities to develop “artificial brains”. A foundation for this was created by the theoretical work of British mathematician Alan Turing and the computer architecture based on it developed by John von Neumann.

Earliest ideas in research on intelligent machines were inspired by

1) research in neurology showing that brains include an electronic network composed of neurons,

2) Norbert Wiener’s cybernetics dealing with control and stability in electronic networks,
3) Claude Shannon’s information theory dealing with digital signals, and
4) Alan Turing’s theory of computation, which showed that any computation can be done digitally.

The paper by Warren McCulloch and Walter Pitts on ”A logical calculus of the ideas immanent to nervous activity” is regarded as the first actual work related to artificial intelligence. They proposed an artificial neural network model, in which each neuron is supposed to be in a binary state - on or off (McCulloch & Pitts, 1943). They also showed that this kind of network is able to learn.

The most notable person in creating the foundation for AI was Alan Turing. In his paper ”Computing machinery and intelligence” published in 1950 he suggested many fundamental research questions of the coming years, including playing games, natural language understanding and translation (e.g., from English to French), theorem proving, and breaking codes (Turing, 1950). He also presented his widely referred definition for evaluating machine intelligence, i.e., the Turing test introduced in Section 2.1.

Norbert Wiener, the pioneer of feedback-based cybernetics, mathematician and philosopher (Wiki-Wiener) presented that all intelligent activities are based on feedback mechanisms. In a feedback all earlier information produced by the system will have an effect on all later outputs (Wiki-Feedback). He suggested that it might be possible to simulate this kind of mechanisms with machines (Wiener, 1948).

Mathematician Claude Shannon had also significant contributions to the birth of AI with his pioneering research on information theory and foundations of digital signals (Wiki-Shannon). In 1950 he published an article "Programming a computer for playing chess". He presented that if all possible move combinations of chess pieces would be completely examined, then a typical chess game would include about $10^{120}$ possible moves – i.e. an enormous number of moves. With this Shannon demonstrated that it is necessary to use *heuristics* (“rules of thumb”) for finding the best possible solution.

Already in the early 1950s the first computer programs for AI problems were written. Among the best known are ”Logic Theorist” (Wiki-Logic) dealing with the theory of logic and the game program for checkers (Samuel, 1959). The first one introduced by Newell, Simon and Shaw in 1956 was using tree search in problem solving and was able to find proofs of theorems for simple proposition logic. Logic Theorist is considered as the first real AI program. In his publication on checkers Arthur Samuel was the first one to use the term *machine learning.*

In robotics area the science writer Isaac Asimov published three laws of robotics in 1950: 1) a robot cannot hurt people, or when resting it does not allow any person to make harm, 2) robot must obey orders given by a human, if they are not in conflict with the first law, and 3) robot must protect its own existence as long as this is not in conflict with rules 1 and 2. These laws of Asimov have not prevented developing robots for military purposes.

**1956 - end of 1960s: Time of great expectations.** The meeting held in 1956 at Dartmouth College, USA, is considered as the beginning of AI as a scientific field. Many future pioneers of AI research attended. When proposing this meeting in 1955, cognitive scientist John McCarthy used the term *artificial intelligence* for the first time. The objective of the meeting was to find out how to create a machine that would think like a human, would be capable of abstract thinking, problem solving and improving itself.

With the name Artificial Intelligence McCarthy wanted to emphasize the neutrality of this new research field and its differences to the narrow automata theory. He also wanted to avoid mixing AI to feedback-based cybernetics known from analog control theory. It is also speculated that one reason for choosing this name was that McCarthy wanted to avoid the acceptance of self-confident Norbert Wiener as a guru of AI, or having to argue with him.

In his article published in 2011 Ronald R. Kline introduces new, unpublished information about the birth of Dartmouth meeting and complex interactions between cybernetics, automation and artificial intelligence in the 1950s (Kline, 2011). History of cybernetics, as well as connections between AI and war history are covered in his book ”Machines: A Cybernetic History” (Kline, 2016).

We consider that McCarthy’s interpretation about AI as a new independent research field, strongly supported by computer science, is correct. The later progress of AI, its emphasis in software, huge investments to computing research and wide potential for applications support this choice.

In 1958 Prof. McCarthy developed LISP language which supports computing with symbols. For a long time it was the most important programming language in AI. Currently the most important one is Python developed in the Netherlands in the late 1980s by Guido van Rossum.

In the same year McCarthy published an article about programs with common sense (McCarthy, 1958), introducing Advice Taker program, which tackled general problems in the world. For example, it was capable of planning how to drive to the airport based on a few simple logical axioms. The program could accept new axioms, i.e. new knowledge about different tasks without re-programming. Therefore, it can be considered as the first knowledge-based system, which includes the central elements of knowledge representation and reasoning (Negnevitsky, 2011).

At that time is was often presented that after 25 years, around 1981, people can focus on free time activities, because machines will do most of the work. It was believed that intelligent activities are mainly based on smart reasoning, which could be without major problems implemented as computer programs. This kind of optimism continued until the end of 1960s.

McCarthy predicted that a breakthrough will be achieved in 5-500 years. He never retreated from this opinion (Web-McCarthy).

In 1957, Herbert A. Simon, an economist and later a Nobel laureate, predicted that machines will win humans in the game of chess in ten years – but this took 40 years (Wiki-Simon). Cognitive scientist Marvin Minsky predicted in 1967 that during one generation the problems of AI have been largely solved (Wiki-Minsky). Minsky, McCarthy and Simon were leading pioneers during the first decades of AI research.

Allen Newell and Herbert A. Simon (1961, 1972) developed General Problem Solver (GPS), a computer program that imitated human problem solving. It was the first program that tried to separate problem solving method from the data by using so called means-ends analysis. The given problem was defined as a set of states. In analysis, the difference between the current state and goal state was computed and the most suitable operator was then used for the next step. GPS was based on formal logic and search, which made it computationally very complex, and impractical for real world problems.

In 1969 Herbert A. Simon published his book ”The Sciences of the Artificial”, which emphasized the importance of representations (Simon, 1969).

From the beginning AI has been applied to solving mathematical problems that include symbolic computations. In the early 1960s the first computer program for calculus was introduced. Based on it, the MACSYMA software system for professional use was released in 1978. It could be used for solving many types of mathematical problems. With it, or corresponding newer software packages such as Maple (1984-) or Mathematica (1988-), the user only needs to define the mathematical problem to be solved in symbolic form. This technology is no longer considered as AI as it has been in routine use by mathematicians and engineers for a long time.

As an indication that AI can solve tasks requiring some form of intelligence was the ANALOGY program introduced in 1963. It could solve pattern recognition tests used when estimating human intelligence. Its principle was to go through by search different changes in the patterns such as addition, deletion and rotation.

Frank Rosenblatt introduced a neural network called Perceptron (Figure 2.7) (Rosenblatt, 1958). Following the optimism of that time he predicted that Perceptron will soon enable machines to walk, talk, make decisions and even do language translations.

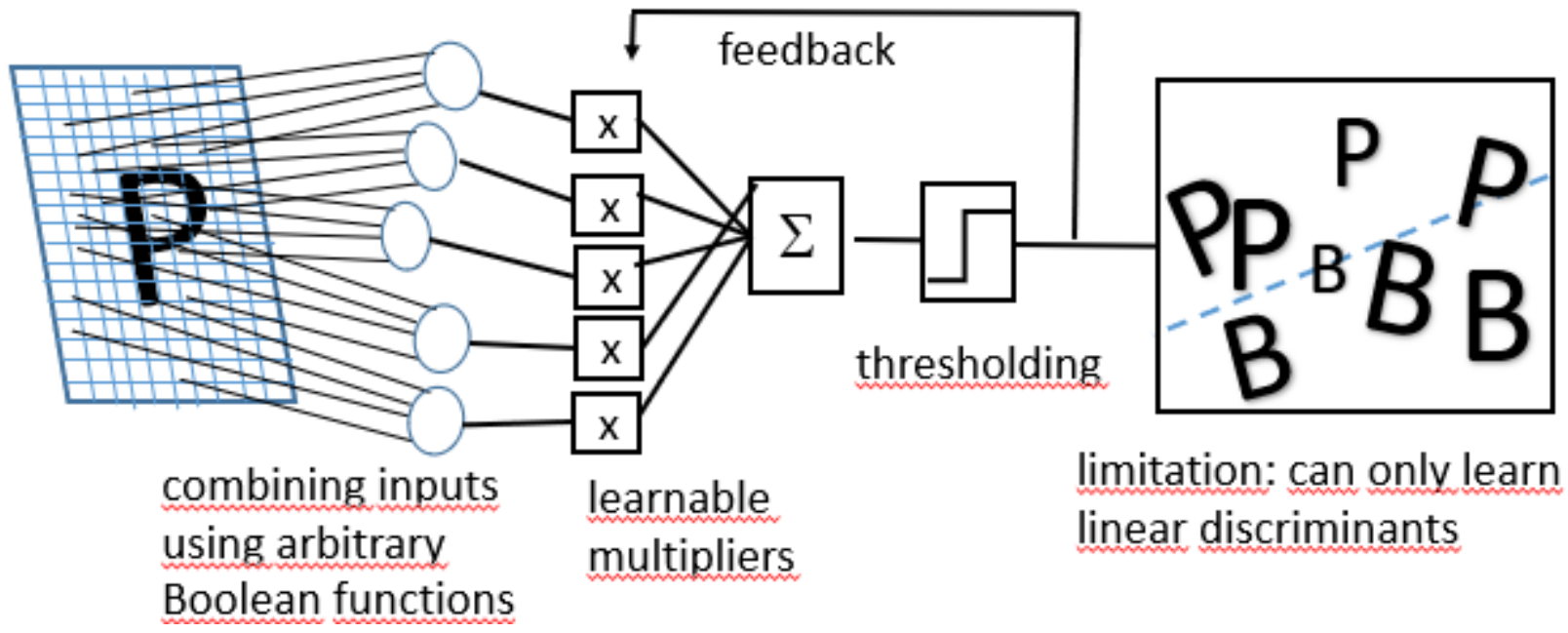

Figure 2.7. Perceptron neural network.

This principle was studied actively nearly the whole 1960s. However, Minsky and Papert (1969) showed in their book Perceptrons that the method has serious limitations: it can only use linear discriminants to separate features describing different objects. Therefore, the great expectations were far too optimistic. Minsky and Papert concluded that the data representations obtained with neural networks are not enough for intelligent actions. They knew that these limitations did not apply to multilayer networks, but there was no learning algorithm available for this purpose. This led to the end of nearly all neural network research for about ten years.

An interesting note about the progress in 1960s is that in 1965 Prof. Lotfi Zadeh published an article on ”Fuzzy sets”, which became very famous about twenty years later by creating foundation for numerous applications of fuzzy logic to different types of intelligent devices and systems (Zadeh, 1965).

In 1961, the first mass produced industrial robot manipulator, Unimate 1900, was put in operation. It was assembling cars at a General Motors plant (Paukku, 2017).

Also some computer vision related research was carried out in 1960s. A famous example of the early work, and poor technology prediction, is a summer project at Massachusetts Institute of Technology (MIT) in 1966. The key problems of computer vision were supposed to be largely solved in a summer project of a PhD student. Research results obtained in 1960s dealt, for example, with methods for analyzing 2-D images.

Closely related to the activities of our group, research on image texture analysis, dealing with gray scale variations caused by properties of objects' surfaces, began in the 1960s. Bela Julesz presented his research on texture discrimination based on second order statistics. A significant milestone in computer vision was the PhD dissertation of Larry Roberts finished in 1960 at MIT titled "Machine Perception of Three-Dimensional Solids". In his work he developed methods for detecting 3-D geometric information from 2-D scene images.

Face recognition research was also started during this decade, and continues today as one of the central application domains of computer vision and AI. First research was initiated in 1964 by Woodrow Bledshoe and the first PhD thesis was finished in 1972 by Takeo Kanade (Jain, 2013). Kanade from Japan became later a world-renowned scientist in computer vision and robotics at the Carnegie Mellon University in USA.

**End of 1960s – beginning of 1970s: Time of disappointments – first AI winter.** During this period much critique was presented against AI. The researchers had not been able to predict how difficult it would be to develop real applications. The too optimistic promises were not fulfilled, and the funding for research collapsed. This is called *the first AI winter*. Among the reasons for this winter were, for example, too small capacity of the computers and exponential growth of computational complexity when moving from "toy problems" to real problems.

Researchers mainly focused on general large-scale problems, in which only little, or no knowledge at all about the problem was used. Different search methods were mostly used in problem solving. This led to the explosion of computational complexity in real world problems, for example, in automatic translation from one language to another.

In the 1960s and 1970s, however, first programs for understanding natural languages, having very limited vocabulary and language structure, were developed. In 1964-66 the ELIZA program developed by Joseph Weizenbaum at MIT simulated discussion in English by using a simple pattern (a word, part of text) matching and replacement method. With ELIZA a user might believe that the machine understands something about the discussion.

Other attempts included the STUDENT program introduced by Daniel Bobrow in 1967 aimed at solving high-school algebra problems described with a natural language. In 1982 Terry Winograd developed SHRDLU for describing tasks in a simulated blocks world, and Warren ja Pereira's CHAT-80 for answering questions dealing with geography.

In the late 1960s Roger Schank, one of the leading AI researchers, developed a model for natural language understanding called "conceptual dependency theory" (Schank & Abelson, 1977). He was also a pioneer in case-based reasoning, in which the solution to a new problem is based on solutions developed for earlier resembling problems.

A foundation for genetic algorithms was created already in the early 1970s. John Holland developed an algorithm for handling artificial chromosomes with operations for selection, crossover and mutation. (Holland, 1975). At around the same time (beginning already in the1960s) the basis for evolutionary computing was invented, first by Ingo Rechenberg and later also by Hans-Paul Schwefel (Rechenberg, 1973), (Schwefel, 1995). Both genetic algorithms and evolutionary computing can be applied to many, earlier unsolved complex problems that need nonlinear search and optimization.

Stanley Kubrick's science fiction film ”2001: A Space Odyssey”, based Arthur C. Clarke's book, introduced the HAL 9000 computer, which inspired research on intelligent human-computer interfaces in the early 2000s. HAL was able to understand natural language, even by reading from lips, could speak, and interpret human actions and emotions.

In machine vision, research included methods for analyzing 2-D images, for example, to detect edges and segmenting images into regions. Among the first applications was automatic recognition of characters from printed text.

**1970 - end of 1980s: Era of knowledge engineering.** Research on knowledge engineering and representing application-specific narrow knowledge with rules began in 1970s. The implementation of a rule-based expert system includes separating application domain knowledge and rules used for problem solving through inference (Figure 2.3). Normally these kinds of systems should be able to give justification for the decisions.

In Marvin Minsky's article "A framework for representing knowledge" an important knowledge representation principle called *frame* was proposed, in which knowledge is divided into sub-structures called *stereotypic situations* (Minsky, 1975). The frames, originally developed from semantic networks, also belong to called structural knowledge representations. They were regularly used in the knowledge-based systems of the 1970s. The structural systems collect facts from chosen object or action types into taxonomic hierarchies, resembling taxonomies in biology (Russell & Norvig, 2010).

Among the best known first expert systems were PROSPECTOR (Hart and Duda 1977) - a tool supporting mineral exploration, DENDRAL developed for structural analysis of organic chemical structures in mass spectroscopy (Buchanan and Feigenbaum 1978), and MYCIN - a system developed for the diagnosis of infection diseases (Buchanan and Shortliffe 1984).

The development of expert systems was commercialized in the early 1980s, with too optimistic expectations of rapid success.

The goal was to implement computer programs imitating human reasoning to solve dedicated problems. The areas of application included medical diagnosis, exploration of soil minerals, studying oil wells, fault diagnosis of various machines, and tools for investigating credit standing of customers applying for loans.

The first commercially successful expert system was R1 (also called XCON) developed by Digital Equipment Corporation (DEC), at the time known for its VAX and PDP computers. R1 helped to configure computer systems according to the customers' orders, providing apparently significant savings for DEC. Nearly every major company in USA had an AI team to investigate expert systems (Russell & Norvig, 2010) – resembling today's AI hype!

In 1984 Douglas Lenat started his Cyc project dealing with modeling common sense knowledge. It is currently known as the world's longest AI project (Wiki-Cyc) with its first stable version 6.1 released in November 2017!

In 1982 Japan started a ”Fifth Generation” project, with an aim to develop in ten years massively parallel intelligent computers that would support the use of Prolog logic programming language. Following this, USA and UK started their own AI programs: MCC (Microelectronics and Computer Consortium) and Alvey (1983-87), but none of these achieved their ambitious goals.

The basic research carried out by Matti Pietikäinen at the University of Maryland in 1984-85 was supported by the Autonomous Land Vehicle Project, which belonged to DARPA's (Defence Advanced Research Project Agency) Strategic Computing Initiative (SCI, 1983-1993). The SCI program focused on developing advanced computing devices and artificial intelligence. The objective was to achieve complete machine intelligence in ten years, allowing computers to execute ten billion instructions per second with an ability to see, hear, talk and think like humans do! Integrating multiple technologies, a machine would compete with the human brain (Wiki-SCI). These very ambitious goals were not achieved, and in part this contributed to the beginning of the next AI winter.

At the same time the German Bundeswehr University achieved better results in autonomous driving than the DARPA program. In 1987, the robotic car developed by Professor Dickmanns and his team was able to drive at a maximum speed of 96 kilometers/hour on an empty road with intersections. His solution was based on advanced control engineering, and in a way honored Norbert Wiener's ideas that intelligent behaviors are based on feedbacks (Dickmanns & Graefe, 1988).

However, as a ”side product” of DARPA's program an AI-based tool was developed for planning logistics. It has been estimated that this tool helped save billions of dollars for planning the logistics for vehicles, freight and people during the Iraq War in 1991. This was more than all DARPA's funding for AI before that. In fact, this kind of situation is typical for the AI research. Highly ambitious goals are set, which means that much better software development tools and devices than existing ones are required. In this way the AI research has had a great influence on the progress of computing field in general, but it has not been given much credit for such developments. For a future-oriented research like AI, very powerful computing facilities and software are needed, and in this way the research helps develop new tools and computing environments for general use.

In the 1970s, statistical pattern recognition took the leading role in object classification (Duda & Hart, 1973), and remained as such until recently. Neural networks can be viewed as a massively parallel computing architecture for statistical pattern recognition.

Neural networks research experienced a renaissance in the1980s. New types of networks were introduced, removing problems of the earlier schemes. Among the best-known ones were Grossberg's network based on self-organizing adaptive resonance theory (1980), Hopfield's network (1982), multilayer-perceptron based on backpropagation (Rumelhart et al., 1986), and Teuvo Kohonen's self-organizing network (SOM) from 1982 for unsupervised classification (Kohonen, 2001). From today's viewpoint, the multilayer-perceptron was of special importance, because is laid foundation for more recent deep learning networks.

Interestingly, the first modern version of backpropagation algorithm (i.e., automatic differentiation) used for training multilayer neural networks was proposed as early as 1970 by Seppo Linnainmaa in his Master's thesis for University of Helsinki. Paul Werbos was the first one to apply backpropagation algorithm for neural network training in 1974.

Kunihiko Fukushima introduced the use of convolution operators in his paper on Neocognitron published in 1982 (Fukushima & Miyake, 1982). This was inspired by the research of David Hubel and Torsten Wiesel (1962) dealing with the structure of the visual cortex of mammals. Later Hubel and Wiesel were awarded with a Nobel Prize. Yann LeCun et al. (1989) applied backpropagation principle to the convolution layers of a neural network, with an application in character recognition. This method created basis for the current convolutional neural networks.

The roots of reinforcement learning are also in the 1980s, because Barto et al. (1983) introduced a method based on this principle for challenging control engineering problems. In reinforce-ment learning a system works in its environment receiving positive or negative feedback from its actions, and in this way learns little by little a suitable strategy for the actions.

The first AI boom in Finland began in the 1980s. In 1984 the Technology Development Center of Finland (Tekes) started its FINPRINT technology program on information technology. Its objective was to reach and preserve Finland's international competitiveness, especially in hardware and software. From its five sub-programs two were related to AI: 1) Artificial intelligence and 2) Applications of pattern recognition. The first one focused on knowledge engineering and expert systems, and the second one on speech communications, digital image processing and machine vision, and applications of medical engineering. Our research group participated in the second sub-program.

Research on speech recognition has been done already since 1950s, but more significant research started in the1970s. In USA, the research agency DARPA funded a speech understanding research program from 1971. One of its goals was to recognize speech from a corpus of at least 1000 words. DARPA was unhappy with the results and ceased funding the program.

In the late1970s, a team led by James Baker from Carnegie Mellon University, which had participated in the speech understanding program, was first to apply Hidden Markov Models (HMM) to speech recognition. HMM is based on Markov chains proposed in the 1960s by Lennard Baum. With HMM, knowledge based on acoustic, language and syntax information of the speech can be combined into a uniform model based on probabilities. The use of HMM models revolutionized speech recognition for decades (Wiki-Speech).

Paul Ekman introduced in 1978 a Facial Action Coding System (FACS), which had a great impact on facial expression analysis research (Ekman & Friesen, 1978). Later he presented significant improvements to the FACS (Ekman et al., 2002).

Computer vision research was getting wider and deeper in the 1980s. New methods were developed for analyzing images, image sequences (videos) and three-dimensional (3-D) images. Applications of machine vision in industrial automation were increasingly studied. Among the first applications were visual inspection, product sorting and assembly tasks, for example, in car manufacturing and electronics industries. In his book *Vision* David Marr from MIT proposed a model on how a description of 3D environment is created in human visual system through three phases: the primal sketch (2-D), intermediate (2.5-D) description, and the 3-D view-independent description (Marr, 1982). This model had a great influence on later computer vison research.

**1988 - 1993: The second AI winter.** During this period the funding sources for AI research and development were largely closed for several years due to the earlier major setbacks. The great expectations on expert systems failed and only very few productive systems were developed. It was easy to implement simple systems working in a "toy world", but generalization to real-world problems was very hard. Furthermore, the computational complexity of larger systems was overwhelming.

Most profits from the expert systems were obtained by the producers of development tools. One such company was Symbolics, Inc, developing computers specialized for processing symbolic information and the LISP language. We also got such a computer for our research and teaching.

The AI pioneer Marvin Minsky proposed that in the development of intelligent systems one should utilize both connectionist neural networks and symbolic processing (Minsky, 1990).

**End of 1980s - beginning of 2000s: Establishment as a research field.** Artificial intelligence started to establish its position as a field of science. The hypotheses created had to be tested carefully, the results had to be analyzed with statistical methods, and it was necessary to compare results to the others in the scientific community. The use of models based on probabilities increased, including Bayesian statistics and Bayesian networks. Rodney Brooks (MIT) was the frontrunner in the so called "Nouvelle AI", in which, for example, some intelligence for a robot can be built by combining, over time, simple behaviors (for example, "move forward", "avoid obstacles") to more complex actions (for example, how to behave with moving objects).

Methods based on soft computing were also investigated. In addition to neural networks, there was also growing interest on fuzzy logic introduced already in 1965 by Professor Lotfi Zadeh, as well as on genetic algorithms (Figure 2.8), and evolutional computing roughly imitating biological evolution.

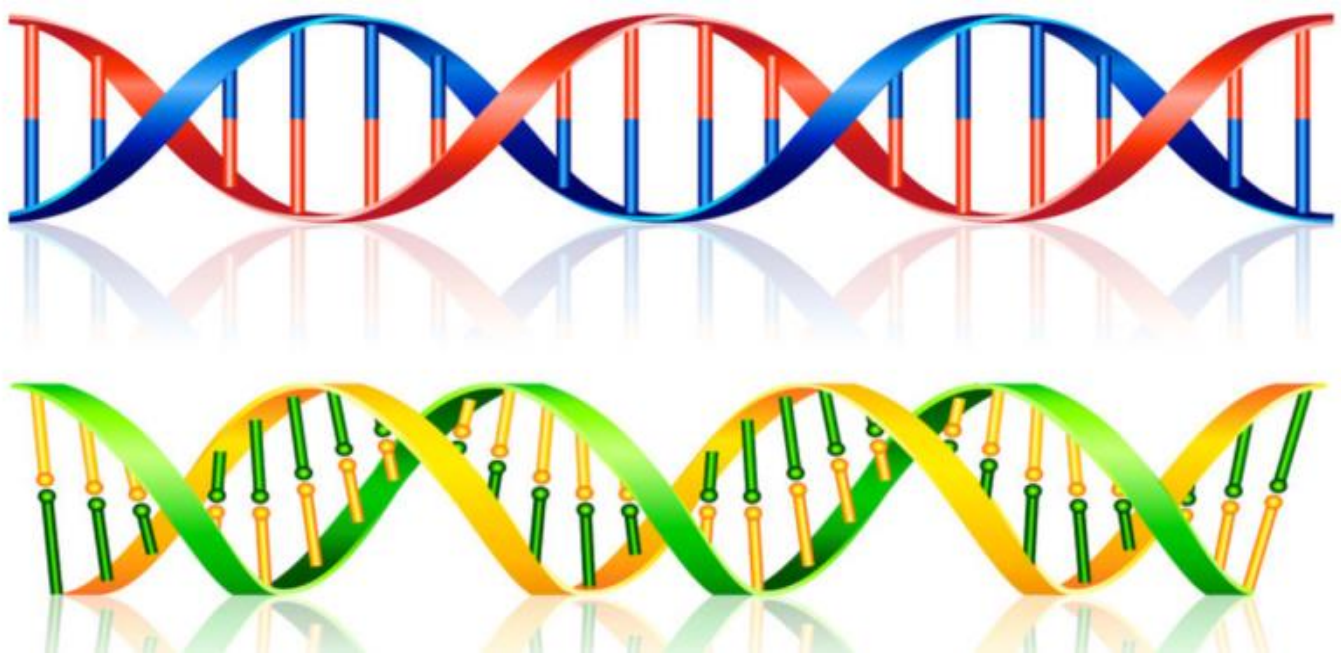

Figure 2.8. Genetic algorithms are using a simplified model from biological evolution. (© 123RF)

In deep neural networks, LeCun and Bengio (1995) presented a complete description of the convolutional neural network

(CNN). LeCun et al. (1998) applied CNN to the problem of recognizing handwritten characters.

In 1997, IBM's program on Deep Blue computer won the world's best chess player Gary Kasparov. This achievement received huge publicity and showed that in limited tasks a machine can beat humans.

The first world-championship competition of simulated soccer robots (RoboCup) was arranged in 1997. Jukka Riekki's team from the University of Oulu ranked 13th among 29 participants (Wiki-Robocup). In 1998 Sony introduced the prototype of a walking robot companion (dog) named AIBO, and its commercial version was released in 1999 (Wiki-AIBO).

Rosalind Picard published book ”Affective Computing”, creating the basis for the new scientific field also known as emotion AI (Picard, 1997).

In computer vision, the research was moving from methods working in controlled conditions, for example under uniform lighting, to solutions aimed for uncontrolled indoor or outdoor conditions. The eigenface method developed by Matthew Turk and Alex Pentland in 1991 was the most widely used method for face recognition.

In speech recognition, research was moving from *ad hoc* methods without any theoretical foundation to Hidden Markov Models (HMM) based on a strong mathematical theory. Many AI problems were approached from an agent-based viewpoint outlined in Section 2.1.

The milestones of our own research included inventing the basic Local Binary Pattern method (Ojala et al., 1996) and the first steps of its further development, and the method for geometric camera calibration for 3-dimensional (3-D) computer vision (Heikkilä & Silvén, 1997).

The Machine Vision Technology Programme 1992-96 funded by Tekes and industry was very important for developing Finnish machine vision applications in industry - and for our own research (Pau & Savisalo, 1996).

**Early 2000s - 2012: Intelligent agents, big data, new application areas.** Research on intelligent agents was increasing. Among the goals was to get rid of focusing on single problems and move towards intelligent systems composed of several parts. Internet has played a key role in the development of intelligent agents. AI technologies can be found, for example, in chatbots, search engines, and recommender systems advertising different commercial products.

In 2005 and 2006, the DARPA Grand Challenge and DARPA Urban Challenge competitions for self-driving vehicles were arranged. The winner of the first competition was the STANLEY robotic vehicle built at Stanford University. It was driving autonomously on a route of 132 miles in the desert at the speed of 22 miles per hour. In the latter competition, the winning BOSS vehicle developed at Carnegie Mellon University drove safely in the traffic at an air force base, following the traffic rules and avoiding pedestrians and other vehicles. Motivated by these results, Google started the development of a self-driving car in 2009.

In 2002, iRobot company introduced its Roomba robot vacuum cleaner, which was autonomously able to clean a whole apartment. Already in 1996 Electrolux had started marketing its own autonomous vacuum cleaner, but it was not capable of performing equally demanding tasks. In 2004 NASA's robot Mars rovers Spirit and Opportunity started to work independently on the surface of Mars (Wiki-Mars).

During the whole history of computing research, the main emphasis had been in algorithm development. However, already in the beginning of 2000s it was realized that in many problems the amount of data available is more important than the algorithms (Russell & Norvig, 2010). At the same time the capacity and speed of computers was continuously growing, making it possible to use more and more data.

With Internet is was possible to easily get big masses of data. The technology leaders of Internet, at that time especially Google and Microsoft, started significant investments into machine learning, computer vision, and their applications.

The Support Vector Machine (SVM) proposed already in the 1990s became a leading approach in statistical classification (Cortes & Vapnik, 1995). Its key idea is to project a feature vector to a higher dimensional feature space, in which the discrimination of different classes is easier.

Geoffrey Hinton et al. (2006) introduced a principle for unsupervised layer-wise training for a deep model and Honglak Lee et al (2009) presented how to use this kind of layerwise training for a CNN network. These were among the last steps towards making a breakthrough in deep learning in 2012 by Hinton's team (Krizhevsky et al., 2012).

The central figure in the research on causalities (i.e., the relationship between cause and effect) is Judea Pearl. In 2000 he published the first edition of his book dealing with modeling and use of causalities in reasoning (the second edition in 2009) (Pearl, 2009). He is also regarded as a key person in artificial intelligence based on probabilities and in developing Bayesian networks often used in reasoning.

In 2011, i.e. before the breakthrough in deep learning, IBM's Watson AI engine was able to win humans in the demanding TV quiz show Jeopardy.

Research on computer vision became wider and deeper. The Viola & Jones method proposed in 2001 was a milestone in real-time face detection from images. New effective methods for measuring image properties were developed, including SIFT, LBP and HOG, which were widely used in different types of applications. So-called bag-of-words (BoW) principle became the state-of-the-art approach for representing textured images. The LBP histogram (see Chapter 7) is a simple example of BoW representation (Liu et al., 2019).

Recognition of faces and human actions became an increasingly important research area. These are central for many applications when computing transforms from machine-centered to more human-centered. The importance of machine learning increased when it was possible to use big amounts of data. Computer vision started to be used in natural environments in our everyday lives, such as smart phones, biometric recognition, video surveillance, and Internet search engines.

Very significant scientific breakthroughs related to AI were achieved in Finland in the late 1990s and early 2000s. Perhaps the best known of these are given below.

Professor Erkki Oja's team developed methodology for independent component analysis (ICA), in which a multidimensional signal is divided into its additive sub-components (Hyvärinen et al., 2001). ICA is a special case of the blind source separation, for example in the so called "cocktail party" problem we could try to hear a person's speech in a noisy room. The Local Binary Pattern method and its application to face recognition led to major breakthroughs in the early 2000s (Chapter 7) and to the first method for recognizing spontaneous facial micro-expressions in 2011 (Chapter 9). Professor Samuel Kaski's team developed various dimensionality reduction methods for data visualization (Venna et al., 2010). The frequently cited image noise reduction method developed by Professor Karen Egiazarian's team can also be considered as an AI method (Dabov et al., 2007).

## 2.6 A Summary of AI History

Table 2.2. AI from its birth to early 2000s

| | |
|---|---|
| **Birth of AI (- 1956)** | Beginning of neural networks: McCulloch & Pitts (1943) A logical calculus of the ideas immanent in nervous activity |
| | Norbert Wiener considered importance of feedback in intelligent actions (1948). |
| | Foundations for AI: Turing (1950) Computing machinery and intelligence |
| | Isaac Asimov presented three laws of robotics (1950) |
| **Time of great expectations (1956 – end of 1960s)** | Dartmouth College AI meeting (1956)<br>LISP language (McCarthy, 1958) |
| | Advice Taker – a reasoning program (McCarthy, 1958) |
| | Perceptron neural network (Rosenblatt, 1958) |
| | General Problem Solver (GPS) (Newell & Simon, 1961) |
| | Face recognition research begins (1964-) |
| | Fuzzy sets (Zadeh, 1965) |
| **Time of disappointments – first AI winter (end of 1960s – beginning of 1970s)** | Too generic problems, methods not suitable for real-world problems, mainly search methods – no domain knowledge employed Neural network research almost ends: Minsky & Papert (1969) Perceptrons |
| | First steps towards natural language analysis: ELIZA, STUDENT, SHRDLU, CHAT-80 |
| | Stanley Kubrick's film "2001: A Space Odyssey" introduced intelligent HAL 9000 computer (1968) |
| **Period of knowledge engineering (1970 – end of 1980s)** | Minsky (1975) A framework for representing knowledge |
| | First rule-based expert systems: PROSPECTOR, DENDRAL, MYCIN, XCON etc. |
| | The Cyc project modeling common sense knowledge begins (1984-) |
| | Prolog language for logic programming |
| | Statistical pattern recognition comes to major role in object classification |
| | Return of neural networks: Self-organizing adaptive resonance theory (Grossberg), Hopfield's network (Hopfield), self-organizing map (Kohonen) |
| | Multilayer perceptron using back-propagation (Rumelhart et al., 1986), convolution in neural networks (Fukushima, 1982), (LeCun et al., 1989) |
| | Reinforcement learning for control engineering problems (Barto et al., 1983) |
| | Genetic algorithms and evolutionary computation (Holland, Schwefel and Rechenberg) |
| | Facial action coding system FACS (Ekman & Friesen, 1978) |
| | Model of human visual system for interpretation of 3-D scene images (Marr, 1982) |
| | Progress in speech recognition: E.g., use of Hidden Markov Models (HMM) |
| **Second AI winter (1988-1993)** | Great expectations on expert systems failed: real-world problems are too difficult |
| | In intelligent systems both connectionist neural networks and symbolic computing is needed (Minsky, 1990) |
| **Establishment as a scientific discipline (end of 1980s – beginning of 2000s)** | Bayesian statistics an Bayes networks (Pearl etc.), "nouvelle" AI (Brooks), fuzzy logic (Zadeh), artificial life and genetic algorithm |
| | Progress in deep neural networks (LeCun & Bengio, 1995), (LeCun et al., 1998) |
| | IBM's Deep Blue wins Gary Kasparov in chess (1997) |
| | Sony's autonomously walking robot dog AIBO (1998, 1999) |
| | Creating foundations for Affective computing (Picard, 1997) |
| **Intelligent agents, big data, new application areas (beginning of 2000s – 2012)** | Research on intelligent agents broadens: Internet gets widely used, AI technologies e.g. in chatbots, search engines, and product recommendation systems |
| | Demonstrations with self-driving vehicles: DARPA Grand Challenge and DARPA Urban Challenge; Google's vehicle development starts |
| | Data comes to central role instead of algorithms |
| | Deep neural networks progress: Unsupervised layer-wise training (Hinton et al., 2006), (Lee et al., 2009) |
| | Support vector machine (SVM) has leading role in statistical classification |
| | Progress in analyzing cause-effect (causality) problems (Pearl, 2009) |
| | Significant AI results from Finland: Independent component analysis (Hyvärinen et al., 2001); Local Binary Pattern and its application to face recognition (Ojala et al., 2002), Ahonen et al. (2004, 2006); Dimensionality reduction methods (Venna et al., 2010); Image noise removal (Dabov et al., 2007); Recognition of spontaneous micro-expressions (Pfister et al., 2011) |
| | Bag-of-words principle becomes mainstream is representing images with textures |
| | IBM's Watson AI wins humans in TV quiz show Jeopardy (2011) |
| | Breakthrough in deep learning: (Krizhevsky et al., 2012) |

## 2.7 References

Barto AG, Sutton RS & Anderson CW (1983) Neurolike adaptive elements that can solve difficult learning control problems. IEEE Transactions on Systems, Man and Cybernetics 13:834-846.

Bishop CM (2006) Pattern Recognition and Machine Learning. Springer, 738 p.

Cortes C & Vapnik VN (1995) Support vector networks. Machine Learning 20 (3):273-297.

Dabov K, Foi A, Katkovnik V & Egiazarian K (2007) Image denoising by sparse 3-D transform-domain collaborative filtering. IEEE Transactions on Image Processing 16 (8):2080-2095.

Dickmanns E & Graefe V (1988) Dynamic monocular machine vision. Machine Vision and Applications 1:223-240.

Duda RO & Hart PE (1973) Pattern Classification and Scene Analysis. Wiley-Interscience, 482 p.

Duda RO, Hart PE & Stork DG (2001) Pattern Classification, 2nd Edition. Wiley Interscience, 654 p.

Ekman P & Friesen W (1978) Facial Action Coding System: A Technique for the Measurement of Facial Movement. Consulting Psychologists Press, Palo Alto.

Ekman P, Friesen WV & Hager JC (2002) Facial Action Coding System: The Manual on CD ROM. A Human Face, Salt Lake City.

Fukushima K & Miyake S (1982) Neocognitron: A new algorithm for pattern recognition tolerant of deformations and sifts in position. Pattern Recognition 15(6):455-469.

Heikkilä J & Silvén O (1997) A four-step camera calibration procedure with implicit image correction. Proc. IEEE Conference on Computer Vision and Pattern Recognition, June 17-19, San Juan, Puerto Rico, 1:1106-1112.

Hinton GE, Osindero S & Teh YW (2006) A fast learning algorithm for deep nets. Neural Computation 18 (7):1527-1554.

Holland JH (1975) Adaptation in Natural and Artificial Systems. University of Michigan Press, Ann Arbor.

Hyvärinen A, Karhunen J & Oja E (2001) Independent Component Analysis. John Wiley & Sons, 506 p.

Jain AK (2013) 50 years of Biometric Research. Almost ~~The~~ Solved, The Unsolved, and The Unexplored. Keynote lecture. International Conference on Biometrics, 2013.

Kline RR (2011) Cybernetics, automata studies, and the Dartmouth conference on artificial intelligence. IEEE Annals of the History of Computing 33:5-16.

Kline RR (2016) Machines: A Cybernetic History. WW Norton & Company, 432 p.

Knight H (ed.) (2006) Early artificial intelligence projects: A student perspective.

Kohonen T (2001) Self-Organizing Maps, Third Edition, Springer.

Krizhevsky A, Sutskever I &, Hinton GE (2012) Imagenet classification with deep convolutional neural networks. Advances in Neural Information Processing Systems, 1097-1105.

LeCun Y, Boser B, Denker JS, Henderson D, Howard RE, Hubbard W & Jackel LD (1989) Backpropagation applied to hand-written zip code recognition. Neural Computation 1(4):541-551.

LeCun Y & Bengio Y (1995) Convolutional networks for images, speech and time series. The Handbook of Brain Theory and Neural Networks (ed. MA Arbib), MIT Press.

LeCun Y, Bottou L, Bengio Y & Haffner P (1998) Gradient-based learning applied to document recognition, Proceedings of the IEEE 86(11):2278-2324.

Lee H, Grosse R, Ranganath R & Ng AY (2009) Convolutional deep belief networks for scalable unsupervised learning of hierarchical representations. ICML '09 Proceedings of the 26th Annual International Conference on Machine Learning, 609-616.

Liu L, Chen J, Fieguth P, Zhao G, Chellappa R & Pietikäinen M (2019) From BoW to CNN: Two decades of texture representation for texture classification. International Journal of Computer Vision 127(1):74-109.

Marr D (1982) Vision: A Computational Investigation into the Human Representation and Processing of Visual Information. San Francisco: W. H. Freeman and Company.

McCarthy J (1958) Programs with common sense. Proceedings of the Symposium on Mechanisation of Thought Processes, vol. 1, London, 77-84.

McCulloch WS & Pitts W (1943) A logical calculus of the ideas immanent to nervous activity. Bulletin of Mathematical Biophysics 5, 115-137.

Minsky ML & Papert SA (1969) Perceptrons. MIT Press, Cambridge, MA.

Minsky ML (1975) A framework for representing knowledge. The Psychology of Computer Vision (ed. P Winston), McGraw-Hill, New York, 211-277.

Minsky ML (1990) Logical vs. analogical or symbolic vs. connectionist or neat vs. scruffy, In Artificial Intelligence at MIT, Expanding Frontiers, Patrick H. Winston (Ed.), Vol.1. MIT Press, 1990. Reprinted in AI Magazine, Summer 1991.

Negnevitsky M (2011) Artificial Intelligence – A Guide to Intelligent Systems, Third Edition. Pearson Education Limited, 479 p.

Newell A & Simon HA (1961) GPS, a program that simulates human thought. Lernende Automatten (ed. H Billing), R Oldenburg, Munich, 109-124.

Newell A & Simon HA (1972) Human Problem Solving, Englewood Cliffs, NJ: Prentice-Hall.

Ojala T, Pietikäinen M & Harwood D (1996) A comparative study of texture measures with classification based on feature distributions. Pattern Recognition 29(1):51-59.

Pau LF & Savisalo H (1996) Machine Vision Technology Programme 1992-1996 – Evaluation report. TEKES Technology Programme Report 16/96, 34 p.

Paukku T (2017) Tekoälyn viisasten kivi on superäly (The philosopher’s stone is super AI). Helsingin Sanomat 20.6.2017.

Pearl J (2009) Causality: Models, Reasoning and Inference. Cambridge University Press, 484 p.

Picard R (1997) Affective Computing. MIT Press.

Pietikäinen M (1992) Tekoälyn kehittäjillä mittavia haasteita (Developers of AI face major challenges). AKTUUMI 4:23-28.

Rechenberg I (1973) Evolutionsstrategien – Optimisieriung Technischer Systeme Nach Prinzipien der Biologischen Information. Frommann- Holzbook-Verlag. Stuttgart.

Rosenblatt F (1958) The perceptron: A probabilistic model for information storage and organization in the brain. Psychological Review 65(6):386-408.

Rumelhart DE, Hinton GE & Williams RJ (1986) Learning internal representations by error-propagation. Parallel Distributed Processing: Explorations in the Microstructure of Cognition (eds. DE Rumelhart & JL McClelland) Volume 1, Issue 6088. MIT Press, Cambridge, 318-362.

Russell S & Norvig P (2010) Artificial Intelligence: A Modern Approach, 3rd Edition. Pearson, 1152 p.

Samuel A (1959) Some studies in machine learning using the game of checkers. IBM Journal of Research and Development 44(1.2).

Schank RC & Abelson RP (1977) Scripts, Plans, Goals, and Understanding: An Inquiry into Human Knowledge Structures. Hillsdale, NJ : Lawrence Erlbaum Associates, 256 p.

Schwefel H-P (1995) Evolution and Optimum Seeking. John Wiley, New York.

Simon HA (1969) The Sciences of the Artificial. The MIT Press, 1969.

Turing AM (1950) Computing machinery and intelligence. Mind 59:33-46.

Venna J, Peltonen J, Nybo K, Aidos H & Kaski S (2010) Information retrieval perspective to nonlinear dimensionality reduction for data visualization. Journal of Machine Learning Research 11:451-490.

Wiener N (1948) Cybernetics: Or Control and Communication in the Animal and the Machine. Hermann & Cie & MIT Press; 2nd revised ed. 1961.

Zadeh L (1965) Fuzzy sets. Information and Control 8(3):338-353.

Zhou Z-H (2018) An Exploration to Non-NN Style Deep Learning. Keynote lecture, ICPR 2018, Beijing.

Web-History2: The History of Artificial Intelligence, Harvard University

Web-History3:The History of Artificial Intelligence, University of Washington

Web-McCarthy: John McCarthy: Computer scientist known as the father of AI, Independent 1.11.2011

Wiki-AIBO: AIBO

Wiki-Cyc: Cyc

Wiki-History1: History of Artificial Intelligence

Wiki-Logic: Logic Theorist

Wiki-Mars: Mars Exploration Rover

Wiki-Minsky: Marvin Minsky

Wiki-Robocup: Robocup 2D Soccer Simulation League

Wiki-SCI: Strategic Computing Initiative

Wiki-Simon: Herbert A. Simon

Wiki-Shannon: Claude Shannon

Wiki-Speech: Speech Recognition

Wiki-Wiener: Norbert Wiener

Wiki-Feedback: Feedback

Wiki-Shannon: https://en.wikipedia.org/wiki/Claude_Shannon

# 3 Artificial Intelligence Representation Methods

## 3.1 Introduction

Herbert A. Simon (1916-2001), a Nobel laureate in economics who studied problem solving processes in human beings and artificial intelligence, emphasized in his book “The Sciences of the Artificial” (1969) that “solving a problem simply means representing it so as to make the solution transparent”. Similarly, Albert Einstein has been claimed to have said "if I had an hour to solve a problem, I would spend 55 minutes to diagnose it and five minutes to solve it".

A major challenge in implementing systems that perform functions that are considered intelligent is to represent problems and their parts in such forms from which they can be automatically solved. The task can be, for example, a medical diagnosis, car lane keeping, language translation, or home robotics. Suitable representations have been sought, e.g., from neuroscience, mathematics, formal logic, and imitated human intelligence. The choice of a proper representation is thought to lead to an easier solution. For example, a writing is a representation that substantially facilitates communication between people: the phonemes, syllables, or whole words are coded by letters. The ability to read and write are central to human learning, so it is natural to look for similar solutions for learning capable artificial intelligence.

Key issues in implementing an intelligent lane guard or robot include actively collecting data from the environment, transforming it into meaningful information, and appropriate action based on its interpretation. This problem of information representation and modeling has been solved by both symbolic and data-driven methods (Russell & Norvig, 2010). The latter, usually implemented by connectionist neural networks, represents the current mainstream of artificial intelligence - with a primary interest in developing learning algorithms and handling data masses. One explanation for the popularity of data-driven methods is the available easy-to-use tools that allow even for persons not familiar with statistical methods to go forward with their own data. These are offered by, among others, by Google, Microsoft, Facebook and Amazon.

Representations still play a significant role. Boundaries can be drawn based on whether a representation is understandable for humans, or only suitable for automatic processing. Ideally, an artificial intelligence application is built on data and knowledge as well as application-independent algorithms based on them (Figure 3.1). However, to date, the algorithms and ways of representing knowledge have not been independent. This is due to the gap between the representations and algorithms of symbolic and data driven solutions. This is a major obstacle to the development of man-made versatile artificial intelligence.

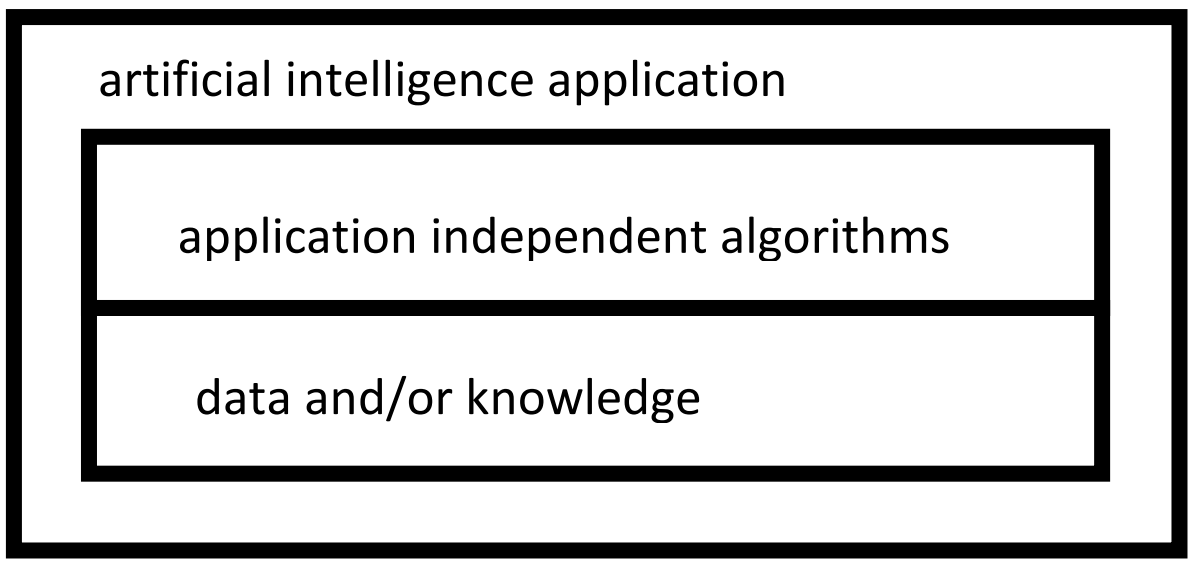

Figure 3.1. A typical artificial intelligence application.

## 3.2 Symbolic Artificial Intelligence

Symbolic Artificial Intelligence seeks to present knowledge as tangible elements, such as edges, angles, objects, their interrelationships, and rules that describe the conditions and effects of actions. The information is typically represented as text strings. If the presentation of data is successful, functionality that can be considered intelligent can be implemented, for example, by logic programming. This can be considered to mimic the corresponding human deductive problem solving.

Let's look at a simple example that finds the relationship between generations, when the following direct parents are known: Alan is Ann's father, Elizabeth is Ann's mother and Ann is Eric's mother. In Prolog (Programmed Logic) (Wiki-Prolog) this could be expressed

```
father (Alan, Ann)
mother (Elizabeth, Ann)
mother (Ann, Eric)
```

where the words "father" and "mother" are so called predicates. We are still defining parenting relationships with conditional statements, which are the rules that underlie reasoning. Here the capital letters A, B and X represent variables that can be replaced by any fact

```
grandparent (A; B) :- parent (A, X), parent (X, B)
parent (A, B) :- father (A, B)
parent (A, B) :- mother (A, B)
```

Now the Prolog interpreter can be asked "?-grandparent (Z, Eric)", and the obvious answer is given to the person by automatic reasoning. In the process the Prolog interpreter matches the predicates to the conditional statements, finding all the results that meet the conditions:

```
Z = Elizabeth
Z = Alan
```

The challenge of symbolic artificial intelligence is to acquire and build the necessary knowledge base. In our example, the information could have come from the population register, or from data collected by a genealogist. In a more general case, the learning process of the rules should be learned.

Essential aspects of presenting knowledge symbolically and implementing knowledge systems are
**syntax**: the rules of human and machine-understandable ways to create descriptions
**semantics**: meanings defined according to syntactic rules
**ontology**: A knowledge structure that describes concepts and their semantic relationships.

Due to shortcomings in automatic learning, symbolic artificial intelligence still requires manual work to gather information and prepare rules. Douglas Lenat's Cyc project modeling common sense knowledge, started in 1984, is an example of the challenges of this solution. It has so far required over 1,000 man-years to implement over one million rules, according to some sources more than 20 million (Wiki-Cyc). This has not been enough to make the system capable of self-directed learning.

Compared to the estimated 40,000 person-year development effort of Linux kernel, the 200,000 person-years consumed by the Manhattan project to create nuclear bomb, or the construction of the Kheops pyramid, which needed an estimated 3 million person-years, the Cyc project is not huge. Nevertheless, more investments are unlikely, since no one can say how many rules should be built before Cyc starts to produce rules on its own and develops into super-intelligence, if ever.

In the early days of artificial intelligence, efforts were made to represent knowledge with natural language for logic programming. Cyc is the only one of those developments, which has not remained a small-scale demonstration or a narrow single application focused realization. It has consistently been one of the most controversial developments in artificial intelligence research, seen not only as a great opportunity, but also as a complete disaster. Depending on the observer, either too much has been invested into a dead-end, or too little to an important service for the humanity. In any case, symbolic representations may turn invaluable if machines need to justify their decisions.

Advances in machine perception capabilities may, in the future, allow the acquisition of the symbolic information needed by systems like Cyc. Developers expect future computer vision methods will be able to extract accurate depictions of everyday situations and spatial relationships from the pixels in the images. In that case, thanks to Cyc's existing rule base, genuinely smart cooking and cleaning home robotics could be quickly made possible.

An example of the difficulty of predicting the technological future over short to medium term is the "object profile" introduced in the 1999 MPEG-4 video encoding standard. It took advantage of the predicted “soon to be reached breakthroughs" of artificial intelligence and pattern recognition. The idea was to encode the moving objects and related changes separately from the unchanged background. However, it is only now 20 years later that the necessary knowhow is available. At the same time other video coding methods have improved, and the "object profile" is not computationally viable.

A significant strength of symbolic artificial intelligence is that the representations, reasoning chains and results are in a form that can be understood by humans. Thus, erroneous operations can be traced and corrected. The inability of the inference machine to understand meanings is a side issue here.

Symbolic artificial intelligence has been criticized based on the famous Chinese room argument (Wiki-Searle), since it is not considered to have an understanding of the meanings: a message is sent to a person under the door in Finnish and must be translated into Chinese. If he doesn't know Chinese, or Finnish, or neither language, he uses Google Translate to find his answer, print it out on paper, and slip it under the door to the questioner.

### 3.3 Data-driven Artificial Intelligence

Data driven artificial intelligence is easier for application developers, because it does not refine data into semantically meaningful descriptions. Instead, the system builds and optimizes its own representation, for example, based on input images and recognizes the objects in new images. This is very attractive because training can be automated.

Unfortunately, this comes at a price: the built-in representation may be incomprehensible to humans, which can make it extremely difficult to find explanations for system functions and errors. For example, one layer of a translation machine's neural network can have millions of coefficients, each of which has some meaning to the whole. The Chinese room argument justifies not to consider such technology as artificial intelligence. On the other hand, how many bilingual Finnish and Chinese speakers can explain how they make quick translations?

In data-driven artificial intelligence, reliability is pursued by providing huge amounts of data during the learning phase so that all options are covered. Similarly, learning Chinese (or Finnish) is a long and laborious process.

Although artificial neural networks are often claimed to represent inductive reasoning that builds structures, i.e. symbols, from their observations, their capacity is still quite limited. For example, making three dimensional interpretations from two dimensional images fed to a neural network is an unsolved challenge. This alone is a barrier to numerous everyday applications that are deceptively easy for humans. Importing physical knowledge appears to be often easier by symbolic means, in practice, by mathematical models of reality measured by sensors.

## 3.4 Symbolic + Data-driven Artificial Intelligence

There is still a significant gap between symbolic and data driven solutions that has been investigated over decades. Based on biological findings, David Marr (1945-1980), in a book posthumously completed by his fellow researchers (Marr, 1982), outlined the procedure depicted in Figure 3.2 for constructing three-dimensional models from two-dimensional images. It is a combination of data-driven solutions and symbolic representations.

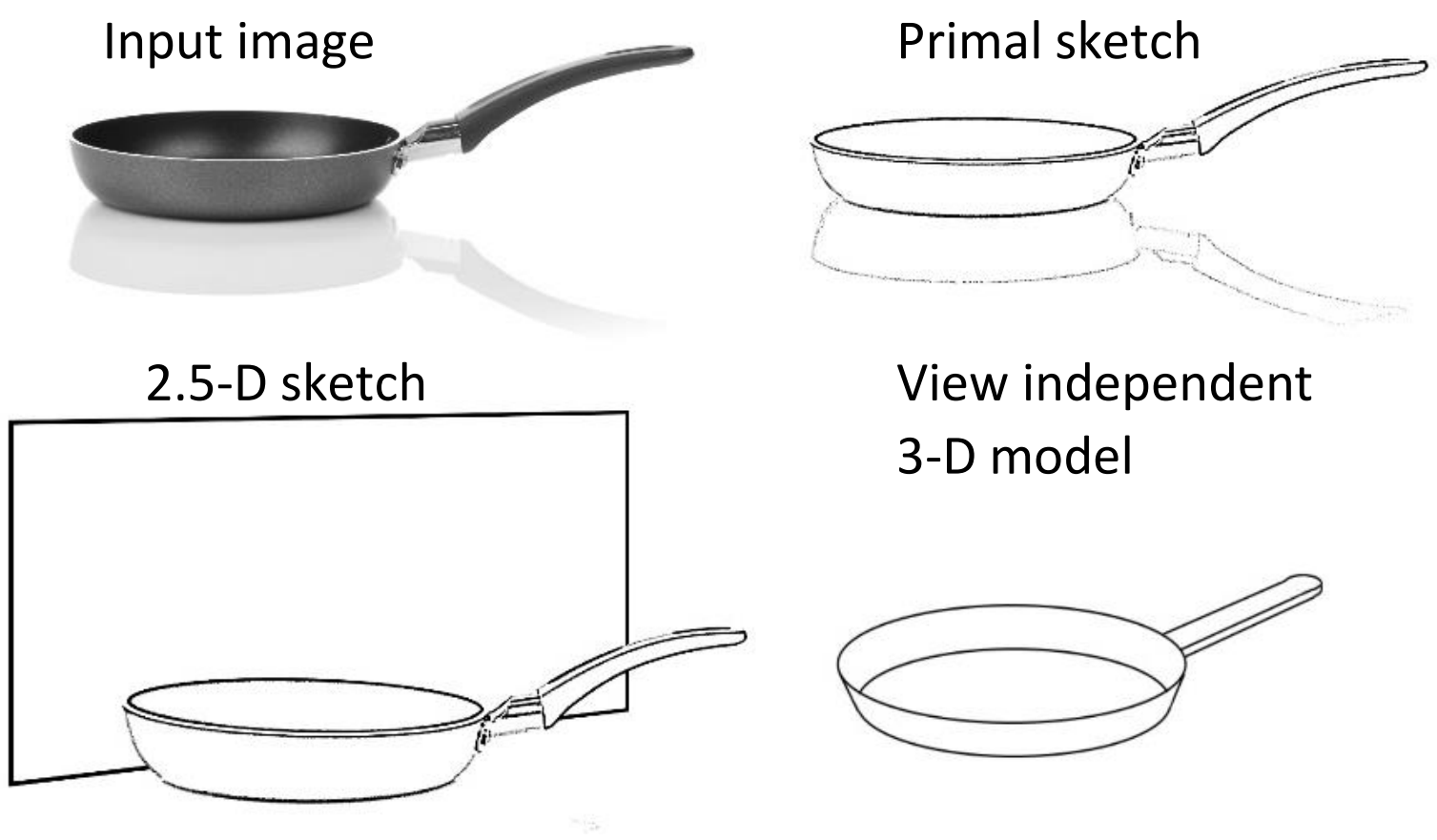

Figure 3.2. Progress of analysis from primal sketch to three-dimensional model. (© Tuomas Holmberg)

1. Creating a primal sketch consisting of a view with recognized edges, areas, corners, etc., reminiscent of the nature of the view produced quickly by the artist.
2. Proceed with producing a 2.5-D sketch, including depth cues provided by analysis of texture and shadows in the primal sketch.
3. Produce a three-dimensional, view-independent model of the view.

To accomplish this, Marr proposed a three-level solution consisting of

1. implementation level, such as the camera and the computer, that mimic the neural system. Current understanding suggests that artificial neural networks can play a role in this.
2. algorithm level that controls and modifies presentation, for example, based on e.g. reasoning rules and physical knowledge; and
3. computational level to select the tasks to be solved and guide the activities. This makes the system work for some purpose.

Decades later, with advances in machine learning based on neural networks triggering the latest excitement on artificial intelligence, renowned neuroscientist Tomaso Poggio (2010) proposed to add to Marr's framework

4. learning to get solutions without detailed programming.

This addition has not, at least so far, led to the capability to create three-dimensional models that are independent of the viewer's position. Representation issues are therefore still core challenges in creating intelligent systems. The current field of neural computation is probably only at the level of creating Marr's primal sketch, and it is difficult to estimate what the next truly significant step could be. The late introduction of machine learning into the canonical model may be a sign that other essential components may also be missing.

AI systems have regularly been built starting from zero knowledge. It has been hypothesized that based on human provided basic knowledge and preliminary algorithms the systems will at some point get ability to learn by themselves to use observations, questions, actuators - and information from the Internet. In these scenarios for more than half a century the researchers have believed that "the solution is around the corner".

One potentially fruitful research path is Bayesian program learning, which automatically learns to construct an algorithm that can produce input data even based on a single sample (Lake et al., 2015). The method has been demonstrated to learn handwritten characters, but could also work with speech signals. The key prerequisite is modeling of application knowledge and uncertainty. From the same basis, symbolic methods based on logic programming and grammars have failed to create artificial intelligence that learn from the examples. Employing neural computing and massive amounts of training data has been a faster route to practical applications.

It would be most natural if machine intelligence used the same representations and concepts as humans, that is, employed the same semantics. Communication with machines would then be in natural language and with the same meanings. Symbolic artificial intelligence has therefore an undeniable role when dealing with people in their environments. On the other hand, data-driven artificial intelligence suits to recognition tasks, acting as eyes and ears, and processing signals produced by the other senses. Together the symbolic and data-driven approaches allow for higher levels of functionality.

Data-driven connectionist machine intelligence has some ability to recognize objects on the dining table in the kitchen, such as spoons, forks, plates, beverage glasses, napkins, etc., contributing to the creation of a symbolic representation. Based on this information, a home robot based on symbolic knowledge and reasoning rules could generate an action sequence to first gather the cutlery from the plates, always proceeding to unobstructed objects to be picked-up and put to the dishwasher.

Combining of multiple sources of information and elementary actions into goal-oriented behavior is overwhelmingly challeng-ing without “understanding” their relationships and dependencies; pattern recognition alone is not enough. Thus, artificial intelligence is not (at least not yet) able to compete with humans in acting in everyday environments. No current artificial intelligence implementation can read the instructions in the dishwasher manual, and then act accordingly.

Almost all the everyday problems involve the need for good understanding of three-dimensionality and practical uncertainties, which are modeling challenges for both representations and methodology. People use invariant representational techniques that enable recognizing the same object from multiple viewpoints, even when partially occluded, although it has been seen only once. Unfortunately, we don’t know how we do it.

## 3.5 Representation of Information

The successes of data-driven connectionist approach have attracted developers and researchers to focus almost exclusively on the application of deep neural networks. This solution is well suited to situations where large amounts of data are available as categorized raw observations, such as images and speech signals. In other cases, however, it may be highly justified to consider the use of other representations and methods.

In light of current understanding, procedural program code is not an optimal way of representing knowledge already due to the manual maintenance effort. In addition, system developers, data and knowledge holders are typically different individuals who need to communicate with concepts that all parties understand.

### *3.5.1 Physical knowledge*

For centuries, mankind’s brightest individuals have built understanding of the laws of nature. Examples of the results include Newton’s basic mechanics, Einstein’s theory of relativity, Maxwell’s equations, etc. The experimental particle physics work continues, e.g., by researchers working at CERN.

Such scientific knowledge is independent of the mainstream technology of the time but has enabled its continuous renewal. It is probably unrealistic to assume that we have arrived at the end of scientific development with artificial intelligence and everything we have learned before can be rejected? The key to future better artificial intelligence can well be in the utilization of previous, proven knowledge with new automatic learning methods.

**Simulations** based on physical knowledge imitate reality and can provide data to a learning system. For example, simulations help collaborating robots to learn to avoid collisions without destructive experimentation, or large numbers of virtual photorealistic kitchens could be created to teach neural networks variants of everyday reality. Together, physical models and simulations have significantly speeded up the digital revolution (Figure 3.3).

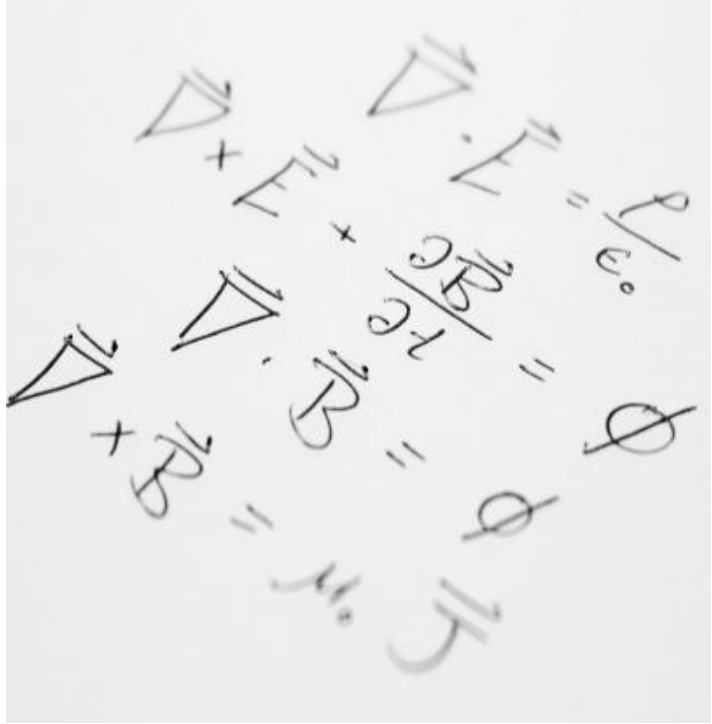

```
ARCHITECTURE Behavior OF Sensor IS
BEGIN
PROCESS(SensorReading)
  BEGIN
    CASE SensorReading IS
       WHEN "00" => s <= y;
       WHEN "11" => s <= y;
       WHEN OTHERS => s <= z;
  END CASE;
END PROCESS;
```

Kuva 3.3. Maxwell's equations and part of an integrated circuit's behavior model written in VHDL language. (© 123RF)

Behavioral models that accelerate the design of integrated circuits are mathematical descriptions of their properties. Because the models can be produced in arbitrary quantities, they can be used through simulations to learn whether a solution is feasible, and to predict the performance of the design. Such advanced functionalities that could be considered artificial intelligence have been extensively incorporated in design automation in all fields of technology.

### *3.5.2 State representation*

Symbolic information is often modeled as a state representation, where the solution to the problem is found through search. Roughly, the principle is to establish the states and possible state transitions or actions in the problem, and then find the path from the initial state to the desired end state by using the allowed actions. Each operation changes the state of the state space. Below we consider a small warehouse where pallet A needs to be moved to the loading dock (Figure 3.4).

The pallets A, B, and C can only be moved at right angles and at a time only to the next position. Each possible pallet location is labeled by numbers 1-5, and the initial state is ABC00, where 0 represents the empty space, A, B and C occupying positions 1, 2, and 3, respectively.

Initially, position transfers 2-> 4 or 3-> 4 are possible, resulting in respective new states A0CB0 or AB0C0. A human is able to see the solution to this simple transfer problem directly, but automated implementation requires either a ready-made search algorithm or logic programming.

The A* algorithm is probably the most widely known heuristic search algorithm. Powerful and under certain conditions optimal, A* is very popular in a variety of route search problems. The existence of this algorithm, which was originally developed for mobile robot guidance for environments containing obstacles, is often a motive for modeling problems into graph representations.

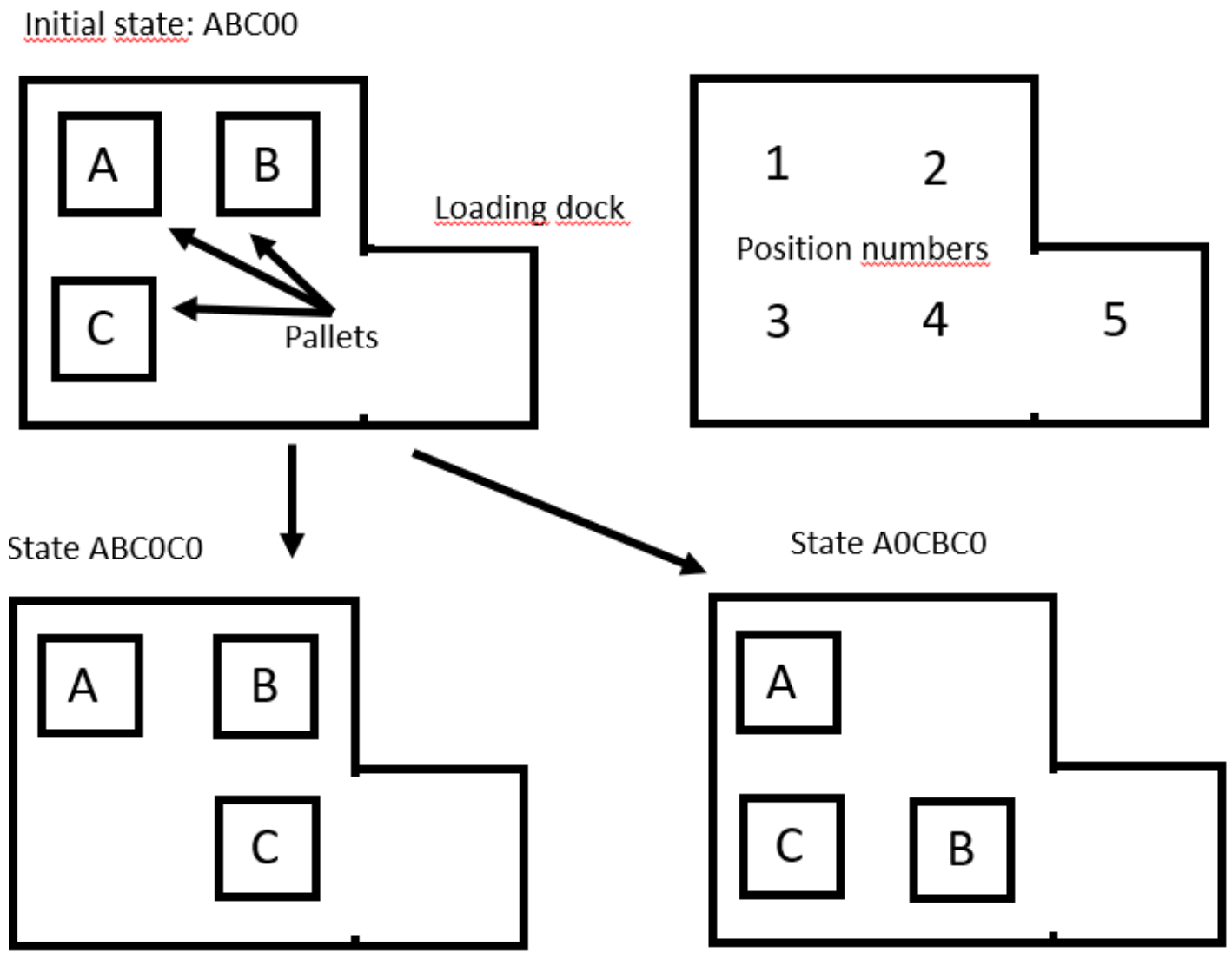

Figure 3.4. Example of modeling a pallet transfer task.

In the A* algorithm, each state A, B, etc., is described as a node as shown below. The number at each state transition is the effort of reaching the target node, e.g., the cost of moving to a new location. The idea behind the algorithm is following:

4. At each moment, we consider moving to every possible target node $x_i$
5. the heuristic function $f(x_i) = g(x_i) + h(x_i)$ evaluates the work to the target node
6. $h(x_i)$ is an estimate of the work from node $x_i$ to the target node and

- $g(x_i)$ known effort so far required to move to $x_i$
- next, move to the node through which $f(x_i)$ gives the shortest route

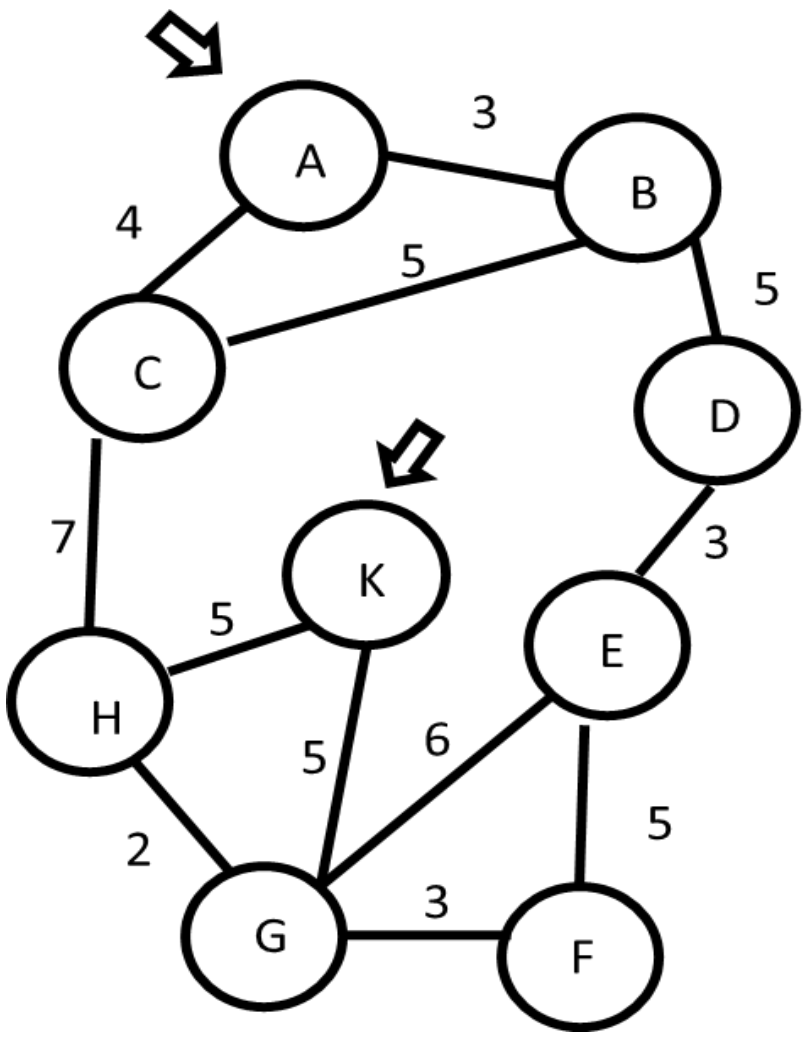

A* search is optimal if $h(x_i)$ never overestimates the effort to reach the target node, so the estimate must be optimistic.

In case of the graph above we search for the shortest route from node A to the destination node, in this case $x_i$ = K. We assume that as the search progresses, the distance to the destination always decreases by at least step 1. Thus, at the first four search levels our prediction for the distance to the target node as $h(x_i)$ = 4,3,2,1, respectively.

At node A: can expand to nodes C and B, $h(x_i) = 4$

- for node C, $f(x_i) = 4 + 4 = 8$
- for node B, $f(x_i) = 3 + 4 = 7$

→ select B, because of shorter distance prediction

At node B: can expand to nodes C and D, now h ($x_i$) = 3
• for node C, f ($x_i$) = (3 + 5) + 3 = 11
• for node D, f ($x_i$) = (3 + 5) + 3 = 11
→select D at random as there is a tie
At node D: can expand to node E, h ($x_i$) = 2
• for node E, f ($x_i$) = (3 + 5 + 3) + 2 = 13
• the forecast of the route through earlier node C is shorter, so
→select C next
At node C: can expand to node H, h ($x_i$) = 3 (notice search level!)
• for node H, f ($x_i$) = (4 + 7) + 3 = 14
but this is longer than the prediction through node E, so
→ select E next
At node E: can expand to nodes G and F, h ($x_i$) = 1
• for node G, f ($x_i$) = (3 + 5 + 3 + 6) + 1 = 18
• for node F, f ($x_i$) = (3 + 5 + 3 + 5) + 1 = 17
→ select H next because it provides route forecast of 14
At node H: can expand to nodes K and G
• for node G, f ($x_i$) = (4 + 7 + 2) + 2 = 15
• for node K, f ($x_i$) = (4 + 7 + 5) + 2 = 18
→ but now we find that target node K has been achieved!
• the distance from departure node is 16. As the nodes G and F have an optimistic route length prediction of 18, they can't be on a shorter route if the heuristic function is optimistic

### *3.5.3 Modeling games as state representations*

The goal of game participants is to beat their opponents. Most games can be modeled as game trees that are graphs in which the nodes are game situations (states) and the transitions between the states are player actions such as moves.

The presentation of game states varies, for example, from discrete chess board situations to statistical models. In case of simple games the entire game trees from the initial situation to the final solution could be modeled. Then, the player starting the game could be able win every time or ensuring at least a draw.

However, in many games, e.g., chess and poker, the state space is very large. As a result, simulations are of great importance in teaching artificial intelligence to play such games. An automated system can learn winning strategies with simulation producing situations at tremendous speed, enabling to go through large numbers of entire game sessions with the neural network serving as a modeler.

In two-player games both players try to maximize their own and minimize the opponent's advantage. This is called the minimax strategy, where in the max node of the game, player 1 tries to maximize his or her advantage and in the min node, player 2, in turn, tries to minimize the advantage of player 1.

Consider the minimax strategy using the game tree in Figure 3.5, where it is possible to reach a draw (D = 0), lose (L = -1), or win (W = +1). Each node is marked with the minimum or maximum available, depending on the level involved in the game tree.

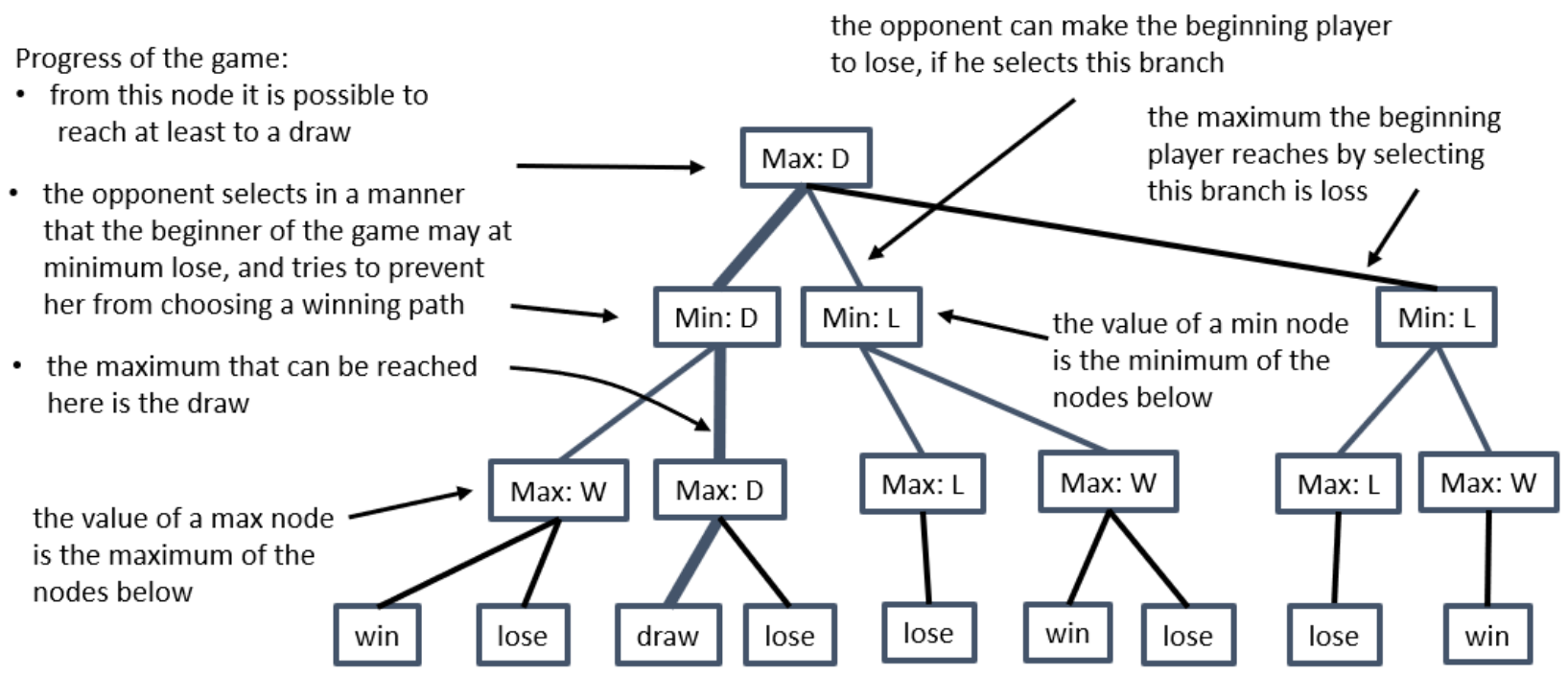

Figure 3.5. A minimax game tree.

The beginner (player 1) has modeled the entire game tree all the way down to the leaf nodes and traced the game from its last nodes backwards, maximizing at the max nodes and minimizing at the min nodes. Generally, the entire search tree cannot be produced due to computational complexity or game-related randomness, so it may be truncated, for example, by merely searching a few layers down from the current state.

The beginner of the game notices that by choosing the rightmost branch of the tree, he or she can at most lose, since at the min level the opponent can from the available two branches choose the alternative leading to the loss. The same applies to the center branch. From the left branch player 2 can steer the game to a path leading to a possible loss, but at the last max level the player 1 can select a draw.

In the minimax strategy, the values of the end nodes can be set in an arbitrary, though still a rational, way. In world politics we have witnessed a game in which the president of a big nation has sought to force others to play according to his unilaterally defined rules.

In the following game tree, the numerical values represent the DJT's "tis-for-tat" emotional reward values, while the minimizing player aims at financial stability. The starting party, DJT, is the maximizing player (Figure 3.6).

DJT states that there is no point in joining the TTIP transatlantic agreement presented in the right branch because the minimizing player (EU) will not later accept dictation agreements. Instead, the best outcome is to increase customs duties, as it can try to force individual EU countries into bilateral trade agreements.

Such real-world games, involving unpredictable or augmented information, are dynamically shaped by changing situations. On the winning path of this tree, too, DJT has forgotten that individual EU countries have no authority to negotiate trade agreements.

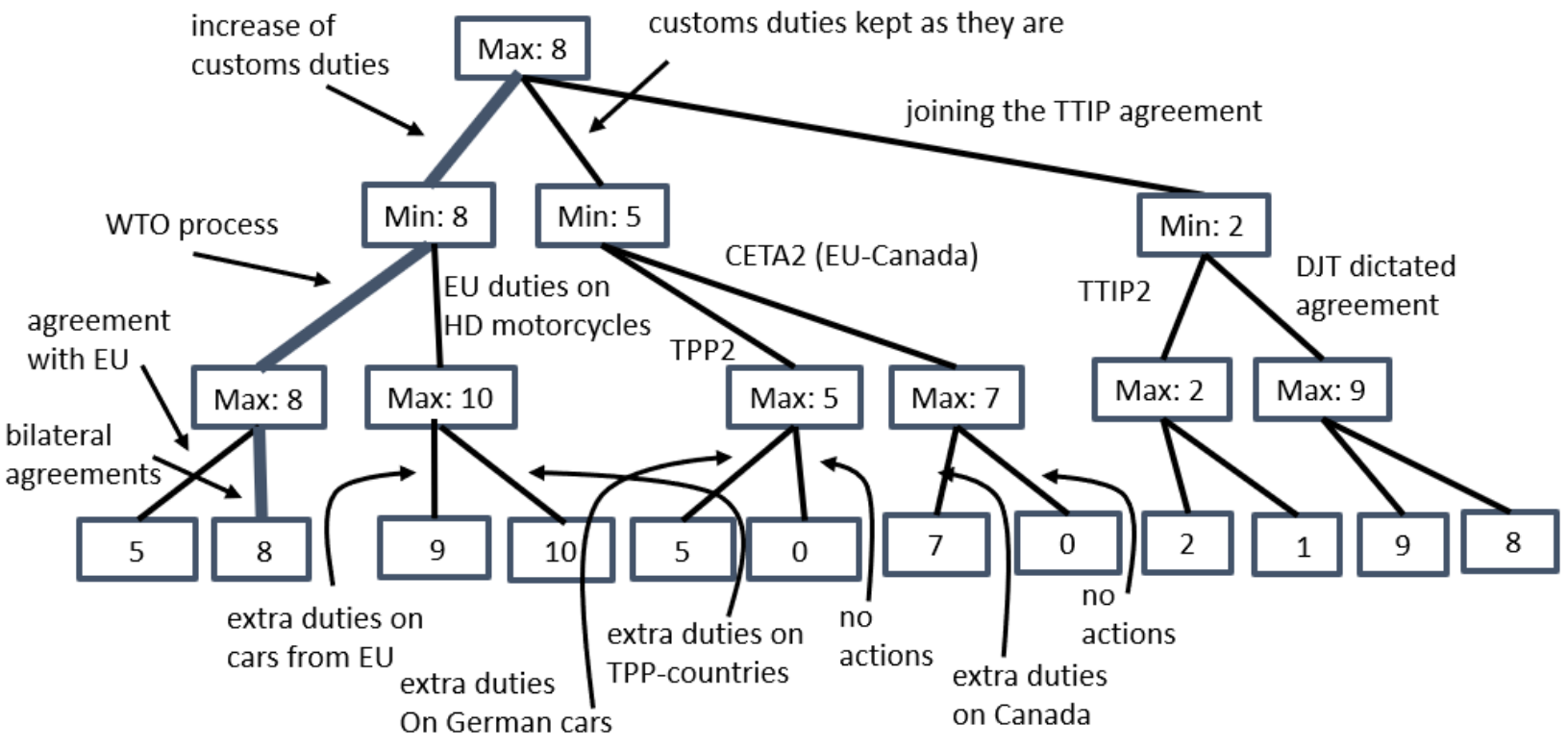

Figure 3.6. A global politics minimax example.

### *3.5.4 Feature representation*

Features are computationally extracted information from data, and are most versatile  representations and popular especially for interpreting data from sensors. For instance, speech features can be spectrograms, while features extracted from images might be be texture and color histograms.

Typical feature representation is a table in which each data sample is described by, for example, 10 features. In that case, the dimensionality of the data is 10.

In the past, a significant part of pattern recognition research concentrated on “feature engineering”, that is, designing features to be extracted, e.g.,  from financial data, medical images, or any other samples that are  in digital form. The objective has been to find as few features as possible to efficiently and accurately describe the target phenomena.

However, extraction of features from data requires strong assumptions. For instance, image preprocessing algorithms must often cope with the problem of segmenting numbers apart from manually filled official forms. This is often complicated by patterned backgrounds that are irrelevant for reading the content.

Consequently, lots of of research on image processing has been carried out, resulting in a large number of feature detectors with increasing discrimination capabilities. An example is the Local Binary Pattern (LBP) method presented in Chapter 7.

For example, information about corners, edges, contrasts, object circumference, number of holes, etc. can be extracted from the image. Sometimes syntactic or structural methods are needed. One possible such description of number 4 with  links connecting terminal, branch, and corner points is in Figure 3.7.

The attractiveness of structural representations is in the provided capability to (at least roughly) reconstruct the original target.

This facilitates the development of further recognition and analysis algorithms, since the presentation is easily understood by humans. As such, syntactic and structural methods are a step in the direction of symbolic artificial intelligence.

feature points: [[A, terminal, 6, 7], [B, corner, 6, 14],
[C, branch, 16, 14], [D, terminal, 14, 24],
[E, terminal, 18, 8]]
links: [[A,B], [B,C], [C,D], [C,E]]

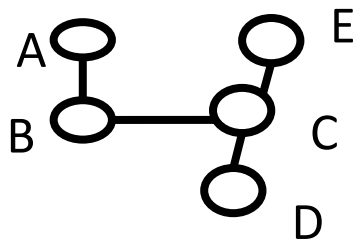

Figure 3.7. A structural description.

Neural computing methods and machine learning have dramatically changed the attitude toward features. Essentially, low-level features can be learned automatically, and no interpretations that a human can understand are needed. This, in turn, has been a step toward data-driven artificial intelligence.

For example, each pixel in an input image can be considered to be a feature. In the case of samples of handwritten numbers extracted from the MNIST database, the dimension of the vector is then 784 (see Figure 3.8) as the images consist of 28-by-28 pixels. The feature values ranges can be from 0 to 1, or from 0 to 255, if 1 or 8 bit pixels are used, respectively.

28x28 pixel 8-bit grayscale image
feature vector: [230 225 ... 15 17 ... 220 204 ...]
Alternatively
28x28 pixel 1-bit binary image;
feature vector: [1 1 1 ... 0 0 0 ... 1 1 ...]

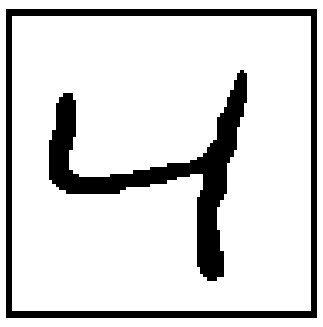

Figure 3.8. A feature vector representation of an image.

We may realize that as a single HDTV image has 2,073,600 pixels, each video has a dimension of over two million multiplied by the number of frames. Indeed, in related machine learning problems the dimensionality is enormous.

In the case of images, convolutions have been found to be effective feature extractors. The principle is to multiply the pixels in small neighborhoods in the image by corresponding elements in the convolution template and summing the results.

In images the closely-spaced pixels are highly correlated, which information is utilized by convolutional neural networks (CNN). The CNNs aim at learning automatically the convolution templates that most efficiently characterize inputs or differentiate between different categories of input data. In such cases, the purpose is often to identify objects, for example, to find people, cars and animals in images. The recognition systems are trained by pre-classified samples that represent the target categories.

In Figure 3.9 the convolution operation has resulted in feature value -10 at the upper left corner, somehow characterizing the local gradient in the input image.

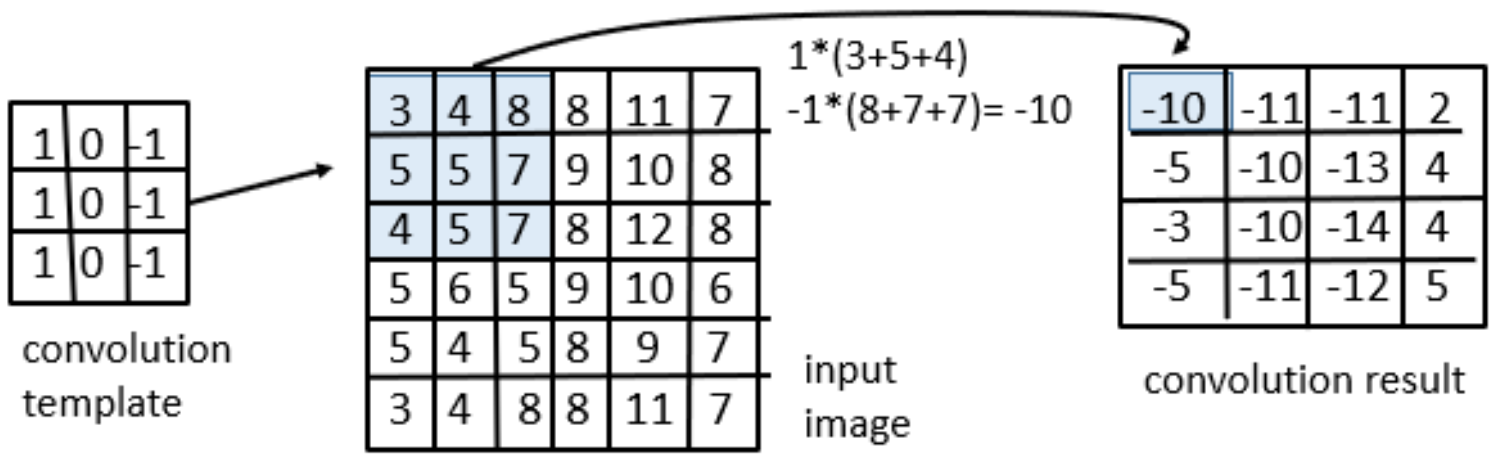

Kuva 3.9. Example of computing convolutions.

As the result of learning, it is generally observed that convolutions tend to produce feature information related specifically to edges, ridges, and corners. In multilevel neural networks the following levels combine these observations. For example, after being trained a CNN might provide specialized recognizers for handwritten numbers.

Much of the research in the field is conducted using standardized, very large test databases, such as ImageNet (Russakovsky et al., 2015) and MNIST (LeCun & Cortes, 2010). If the available data is more limited, the development of a new application may result in a significantly lower recognition accuracy, despite using state-of-the-art methodology.

### *3.5.5 Feature representation enabling reconstruction*

Auto-encoder is a neural network category capable of building a feature representation from which the inputs can be reconstructed (Rumelhart et al., 1986). For example, below in Figure 3.10 images are fed to the network with the aim of finding a representation that minimizes the reconstruction error.

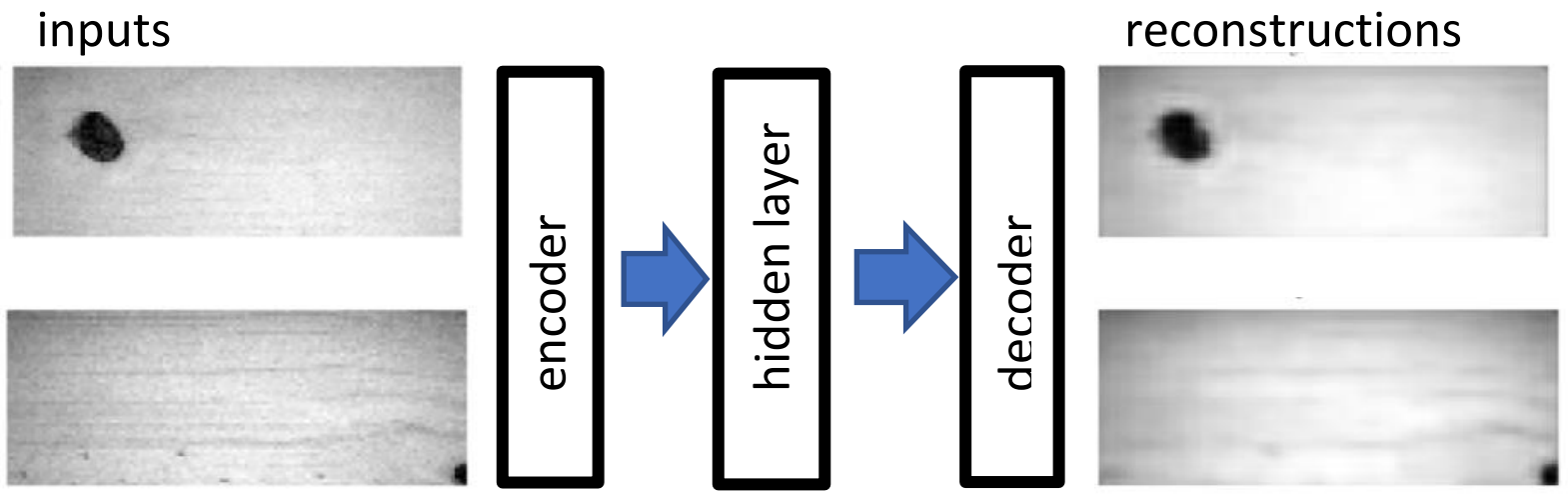

Figure 3.10. Principle of auto-encoder.

The auto-encoder consists of encoding and decoding blocks, with a compressed representation layer in between. That hidden layer can also be rightly called a "feature layer". The encoder seeks to learn a representation from which the decoder can produce a good approximation of the original input. Once the network has been trained, the decoder is removed, and the encoder is used to calculate the features.

Auto-encoders are often used as part of deep learning convolutional neural networks, but they also have applications as such. One of them is the synthesis of face images. By teaching an auto-encoder with a large number of faces, suitable inputs can be used to produce entirely new ones that represent the interpretation of the inputs.

Although auto-encoder is an unsupervised learning scheme, it does not achieve the capability to reconstruct everything. For instance, if images of hands are used as inputs, feeding a face image will result in a hand-like result. Only balanced training with both categories will produce the desired results. They are far from being able to automatically - and perhaps spontaneously - learn meanings from the environment.

Training of neural networks often requires their parameters to be modified and proper performance metrics to be used. This also applies to auto-encoders, and poses substantial challenges in building practical applications, e.g., for medical imaging in which unbalanced training samples are common, only a small minority representing cases to be recognized.

### *3.5.6 Dimensionality reduction and visualization of feature space*

In the processing of multidimensional data, support is often sought from visualizations that make the structure of the data and its possible imbalance easier to see by humans. A typical objective is to produce a two-dimensional view of data from which the human can make conclusions about its possible categorization (Venna et al., 2010).

In addition, the performance of many machine learning algorithms degrades with the increasing number of dimensions unless the amount of training samples is significantly increased. This phenomenon is known as the "curse of dimensionality," which is due to the sparsening of data samples in the feature space due to additional dimensions. As a result, the accuracy of the classification may suffer (Figure 3.11).

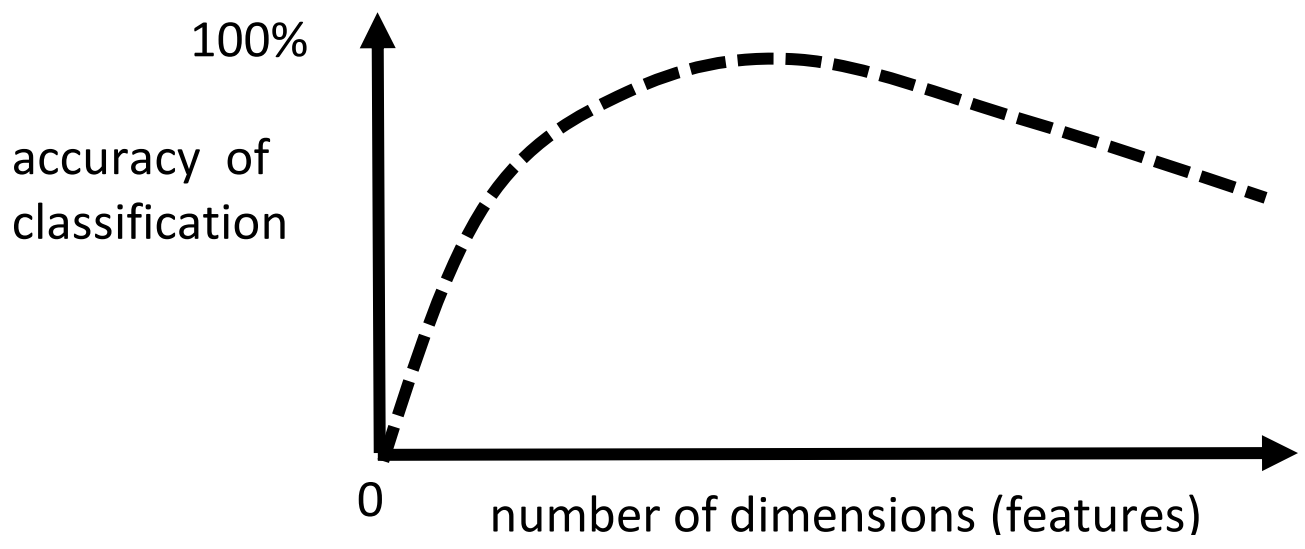

Kuva 3.11. Curse of dimensionality.

There is a growing number of dimensionality reduction methods to meet the human capability to judge data from two or at most three-dimensional representations. Typically, such mappings are intended to keep the relative distances or topology of the neighborhoods in large-scale data. An auto-encoder can also be considered as a method for reducing dimensionality, although it does not seek to maintain any single property.

The simplest dimensionality reduction scheme is identifying the features that best distinguish categories of data. For example, out of ten features, the two or three that lead to the smallest recognition error are selected. However, this procedure is only applicable to supervised learning situations, in which the sample categories are known in advance.

In the absence of category information, a suitable linear or non-linear projection or mapping algorithm may be used to compress the high-dimensional data into lower dimensions. This procedure can also be employed with supervised learning cases.

The manifold concept is associated with dimensionality reductions. For example, while standing in a parking lot, we observe that our immediate environment is flat, even though we actual stand on the round earth. In other words, we percept earth as two-dimensional, but we are still on the surface of a three-dimensional sphere. Thus, our easy-to-understand two-dimensional manifold is in three-dimensional space.

Mathematically, the dimension of a manifold is the number of independent parameters needed to determine a point on it:

- in one-dimensional case, manifolds are straight and curved, such as circles, ellipses, hyperbolas, that is, any curves. Also, space curves whose parametric equations have the form [x, y, z] = [f (t), g (t), h (t)] are one-dimensional, since t is the only independent parameter.
- two-dimensional manifolds are surfaces such as planes, cylinders, ellipsoids, toroids. These include only surfaces, not their possible insides.
- three-dimensional manifold, which locally appear as three-dimensional Euclidean space, is significantly more difficult for a human to perceive, let alone a four-dimensional manifold, which is a conceptual framework of relativity.

We can concretize this by drawing on the paper a two-dimensional xy-coordinate system, marking arbitrary data points in it, and folding the sheet into a paper ship. As a result, the data points are now in three dimensions. By dismantling the ship back into a flat sheet, we reduce dimensionality.

For dimensionality reduction both linear and non-linear methods exist. The idea of the linear schemes is to find such projections of data that expose the variations of the data from largest to smallest.

An example of non-linear dimensionality reduction is in Figure 3.12 that demonstrates the Isometric Mapping (IsoMap) method (Tenenbaum et al., 2000). The colored samples help to see the mapping of 3-D data representation on the left onto a 2-D manifold on the right.

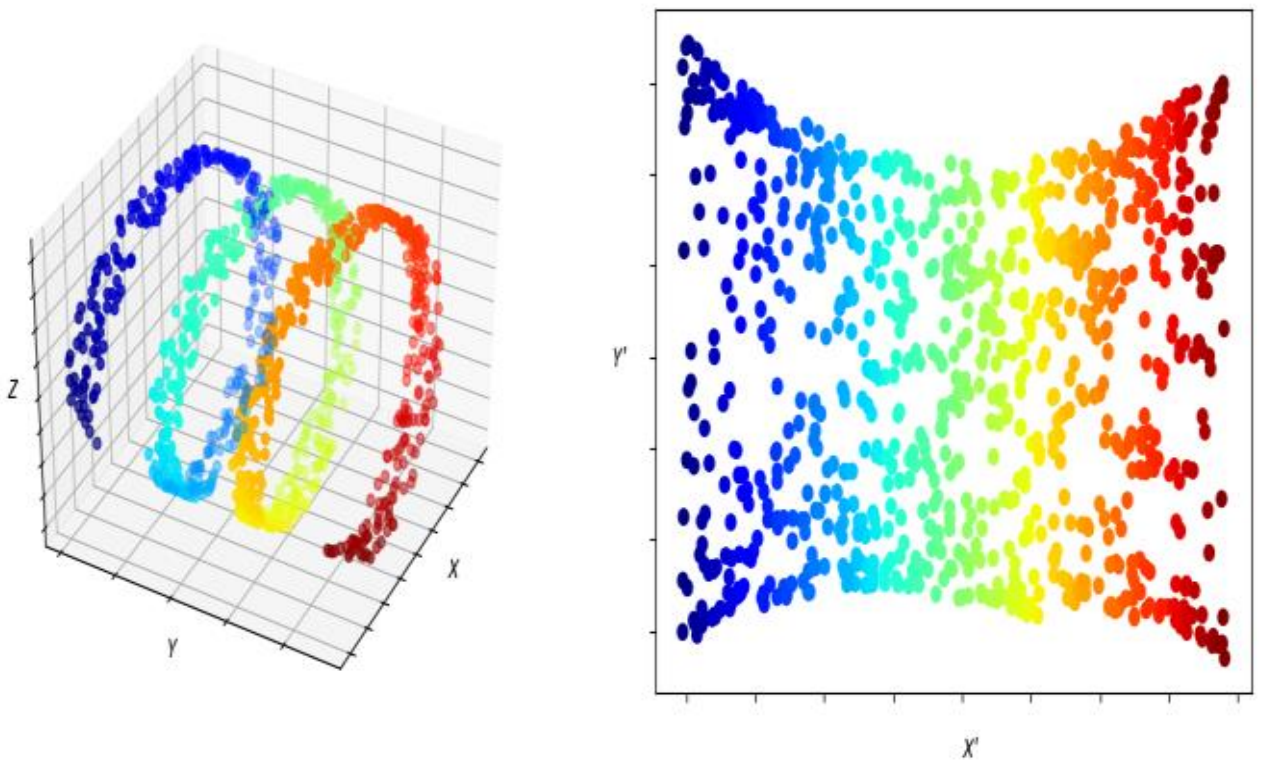

Kuva 3.12. Three-dimensional “coil spring data” and its two-dimensional description. (© Tuomas Holmberg)

Principal Components Analysis (PCA), also known by names Hotelling or Karhunen-Loève Transform (Wiki-PCA) is the most popular linear dimensionality reduction method. It is usually used as the first experiment for unknown data before moving to non-linear dimensionality reduction.

The principle of PCA is to determine such orthogonal base vectors for multidimensional data that the data can be reconstructed without major information loss using just the most significant ones of the vectors. The method has often been used for reducing the number of features as classification becomes cheaper, and the results can improve, despite the loss of information, as the data becomes denser in the feature space!

Figure 3.13 on the left shows a two-dimensional feature space in which the data samples are located as elongated structure. After determining directions for optimal viewing of the maximum and the minor variations, those can be set as new coordinate axes of the feature space.

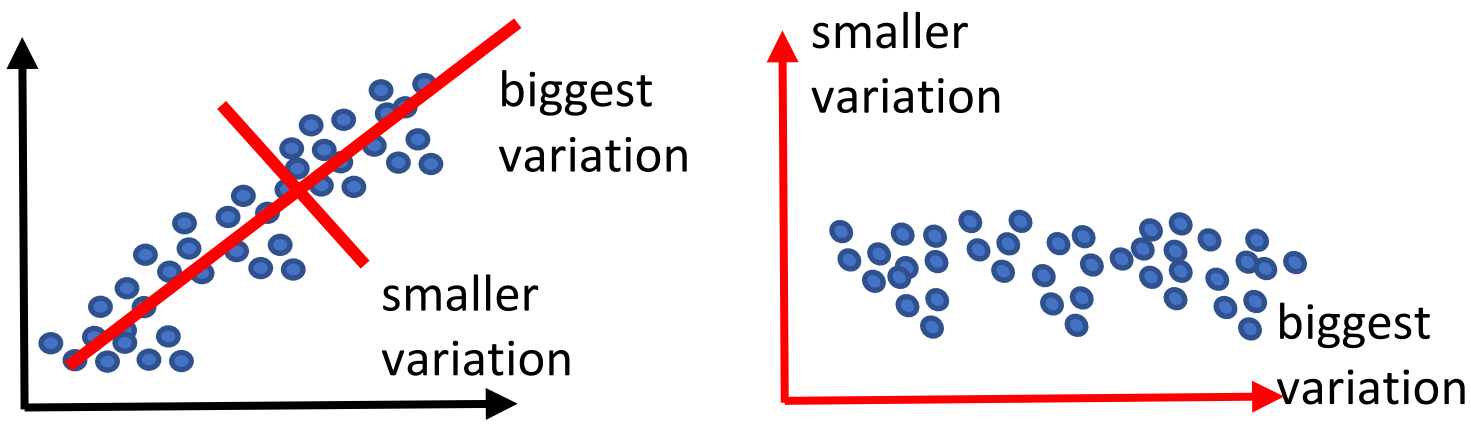

Kuva 3.13. Principle of principal component analysis (PCA).

In the above case, the number of dimensions didn’t drop. However, it is easy to see that the clusters in the data could probably be discriminated by just selecting suitable value ranges from the biggest variation axis.

Successful use of PCA requires linear correlations in the sample data. As a simple method, it is popular and often gives convincing visualizations. Independent Component Analysis (ICA) is another method that can uncover hidden data dependencies. However, its application to reducing dimensions is not easy.

In many cases, machine learning applications need to deal with very high dimensional data. For example, the handwritten number samples contained in the MNIST database are small 28x28 pixel images, but the dimensionality is already 784. PCA is seldomly powerful with this kind of problems. Reducing the dimensionality by using higher-level features instead of pixels can be more effective, but from the human point of view too many of them may be needed.

Nonlinear dimensionality reduction techniques aim to produce visualizations that are easy to understand for humans. These are just one-way methods: unlike with PCA, the original data can't be reconstructed from the results. On the other hand, they seek to bring the neighborhoods and densities of the high-dimensional representation for human viewing based on of similarity measures.

Figure 3.14 shows 2-D and 3-D visualizations of MNIST handwritten number image data produced using t-SNE method (van der Maaten & Hinton, 2008). The points corresponding to the numerical values have been colored after the dimensionality drop. It is interesting to note that 3-D visualization in this case does not appear to bring additional information for a human analyst. Two dimensions suffice, and the clusters could be separable even without coloring. However, now we may notice that some data samples fall into incorrect apparent clusters.

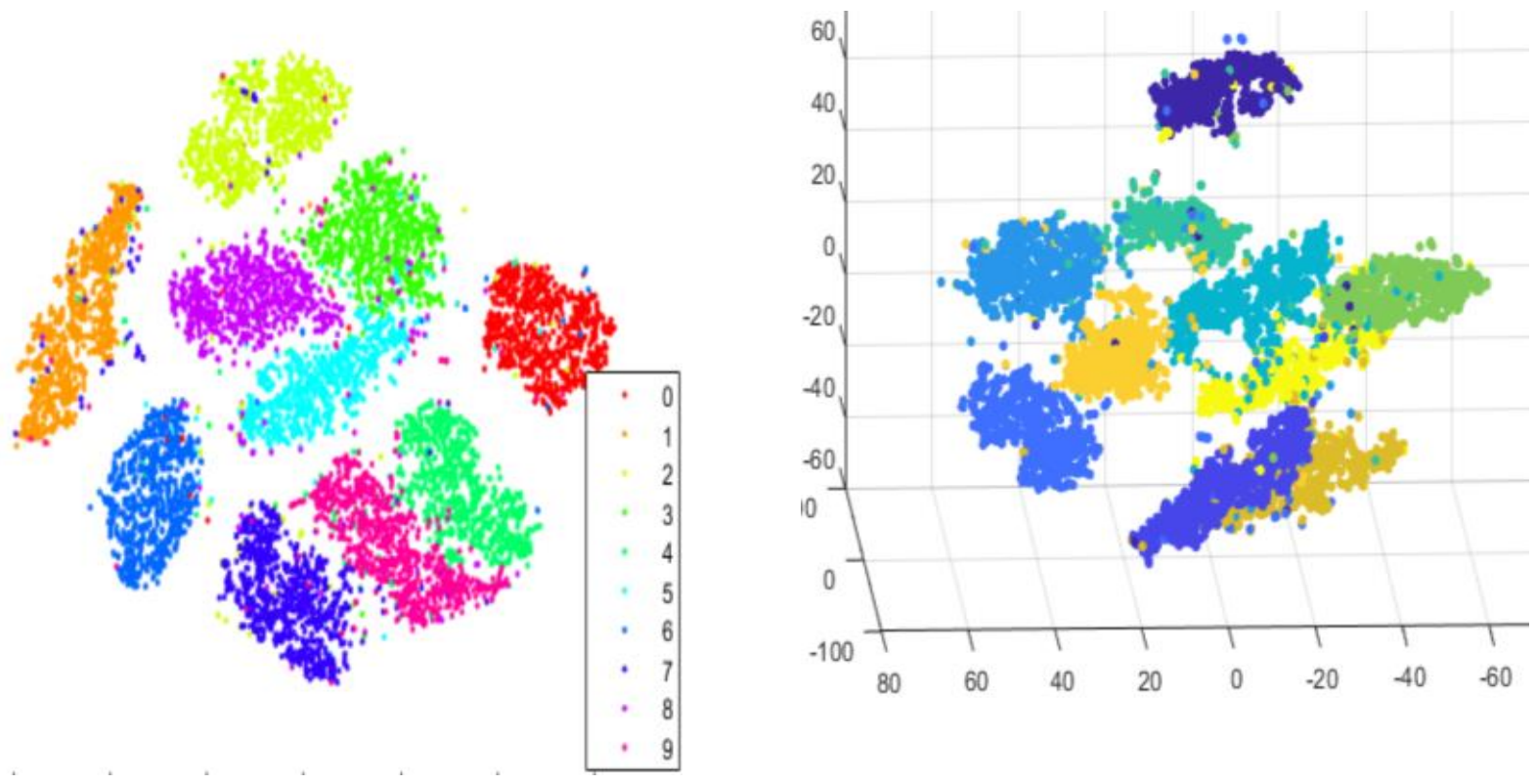

Figure 3.14. 2-D and 3-D descriptions of MNIST data using the t-SNE method. (© Tuomas Holmberg)

Multi-Dimensional Scaling (MDS) is one of the oldest methods for dimensionality reduction (Torgerson, 1958). Its input data are similarities between all possible pairs of sample data calculated, e.g., as Euclidean distances from the features. The result is a similarity matrix, each row representing the distance of a sample to every other one in the sample set. Usually a two-dimensional projection is then computed, maintaining the distances between the sample pairs as close to the original ones as possible.

Figure 3.15 shows defects of lumber mapped into a two-dimensional representation using the MDS method (Niskanen, 2003).

In this case, we notice that it is difficult to see distinct structure. This is explained by the principle of MDS to compute the distance from each data sample to all others. Depending on the intended use, this is either a desired or undesirable feature.

At its simplest, MDS is produced by performing PCA to the similarity matrix. In practice, PCA can be considered as one variant of MDS. The difference is that PCA is a method of reducing dimensionality, while MDS is a mapping technique.

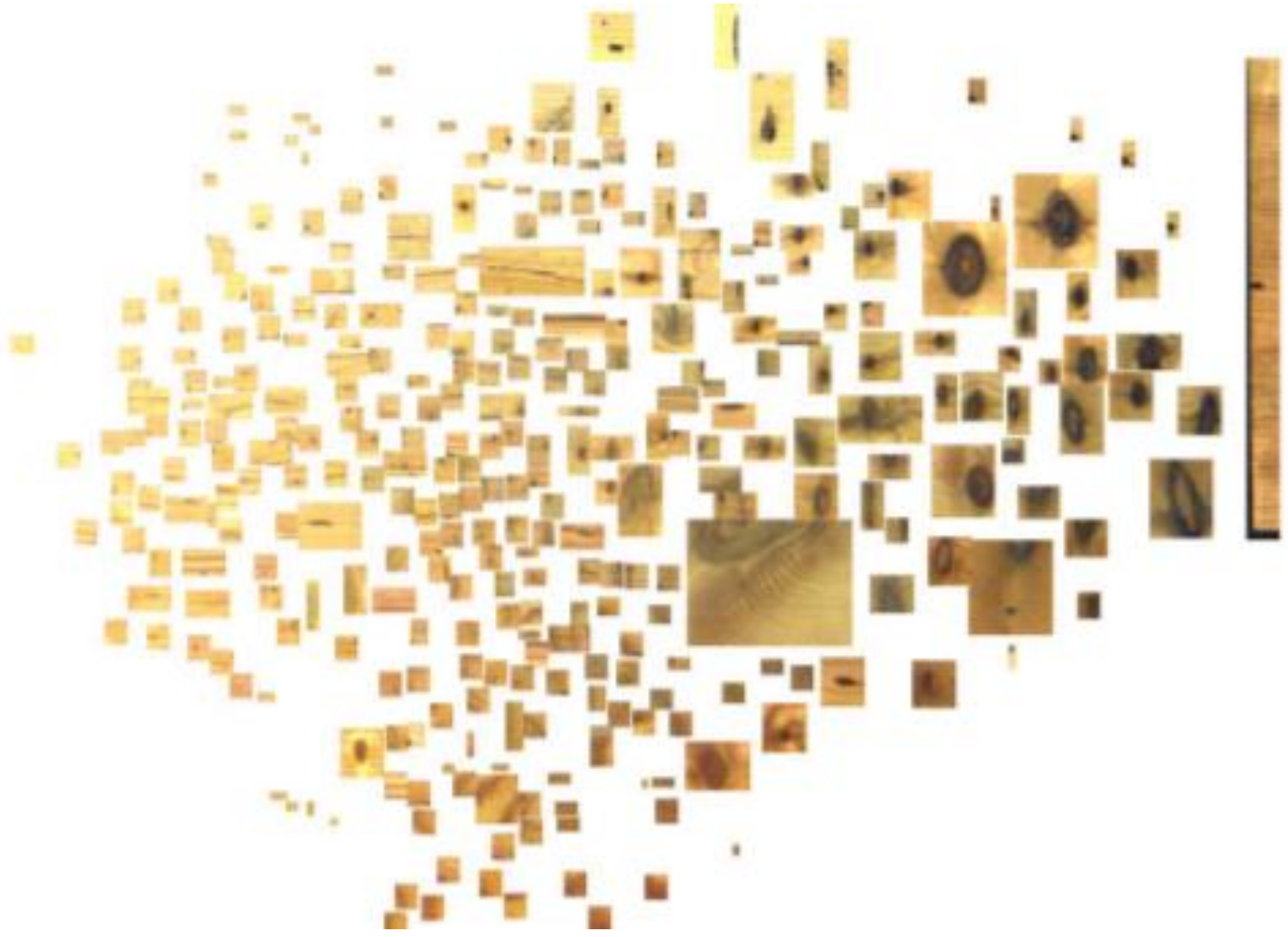

Kuva 3.15. Multidimensional scaling. (© Acta Univ. Oul.)

Niskanen M (2003) A Visual Training Based Approach to Surface Inspection. Acta Univ. Oul. C 186, 125 s. http://jultika.oulu.fi/files/isbn9514270673.pdf

It is often more useful to uncover structure of data by using methods that consider neighborhoods. Among the best known of these are Isomap (Isometric mapping) (Tenenbaum et al., 2000) and LLE (Roweis & Saul, 2000). Recently, t-SNE (t-distributed Stochastic Neighbor Embedding), based on neighborhood density functions and information theoretic minimization of target function (van der Maaten & Hinton, 2008), and UMAP (Uniform Manifold Approximation and Projection) that views manifolds as fuzzy topological structures (McInnes et al. 2018), have gained popularity.

Many problems with high dimensionality do not have a clear category structure. In such a case, even a human can't completely classify samples. For example, in many quality control applications, changes from flawless material to defects or another quality class are continuous. Similarly, for example, changes in a person's facial posture and emotional states may be continuous-valued.

Non-linear dimensionality reduction methods provide an opportunity to analyze even in such cases. In Figure 3.16 LLE method has been used for wood samples (Kayo, 2006) representing a continuum of appearances.

The self-organizing map (SOM) can also be considered as a method for dimensionality reduction (Kohonen, 2001). The principle is that vectors in a two-dimensional map learn the characteristics of input data. A practical advantage of SOM is the opportunity to produce rectangular visualizations that are convenient for various user interfaces.

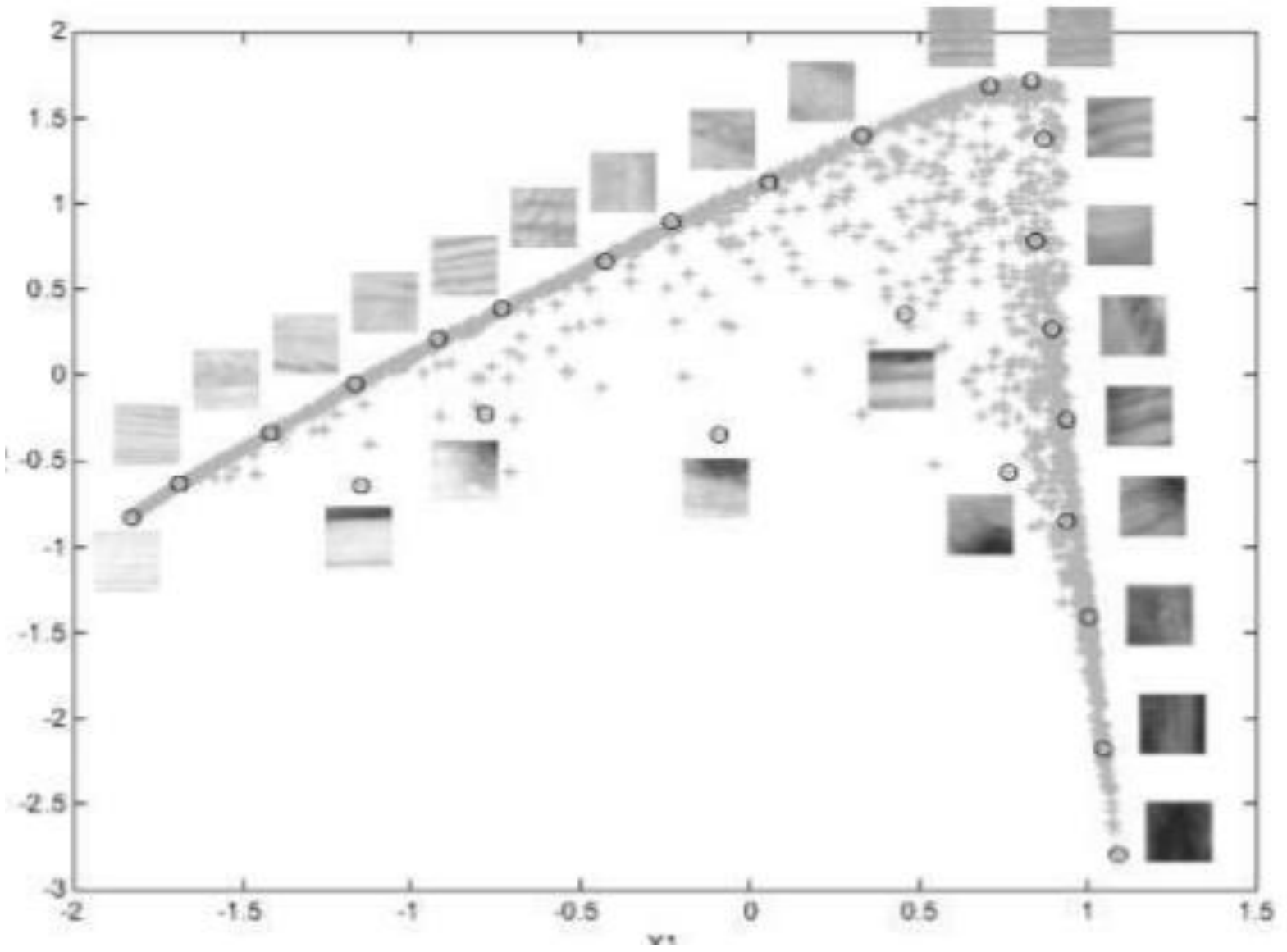

Figure 3.16. Continuity of wood samples visualized by the LLE method. (© Acta Univ. Oul.)

Kayo O (2006) Locally Linear Embedding Algorithm. Acta Univ. Oul. C 237, 120 s. http://jultika.oulu.fi/files/isbn9514280415.pdf

Figure 3.17 depicts the material variation of wood in a SOM. This kind of visualization enables a human to define and modify boundaries for categories in a graphical manner.

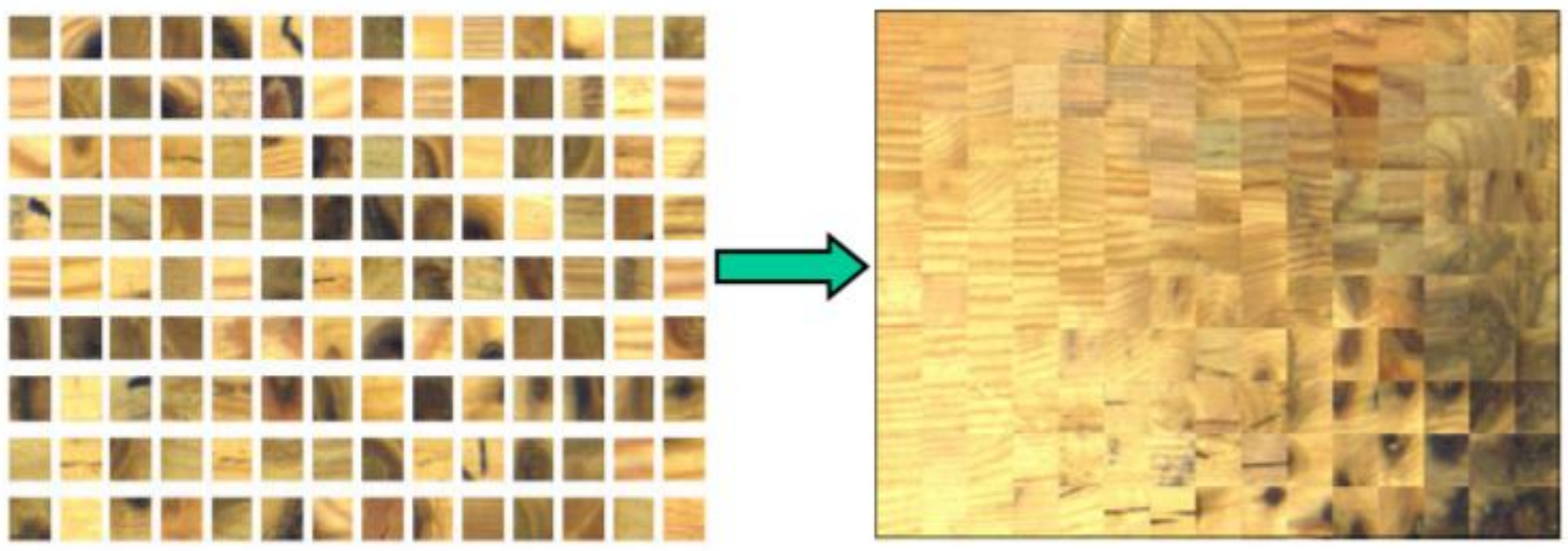

Figure 3.17. Visualization of wood material variation using a self-organizing map. (© Acta Univ. Oul.)

Niskanen M (2003) A Visual Training Based Approach to Surface Inspection. Acta Univ. Oul. C 186, 125 s. http://jultika.oulu.fi/files/isbn9514270673.pdf

An all-round dimensionality reduction method that fits every purpose has not yet been developed. The current methods have their own drawbacks, but also strengths that can be utilized when the nature of data is understood. For example, self-organizing map models the multidimensional probability density. As such, the method leaves little space for the rare categories when trained with unbalanced material. However, it can be very useful in exposing the variation in the sample material. This challenge is demonstrated in Figure 3.18 (Niskanen, 2003).

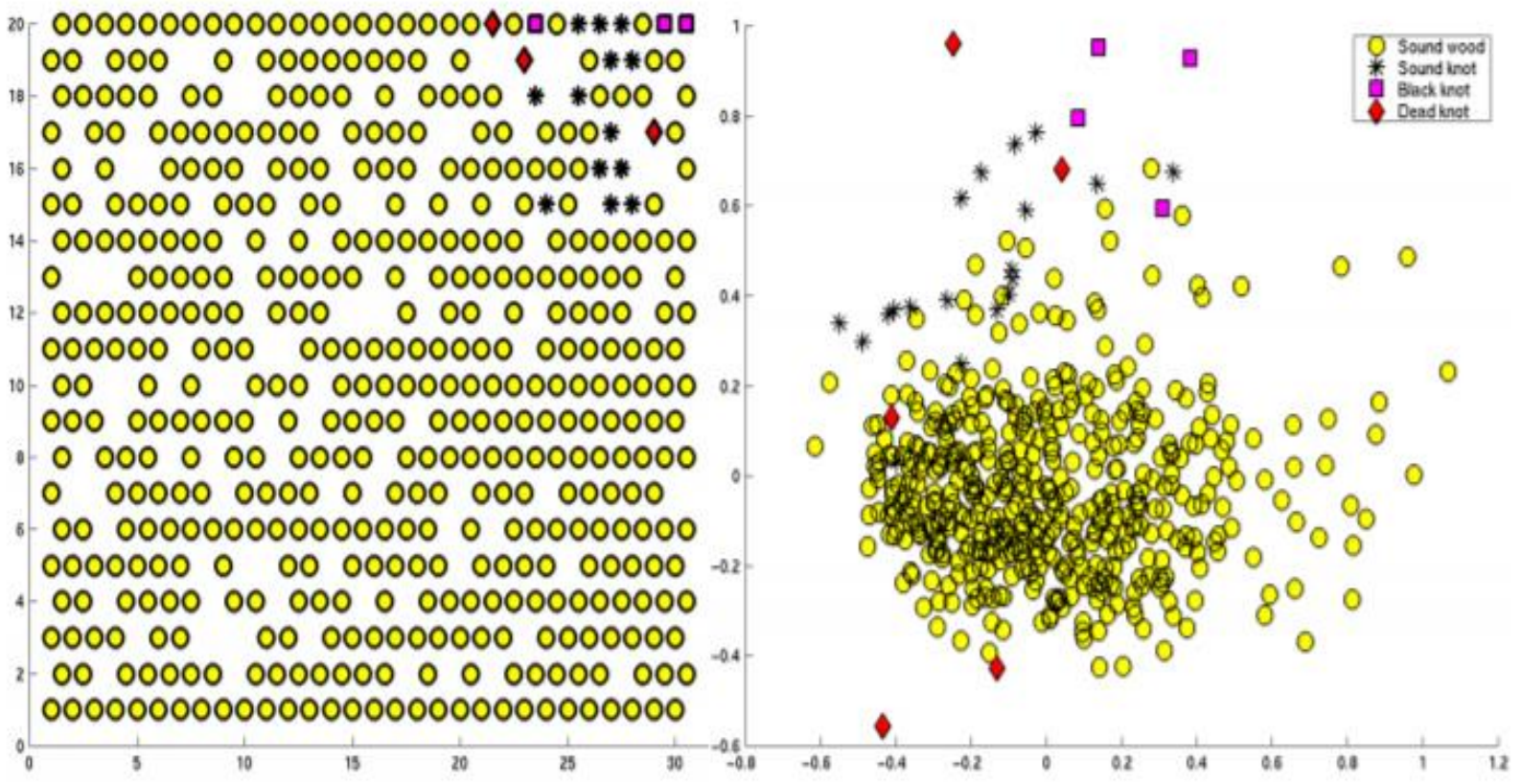

Figure 3.18. SOM and Isomap visualizations of unbalanced data. (© Acta Univ. Oul.)

Niskanen M (2003) A Visual Training Based Approach to Surface Inspection. Acta Univ. Oul. C 186, 125 s. http://jultika.oulu.fi/files/isbn9514270673.pdf

On the left, the SOM has learned the probability density of wood samples, allocating very few nodes for rare "classes". Only a few nodes are left for the knots and modeling their variation, although they are of interest. On the right, Isomap shows knots as distinct outliers. Obviously, it is better suited to uncover data imbalances, and can be used for data exploration to produce more balanced training sample sets.

## 3.6 What Matters: Presentation, Data, or Algorithms?

The focus of artificial intelligence application designers and researchers has changed. From an application point of view, it is secondary how a problem is solved. The total effort required to achieve the solution is much more important for the developers.

Feature engineering as human effort was laborious and resulted in extensive manual algorithm tuning. Now, with neural networks, the focus is on acquiring data and finding the parameters that give the best performance. However, the algorithms may still take a crucial role if data is limited, it is expensive to acquire, or other knowledge of the problem area is available, such as physical models.

## 3.7 References

Kayo O (2006) Locally Linear Embedding Algorithm. Acta Univ. Oul. C 237, 120 p.

Kohonen T (2001) Self-Organizing Maps, Third Edition, Springer.

Lake BM, Salakhutdinov R & Tenenbaum JB (2015) Human-level concept learning through probabilistic program induction. Science 350(6266):1332-1338 1332.

LeCun Y & Cortes C (2010) MNIST handwritten digit database.

Marr D (1982) Vision: A Computational Investigation into the Human Representation and Processing of Visual Information. San Francisco: WH Freeman and Company.

McInnes L, Healy J & Melville J (2018) UMAP: Uniform manifold approximation and projection for dimension reduction. arXiv: 1802.03426v2, 51 p.

Niskanen M (2003) A Visual Training Based Approach to Surface Inspection. Acta Univ. Oul. C 186, 125 p.

Poggio T (2010) Afterword to Marr's Vision and Computational Neuroscience. The MIT Press.

Roweis ST & Saul LK (2000) Nonlinear dimensionality reduction by locally linear embedding. Science 290(5500):2323-2326.

Rumelhart D, Hinton G & Williams R (1986) Learning representations by back-propagating errors. Nature 323:533-536.

Russakovsky O, Deng J, Su K, Karause J, Sathees S, Ma S, Huang Z, Karpathy A, Khoslta A, Bernstein M, Berg AC & Li F-F (2015) ImageNet large scale visual recognition challenge. International Journal of Computer Vision 115(3):211-252).

Russell S & Norvig P (2010) Artificial Intelligence: A Modern Approach, 3rd Edition. Pearson, 1152 p.

Simon HA (1969) The Sciences of the Artificial. The MIT Press, 1969.

Tenenbaum JB, de Silva V & Langford JC (2000) A global geometric framework for nonlinear dimensionality reduction. Science 290(5500):2319-2323.

Torgerson, WS (1958) Theory & Methods of Scaling. New York: Wiley.

van der Maaten, LJP & Hinton GE. (2008) Visualizing data using t-SNE. Journal of Machine Learning Research. 9:2579-2605.

Venna J, Peltonen J, Nybo K, Aidos H & Kaski S (2010) Information retrieval perspective to nonlinear dimensionality reduction for data visualization. Journal of Machine Learning Research 11:451-490.

Wiki-Cyc: Cyc

Wiki-PCA: PCA

Wiki-Prolog: Prolog

Wiki-Searle: Chinese room

# 4 Machine Learning

## 4.1 Introduction to Data-driven Machine Learning

The most significant recent advances in artificial intelligence have been achieved through machine learning (Goodfellow et al., 2015), (Duda et al., 2001), (Bishop, 2006). A popular strategy is to feed large amounts of raw data into neural networks in a "brute force" style. In this scheme, the role of algorithms as the basis of "intelligence" shrinks, as shown in Figure 4.1, and developers are more interested in modeling the data and related solutions. Based on the modeled data, it is possible to conclude its most probable categories or to make predictions. For example, the contents of input images can be categorized as cats or dogs, while the control parameters of the autonomous car are adjusted based on the location of the roadside.

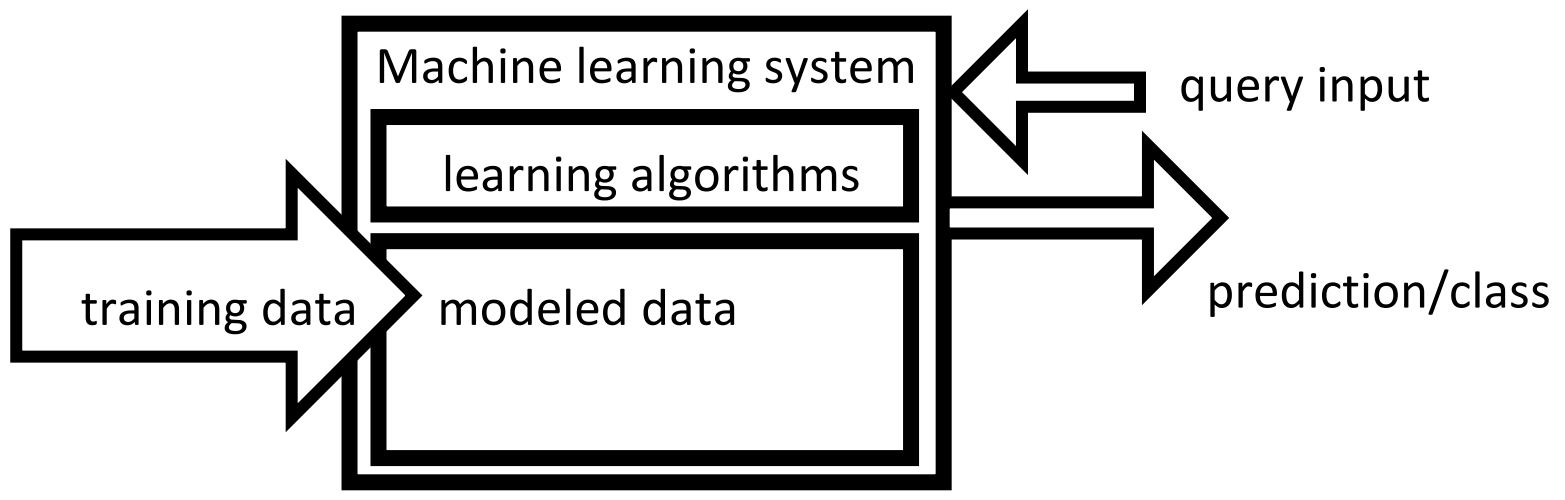

Figure 4.1. Machine learning focuses on modeling the data.

The success of the data-driven approach has undoubtedly undermined faith in presentations and algorithms and may have surprised many pioneers in the field. Especially, success has been achieved with problems where finding purely algorithmic solutions has been difficult, such as speech and face recognition. In many cases, machine learning is close to statistical modeling, data mining, optimization or retrieval. The manual development of features and models by humans can be characterized as search or optimization that can now be automated by machine learning.

The popularity of data-driven machine learning stems from the flood of multidimensional data following digitalization and the widespread emergence of problems that are difficult for people to understand. One can somehow visualize problems of two or at most three variables. For example, in information modeled for symbolic artificial intelligence, this is manifested as laborious determinations of simple cause-effect and spatial relationships. Data-driven machine learning and its easy-to-use tools available have brought many previously difficult problems solvable by almost anyone.

Car lane guards, speed limit sign recognition, and Internet image search engines are evidence of the power of this solution. They all have the same methodological core behind them, but are trained with different image materials. The purpose of machine learning is to avoid writing algorithms or rules for each application individually. However, the architectures of neural system implementations are often different (e.g., the number of layers, connections of neurons, activation functions), depending on the requirements of the given application problem. As shown in Figure 4.2, machine learning algorithms can be roughly divided into three methodology classes, each of which will depend on the nature of the data and the problem.

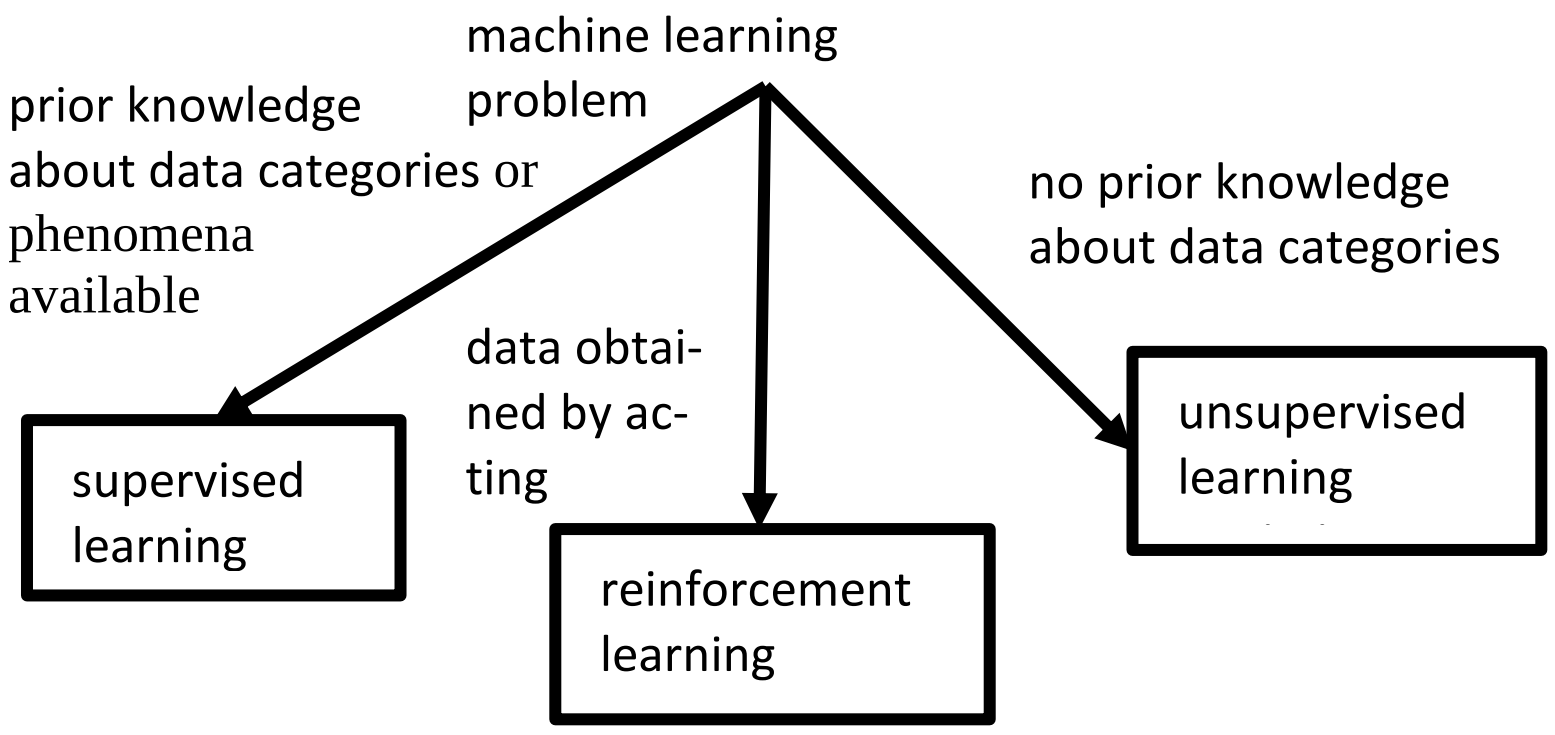

Figure 4.2. Categorization of machine learning methods.

In supervised learning, learning takes place with inputs of known categories (Duda et al., 2001). After learning, new feeds should be able to be categorized into those categories. So training is done by examples. For example, a text translation solution can be taught using different language versions of the same materials, that is, through human supplied examples. In the same style, one can learn how to correct spelling and grammatical errors in human typing. Learning can continuously improve performance by entering new data.

The problem with supervised learning can be misclassified training samples that drop accuracy. This problem is not present in **unsupervised learning**, where categories are not known in advance, and may not even exist in the human-recognized sense (Duda et al., 2001). The principle is to construct, based on the inputs, a description of the data structures where the similar inputs are located close to each other. Dimensionality reduction methods and auto-encoders discussed in Chapter 3 are methods of unsupervised learning.

In **reinforcement learning**, the learning machine explores the operating environment, where each action receives positive or negative feedback, for example, through situational analysis based on new information generated by sensors (Sutton & Barto, 1998). The goal is to find the solution with the most positive feedback. It is also possible to implement forgetting in reinforcement learning, which is often difficult in other learning methods. Continuous reinforcement learning can mean continuing after the first solution is found, seeking new, possibly better, solutions with ever-increasing feedback.

In addition to the above, there is **semi-supervised learning** (Chapelle et al., 2006), where training takes place through a combination of unsupervised and supervised learning. In this case, the methods of reducing the dimensionality can give an idea of the structure of the data and identify possible categories. Thereafter, machine learning can ensure human categorization of interpreted data samples that fall close to the expected class boundaries.

Semi-supervised learning is supported by the challenge of producing a large amount of pre-classified training material. In addition, when made by man, it is also prone to error. In addition to occasional mistakes, man is prone to systematic errors. Correcting the latter is later difficult or even impossible without laborious re-examination of the entire material. Hence, semi-supervised learning aims to reduce human work.

Typical useful data is always somehow structured. The task of machine learning is to model the data to find structures and finally act upon them when identifying objects, producing predictions or measures. Often, the strategy is to seek simpler structures first and then more complex ones. As a result, the boundary between supervised and unsupervised learning is a line drawn in water.

## 4.2 Processing of Data

The development of machine learning applications is based on data and must be handled with careful scientific work practices. Namely, errors made with data sets can easily lead to solutions that work with the original data, reaching, for example, 95% accuracy, or even higher. In practice, however, accuracy may drop below 80%.

In the case of supervised learning, data samples are typically divided into three parts (Duda et al., 2001):

**Training samples** are material that is used to train the selected regression model or classifier.

**Validation samples** are data that measure the prediction error of a trained solution. This indicates the suitability of the regression model and the classifier, as well as the quality of the training, which depends on, e.g., the size of the training material. Tests performed with validation samples can lead to changes in regression models and classifiers. At the same time parameters are optimized.

The **test samples** are completely segregated data that is eventually added to the game when the expected optimum has been reached during the training and validation steps. The test set is intended to represent use in the actual application. However, there is a risk in the pre-selected materials that the actual application may deviate from the conception given by it.

The sharing of available data depends on the application. If the regression model is already known, it is often done in 50:25:25 or 25:50:25 ratios. If the model to be used is already known (e.g. linear regression has already been selected), the data can be divided into randomly repeating training and test sets, for example, in the ratios 50:50, 70:30, 90:10, or even 99:1.

In practice, sample data is always multidimensional and the value ranges of its various features may differ. In turn, learning the structures of data often requires, depending on the methods, distance computing in the feature space. As a result, the features may need to be normalized so that the general distance norms would function appropriately.

For example, in the case of the data samples in Figure 4.3, when using Euclidean or block distance, the feature $x_1$ dominates the larger value range. This may not be appropriate. It may be reasonable to assume that the smaller value range is more significant in relation to the larger differences.

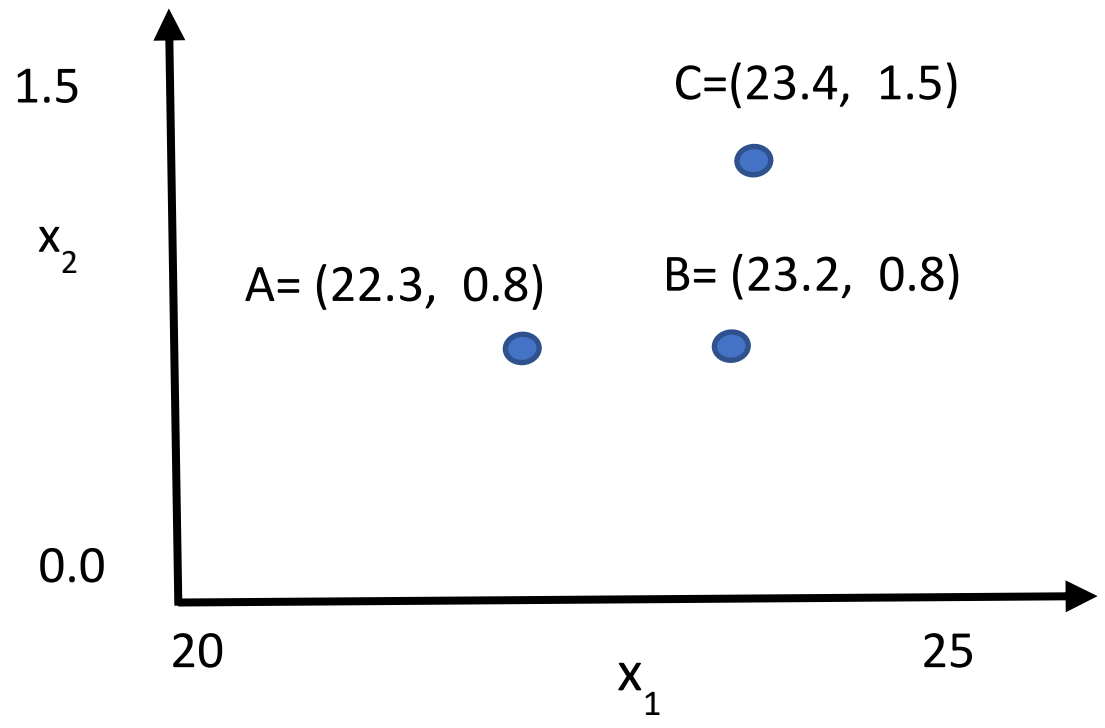

Figure 4.3. Space spanned by features of different value ranges.

Common normalization methods are to set the mean of each feature to zero and variance to 1. This is necessary, e.g., in neural computation and perceptrons, and useful in certain regression techniques and classification using support vector machines (SVM).

A common big mistake is to normalize the entire sample set at the beginning of the design process before dividing it into training, validation and test sets. The correct procedure would be to determine the normalizing factors from the training data alone and to use this information before dividing into the validation and test sets.

### 4.3 Supervised Learning

In the case of supervised learning, there is either a regression or a classification problem as shown in Figure 4.4. The regression problem predicts the value of a continuous variable, for example, 1.27, 478.67, 34576.4, etc. The classification problem, in turn, predicts the value of a discrete variable, e.g., 1, 2, 4, cat, dog.

Both are predicted to be dependent on the input variables X = ($x_0$, $x_2$, $x_3$,…) that explain the value of the dependent variable Y. In each case, the inputs X may be discrete or continuous. In the case of a regression problem, a continuous variable can be predicted from the discrete inputs.

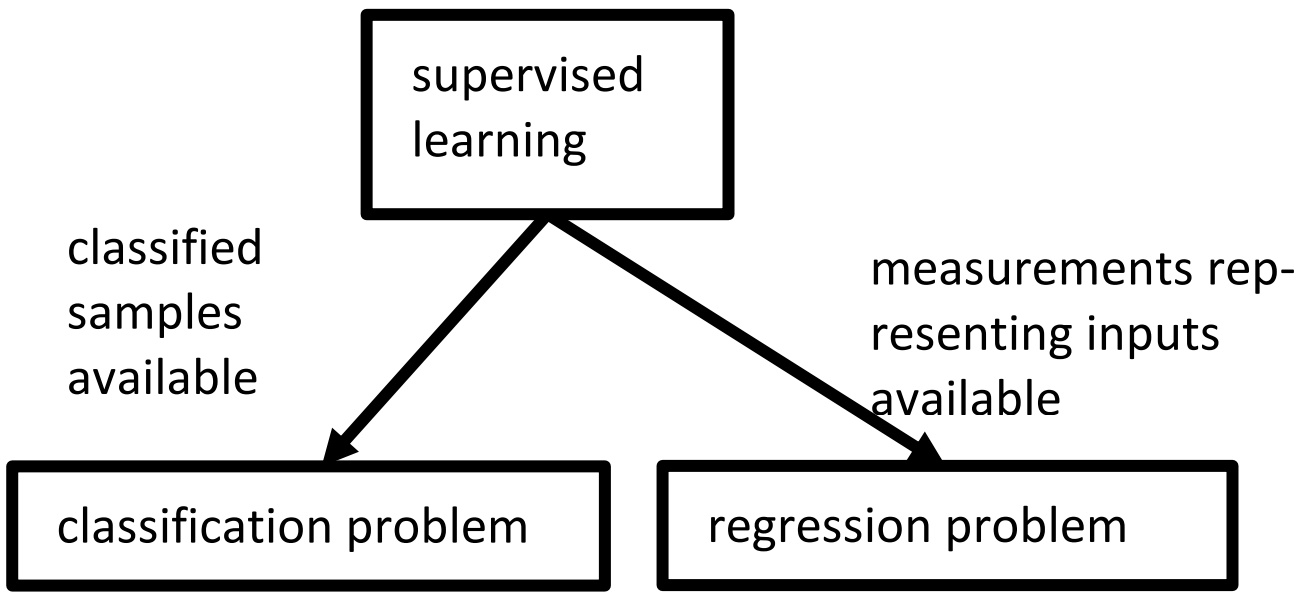

Figure 4.4. Methods of supervised learning.

Regression and classification are closely related. The key difference is that the regression identifies a model for the data structure and the classification identifies the structure to which the data in the model belongs. Similar methods are used in both. If the model and the structures are not understood, then it is a problem of unsupervised learning.

Typical regression problems include, e.g., determining the price of a used car. Inputs at that time are make, year model, mileage, and price as a dependent variable. One regression model created for this purpose with construction descriptions can be found, for example, in the reference (Matikainen, 2017). Predicting the remaining distance of an electric car is clearly a regression problem, as is archaeological sample's the so-called radio-carbon timing based on C-14 isotope concentration.

A frequently occurring special case is logistic regression in which the dependent variable is the natural logarithm of the risk of an event Y, which results in the analysis situation becoming a normal regression model (Peng et al., 2002). In a linear case, such a model has the form

$$\ln [P (Y = 1) / (1-P (Y = 1))] = aX + b.$$

Here, Y can only get two values: either Y happens or it does not. The probability of the occurrence is P (Y = 1) and the opposite is 1-P (Y = 1). Explanatory variables X can be anything, discrete or continuous variables that somehow influence the probability of an event. In everyday life, applications of logistic regression are encountered, for example in bettings organized by Veikkaus in Finland.

Many medical diagnoses, on the other hand, are classification problems where the patient's symptoms, physician findings and laboratory test results are the explanatory variables. Face recog-nition in images is also a problem of classification. After learning the pre-categorized samples, the system is able to identify categories of new samples based on their feature information.

Big Internet companies like Google and Facebook have succeeded in giving classification of training materials as a voluntary service unknowingly done by their users. At the same time users often reveal their own interests.

### *4.3.1 Regression models: learning to predict*

After recognizing the regression problem, the first attempt is usually a linear model

$E(Y|X) = f(X, \beta) = aX + b$,

where a is a parameter vector, b is a scalar, and $\beta = [a, b]$. In linear regression between two scalars x and y, the model has the form

$E(y|x) = f(x, \beta) = ax + b$,

where a and b are parameters and $(x, y) = [(x_0, y_0)\ldots(x_n, y_n)]$ pairs (explanatory variable, dependent variable).

The error criteria used to estimate the parameters β between the model prediction and the measured variable are usually either the squared sum of the error (L2 norm) or the sum of the absolute values of the errors (L1 norm). The linear regression of the two scalars using the L2 norm gives the parameters

a = covariance (x;y) / variance (x) and

b = mean (y) - a * mean (x)

An example of linear regression in Figure 4.5 is a model for the determination of the strength of lumber from modulus of elasticity obtained by bending tests leading to fracture.

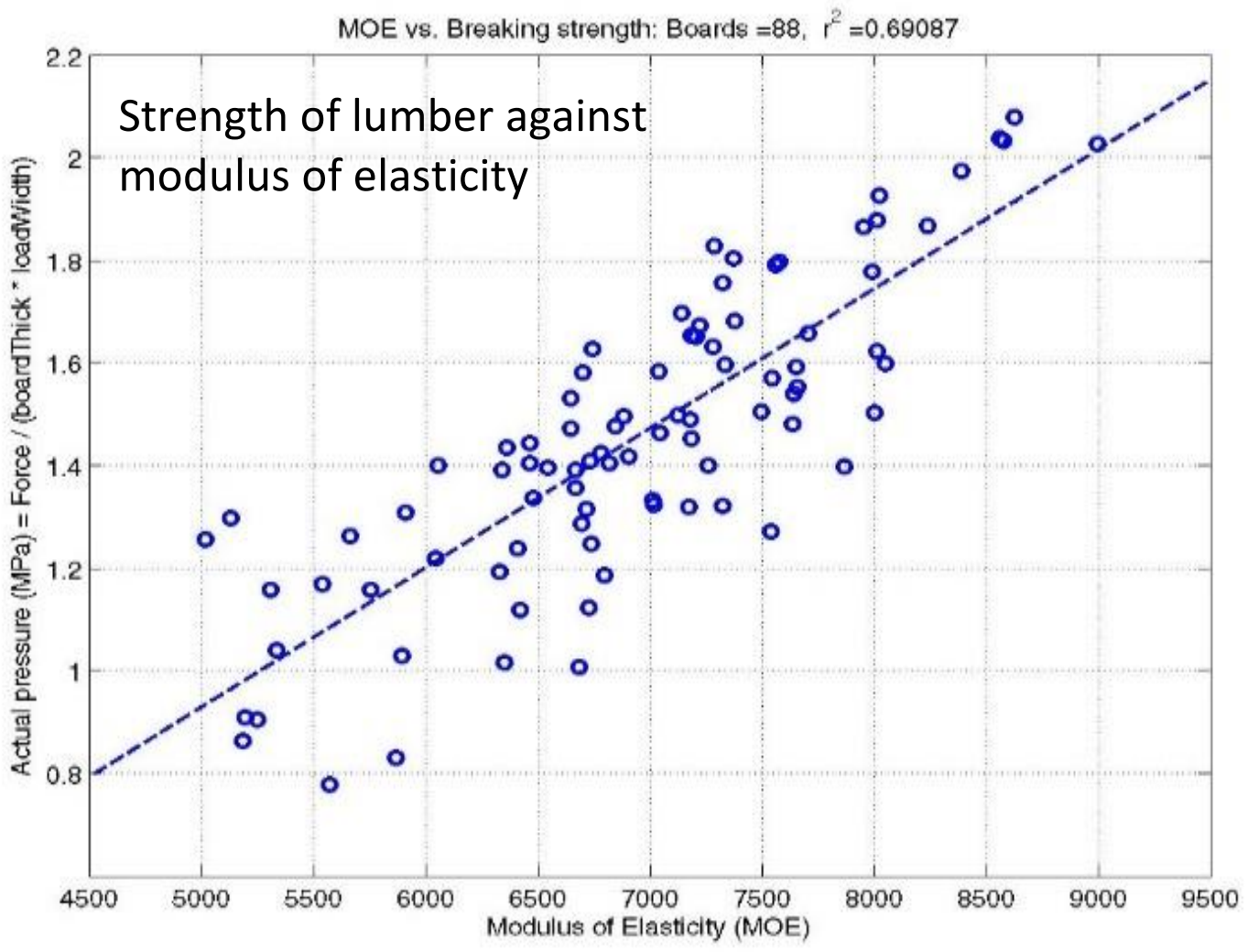

Figure 4.5. Predicting the breaking strength of lumber through a linear regression model.

We note that the data points do not obediently hit the regression line, so the strength predictions carry the risk of classifying the material as too strong or weak. The latter means selling good timber at a lower price, the former can dangerously weaken the structures.

We do not see clear class categories in the picture, although the boards are sold in strength classes such as C18, C24, etc. This contradiction stems from the design rules and the strict human classification criteria that were once followed. Many other actual regression problems have been interpreted as categorization for human use.

We find that linear regression is not necessarily the best predictive model for sawn timber strength. Often a better procedure, when the variables Y to be described are continuous or *ordered* discrete values, is to use the so-called regression trees for learning the prediction model (Breiman et al., 1984). Then, the model is built by recursively splitting the data space into parts.

Figure 4.6 shows an example of a simple regression tree in which the input variables $x_1$ and $x_2$ span the feature space. A simple regression model is fitted to the data by dividing the value range of the dependent variable Y, which is explained in pieces. Splitting ends when the desired error criterion is met. The value ranges of Y are color coded and the red lines in the right figure represent their respective ranges of variables $x_1$ and $x_2$.

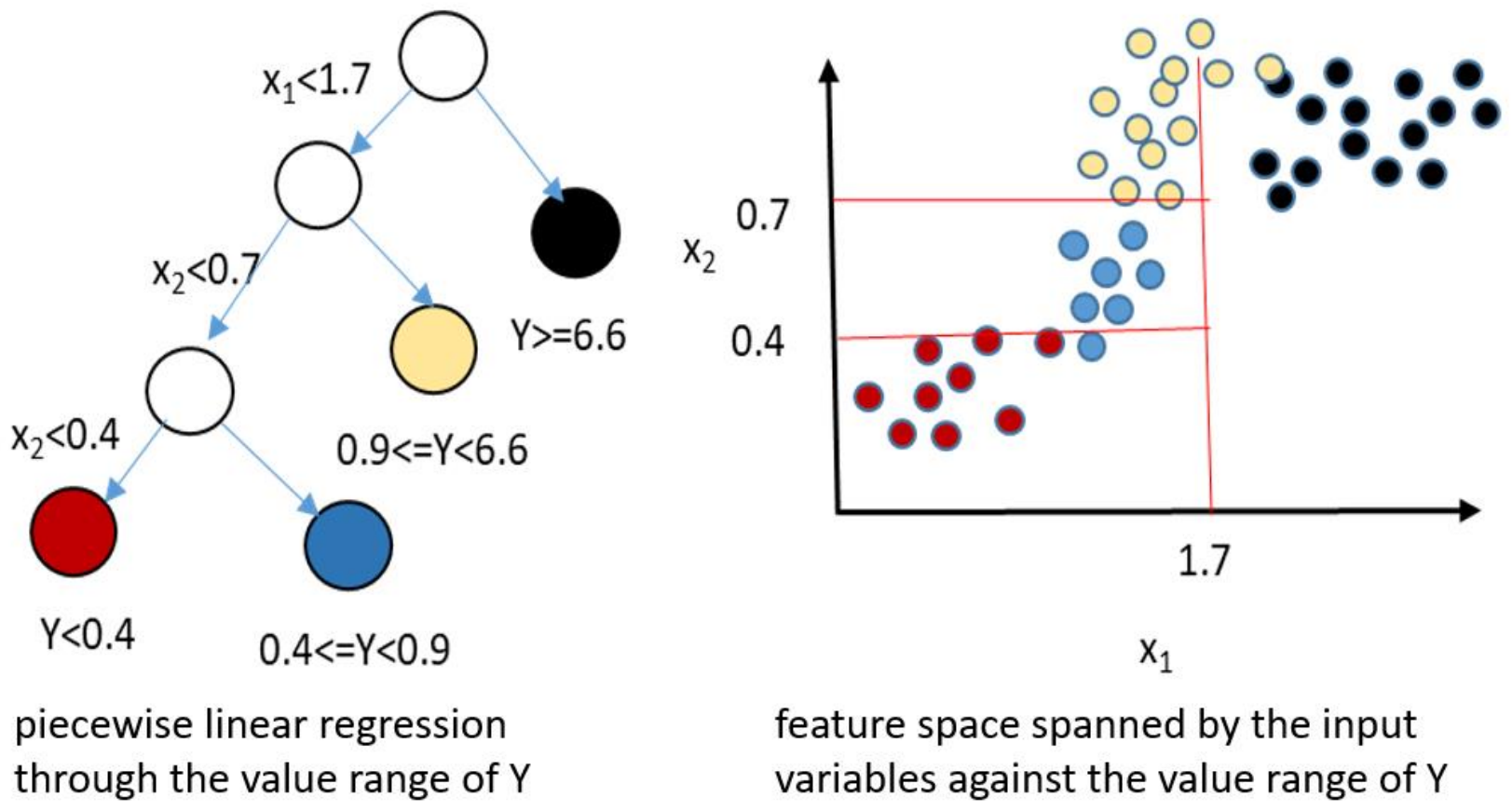

Figure 4.6. Regression tree and corresponding feature space.

Each leaf node in the regression tree corresponds to one part of the regression model. In the case of a new sample, the values of the declarative variables $x_1$ and $x_2$ proceed from the root to the corresponding leaf node and calculate the value of Y using the regression model it contains. Regression trees do not require normalization of input data, which in itself is a significant practical reason for the popularity of the method.

Random forests are popular solutions for supervised learning in regression and classification, because of the accuracy achieved. Their principle is to randomly take samples from training data and generate a large number of regression trees. The prediction given by each tree to the explanatory variable is combined with the final result. When using regression trees, it is essential to ensure sufficient training material.

Neural networks are capable of modeling non-linearities, so their application to regression models is very attractive. Because of their large number of internal parameters, there has to be a large amount of data, regularly more than for regression trees. For small amounts of data, even linear regression due to its small number of parameters can lead to a better model than that provided by neural networks.

Both regression and classification carry the risk of **over- and underlearning** (also called over- and underfitting) (Bishop, 2006). If the model has too many parameters, it may work correctly for the training data but will fail to predict with other samples. For example, the regression tree suffers from overlearning if the data space is too fragmented. Then, each leaf node regression model may predict on the basis of only one training sample and give good results only on data identical to the training samples.

Figure 4.7 illustrates the problems. In overlearning, the model fits too tightly into the training data. In the case of underlearning, however, the training data does not represent the actual use of the model.

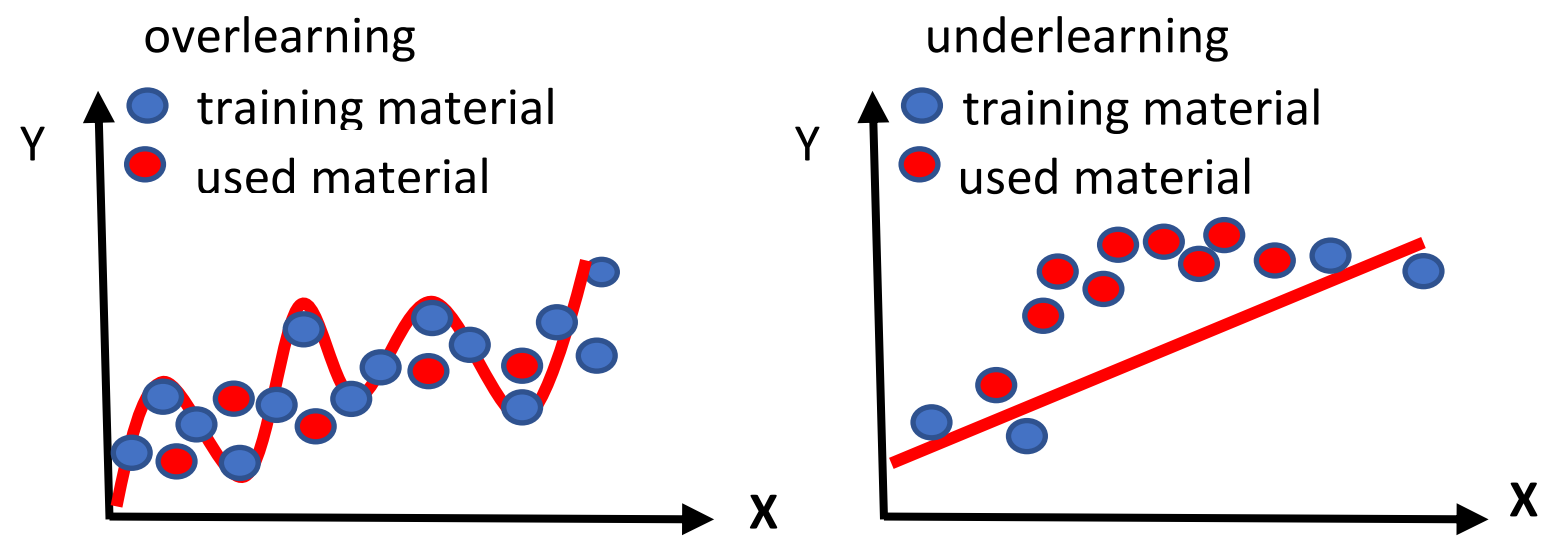

Figure 4.7. Over- and underlearning.

Beyond regression trees, overlearning is also a particular problem when using neural networks if there is too little training data. The random forest method does not suffer from overlearning, which is another reason for its popularity.

Underlearning means that the resulting model is not sufficient to mimic the structure of the data. In a typical underlearning situation, linear regression is chosen as the technique, but the relationship between the input and dependent variables is non-linear. As a result, the accuracy of the predictions remain poor.

Usually, both over-and underlearning is caused by too little training data or has represented only part of the actual data space. In practice, each model performs worse in reality than with the material used during development.

## 4.4 Learning Classification

In classification problems, classes of training samples are known, with the minimum number of classes being two. Members within each class are similar by some measure. Typical classification problems include, for example, heart rate recognition from an ECG or PPG signal and junk e-mail recognition. There are a number of classification algorithms, the most popular of which are k-nearest neighbors (kNN), naive Bayes, combination classifiers, gradient boosting, support vector machines (SVM), neural networks, and random forests (Duda et al., 2001), (Bishop, 2006).

### *4.4.1 Random forest classification*

Random forest classification is one of the best classification methods. In it, the classification tree is constructed in a root-to-leaf style by dividing data samples into similar subsets (Liaw & Wiener, 2002). Thus, more than one leaf node can represent one class. The random forest procedure is an excellent compromise in the classification between sample size requirements and the performance to be achieved. Above, the goal of random forest regression was to have a continuous value, in which the model was adapted to the explanatory variable step by step with each declarative variable.

### *4.4.2 Naïve Bayes classification*

The Naïve Bayes classification is a simple yet performative method of supervised learning (Bishop, 2006). It is also suitable for use with many features. It only needs a priori probabilities for each class and conditional probabilities for features $x_i$ to get each of its possible values for a given class.

**Example of naive Bayes classification**

Consider simplified spam detection. You select the repetitive phrases in the emails in Table 4.1 below, some of which relate to elections, as the training material.

Table 4.1. Phrases in emails

| Phrase | Class |
|---|---|
| Apply for loan immediately | Garbage = G |
| Get a loan immediately | Garbage |
| Loan | Garbage |
| First loan without interest | Garbage |
| Unite the population | Election matter = E |
| Change immediately! | Election matter |

The requisite background information is often obtained from human-graded training samples such as the above.

In the case of an unknown sample, the *a posteriori* probabilities of a feature vector X consisting of its features are calculated for each category. The new sample is then classified into its highest probability class. Assuming that the occurrence of spam and election phrases represents the actual frequencies of the corresponding messages, *a priori* probabilities are obtained

for garbage P (G) = 4/6 = 2/3 and
for election matter P (E) = 2/6 = 1/3.

We decide to choose the words that appear in the phrases of G and E categories. There are four junk phrases, each with the word "loan" in it. So the probability of the word "loan" in spam is

P (“loan” | G) = 4/4 = 1

The word "immediately" appears in two junk phrases, so we get

P (“immediately” | G) = 2/4 = 1/2

Still counting the number of words we find

P ("first" | G) = P ("get" | G) = P ("apply" | G) =
P (“without” | G) = P (“interest” | G) = 1/4 and
P (“unite” | E) = P (“population” | E) =
P (“change” | E) = P (“immediately” | E) = 1/2.

Now the email comes with a clear troll content, "population change immediately!". The posteriori probability of interpreting this phrase as spam is

P (class = G | X = "population" "change” “immediately") =
P (category = G) * [P ("population" | G) *
P ("change" | G) * P ("immediately" | G) =
2/3 * (0/4 * 0/4 * 1/2) = 0

We find the interpretation to be an election message because it has a higher posteriori probability:

P (class = E | X = "population" “change" " immediately ") =
P (class = E) * [P (" population"| E) *
P ("change" | E) * P ("immediately" | E) =
1/3 * (1/2 * 1/2 / * 1/2) = 1/24

We see the 0-coefficients above in cases where there was no observation in the training materials, that is

P (“population” | G) = P (“change” | G) = 0/4.

This wasn’t a problem this particular time. But the next email contains the message “Change immediately! Vote!”, where the word “vote” is not in the training material.

Now our classifier concludes

P (class = G | X = "change" "immediately" “vote ") =
P (category = G) * [P ("change" | G) * P ("immediately" | G) *
P ("vote" | G)] = 2/3 * (0/4 * 0/4 * 0/4) = 0 and

P (category = E | X = "change" "immediately" "vote") = P (class = E) * [P ("change" | E) * P ("immediately" | E) *
P ("vote" | E)] = 1/3 * (1/2 * 1/2 * 0/2) = 0

Both options, garbage and election content, had posterior probabilities of 0! This by no means can be a correct conclusion.

This problem could be solved by the so-called Laplacian smoothing, whereby 1 is added to each number and to the divisors the number of possible features (different words in our case) in order to avoid with certainty the rise of values over 1 (Russell & Norvig, 2010).

We now count the number of different words in category G (seven) and category E (four). In the feed "Change immediately! Vote!", the word "vote" is additional word, but it has not been used in training.

We now obtain the conditional probabilities for the words used in the teaching, as shown in Table 4.2, which, for the sake of clarity, are presented as calculations.

Table 4.2. Conditional probabilities of phrase words when using Laplacian smoothing

| Word | **P(word \|G)** | **P(word \|E)** |
|---|---|---|
| apply | (1+1)/(4+7) | (0+1)/(2+4) |
| get | (1+1)/(4+7) | (0+1)/(2+4) |
| immediately | (2+1)/(4+7) | (1+1)/(2+4) |
| without | (1+1)/(4+7) | (0+1)/(2+4) |
| loan | (4+1)/(4+7) | (0+1)/(2+4) |
| first | (1+1)/(4+7) | (0+1)/(2+4) |
| interest | (1+1)/(4+7) | (0+1)/(2+4) |
| unite | (0+1)/(4+7) | (1+1)/(2+4) |
| population | (0+1)/(4+7) | (1+1)/(2+4) |
| change | (0+1)/(4+7) | (1+1)/(2+4) |

We are now repeating previous calculations. Of course, the *a priori* probabilities are the same as in the first example:

P (G) = 4/6 = 2/3 and
P (E) = 2/6 = 1/3.

For the previously unseen word "vote", we now get the following non-zero conditional probabilities

P ("vote" | G) = (0 + 1) / (4 + 7) and
P ("vote" | E) = (0 + 1) / (2 + 4),

so the posterior probabilities interpret the phrase as part of the election data, since it is larger for the latter in the following calculation:

P (G | X = "change" "immediately" "vote") = P (G) * [P ("change" | G) * P ("immediately" | G) * P ("vote" | G) = 2/3 * (1/11 * 3/11 * 1/11) = 1.5 * $10^{-3}$

P (E | X = "change" "immediately" "vote") =
P (E) * [P ("change" | E) * P ("immediately" | E)] * P ("vote" | E) = 2/6 * (2/6 * 2/6 * 1/6) = 6.2 * $10^{-3}$

The method is called naive Bayes, as it assumes that each feature is independent of the others. In this case, each feature individually affects the posterior probability. For example, when a store's "smart scale" identifies the vegetables and fruits, shape, size and color each contribute their own in classifying the produce as cucumber, banana or orange.

Due to the assumption of independence, probabilities can be calculated for each feature separately, and do not require normalization. The naïve Bayes classifier can be shaped to operate with iterative training, making it also suitable for applications requiring ever-growing number of training samples.

The naïve Bayes classification has been found to work well in many applications where the features are not independent. However, this is not true for regression problems (Frank et al., 2000).

### *4.4.3 kNN classification*

The kNN classifier is popular because of the ease of adding new training samples. Below in Figure 4.8 the value k=5 is used.

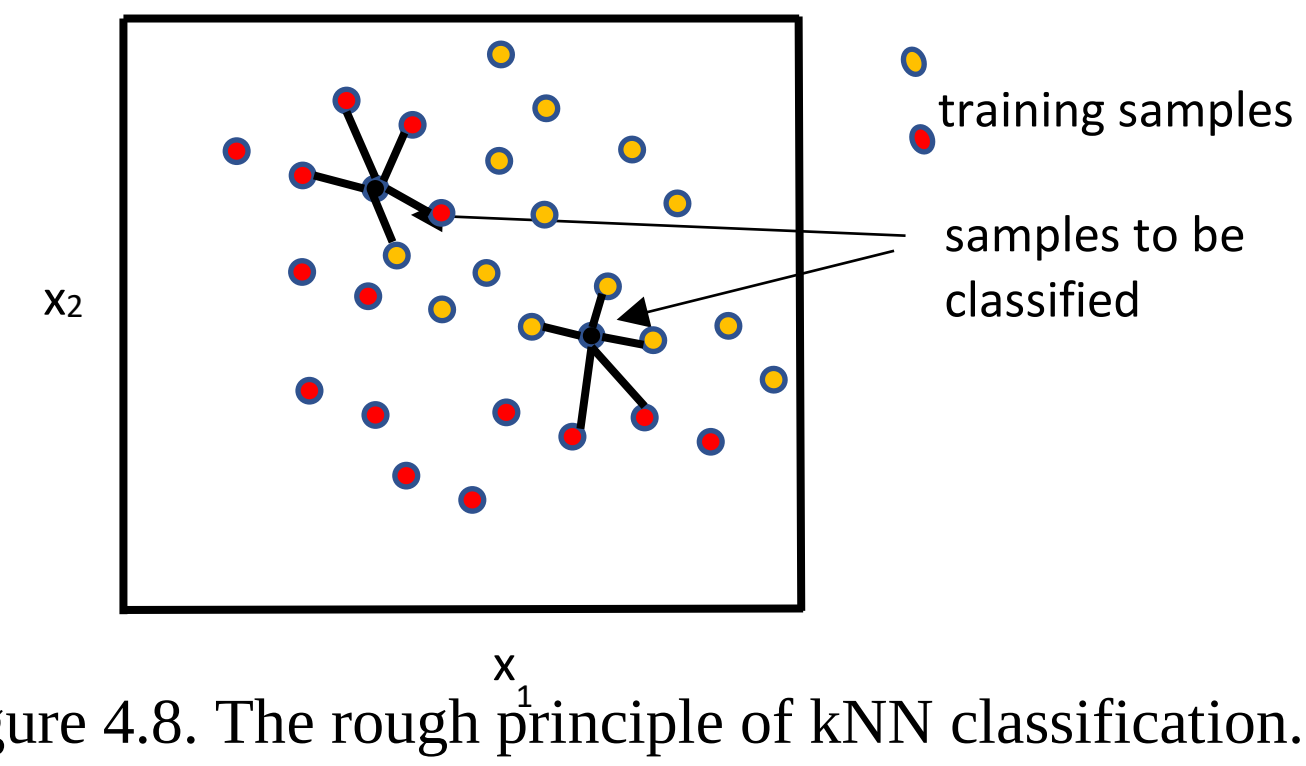

Figure 4.8. The rough principle of kNN classification.

The principle is to find the closest match to the training samples for each class to be classified, from which the most frequently occurring class is selected (Duda et al., 2001). The challenge in kNN classification is the normalization of features and the selection of a suitable distance measure, for example, in the case of the words that were features in the previous naive Bayes example. In addition, it suffers from problems with unbalanced teaching materials where some classes are underrepresented.

### *4.4.4 Ensemble classifiers*

Ensemble classifiers, principle in Figure 4.9, are one solution to unbalanced training sets. Their original purpose has been to use more or less parallel learning algorithms (Boosting) to achieve better results than the individual learner. A random forest classifier can be interpreted as a special case of an ensemble classifier where all classifiers are of the same type.

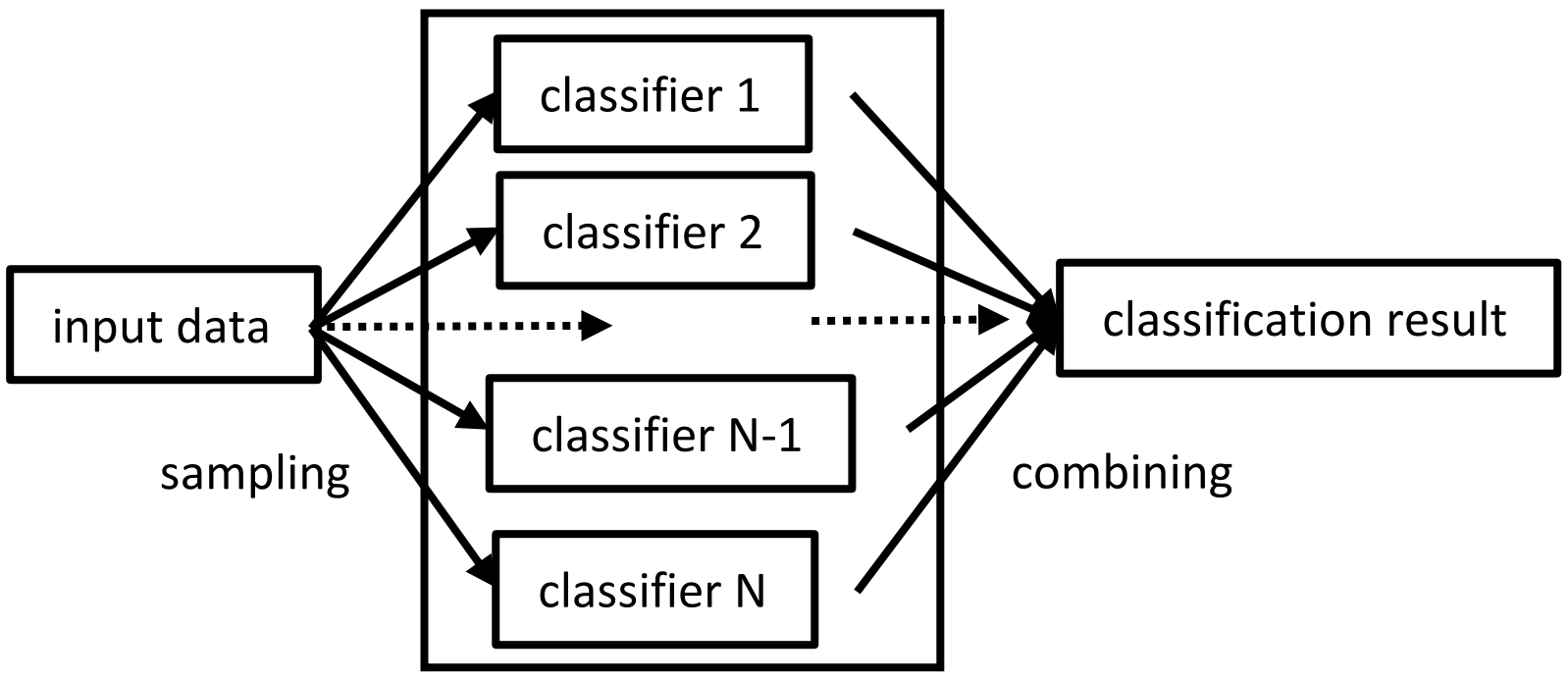

Kuva 4.9. Principle of ensemble classifier.

The concept associated with ensemble classification is "bagging" (“bootstrap aggregation”, Breiman, 1996). Its principle is illustrated in simplified form in Figure 4.10.

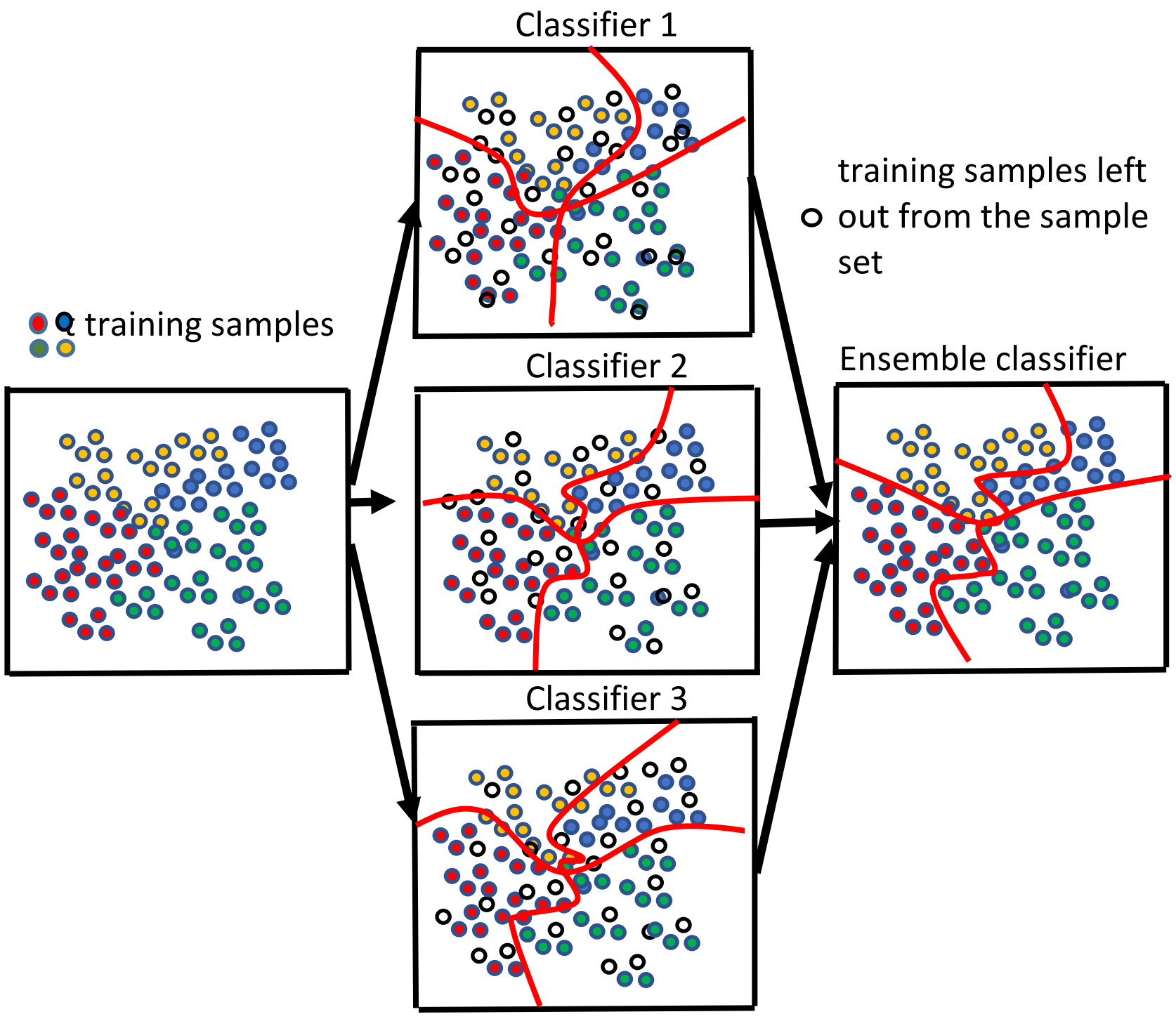

Figure 4.10. Ensemble classifier with ”bagging”.

The idea is to make random sampling of the training data so that selected samples can be selected also in other sample sets. In the picture, the red boundaries are the class boundaries for each classification result.

A classifier is trained for each sample set and the final decision is obtained by combining the classifier results. When necessary, imbalanced training data can be at least partially balanced by sampling proportionally smaller shares from the majority class. The method is also applicable when different features have been calculated for some samples.

### *4.4.5 Gradient boosting*

The accuracy of many composite classifiers can be improved by using advanced machine learning methods. The solutions based on gradient boosting have been very successful in various applications. (Wiki-Gradient).

The boosting is based on the assumption that the so-called weak learner in classification or regression can be modified to be better (Kurama, 2020). The first solution of this type was the AdaBoost that was used in Viola & Jones algorithm to detect faces in images in real-time (Viola & Jones, 2004). AdaBoost is based on a statistical solution developed by Leo Breiman (1997).

Gradient boosting is an algorithm that minimizes the value of a loss function by iteratively selecting a function that points toward a negative gradient.

Gradient boosting consists of three parts (Kurama, 2020):

1. The purpose of the *Loss Function* is to estimate how well the model is able to make predictions for a given data. In the case of regression problems, the loss function can be used to find the difference between the predicted and observed values (e.g., predicting the strength of sawn timber, Figure 4.5). In classification problems, the loss function must help to understand how good that model is in the classification problem under consideration.
2. A *Weak Learner* classifies that data, but does so poorly, perhaps not even better than in a random guess. Typically, weak learners are implemented with simple decision trees.
3. The *Additive Model* is an iterative and sequential way to add new trees (weak learners) step by step. Each iteration should get closer to the final model, i.e. reduce the value of the loss function.

Gradient boosting has proven to be a very good technique for building predictive models suitable for both classification and regression. In addition, various constraints or regularization methods can be used to prevent possible over-matching and to improve performance,

**XGBoost** is a gradient boosting program library (Wiki-XGBoost) developed by Tianqi Chen that is widely used in applications, optimized and distributed. It can be used in different programming languages and in different distributed hardware environments for regression, classification, ranking, and user-defined prediction problems.

The problems with gradient boosting (and XGBoost) are mainly due to the growing number of parameters that require user tuning through which performance improvements are achieved. At the same time, the number of training samples required also increases compared to, for example, the random forest-based method. On the other hand, gradient boosting methods then achieve better results because the random forest parameterization options are smaller.

### *4.4.6 Classification with neural computing*

Artificial, computationally implemented neural networks are also used as classifiers when sufficient teaching material is available. Neural computing is implemented with a series of interconnected layers of neurons, each with optimizable parameters through learning. The computation performed by the individual neurons is very straightforward and the potential of the networks is only fully revealed in massive applications.

Training is done by, for example, importing images of numbers and letters into the network input layer. Next, the data goes to the hidden layer and finally to the output layer that handles it, providing a response to the input. If the response deviates from the desired response, back-propagation algorithm is used to modify the weights used by the network in its calculation. In the case of deep convolutional neural networks (CNN) presented in Section 4.5, the weights are coefficients of convolutional operations.

Learning requires information on the correct response to the feed, for example, whether a cat or dog is in the picture. So it's about supervised learning. The more training samples and rounds, the better the network will learn. Because the search converges slowly, much data is needed, and increasing the amount of data generally improves the accuracy that can be achieved.

**Example of how a Perceptron neuron works:**

We consider the following simple Perceptron in Figure 4.11.

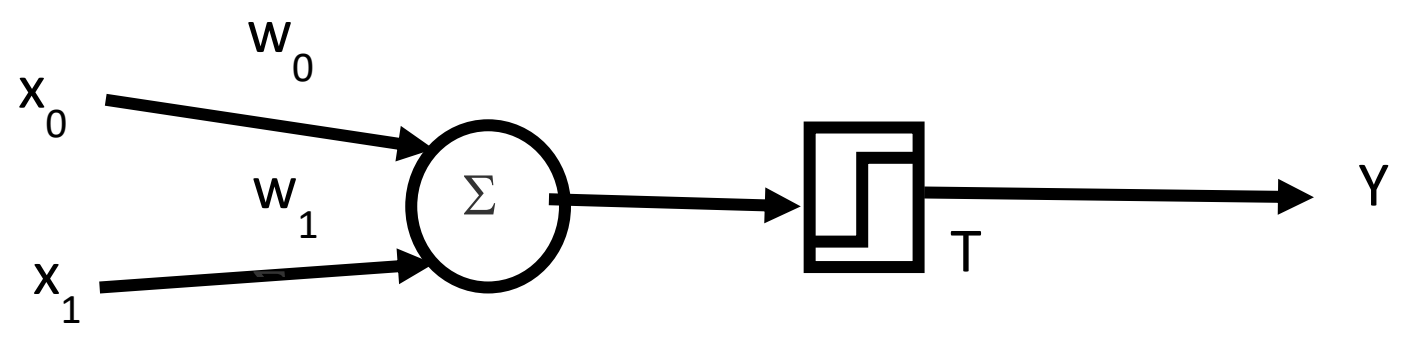

Kuva 4.11. Simple Perceptron.

The parts of the perceptron are the following:

input: $X = (x_0, x_1)$ and
weight factors: $W = (w_0, w_1)$,
which are all real numbers and
output: Y which is binary
$Y = 1$, if $\Sigma w_k * x_k >= T$, and
$Y = 0$, if $\Sigma w_k * x_k < T$, where $k=0,1$

Perceptron learns the weight factors W when presented with the outputs it should produce for the inputs.

The algorithm is as follows:

1. For the input $X = (x_0, x_1)$ calculate the output Y
   $Y = 1$, if $\Sigma w_k * x_k >= T$, and
   $Y = 0$, if $\Sigma w_k * x_k < T$, where k=0,1
2. Action options:
   - If $Y = 0$ , but it should be 1, increase the weight factors $W = (w_0, w_1)$, because the sum $\Sigma w_k * x_k$ is not large enough, also lower the threshold T
   - If $Y = 1$ , but it should be 0, the reduce the weight factors $W = (w_0, w_1)$, because the sum $\Sigma w_k * x_k$ is too large, also increase the threshold T.
   - If Y is correct, nothing is done.
   - Note: only the weight coefficients of the input lines with $x_k = 1$, need to be modified, so if X= (1,0), only weight $w_0$ needs to be changed
   - The weights are updated using equations
     $w_k = w_k +$ learning speed * (right_loutput-Y)*$x_k$ and
     the threshold is updated using the equation
     T = T – learning speed*(right_output - Y).

   If Y is corrrect, no change in values $w_k$ and T
3. Return to step 1

Our interest focuses on the calculation of sum terms and threshold comparisons:

$$w_0 * x_0 + w_1 * x_1 >= T \text{ and}$$
$$w_0 * x_0 + w_1 * x_1 < T$$

This pair of equations means that in a 2-dimensional case with features $x_0$ ja $x_1$ the Perceptron divides the feature space by the line $w_0 * x_0 + w_1 * x_1 = T$. During learning in our two dimensional case the Perceptron checks on which side of this line each sample falls, and changes the equation of the line respectively.

The weight factors W and the threshold value T are updated in a manner that the line moves toward incorrectly classified samples. In a three-dimensional case the feature space is divided by a plane, and in multi-dimensional situations by a hyper plane.

In the coordinate system of Figure 4.12 below is a line
$w_0 * x_0 + w_1 * x_1 = T$, when $w_0 = 0.5$, $w_1 = 0.5$ and $T = 0.5$.

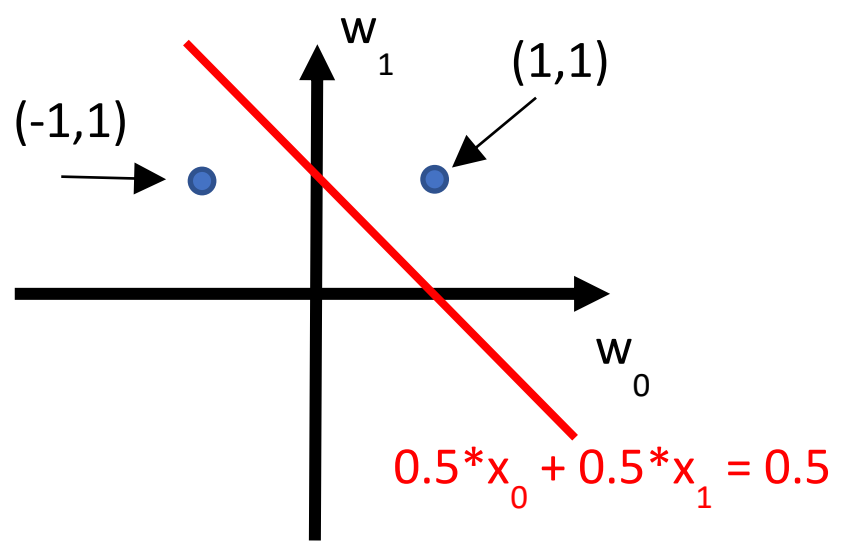

Kuva 4.12. Initial situation of Perceptron learning.

Starting from this initial situation we consider the impacts from new training samples (1,1) and (-1,1). We assume that the respective correct categories for the output are “0” and “1”. The same inputs are usually required to be repeated several times.

We choose learning speed = 0.1 and apply the update equations

$w_k = w_k +$ learning speed* (right_output - Y)*$x_k$

T = T – learning speed*(ríght_output - Y)

Step 1: Get training sample (1,1) for which we know the right output category '0'.

The perceptron equation gives

$w_0*x_0 + w_1*x_1 = 0.5*1+0.5*1 = 1 > T= 0.5$,

according to which the output is Y = 1,

but it should be “0”.

As a result, the parameters W and T are updated

$w_0 = w_0 + 0.1* (0-1)*x_0 = 0.5 - 0.1 = 0.4$

$w_1 = w_1 + 0.1* (0-1)*x_1 = 0.5 - 0.1 = 0.4$

$T = T - 0.1*(0-1) = 0.5 + 0.1 =0.6$

The new discriminant line is

$0.4*x_0 + 0.4*x_1 = 0.6$,

which we see in Figure 4.13 to be closer to sample (1,1) than the original discriminant

$0.5*x_0 + 0.5*x_1 = 0.5$

Step 2: we get a new training sample (-1,1) for which the correct output is known to be “1”. But according to equation

$w_0*x_0 + w_1*x_1 = 0.4*(-1)+0.4*1 = 0 < T= 0.6$

Y = 0. The result is the update

$w_0 = w_0 + 0.1* (1-0)*x_0 = 0.4+0.1*(1)*(-1) = 0.3$

$w_1 = w_1 + 0.1* (1-0)*x_1 = 0.4+0.1*(1)*1 = 0.5$

$T = T - 0.1*(1-0) = 0.6-0.1 =0.5$,

The new discriminant line becomes

$0.3*x_0 + 0.5*x_1 = 0.5$,

which we now see in Figure 4.13 to have turned toward sample (-1.1)

Step 3: We get the sample (1,1) again knowing that the correct output is "0". The Perceptron equation now gives

$w_0*x_0 + w_1*x_1 = 0.3*1+0.4*1 = 0.7 > T= 0.5$, on the basis

of which Y = 1, but it should be "0". As a result the weights W are updated

$w_0 = w_0 + 0.1*(0-1)*x_0 = 0.3+0.1*(-1)*1 = 0.2$
$w_1 = w_1 + 0.1*(0-1)*x_1 = 0.5+0.1*(-1)*1 = 0.4$
$T = T - 0.1*(0-1) = 0.5+0.1 = 0.6$, giving new discriminant line is $0.2*x_0 + 0.4*x_1 = 0.6$, that has increasingly shifted towards the sample (1,1) and now passes through it as illustrated in Figure 4.13.

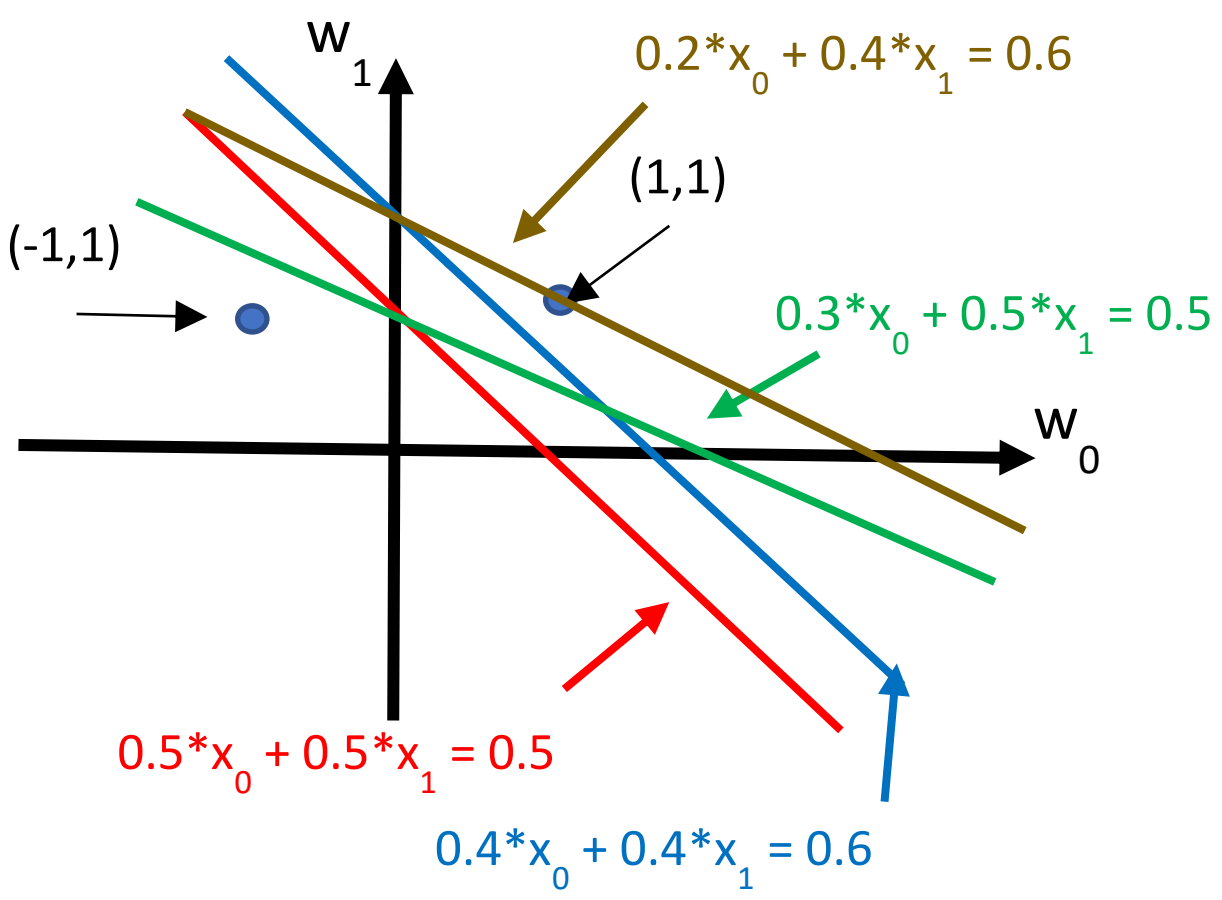

Figure 4.13. Effects of Perceptron's learning steps on the discriminant line.

The ability of a single-layer neural network is limited to learning linear discriminants. Instead, a multilayer network is capable of learning nonlinear cases. Both are trained by entering data samples and comparing the outputs to known classes of samples or other correct values.

In multilayer implementations, the weight coefficients of the neurons in each layer are modified by the so-called back-propagation algorithm and the data samples are again passed through. The training is terminated when the result no longer improves or the desired error rate is achieved.

## 4.5 Deep Learning and Convolutional Neural Networks

With the multilayer perceptron neural network it is possible to obtain non-linear discriminant functions (Figure 4.14). This differs from Rosenblatt's perceptron.

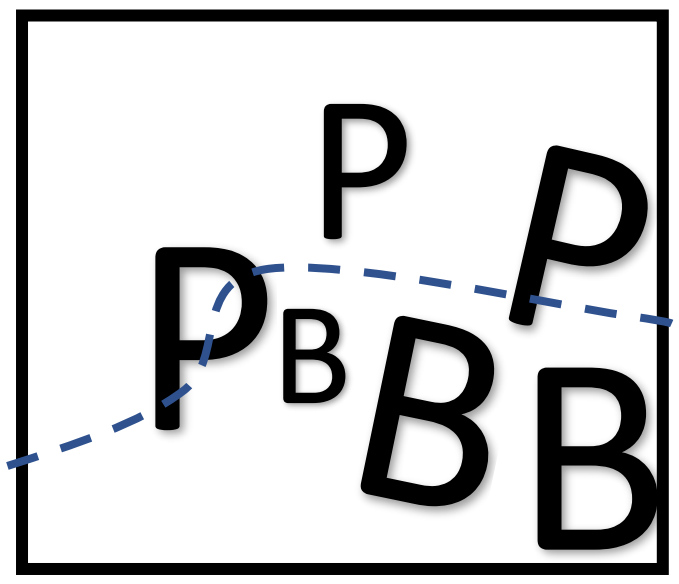

Figure 4.14. A non-linear discriminant.

Neural networks have a layer of input units, including, for example, pixels of the image or words of speech (or features thereof), hidden layers (neurons, nodes) and output units. There are connections or links between neurons. The more layers there are, the deeper the network is called (Goodfellow et al., 2015).

Figure 4.15 shows a **fully connected** neural network with two hidden layers. A network, or one of its layers, is called fully connected when each of its neurons is connected to each of the next layer's neurons. Suppose you want to identify which of the two objects is a cat or a dog. The input of the network is a set of features measured from the object, which can describe, for example, the color and shape of the object. There are two classes in the output, "cat" and "dog", and as many neurons as there are classes. For example, if one wanted to recognize handwritten numbers 0-9, there would be ten neurons in the output.

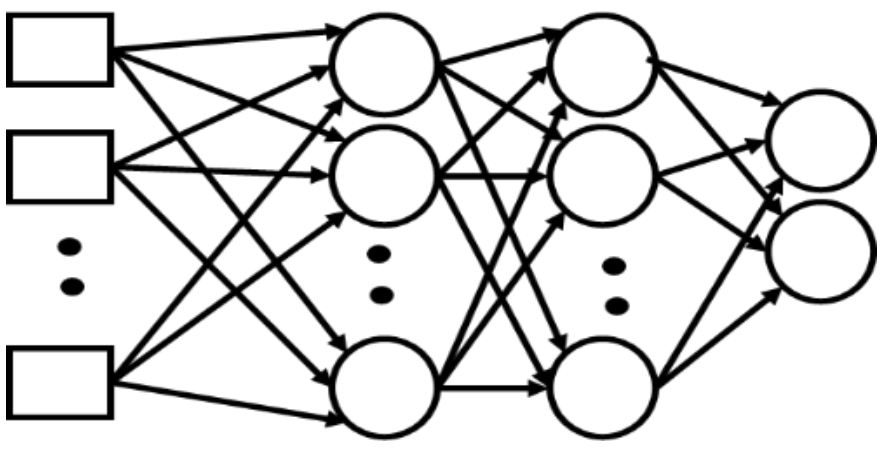

Figure 4.15. Fully connected neural network.

In Figure 4.15, the squares represent the input layer to which the features to be measured are fed, and the circular nodes are artificial neurons in two hidden layers and outputs. The neurons interconnecting links each have their own weight factor. The coefficients express the strength or importance of the neuron in question. The value of the signal transmitted by each neuron is multiplied by the weight coefficient of that link and the adding unit calculates the sum of the weighted signals arriving at a particular neuron.

As shown in Figure 4.16, from the result of the addition unit a non-linear mapping is calculated using so-called activation function f, allowing approximation of an arbitrary function. Typical activation functions are the sigmoid (f (x) = 1 / (1 + exp (-x)) or the step function.

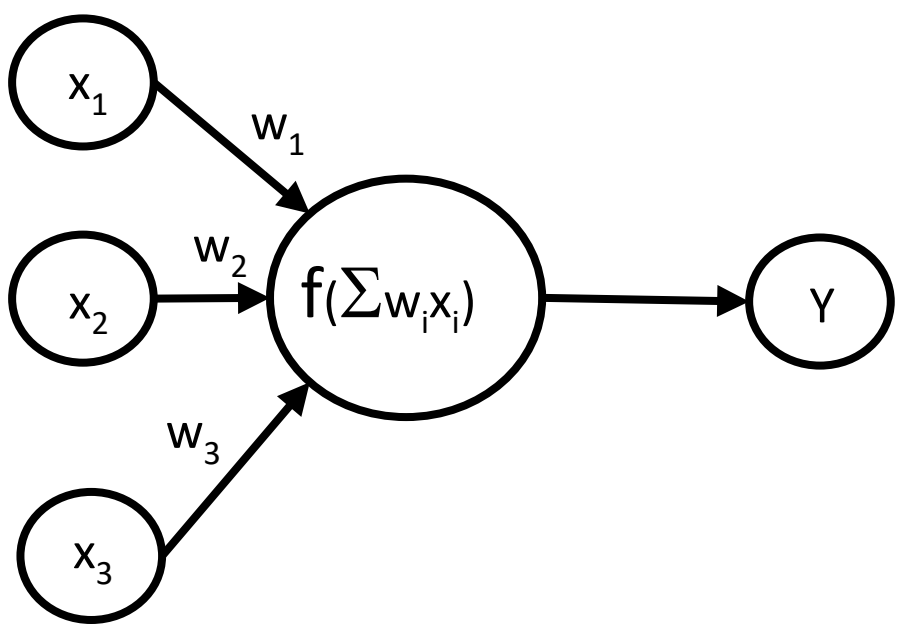

Figure 4.16. Neuron with a non-linear activation function f.

Figure 4.17 shows the general structure of a deep convolutional neural network.Most current deep networks use the convolution familiar to image and signal processing (see Section 3.5.4), which refers to convolutional neural networks (CNN) (LeCun et al., 1989).

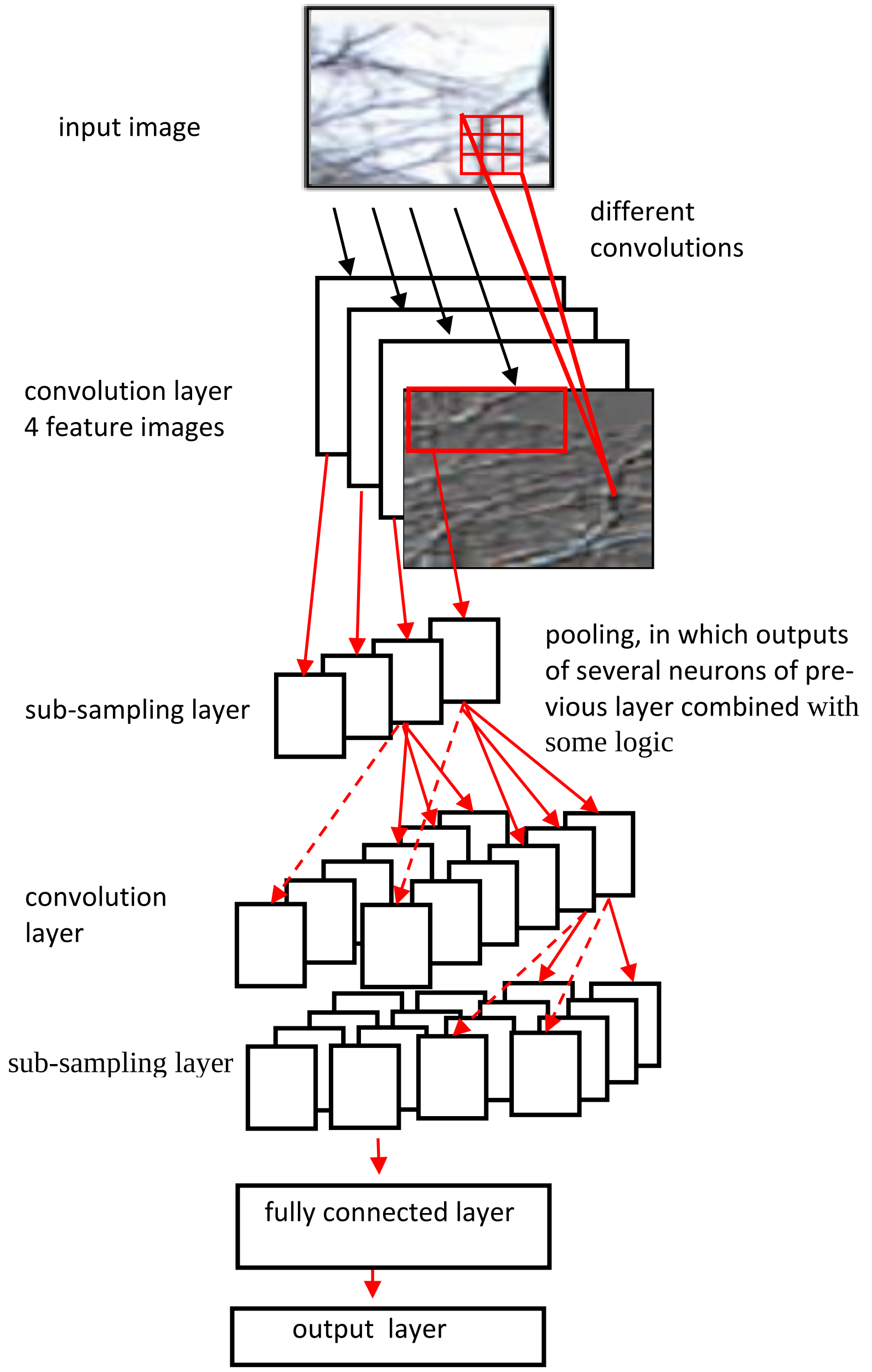

Figure 4.17. Deep convolutional neural network.

Here, the connections between neurons are limited so that the network is made independent of shifts. In this case, the same subject may be in different locations in the image and still be identified as the same. Convolution also reduces the number of required parameters compared to a fully connected network.

In the hidden layers of the deep network, gradually more global intermediate representations can be found from the objects to be recognized. For example, at lower levels edges and lines of regions, and at upper levels more complex parts of the structure of a trained object.

Initially, as described in Section 3.5.4, the convolution for the image fed into the network is calculated. The activation function mentioned above is then applied to the convolution result.

The subsampling pooling operation "diminishes" the resulting feature image after convolution. For example, four adjacent pixels can be replaced by the maximum value of the pixels in a 2x2 window, i.e. each new pixel represents the four previous values as in Figure 4.18. The same operation is performed on all non-overlapping windows in the feature image. Pooling can reduce computational load and also provide coarse-grained information from the feature image. Depending on the implementation, in addition to computational efficiency, pooling can provide, e.g., invariance to shifts in the image.

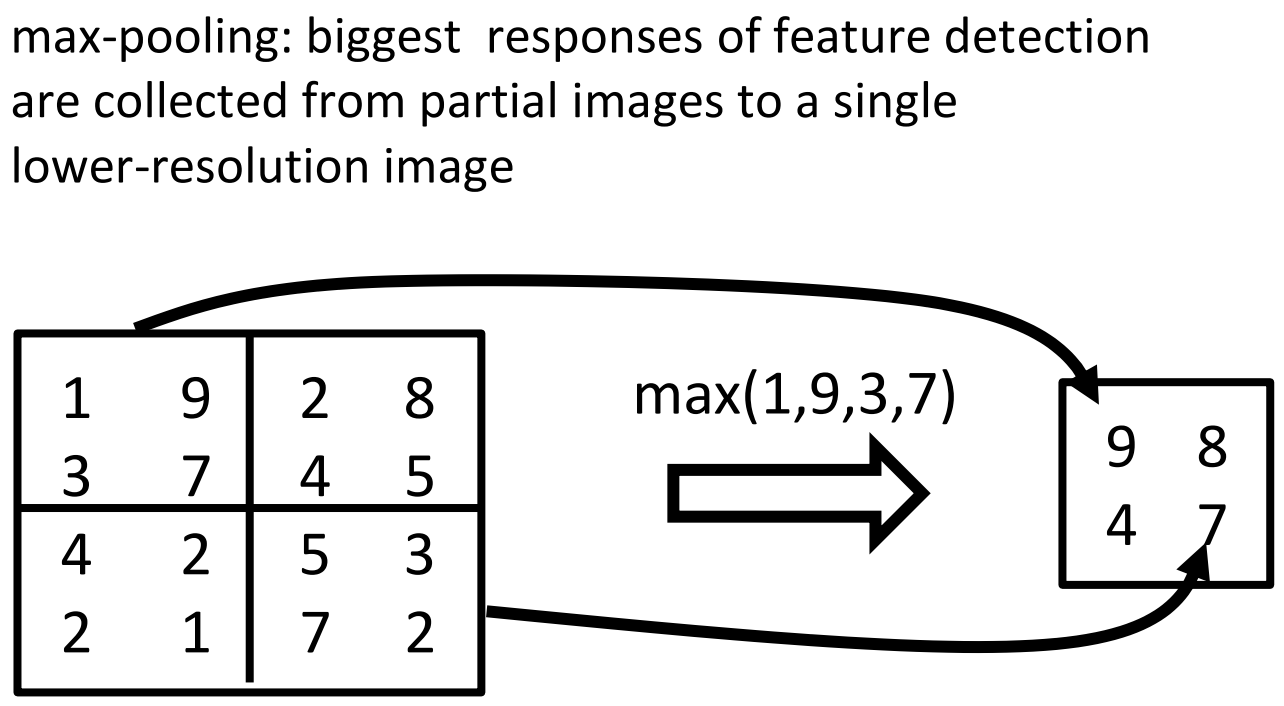

Figure 4.18. Max pooling operation.

Convolution is performed on different layers repeatedly, starting from fine-grained image data to coarser information at the higher levels of the network. Convolution filters are trained to minimize the classification error by the **back-propagation algorithm**.

The back-propagation algorithm based on gradient descent for determining coefficients has proved to be particularly useful. Current deep learning methods are largely based on it (Rumelhart et al., 1986). The method adjusts the neural links of the network during training so that each input sample (e.g., features measured from cat) produces the correct response at output (class "cat").

The name back-propagation comes from starting the process on the output end neurons (two in the above example), finding out the errors in their weighting factors, then looking at the effect of the previous layer neurons on the error, etc. Finally, it is known how each neuron in the network effects on the total error, and then each weighting factor is changed so that the total error decreases most.

By using a large number of training samples from each object to be recognized, the system is learning, i.e., finding the appropriate weighting factors.

The relationship between input and output of a neural network can be described by the term "mapping", meaning that neural networks, and in particular deep networks, learn very well non-linear input-output mappings.

A breakthrough in machine learning was achieved in 2012, when Krizhevsky et al. (2012) demonstrated that by training a sufficiently multilayer CNN neural network with massive amounts of data, it is possible to achieve significantly better classification results than before. They studied the categorization of 1,000 different categories of images using supervised learning with 1.2 million images in the ImageNet database. The so-called AlexNet neural network they used consisted of eight learned layers, the first five of which were convolutional layers and the last three fully connected layers.

The number of training rounds required can be reduced by normalizing the sample data prior to input to the neural network by at least zeroing the mean, or by normalizing the distribution of each feature to mean 0 and variance 1. The correlation between the various features can be reduced for example with Principal Components Analysis (Section 3.5.6).

One of the major benefits of deep networks over traditional methods is that the features and data representation can be learned at many scales directly from the data. Networks also allow for so-called. **End-to-end training**, whereby all parameters of the recognition system can be calculated as a single entity, not step by step. An example would be teaching the CNN network so that the system is able to directly map the information provided by the camera in the direction of travel into self-driving car's control commands.

It has been found that network performance can be enhanced by implementing deeper networks than before, although the limits for this phenomenon are now clearly visible. The 16-layer VGG-Net and the 22-layer of GoogLeNet already achieved much better results than eight-layer AlexNet  in the classification error (Liu et al., 2020).

Some recent networks already have over 100 layers (for example, ResNet and DenseNet). In the GoogLeNet architecture, the number of parameters was reduced to four million compared to AlexNet's 60 million, while the VGGNet, which is publicly available and widely used by the research community, has as many as 139 million parameters. With a small set of parameters, the network can be taught faster and is not as sensitive to overfitting

The number of samples available for teaching can be increased using the so-called **augmentation**. This means artificially manipulating existing samples, for example, calculating versions of different scales, cutting, turning around and rotating.

A deep neural network trained for a particular task can be **fine-tuned** for another task of the same type. For example, in recognizing "cats" and "dogs", the network may have initially been trained to recognize 1000 different classes in the ImageNet database. Now the top layer of the network can be removed and changed to correspond to the two classes, but the lower layers already trained do not need to be interfered, but are marked as non-trainable. After that, the model is easily adapted to the new problem. Section 8.5 provides an example of using a CNN neural network, training sample augmentation, and fine tuning to classify X-ray images.

Research on deep neural networks continues to be intense and numerous new solutions have been developed for a variety of problems and applications. The reference (Liu et al., 2020) provides a comprehensive review of  deep learning based methods and system architectures from the perspective of one challenging key application problem, generic object detection.

How can you then find the most suitable network solution for each application problem? This requires experience with different alternatives, evaluation of alternatives - and plenty of experimentation. Indeed, we might speak of “handcrafted” architectures in the same style as traditional features are often referred to as “handcrafted”.

What then were the reasons why most of the solutions developed much earlier did not lead to a breakthrough in machine learning until the 2010s? The availability of digitized data is perhaps the most important factor. In the past, computers were not as widely used as they are today and there was no need for large amounts of data. It has been said that up to 90% of the available data has been created in the last couple of years. New data comes from music, movies, photos, videos, smart mobile phone cameras and other sensors, medical imaging equipment, video surveillance systems, social media, etc. recorded in digital form (Brynjolfsson & McAfee, 2017).

Another major factor is the huge progress in hardware development. Deep neural network training can take weeks, even with efficient GPU (Graphical Processing Unit) originally designed for computer graphics. Previously, this was not possible. Shane Legg, co-founder of Google's DeepMind unit, said that in 2017, a task that took one day with a hardware device equipped with Google's machine learning TPU (Tensor Processing Unit) would have taken 250,000 years with an Intel 80486 microprocessor in 1990 (Brynjolfsson & McAfee, 2017).

In addition, open availability of systems development tools, such as TensorFlow (Google), Keras, PyTorch, Caffe, and Cognitive Toolkit (Microsoft), has greatly enhanced the use of deep learning. A comparison of the most common tools is presented, for example. in (Opala, 2019), (Wiki-Deepsoft).

An increasing number of deep learning AI applications are being implemented using remote services such as those provided by Google, Amazon, and Microsoft, providing much profits for the companies. There is not enough computing power on users' ordinary laptops and smart phones for training. GPUs designed for high-performance computing can be accessed through cloud services, allowing many smart phones to successfully run applications such as voice recognition and machine vision. The disadvantage, however, is often the need for a continuous Internet connection. For local computing on the user's device, simpler and consequently less accurate solutions are needed. For Nvidia, the market leader in GPUs originally designed for graphical data processing, the immense popularity of deep learning has created a wealth of new markets.

The practical application of deep learning involves a number of risks that companies investing in this technology should take into account. A “white paper” by a team from Peltarion, “Deep learning challenge: Lessons from the field” (Peltarion, 2018), describes various practical problems. Technology challenges include limited transparency, troubleshooting, resource constraints, e.g., due to the lack of AI experts, and testing.

Human challenges include cultural differences, privacy and security, and unintentional feedback loops, allowing the system to adapt to a user model instead of data, for example. Process challenges include evaluating development efforts, managing software and hardware dependencies, monitoring system performance, for example, when incoming data changes, code and support systems needed to "glue" to another system, such as cloud services, and managing experiments required to develop an artificial intelligence system.

The most critical aspect of business impact is the assessment of development effort and resource constraints. Next come the problems related to privacy and security and gluing to other systems.

CNNs are often claimed to function by imitating the information of biological nerves, as feature detectors specialized in details of the trained samples have been found in artificial neural networks. For example, the eye may have its own detector. Similar have been found in the nervous systems of both insects and mammals, which have evolved over millions of years. Now, this kind of evolution is said to take place inside computers within minutes.

This is a very limited view because the human nervous system is a system of specialized parts where the senses and the brain interact. Producing such an entity has so far proved to be quite a demanding challenge. For example, a banana fly has about 100,000 neurons (Zheng et al., 2010) whose “intelligence (?)” gives it the ability to move, get food, and communicate. The ability of man-made artificial neural network based solutions has been significantly weaker, at least so far. Everyone is also familiar with the fact that the fly seems to be learning more cautiously from the first failed swat. Artificial neural networks have not yet been able to do so.

## 4.6 Unsupervised Learning

Often there is no pre-classified data available for applications that could be used for supervised learning. Then, the sample data model is formed by so-called unsupervised learning methods. As shown in Figure 4.19, the solution is then to reduce the dimensionality or clustering, depending on the available prior information. The basis for using both is the observation that high dimensional data associated with the applications are not randomly located in the data space if the feature selection is at least somehow successful.

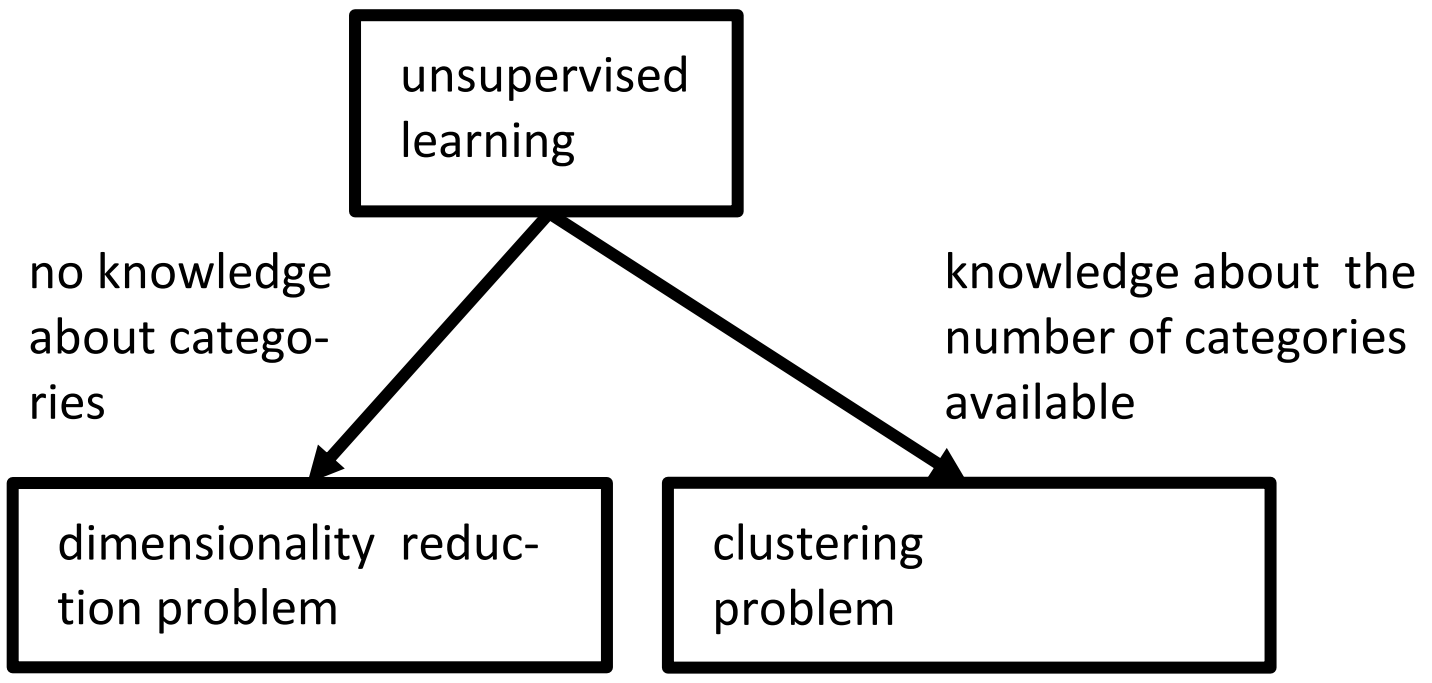

Figure 4.19. Methodological division of unsupervised learning.

Typical applications of unsupervised learning include: video content recommendation solutions, and consumer behavioral data analytics that shape marketing, product mix, and product placement.

Insurance companies analyze their customers with dimensionality reduction and clustering methods for risk analysis and pricing, industrial logistics group orders for transportation by the same means, etc. Unsupervised learning is also used to detect rare deviations from data, e.g., when analyzing medical images and identifying credit card fraud in the financial sector.

The dimensionality reductions presented in Section 3.5.6 provide an estimate of the structure of the data. In turn, clustering can be an effective solution if the number of categories is known at least roughly. Reducing dimensionality is therefore often a precursor step.

To reduce the dimensionality, it is possible to use neural networks connected as auto-encoders (Section 3.5.5), which are excellent for that purpose. However, in clustering the solutions based on them have not so far been successful. Thus, more conventional methods are still used in clustering.

### *4.6.1 Clustering methods*

The principle of clustering methods is to divide the sample data into groups where the samples in each are more similar in terms of selected criteria than between groups (Duda et al., 2001). Therefore, in the solution of the clustering problem, there is a choice of representation, for example, features describing sample data as well as a distance measure to estimate similarity of samples.

In this sense, the clustering and its challenges resemble the kNN classifier in Section 4.4.3: the result may be significantly dependent on the choice of distance measure and features and normalizations.

In addition, there is a need for a clustering criterion that somehow corresponds to our understanding of the structure of the data and a success measure to automate the iteration related to the solution. Initially, the one best suited to the problem is chosen as the clustering method. The first choice is often the so-called cluster-center method, in which the similarity of samples is compared with the cluster center.

Figure 4.20 shows examples of potentially useful clues for human interpretation in four cases of two features. Sample data often has stripe and ring structures, the nature of which depends on the features selected.

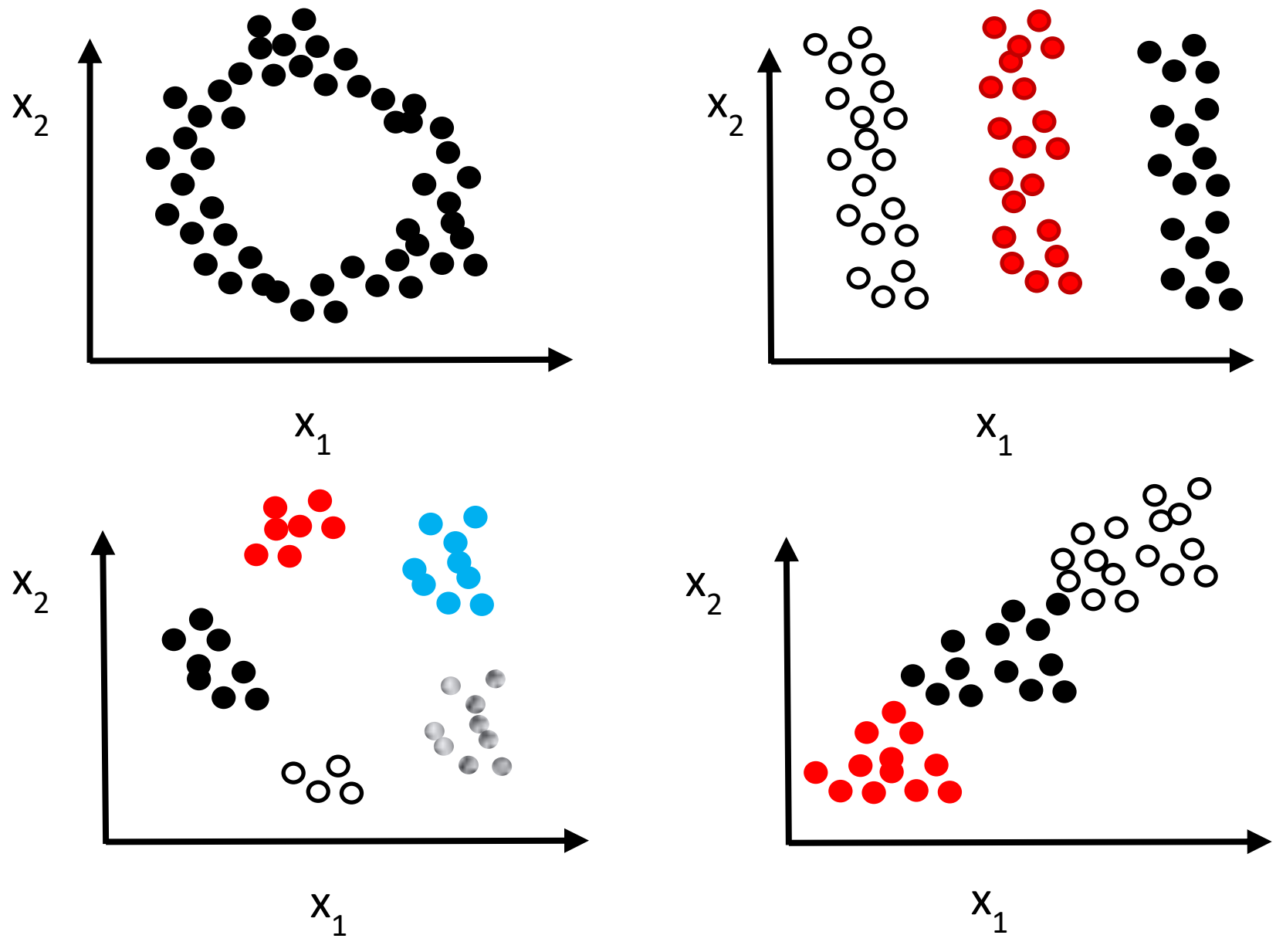

Kuva 4.20. Three different clustering problems.

The popular **k-means clustering** starts from randomly selected initial k cluster centers, iterating in the following steps:

1. Each data sample is placed in the cluster with the closest average value
2. The average of each cluster shall be recalculated
3. Return to step 1 if the average of even one cluster changed, otherwise the result is complete

The method does not always produce a human-pleasing result, as Figure 4.21 illustrates for the previous samples when k = 4.

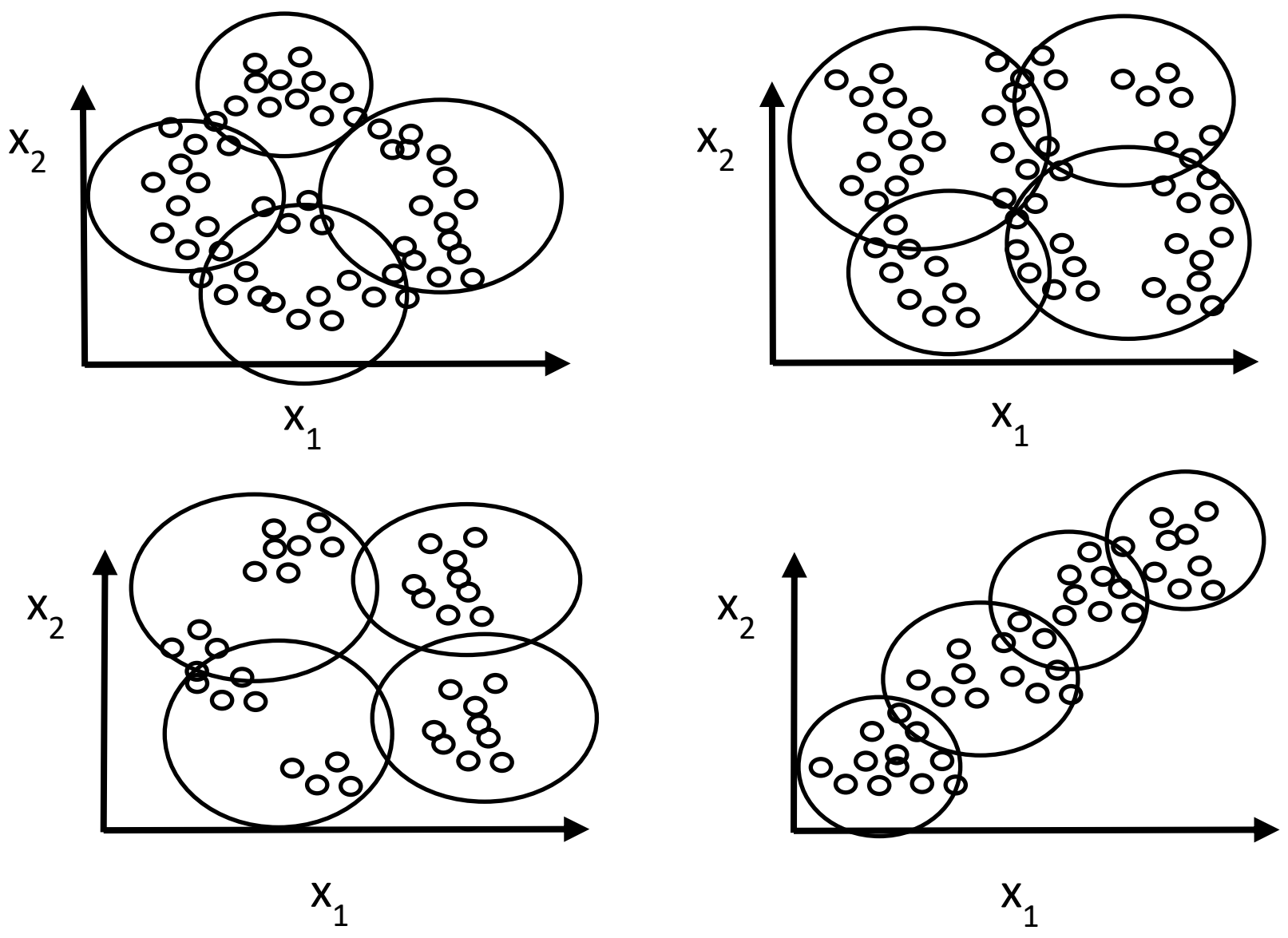

Figure 4.21. Sketched results of k-means clustering.

The number of clusters is an important source of information, as the k-means method always finds k clusters – even if there were more or less in the human mind - as the figure shows. Hierarchical connectivity methods sometimes work better than k-means clustering. Figure 4.22 shows some possible results.

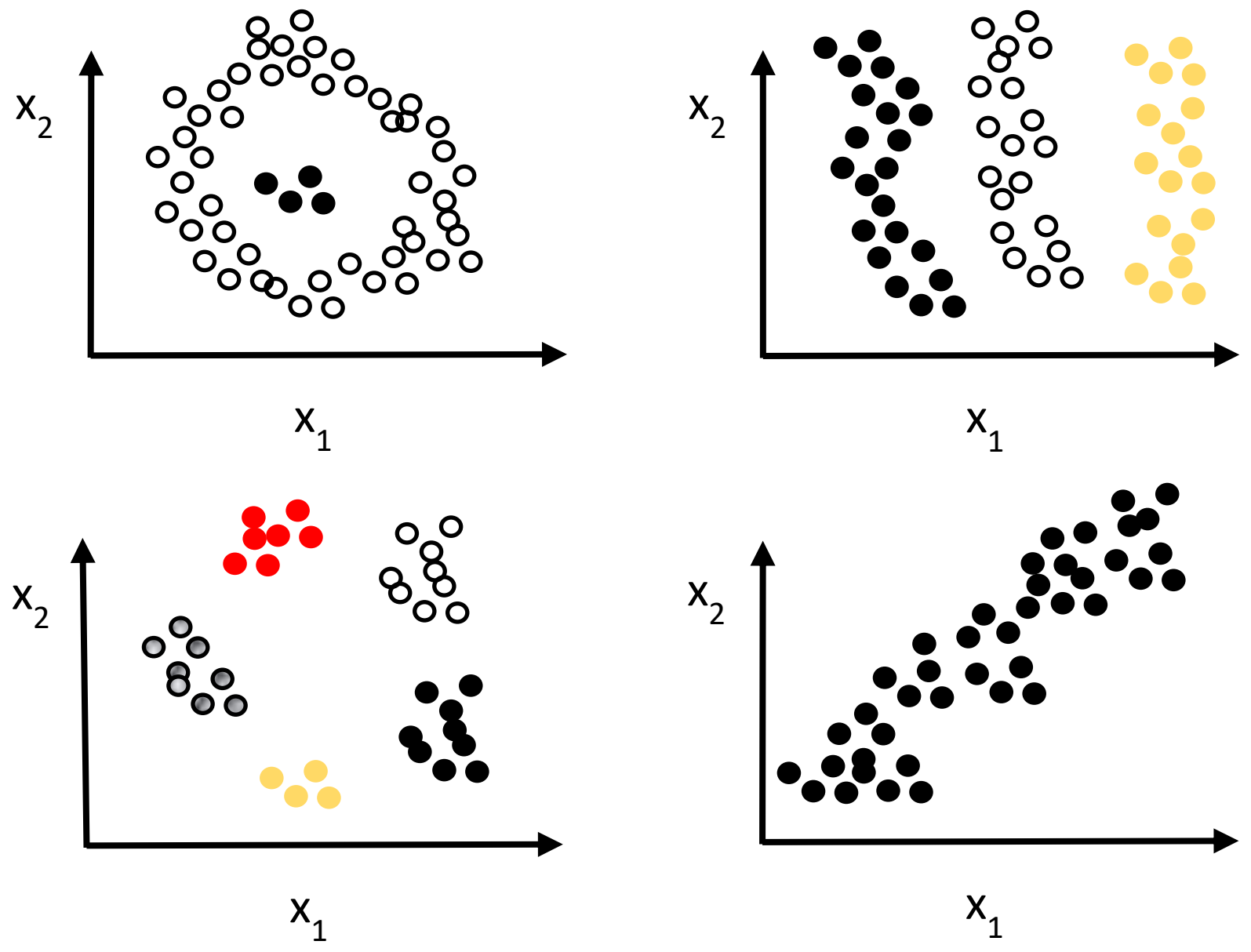

Figure 4.22. Results for hierarchical clustering.

In hierarchical methods, cluster construction proceeds either by splitting data space or starting from individual samples. The data samples located close to one another are assumed to be more similar to those further away. The hierarchical approach has the advantage of being easy to interpret, but the methods are poorly suited to large data samples due to computational costs.

We note that neither the k-means nor the hierarchical method always give clusters that fit the human expectations. At times, even single abnormal samples, or sets of a few data samples, can significantly affect the clustering result, as illustrated in Figure 4.23. The abnormal samples are in the upper right and k = 3 for the k-means clustering.

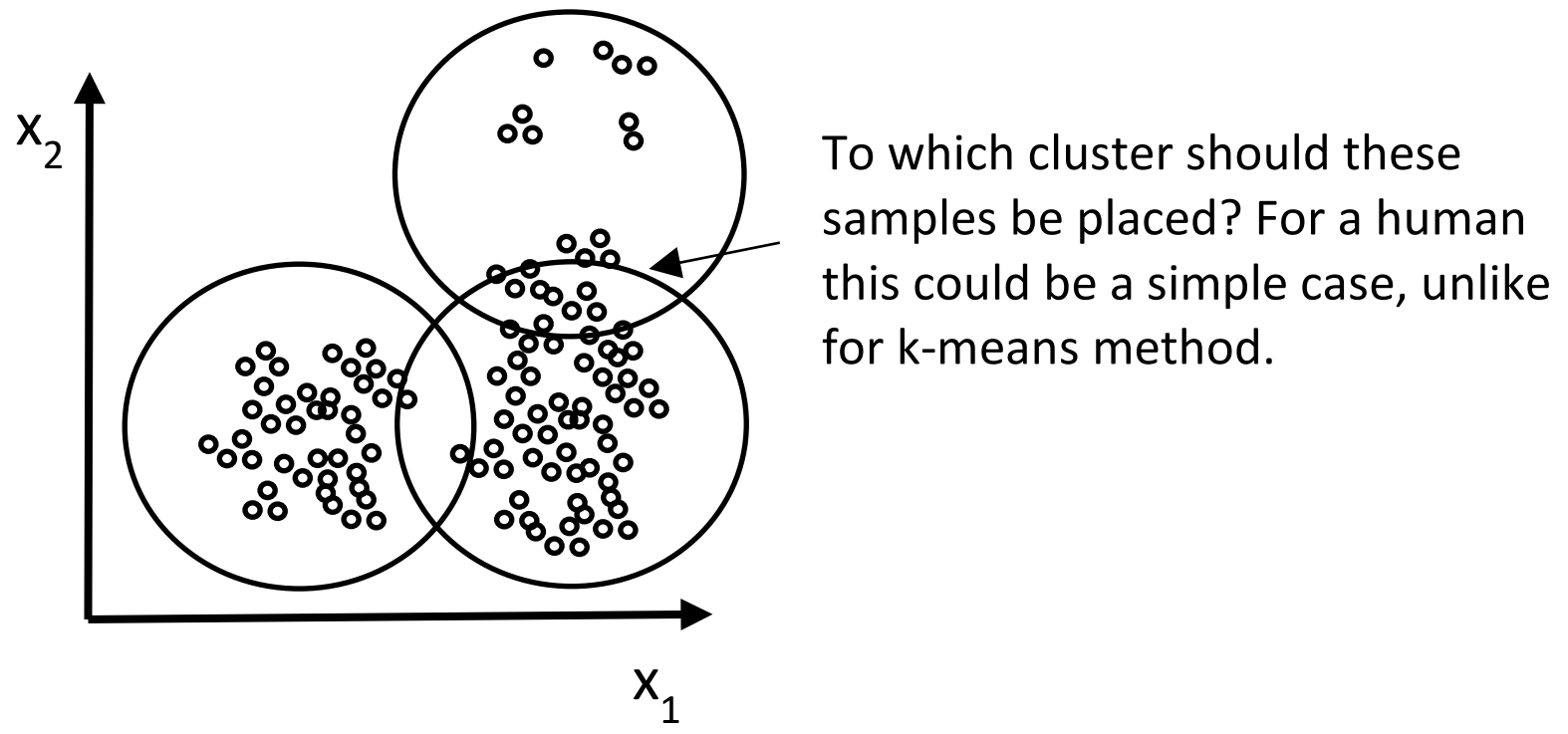

Figure 4.23 Effect of abnormal samples on clustering.

Such a situation can be encountered when the categories contained in the sample data are unbalanced or measurement errors have been included. The discrepancy may also be affected by some lack in representation and may be corrected by finding a suitable new feature.

Divergent samples can lead to problems if the clustering results are used as an input for supervised learning . In the case of Figure 4.23, some of the samples in the cluster on the lower left appear to be categorized by the effect of abnormal samples. Often, the temptation is to remove such ones from the training samples, but similar cases may come up in use cases.

Clustering results are easy to misinterpret. For example, Figure 4.24 shows a k-means clustering (k = 2) for two-car data in winter 2017-18. The data is projected into a 2-D coordinate system.

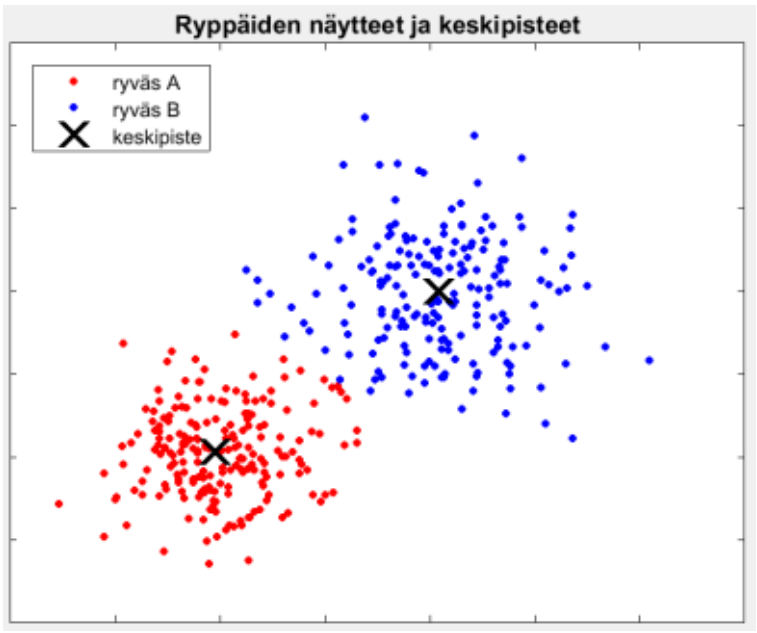

Figure 4.24. Sample data describing the use of two cars.

Initially, there were four features: average fuel consumption, outdoor temperature at departure, travel time, and average speed. It is easy to make the mistake of assuming that clusters represent different cars. In reality, they tell the most about using an engine block heater in the mornings and cold starts in the afternoons.

Sample data is often very large and, with high dimensions it can be difficult to interpret for humans. The results obtained by any clustering method may then be assimilated to pure guesses. Thus, after clustering, the following reviews are always needed to anchor the analytical results into the application:

- are the resulting clusters perhaps a result of chance or do they represent actual data structures?
- is the number of clusters reasonable for the application?
- could the clustering result be any better?

## 4.7 Reinforcement Learning

Many applications do not offer category information or prior data. Instead, they may offer positive and negative feedback on solutions made, so that learning can be based on trial and error.

Reinforcement learning is suitable for such applications to seek the action strategy or solution that receives the most positive feedback (Sutton & Barto, 1998) The goal is continuous learning, also forgetting, which allows coping with changing situations. In Figure 4.25, the agent is an artificial intelligence operating in its environment that can navigate and change its environment, receiving measurement data and feedback on its actions.

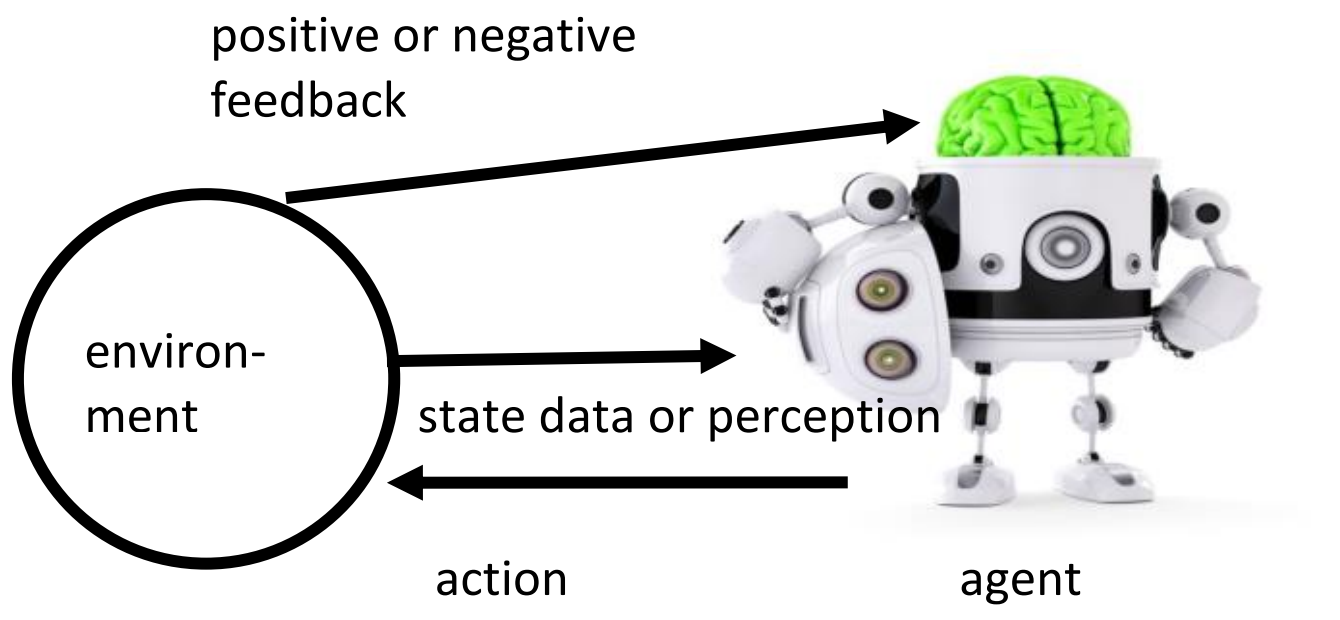

Figure 4.25. Basic reinforcement learning. (© 123RF)

Reinforcement learning lies in the midst of supervised and unsupervised learning, and requires applications that can automatically serve its feedback hunger. The problem is, of course, the large amount of feedback needed and slow learning.

In games, large feedback needs can be satisfied by simulations, which procedure has been used, e.g., in Google DeepMind's AlphaGo program, a combination of Monte Carlo simulation based reinforcement learning and deep neural networks (Section 5.3).

Q-learning is a popular reinforcement learning method (Watkins & Dayan, 1992). It is so-called model-free, not modeling its environment as such, but treating it as a state machine where the agent is always in some state. Each state contains everything about the environment.

The Q function returns the expected feedback for a specific operation when the status information and that operation are entered. Initially, the Q function returns a constant value, such as 0, for all states and operations.

Later, as the agent acquires more information, the Q function in each mode provides better estimates of possible action feedback values. In practice, the Q value is feedback on a longer-term goal, that is, guides you to choose an action toward it. Reinforcement learning observations and actions, in turn, produce Q values.

Consider reinforcement learning in a simplified case in which an android unexpectedly achieves singularity at night in a laboratory. It sets off to search the front door to the stairwell using Q-learning.

The lab layout shown in Figure 4.26 is initially unknown to Android. We also present the problem as a graph where the rooms correspond to the states of Q-learning. The stairwell is one of the states. The door can go in either direction.

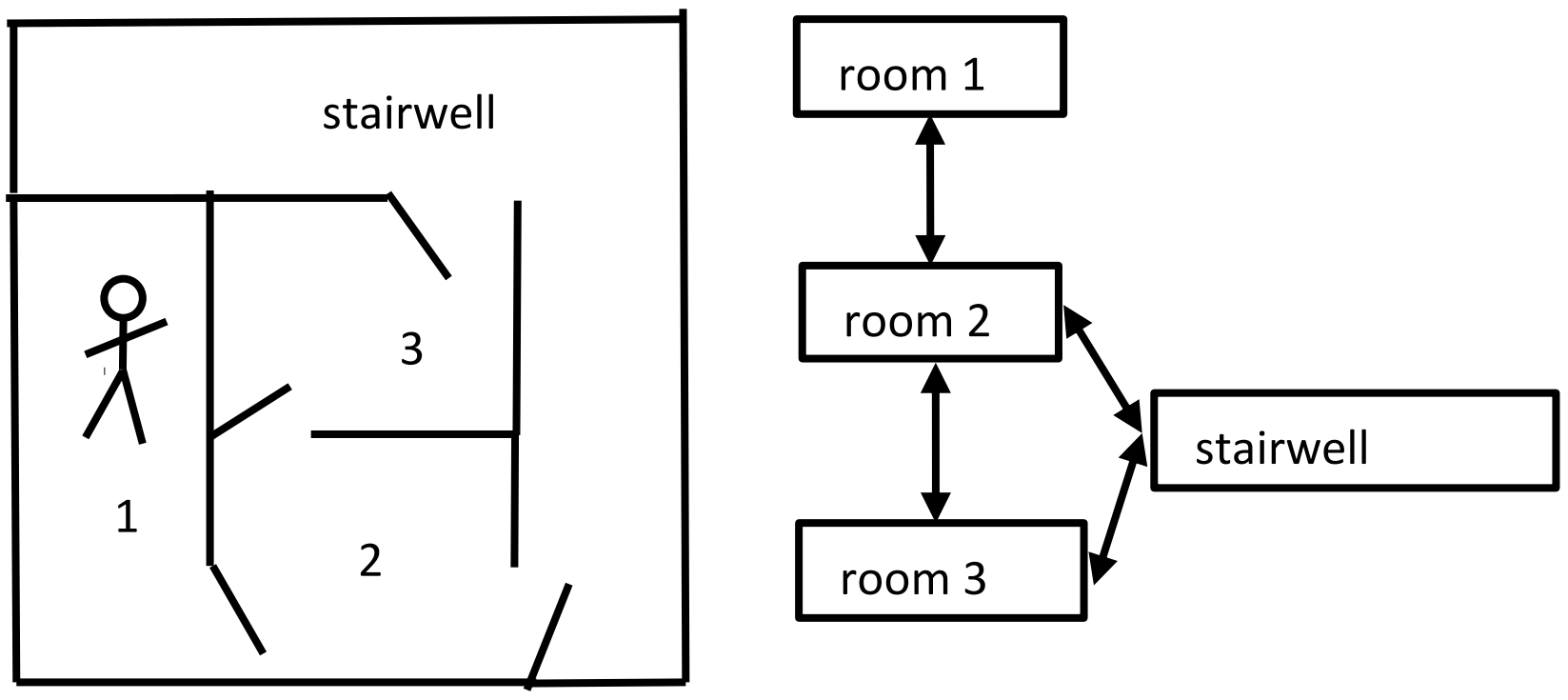

Figure 4.26. Simple application problem for Q-learning.

Transitions through doors from room to room are actions. The access to the stairwell has a feedback value of R = 10 and the access to other rooms is R = 0. The cost for staying in the room and the stairway is R = -1. All Q values are initialized to -1. In practice, the selection of feedback values is arbitrary, as long as the destination has the highest values.

In each of its current states, the android selects one possible action and determines the highest achievable Q value through each possible next state based on the data acquired so far. It updates the Q values with a simple equation

Q (state, action) = Reward (state, action) +
parameter * Max (Q (next state, possible actions))

We set the parameter (approximate) arbitrarily to 0.5.

The transition from room 1 to room 2 is now evaluated as

Q (1,2) = R (1,2) + 0.5 * Max (Q (2,2), Q (2,1),…)),

where R (1,2) is the reward for entering the room 2 through the door. The Q values for transitions from room 2 to the rest are all assumed to be initialized to -1.

The android is not yet fully aware of the potential, but it doesn't matter because of initialization assumptions. For example, initializing Q (2,2) = -1 means that one possible action in room 2 is to stay there.

Since R (1,2) = 0 and all Q values are -1, so

Q (1,2) = 0 +0.5 * (- 1) = -0.5

Of course, at this stage we would have been able to speed-up learning by concluding that if you can go from room 1 to room 2 then you can go also vice versa and Q (2,1) = 0. However, our Android is just beginning to learn. We are now only updating Q values in each state, that is, for room 2, only there.

If the Android stays in room 1, then R (1,1) = -1, Q (1,1) = -1, and Q (1,2) = 0

Q (1,1) = R (1,1) + 0.5 * Max (Q (1,2) = -1 + 0.5 * 0 = -1

The android finds now

Q (1,2)> Q (1,1),

so it moves to room 2. We find that Q (1,2) = -0.5 and Q (1,1) = -1. The latter value has not changed since initialization. Now in room 2, the android detects three doors, and those are associated with the following Q-value updates:

Q (2,2) = R (2,2) + 0.5 * Max (Q (2,1), Q (2,3), Q (2, P)) =
-1+ 0.5 * Max (0, -1, 10)) = 4
Q (2,1) = R (2,1) + 0.5 * Max (Q (1,1), Q (1,2)) =
0+ 0.5 * Max (-1, -0.5)) = -0.25
Q (2,3) = R (2,3) + 0.5 * Max (Q (3,3), Q (3,2), ...) =
0 + 0.5 * Max (-1,…)) = -0.5
Q (2, P) = R (2, P) + 0.5 * Max (Q (P, P),…) =
10 + 0.5 * Max (-1,…) = +9.5

Android now finds that Q (2, P) is the largest of its options, so it moves to the stairwell (P). There, it calculates the following Q values where we use the results calculated above:

Q (P, 2) =
R (P, 2) + 0.5 * Max (Q (2,2), Q (2,1), Q (2, P), Q (2,3)) =
0+ 0.5 * Max (4, -0.25, 9.5, -0.5)) = 4.75
Q (P, 3) = R (P, 3) + 0.5 * Max (Q (3,3), Q (3,2), Q (3, P) =
0 + 0.5 * Max (-1, -1, -1)) = -0.5
Q (P, P) = R (P, P) + 0.5 * Max (Q (P, P), Q (P, 2), Q (P, 3)) =
-1 + 0.5 * Max (-1, 0, 0) = -1

Since Q (P, 2) is the highest, the Android moves back to room 2. There, the Q values are updated again

Q (2,2) = R (2,2) + 0.5 * Max (Q (2,1), Q (2,3), Q (2, P)) =
-1+ 0.5 * Max (-0.25, -0.5, +9.5) = 3.75
Q (2,1) = R (2,1) + 0.5 * Max (Q (1,1), Q (1,2)) =
0+ 0.5 * Max (-1, -0.25)) = -0.125
Q (2,3) = R (2,3) + 0.5 * Max (Q (3,3), Q (3,2), ...) =
0 + 0.5 * Max (-1,…)) = -0.5
Q (2, P) = R (2, P) + 0.5 * Max (Q (P, P), Q (P, 3), Q (P, 2)) =
10 + 0.5 * Max (-1, 0.5,4.75) = +12.375

We now note that the Android has updated a number of Q values, some of them twice, and iteratively continues the update within a simple algorithm. We can find the route by following the highest Q values.

Generally, the applied update equations are slightly more advanced than those used here. In addition, there are loose application-specific control mechanisms for directing the learning.

There are applications for reinforcement learning in financial applications for continuous improvement of stock trading strategies, logistics, Internet services and robotics. For example, the technology behind robots learning to walk is partially based on reinforcement learning. This technique has been used to learn how to play console games with the information provided by the camera, without first knowing the rules, and developing so-called chatbots that are able to keep people engaged all the time.

### 4.8 Recurrent Neural Networks (RNN)

Deep neural networks, as such, are unable to learn the temporal or sequential dependencies of data samples fed at a time. This is problematic, for example, in the development of language-based applications, since interpretations of the following words in the text may depend on the preceding one. In the same way, for example, the following choices of chatbots used in automated customer service depend on previous steps. In addition, in these examples, a single input data must sometimes produce multiple consecutive outputs, or sometimes only a single response may be required from the input sequence.

Such needs for analyzing natural written and spoken language or continuous signals often involve the use of Recurrent Neural Networks (RNN) (Rumelhart, 1986). They can model data with temporal interdependencies, for example, to detect abnormal situations, such as exceptional situations from condition monitoring data produced by industrial rotary machines.

The principle of feedback neural network is shown in simplified form in Figure 4.27. The structure is based on a conventional deep neural network with an added connection from the output layer to the hidden layer and also between the hidden layers. As a result, the neurons in the output layer influence their own inputs through feedback. As a result, the neural network can in principle include information from all previous inputs in its weights, so that it can learn temporally dependent features. This enables time-dependent data to be processed and predicted.

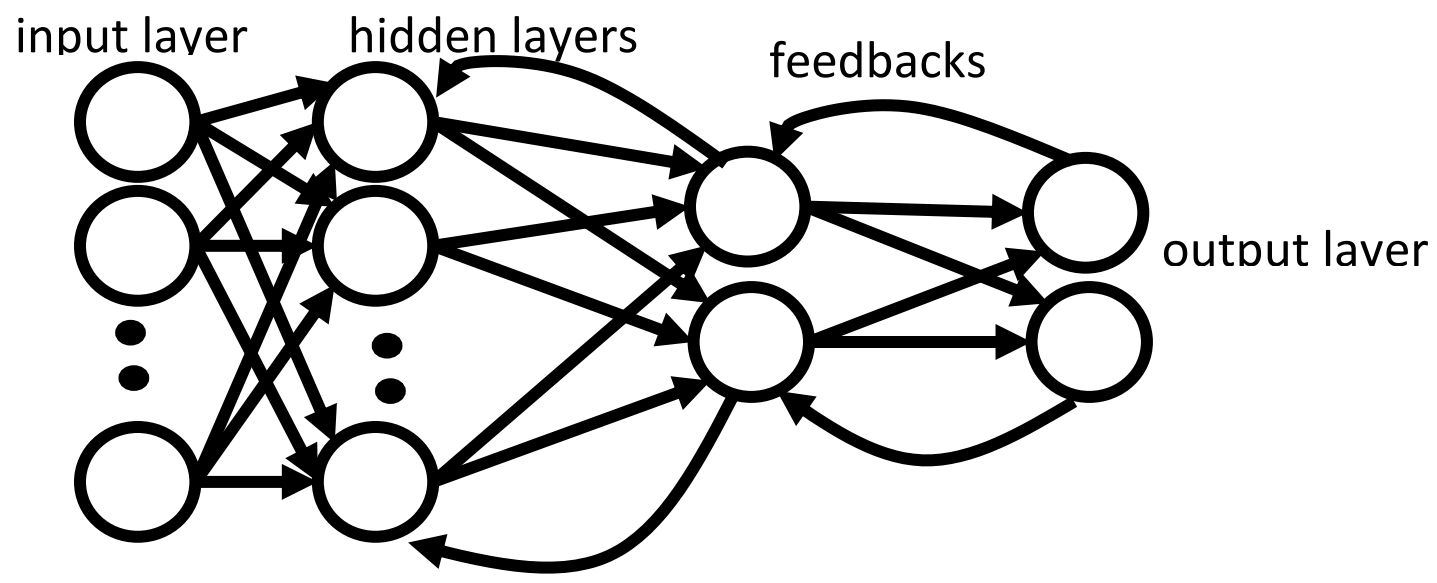

Kuva 4.27. Recurrent neural network.

In their basic form, recurrent neural networks are not capable of modeling data beyond short time series. Unfortunately, the problem is exacerbated by deep neural networks, the capacity of which would in itself be needed for more complex and larger problems.

For the purpose of longer sequences, so-called Long Short-Term Memory (LSTM) networks have been developed (Hochreiter & Schmidhuber, 1997). The term refers to a person's working memory of a few seconds into which information comes from the sensory memory. Working memory data can be stored in long-term memory - or forgotten.

At the core of the LSTM implementation is the so-called node state maintained in the hidden layers to model longer-term dependencies. This allows the RNN to recognize the essential data and to control it with feedback to be remembered. Similarly, less relevant data can be overlooked.

The great advancements in speech recognition are largely the result of LSTM research. Developing such an application requires supervised training, that is, categorized speech samples that are fed to the network by teaching a phoneme or word at a time until the desired recognition ability is achieved. Due to network feedback, a modified version of the back-propagation algorithm (BPTT, Back-Propagation Through Time) is used to modify the neural weights.

The predictive ability of the LSTM method has been demonstrated, e.g., in the automated generation of compositions and handwritten text, in the production of textual descriptions of video content, in the analysis of stock market trends, etc. In practice, the LSTM method or deep convolutional neural networks (CNN) are used as combinations for implementing applications. For example, CNN methods can be used for initial processing, LSTM is used in the intermediate stage, and finally multi-layer perceptron networks.

## 4.9 GAN – Generative Adversarial Network

The General Adversarial Network (GAN), developed by Ian Goodfellow with colleagues, has been of great interest in recent years (Goodfellow et al., 2015), (Wiki-GAN). Figure 4.28 shows the general structure of a GAN (Web-GAN). The principle is that two neural networks (generator and discriminator) compete with each other, most often in a zero-sum game, where one's win is the other's loss.

For a given training set, the GAN learns to generate samples with the same statistical properties as it has. They are used as negative training samples for the discriminator. The discriminator learns to distinguish the wrong data generated by the generator from the right one and punishes it for producing inappropriate results. At the start of training, the generator produces false data and the discriminator quickly learns it to be false. As training progresses, the generator produces more and more false data that the discriminator is no longer able to detect, but begins to classify the false data as true.

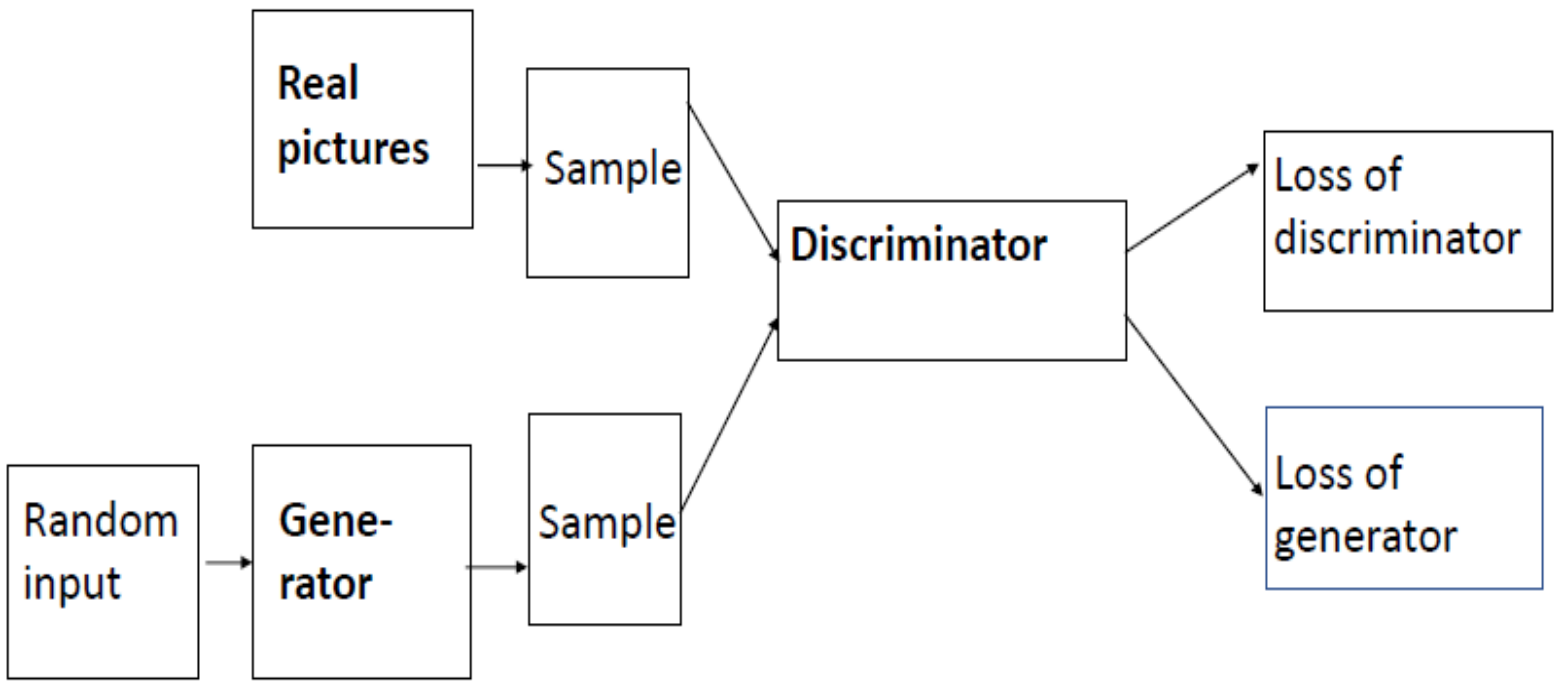

Figure 4.28. Generative Adversarial Network.

GAN trained with photographs is able to create new images that look real to the human eye. In this case, the task of the generator is to generate credible content (images that look natural) and the discriminator evaluates the content (whether the generated image looks natural). Nvidia has used technology based on the GANs to generate new, non-real face images (Paukku, 2018).

The basic idea of the GAN method is based on indirect training using a discriminator, which is also updated dynamically. In this case, the generator is not trained to minimize the distance to a particular image, but to deceive the discriminator. GAN was originally designed as a generative model for unsupervised learning, but it has also proven useful in semi-supervised learning, supervised learning, and reinforcement learning.

## 4.10 Measuring Machine Learning Outcomes

The performance of most machine learning methods deviates from the desires if the training data used is not balanced (Davis & Goadrich, 2006). This means that approximately the same amount training samples should be available for all classes to be identified.

For example, in many medical-related categorization problems, most of the samples available may be normal due to rare occurrences of deviations, in the financial industry there are relatively few credit card frauds for a huge amount of payment transactions, etc.

Thus, most of the data may be so-called negative cases, but an unnoticed problem can be fatal or very expensive. Thus, the cost of misclassification is not the same in all cases.

While access to raw data has become easier with digitalization, data imbalances have become a growing problem. Unbalanced training data is often attempted to be balanced by either under- or oversampling.

In undersampling, a random sample of 10% of the majority classes in the training material is selected for use. If the imbalance is initially in the order of 1:100, balancing can achieve the situation 1:10, but a problem may be the loss of information needed to model the essential data.

Oversampling is used to duplicate randomly selected samples of minority groups, for example in the case of images, by mirroring and rotating, but this may result in so-called over-fitting and over-optimistic impression about the performance of the learned model. Often, for training samples 99% classification accuracy is then achieved, but only for them. With new material, the accuracy may collapse.

Next, assume the problem of classification into positive vs. negative categories, where the positive cases are rare. In a typical application it is not possible to achieve a situation, where the categories are completely separated. In many cases, however, we are able to adjust the decision threshold for the category distributions as illustrated in Figure 4.28.

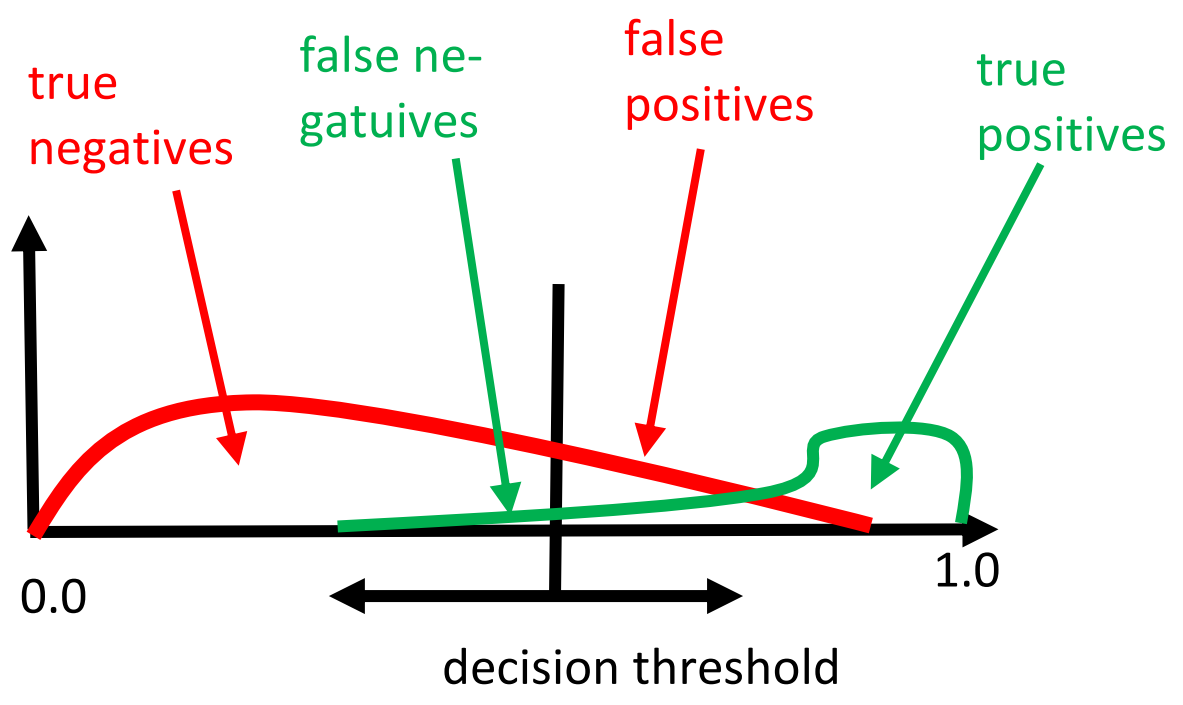

Figure 4.29. Errors in a two category classification problem.

For each value of the decision threshold, we get the

- "true negatives", which are negative samples to the left of the decision threshold;
- "false negatives", which are positive samples but are not classified as negative because they have the features to the left of the decision threshold, and
- “false positives” and “true positives”, which are to the right of the threshold, respectively.

In a balanced case, the result is often represented by a so-called ROC (Receiver-Operating Characteristics) curve, whose name derives from World War II's method of measuring the accuracy of radar observations. Figure 4.30 shows a typical ROC plot. The more area under the curve, the better the result.

The ROC curve measures the proportion of "true positives" to the right of the decision threshold as a function of "false positives" expressed on the same side. The diagonal line represents the result obtained by classifying randomly.

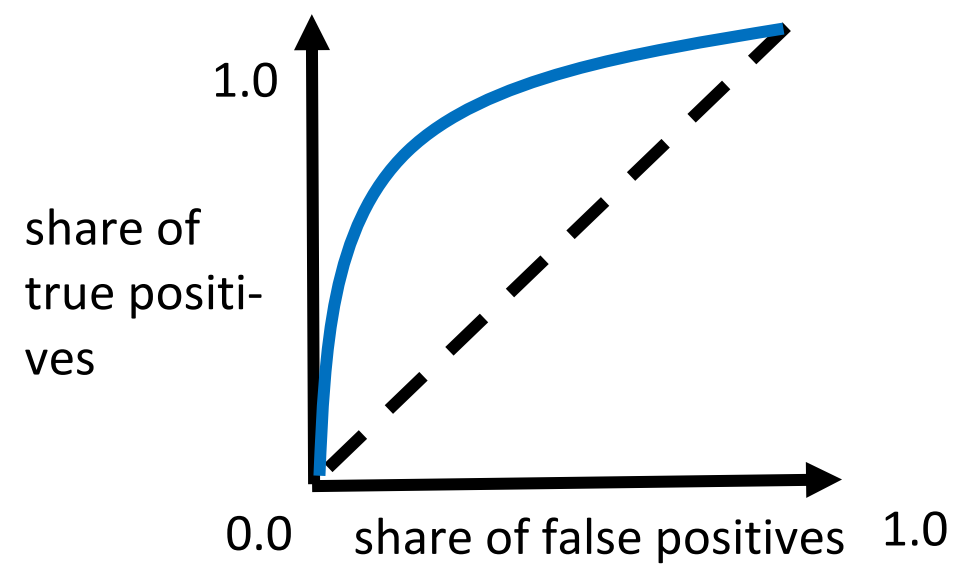

Figure 4.30.Typical  ROC curve.

PR (Precision-recall) graphs are often more illuminating in unbalanced situations where positive samples have a lower proportion of material. Figure 4.31 illustrates the precision-recall results for three different classifiers.

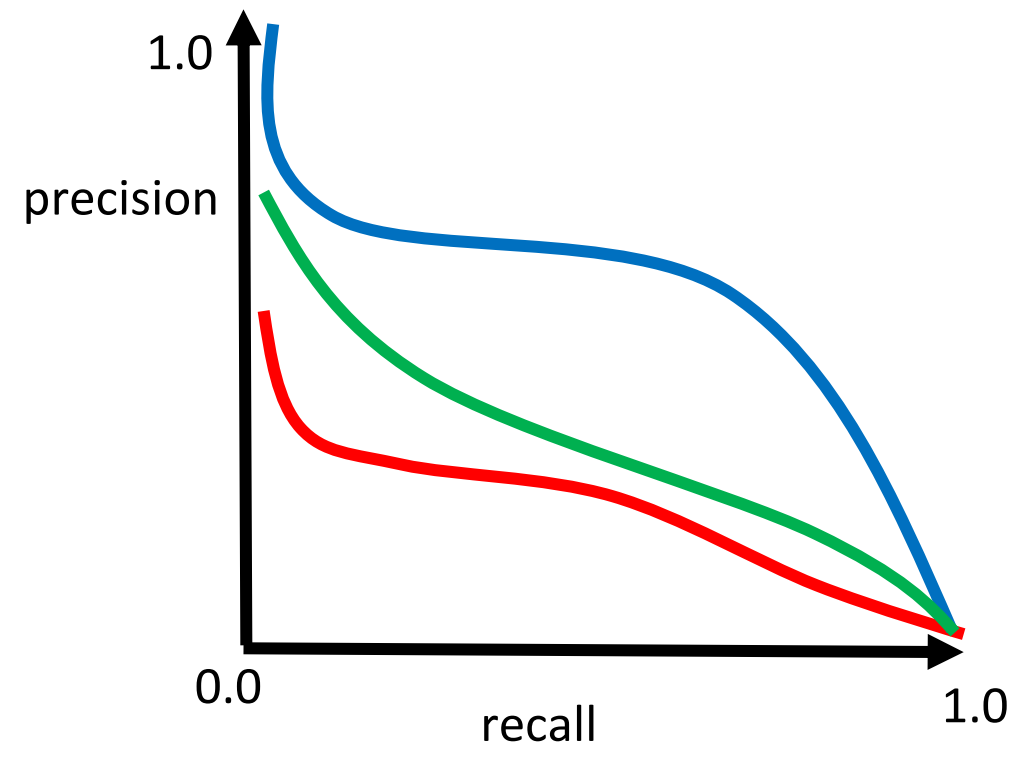

Figure 4.31. Precision-recall curves for three classifiers..

Accuracy measures how much of the results obtained by a given classification algorithm are meaningful, so that relevant information is not buried in the results (Precision).  Recall indicates how much of the relevant information was found in the search.

Because the focus is on the minority category, the "true negatives" representing the majority are not considered.

In the figure the category threshold of each classifier has been lowered to finally obtain all the positive samples. At the same time, however, the accuracy has fallen as the proportion of "false positives" increases. Of course, the lowest of the curves indicates the weakest performance.

Depending on the balance situation, both ROC and PR curves provide important information about the classification performance. Typically, developing an application, you are particularly looking for information on whether the classification is seriously breaking down in any given situation. Alongside this, individual "error" or "accuracy" figures are often less interesting and misleading imbalanced case.

## 4.11 References

Bishop CM (2006) Pattern Recognition and Machine Learning. Springer, 738 p.

Breiman L, Friedman J H, Olshen R A & Stone C J (1984). Classification and Regression Trees**,** Wadsworth & Brooks/Cole Advanced Books & Software, Monterey, CA.

Breiman L (1996) Bagging predictors. Machine Learning 24 (2):123-140

Breiman L (1997) Arcing the edge. Technical Report 486, Statistics Department, University of California, Berkeley CA., 14 p.

Brynjolfsson E & McAfee A (2017) What's driving the machine learning explosion. Harvard Business Review, 18.7.2017.

Chapelle O, Schölkopf B & Zien A (2006). Semi-supervised Learning. Cambridge, Mass.: MIT Press.

Davis J & Goadrich M (2006). The relationship between precision-recall and ROC curves. Proc. 23rd ICML, 233-240.

Duda RO, Hart PE & Stork DG (2001) Pattern Classification, 2nd Edition. Wiley Interscience, 654 p.

Frank E, Trigg L, Holmes G & Witten IH (2000) Technical note: Naive Bayes for regression. Machine Learning 41:5-25.

Goodfellow I, Bengio Y & Courville A (2015) Deep Learning. The MIT Press.

Hochreiter S & Schmidhuber J (1997) Long Short-Term Memory. Neural Computation 9 (8):1735-1780.

Krizhevsky A, Sutskever I &, Hinton GE (2012) Imagenet classification with deep convolutional neural networks. Advances in Neural Information Processing Systems, 1097-1105.

Kurama V (2020) Gradient boosting in classification: Not a black box anymore!

LeCun Y, Boser B, Denker JS, Henderson D, Howard RE, Hubbard W & Jackel LD (1989) Backpropagation applied to handwritten zip code recognition. Neural Computation 1(4):541-551.

Liaw A & Wiener M (2002) Classification and regression by randomForest. R News 2:18-22.

Liu L, Ouyang W, Wang X, Fieguth P, Liu X & Pietikäinen M (2020) Deep learning for generic object detection: A survey. International Journal of Computer Vision 128:261–318. arXiv:1809.02165v4

Matikainen N (2017) Auton arvon aleneminen iän ja käytön myötä. (The decrease of a car's value as a function of it's age and use). MS Thesis, Tampere University, 40 p.

Opala M (2019) Deep learning frameworks comparison – Tensorflow, PyTorch, Keras, MXNet, The Microsoft Cognitive Toolkit, Caffe, Deeplearning4j, Chainer. 10.9.2019.

Paukku T (2018) Kuvan ihmistä ei ole olemassa (The person shown in the picture does not exist). Helsingin Sanomat 27.6.2018.

Peltarion (2018) Deep learning challenge: Lessons from the field, 12 p.

Peng C-Y J, Lee K L & Ingersoll GM (2002) An introduction to logistic regression analysis and reporting. The Journal of Educational Research 96(1):3-14, Taylor & Francis.

Rumelhart D E, Hinton G E & Williams R J (1986) Learning representations by back-propagating errors. Nature 323:533-536.

Russell SJ & Norvig P (2010) Artificial Intelligence: A Modern Approach, Third Edition, Pearson.

Sutton RS & Barto, AG (1998). Reinforcement Learning: An Introduction. MIT Press.

Viola P & Jones M (2004) Robust real-time face detection. International Journal of Computer Vision 57(2):137-154.

Watkins CJCH & Dayan P (1992). Q-learning. Machine Learning 8 (3):279-292.

Zheng Z, Lauritzen JS, Perlman E, Robinson CG, Nichols M, Milkie D et al. (2018) A complete electron microscopy volume of the brain of adult Drosophila melanogaster. Cell, 26:730- 743.

Web-GAN: Overview of GAN

Wiki-Deepsoft: Comparison of deep-learning software

Wiki-GAN: Generative adversarial network

Wiki-XGBoost: XGBoost

Wiki-Gradient: Gradient Boosting.

# 5 Artificial Intelligence Applications

There are very many types of AI applications. A significant part by this time has been associated with machine perception, which allows for the application of pattern recognition and machine learning methods. Speech and image recognition are typical examples.

In various games such as chess and Go (Figure 5.1), artificial intelligence programs have won the best human players. Games usually have a limited number of moving options, randomness plays no or only a minor role, and the operating environment remains unchanged.

Machine learning brings new tools to the analysis of massive data, complementing traditional statistical methods. Areas of application include analysis of the data contained in medical databases and population registers, as well as in business applications.

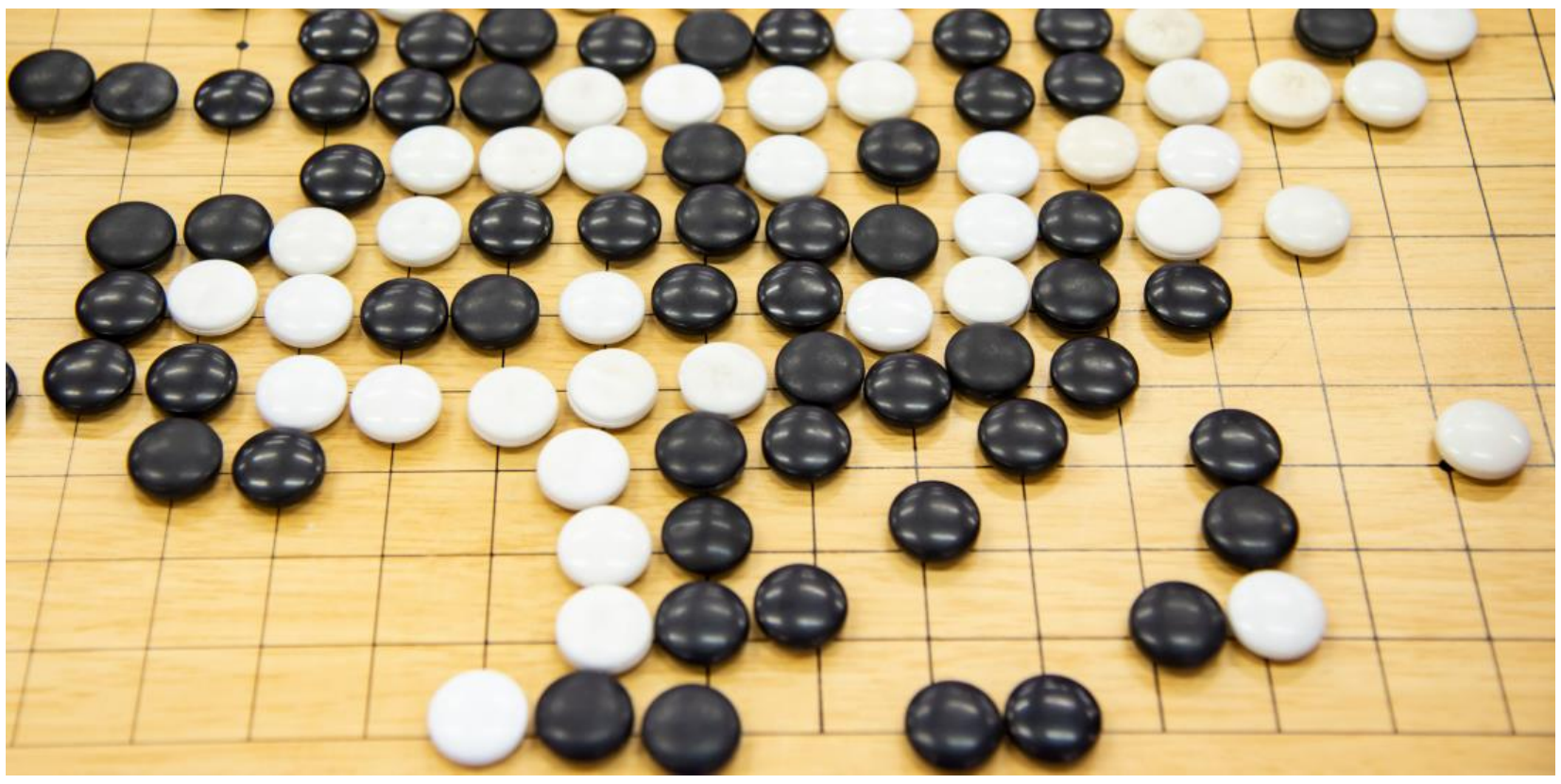

Figure 5.1. In the game Go area is demarcated from the opponent. (© 123RF)

Fully autonomous vehicles are an example of a very challenging situation. The operating environment is only partially observable and constantly changing, the result is also influenced by the actions of other drivers and pedestrians, randomness plays a major role and the number of possible situations is unlimited.

According to Tables 5.1 and 5.2, Russell and Norvig (2010) divide application task environments and their properties into different types using the agent-related principle presented in Chapter 2. However, in some cases, the proposed division is open to interpretation depending on the individual application problem and its solution.

The information produced by the sensor may be wholly or partially **observable** by the agent performing the task. In chess, it is completely discernible, autonomous cars can only partially observe the current operating environment with their sensors.

There is **one agent** performing some tasks, for example, in a crossword puzzle, in image classification, and in speech recognition. There are two players in chess, but in an autonomous car there are **many agents** considering other traffic and pedestrian activities. The same is true for multidirectional dialogue, in which the machine strives for an almost natural conversation with another person or several people.

If the next state of the operating environment is completely dependent on the present state and the next operation, then it is a **deterministic** environment. Chess, as well as image classification or speech recognition by the neural network method, are examples of this. There is **randomness** in poker, backgammon and especially autonomous cars. Table 5.1 illustrates this aspect of categorization.

Table 5.1. Task environments viewed in terms of observability, number of agents and determinism

| **Task environment** | **Observable** | **Agents** | **Determinism** |
|---|---|---|---|
| Crossword puzzle | Fully | One | Deterministic. |
| Chess with a clock | Fully | Many | Deterministic |
| Poker | Partly | Many | Random |
| Backgammon | Fully | Many | Random |
| Speech recognition | Fully | One | Deterministic |
| Image classification | Fully | One | Deterministic |
| Part-picking robot | Partly | One | Random |
| Autonomous car | Partly | Many | Random |
| Multidir.l dialogue | Partly | Many | Random |
| Data analysis | Fully | One | Deterministic |
| Micro-expressions | Partly | One | Random |
| Emotional state | Partly | Many | Random |

The task environment also involves the dynamism of the tasks, the nature of the observations and the periodicity. This is discussed in Table 5.2.

In an **episodic** task environment, the agent's activity can be divided into sequences, each of which comprises detection (perception) and subsequent operation. Image and speech recognition by the classification method are good examples of this. In a **sequential** environment, the current decision can affect all future decisions. Chess, multidirectional dialogue and autonomous cars are good examples.

In a **static** environment, such as a crossword puzzle, an agent can perform a current operation without changing the environment during it. In a **dynamic** environment, the situation is constantly changing, for example in autonomous cars or in multidirectional dialogue. In a semi-dynamic environment, the environment does not change, but the measure of agent performance changes. Chess is clearly semi-dynamic.

Table 5.2. Task environments viewed in terms of episodic, dynamism, and continuity

| Task environment | Episodic | Dynamism | Continuity |
|---|---|---|---|
| Crossword puzzle | Sequential | Static | Discrete |
| Chess with a clock | Sequential | Half-dynamic | Discrete |
| Poker | Sequential | Static | Discrete |
| Backgammon | Sequential | Static | Discrete |
| Speech recognition | Episodic | Static | Continuous |
| Image classification | Episodic | Half-dynamic | Continuous |
| Part-picking robot | Episodic | Dynamic | Continuous |
| Autonomous car | Sequential | Dynamic | Continuous |
| Multidir. dialogue | Sequential | Dynamic | Continuous |
| Data analysis | Episodic | Static | Discrete |
| Micro-expressions | Episodic | Half-dynamic | Continuous |
| Emotional state | Episodic | Dynamic | Continuous |

An environment is **discrete** when a finite number of observations and operations are required to complete the task. Chess and most games are like this. Driving a car is a **continuous** activity. The speeds and locations of both you and other vehicles are constantly changing over time. Also, many of the driver's (agent's) functions are continuous, such as steering wheel and brake positions.

The most difficult applications are those that are *partially observable, with multiple agents, random, sequential, dynamic, and continuous*. Autonomous cars (Section 5.4) and multidirectional dialogue (Section 5.2) are very difficult applications for artificial intelligence based on these features. Also, reliable identification and monitoring of the human emotional state is very challenging (Chapter 9).

This chapter briefly introduces some of the key applications of today's artificial intelligence, except for machine vision, which is covered in Chapters 6-9.

## 5.1 Speech Recognition and Personal Assistants

Significant advances in artificial intelligence and machine learning in recent years have been seen especially in speech recognition. In the past, recognition was too unreliable and background noise made identification impossible. Professor Andrew Ng, a guru of machine learning, has predicted that after raising 95% recognition accuracy to 99%, voice recognition will become our main way of communicating with computers.

This four percent is the difference between disturbingly unreliable and highly usable. Thanks to massive training data and deep learning methods, this distinction is now almost closed. For example, speech recognition on a smartphone has become so reliable over the years that it can be used, e.g., for Google searches. Finnish-language text can be translated into English so well that it is already understandable, although complete translation would require an understanding of the language content. However, there is still considerable room for improvement in translations.

Automatic speech recognition system consists of three main sources of knowledge: 1) acoustic model, 2) phonetic vocabulary model and 3) language model (Fohr et al., 2017). The acoustic model characterizes language sounds, mainly speech phonemes and extra sounds (pauses, breathing, background noise, etc.). The phonetic vocabulary model contains words that the system can recognize with different pronunciation styles. The language model contains information about possible sequences of consecutive words in the language.

Acoustic models include so-called Hidden Markov Models (HMM), which have been the standard method to account for dynamic speech variations. Deep neural networks can be used in conjunction with HMM in acoustic modeling and in the language model to replace the traditional so-called n-gram model, which estimates the next word of a speech or text based on some of the previous words (Figure 5.2).

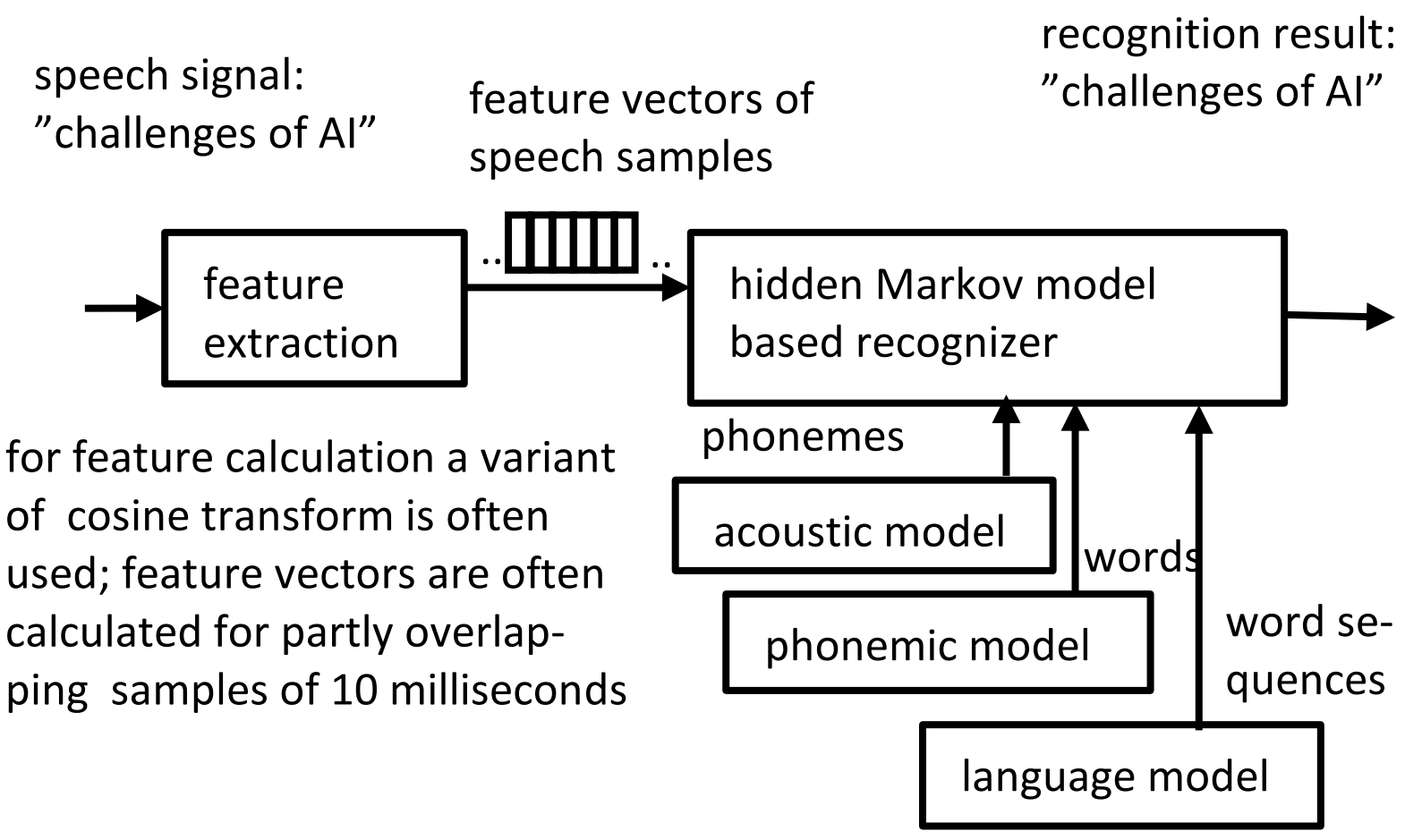

Figure 5.2. Principle of a speech recognition method (n-gram).

In current systems, statistical acoustic models and language models as well as, to some extent, vocabulary and utterance models are estimated using huge audio and text databases. As a result, big Internet companies have had an advantage in development.

Recently, there has been increasing attention to the so-called sequence-to-sequence models. For example, Recurrent Neural

Networks (RNNs) have been used, which are particularly suitable for generating new data or predicting the future state of the system based on past events (see Chapter 4).

The goal is to have fully neural models to replace the three different steps of the conventional method. Training such a system is simpler than the aforementioned three-step method. LSTM (Long Short-Term Memory) models inspired by RNNs use network-based memory cells for temporary storage of data, which makes them better than RNNs for modeling longer sequential data, such as speech or text, because they allow taking into account both short and long term data.

Speech and sound control will have numerous applications, for example, in customer service (Aittokoski, 2018), which is already familiar to us with bank telephone services in relation to what a customer is concerned with. According to Hannes Heikinheimo, CEO of Speechgrinder, a Finnish company that develops Finnish-speaking voice recognition, voice-enhanced customer service applications will include airport check-in, ATMs, grocery shopping, hotel check-in. Also, in cars, voice control is expected to become standard on various functions such as heating or navigation, freeing the driver to concentrate on driving itself.

Thanks to strong advances in speech recognition technology, leading artificial intelligence technology companies such as Amazon, Apple, Google, and Microsoft have developed systems and tools for human-machine voice interaction. The respective products of these and some other companies compete with each other and are often also offered to third parties for their product development.

**Alexa** is Amazon's cloud-based voice service, which is likely to be in use in over 100 million properties. With its tools, application interfaces (APIs), reference solutions and documentation, it is easy to develop voice recognition interfaces for different needs. The Amazon Echo is a very popular voice-controlled speaker that connects to a smartphone via Bluetooth. Other manufacturers have also introduced similar products. The smart speaker allows you to bring home a variety of services with voice control over the cloud and the Internet.

**Siri** is a virtual assistant developed by Apple for its own products that uses voice queries and its natural language user interface to answer questions, make recommendations, and perform various actions via the Internet. **Google Assistant** is a virtual assistant available for mobile devices, smart homes and cars. It can also be used by third parties in their own products. In addition to the natural sound, it is also possible to use the keyboard. The system is capable of searching the Internet, scheduling events and alarms, adjusting a user's device settings, and displaying information from their Google Account. Recent features include capturing and recognizing images taken by the camera of the device, supporting purchases of products and sending money - and song recognition.

**Cortana** is a virtual assistant developed by Microsoft for both Microsoft operating systems and, in part, other operating systems and devices. Cortana can send reminders, recognize speech, and answer questions using Microsoft's Bing search engine.

However, the current assistants still have major shortcomings. The content of the conversation with the machine is very limited, far from normal human-to-human conversation. Emotions also play an important role in human interaction, as noted in Chapter 9. Current personal assistants do not take emotions into account. It is expected that with the development of emotional intelligence, the next generation of assistants will recognize the user's emotional state and take it into account in their responses.

**Project Debater**, developed by IBM since 2012, gives an indication of what is going on. The system is able to argue with an unspecified subject with a master human debater. To do this, it uses a huge amount of articles previously read by the system. It is able to write and speak text-driven data, give a brief description of a given discussion topic, formulate and present a well-constructed speech, even with humor.

The system is also capable of listening to long speech and recognizing the main topics and statements. In addition, it is capable of modeling human-made disputes and dialects (puzzles), and of providing counter-arguments when needed (Wiki-Debater).

In the last demonstration in February 2019, IBM Debater lost to an award-winning human debater on "Should Preschools Be Supported in the United States" (Shankland, 2019). As an advantage over the human debater, the system had access to around 10 billion sentences from news and academic research articles. The weakness was, in addition to the monotone voice, the excessive focus on the system's opening remarks, rather than the opponent's claims, though the listener was left with a real argument.

System’s computing and hardware requirements are far from today's applications. In the future, this technology is intended to be used, for example, to complement human skills and to create more sophisticated human-machine interaction.

Implementations of commercial voice services are built on analysis made in service link centers. They are provided with additional training material by allowing people to interpret unclear cases, which in some cases may violate the privacy of the end user.

## 5.2 Natural Language Processing

Understanding and using spoken and written language, the natural language, is an essential part of human intelligence. People learn a huge amount of new things from childhood through listening to speech, reading texts, and chatting. Natural conversation with the machine should also take place using speech and language - regardless of language. Most Finns would like to talk to the machine in their mother tongue - not English. Translating from one language to another is also a key application area (Figure 5.3).

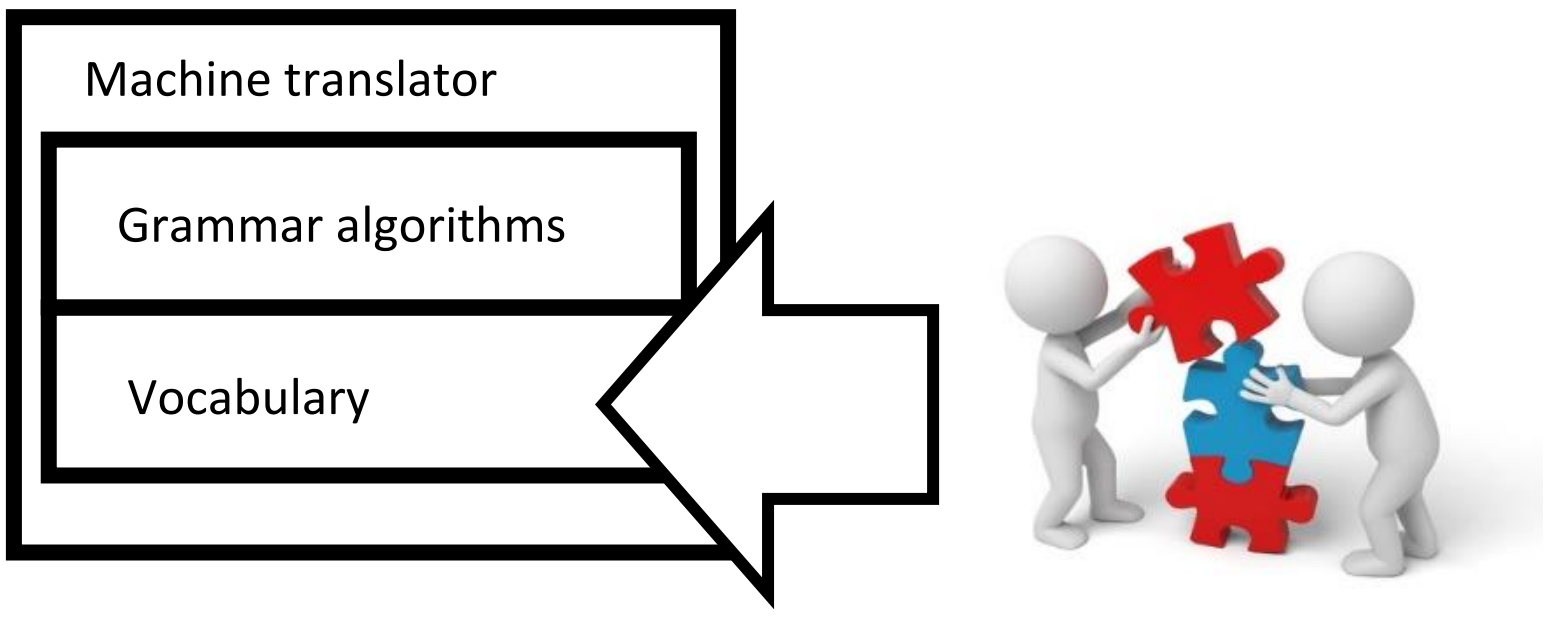

Figure 5.3. Machine translation. (© 123RF)

Natural language processing can be divided into five different tasks: 1) classification, 2) matching, 3) translation, 4) structural prediction, and 5) sequential decision-making (Li, 2017).

In classification, the text is categorized according to its content. Categories can also be opinions, for example, "positive", "negative", "neutral". In matching, strings of texts are matched to see how close they are to each other. This can be applied, for example, in retrieving specific text, answering questions, and search-based one-way dialogue, in which the machine searches the database for the most query-like answers.

In translation, the string is converted to another. This can be applied to language translation, speech recognition, and generic one-way dialogue, where the machine tends to produce new kinds of answers (which may be meaningless!). In structural prediction, a string is described as a structure. For example, the goal may be to find a part of a given sentence, segment individual words, or divide a sentence into semantic parts based on their meaning

The fifth task, the sequential decision-making process, is clearly the most difficult task: it requires actions in the context of a dynamically changing environment. Multidirectional dialogue is one of these. That is, how can we, for example, be able to chat with a personal assistant or an intelligent robot so that the machine understands the content of the conversation and is able to actively participate in it.

Current artificial intelligence based on machine learning is mainly capable of pattern recognition and prediction, but not of abstract or common sense thinking and thus of understanding the meaning of what it recognizes. However, deep learning and the use of massive data have significantly contributed to the study and applications of natural language. Considerable success has been achieved in the above tasks 1-4.

In our daily work, this can be seen especially in machine translators. Google Translate, for example, covers over 100 languages and is already capable of surprisingly good translations when compared to results from a few years ago. Indeed, it is regularly used by foreign researchers in our research group to understand the Finnish text, which is still widely used by university administrators and others. Google's program also helped us in translating this edition of our book from Finnish to English. Complete translation would require an understanding of the content of the text, as would Task 5 in the list above.

## 5.3 Playing Games

The skill of a machine to play chess has been considered as one of the benchmarks when evaluating if a machine is intelligent in solving problems. Machine and human game strategies, especially in the early days of chess development, differed greatly.

The machine's game has been based on evaluating the goodness of the various move options in each game situation (that is, where your own and your opponent's pieces are at that time) by looking through the search tree for the various moving options. The minimax algorithm (see Chapter 3) and pruning away poor-performing transitions are used to reduce the huge search space. The average search tree branching factor is about 35, meaning the player has as many as 35 options for the next move.

In 1997, IBM's DeepBlue winning the world's best player, Gary Kasparov, was able to rate an average of 14 of its own and opponent's move options, ahead of the human players to select the best move. The supercomputer developed by IBM for this purpose was able to estimate 200 million positions per second and its best move estimator already used information from previous games of the Grand Masters.

The game of the best human players, on the other hand, relies heavily on pattern recognition, meaning that the best players remember a large number of previous game openings and game situations, and are able to find the best move without having to go through a wide range of options.

According to an article in Nature magazine (Amidzic et al., 2001), chess grandmasters have learned to remember around 100,000 piece placements on the board, and this is the main reason that sets them apart from regular players. Of course, the best result can be achieved by combining these two strategies. Using deep learning in machine play to take advantage of situations in previous games has been a natural next step.

In 2011, IBM's Watson Artificial Intelligence program won people in the popular TV Jeopardy quiz. This achievement was considered a milestone in artificial intelligence even before the deep learning breakthrough.

In 2017, DeepMind's Alpha Zero program, founded by Google, won the best chess computer software of the time, Stockfish 8 (Paukku, 2017).

In 2016, the same company's AlphaGo Zero won the best human players in the Go board game (Paukku, 2018). Before this, artificial intelligence textbooks (Russell & Norvig, 2010) reported that machines do so poorly in the Go game that the best human players don't bother playing against the machines. The slightly more general-purpose AlphaZero of the same company was able to control chess and Japanese-style shogi, along with Go (Figure 5.4).

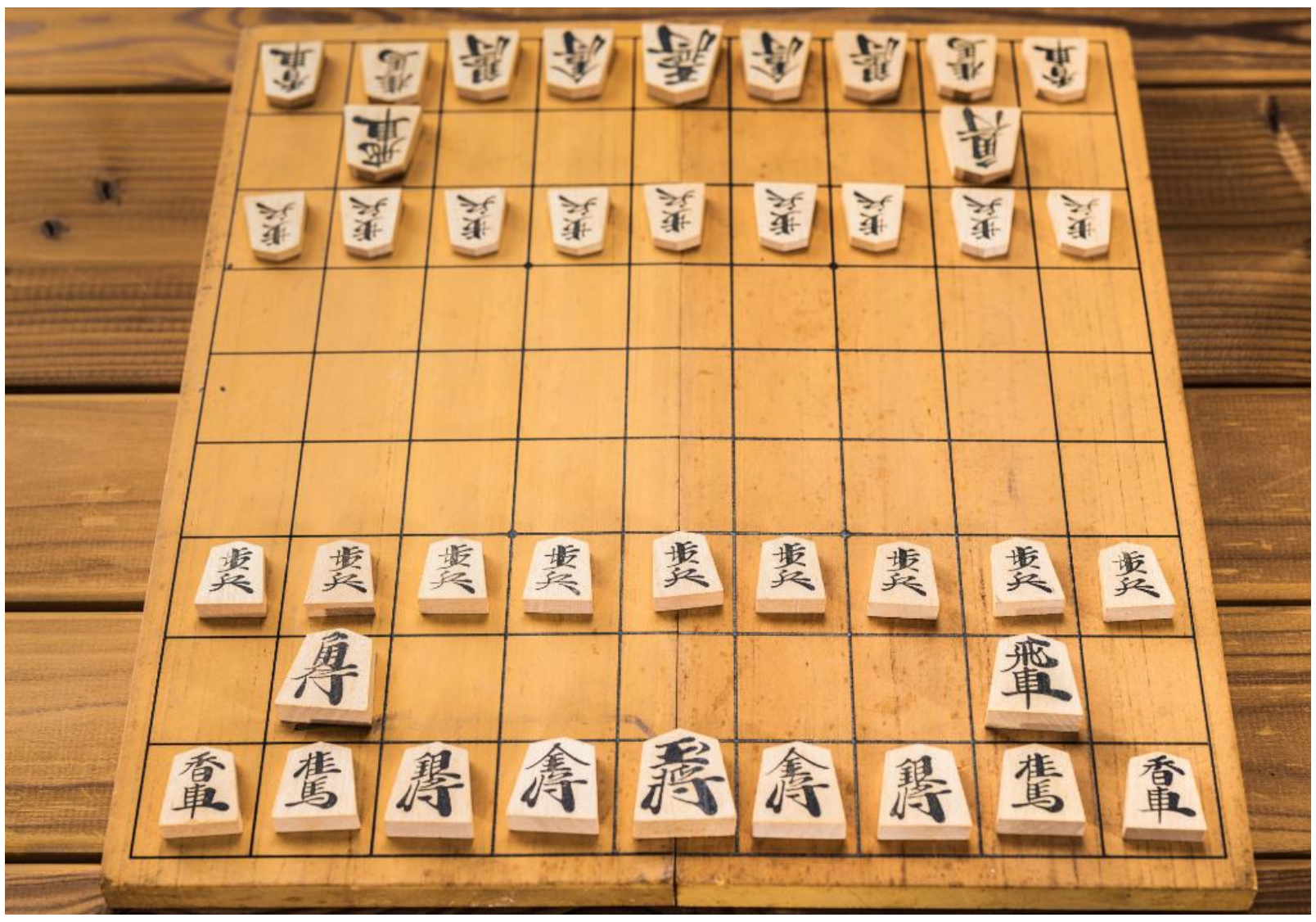

Figure 5.4. Japanese shogi, or general's game. (© 123RF)

The algorithm used by AlphaGo includes machine learning and tree searching. It has been developed through training against both human and machine players. The method is based on Monte Carlo tree search, which is guided by an "evaluation network" and an "operating model network". The latter recommends different game modes, the following of which are evaluated by the rating network. Ultimately, AlphaGo chooses the game mode that is most successful in the simulation.

DeepMind's machines are highly specialized and require a huge amount of computing and training, and their use in real-world applications has not yet been demonstrated. Developed in 2017 at the University of Alberta, Canada, the DeepStack program wins a human in poker, where coincidence plays a huge role (Riley, 2017).

## 5.4 Self-driving Vehicles

The development of self-driving vehicles has been a long time coming. At the time of Matti Pietikäinen's one-year postdoc visit to the University of Maryland in the mid-1980s, there was an Autonomous Land Vehicle project underway on the large DARPA (Defense Advanced Research Agency) strategic computing research program. The aim of the project, coordinated by Martin Marietta, was to make the (military) vehicle run independently, for example on the highway. utilizing for example computer vision (Ake, 1986).

The difficulty of the problem became clear, as already monitoring the roadside in changing circumstances was very challenging, not to mention more complicated tasks. A clear improvement was made a little later when German professor Ernst Dickmanns applied the Kalman filtering, familiar from control theory, to roadside tracking (Wiki-Dickmanns).

Today, self-driving cars can be considered as one of the flagship projects of artificial intelligence (Figure 5.5). Huge sums are being invested in them and it is believed that in the coming years such will be driving in normal traffic.

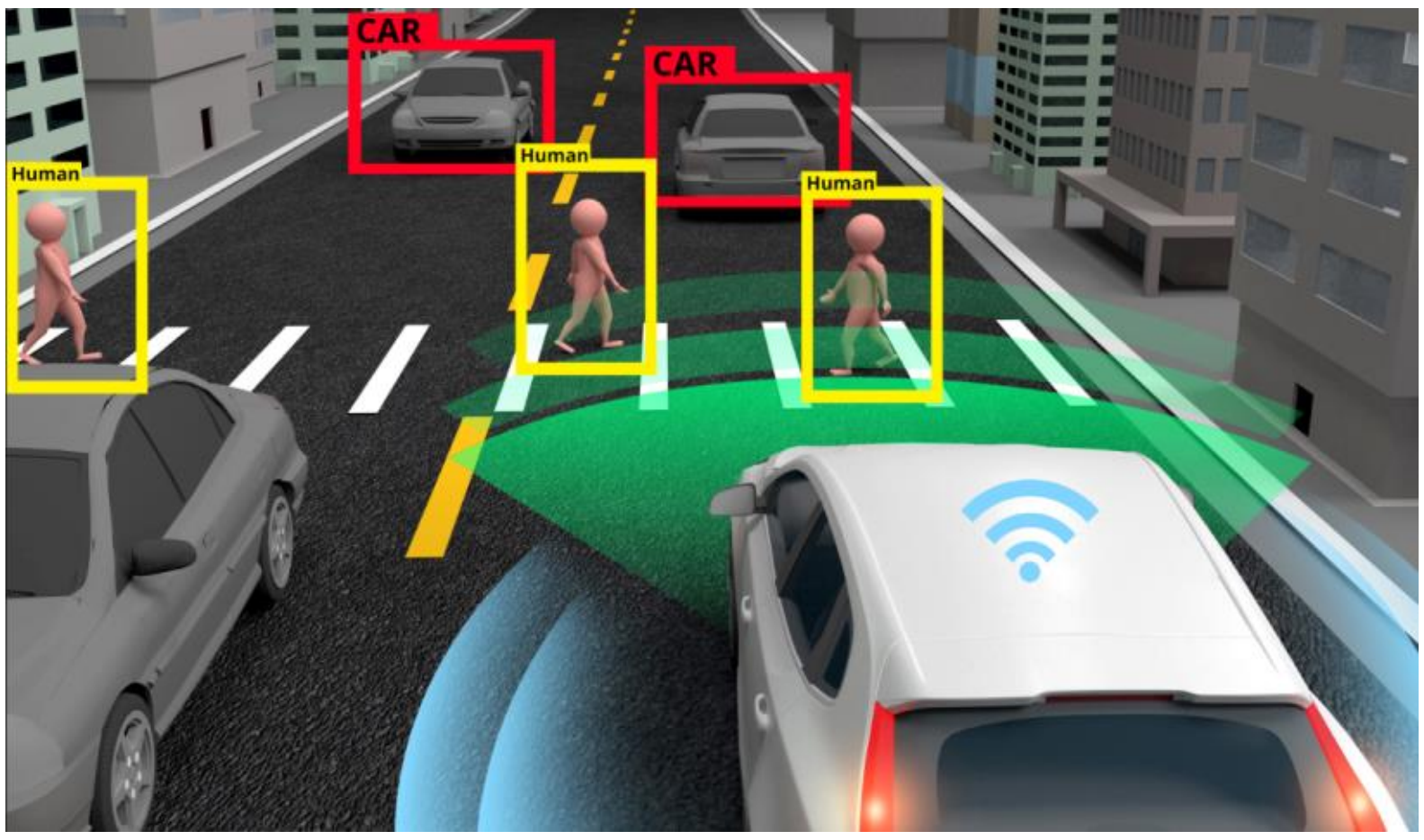

Figure 5.5. Self-driving vehicle. (© 123RF)

Experiences of Google, Uber, Tesla and many automakers have led to successful test drives in limited conditions. Human assistive technology has been developed to stay on the highway and prevent crashes. Inside car cameras can monitor the driver's fatigue. Navigators help you find your desired destination. Road environments have been imaged, e.g., with Lidar radars (Light Detection and Ranging).

However, there are enormous challenges. Fluctuations in the environment and lighting cause major problems in obtaining reliable sensor data. Solutions developed do not necessarily work very reliably in a changing environment, such as in Finland.

How do we get cars to reliably detect pedestrians or animals coming in, predict their actions, and react quickly? How to make autonomous and people-driven cars work flexibly in one traffic? How can such a highly complex technical system be operated with sufficient reliability?

Alongside the technical challenges, there are ethical issues. Who is responsible in the event of an accident? The limits of deep learning are well reflected in this application. How can a system be taught so that it can respond correctly under all possible circumstances? How can it's reliability be guaranteed?

The big problems with machine sensing are well illustrated by an Iltalehti news report from a March 2019, concerning a report from the Tencent Keen Security Lab, which investigated the reliability of the Tesla autopilot by applying unobtrusive labels to the road surface. They made the car turn to the oncoming lane! “According to researchers, Tesla's lane detection works well under normal conditions, but dazzling light, rain, snowstorm or other distractions such as sand and dust cause difficulties” (Känsälä, 2019), (Tencent, 2019).

In August 2017, Professor Rodney Brooks from the Massachusetts Institute of Technology, a pioneer in the development of intelligent robots, wrote an interesting article in the IEEE Spectrum magazine, "The Self-driving Car's People Problem". He looks at problems ahead and predicts how a complete autonomy may be reached sometimes, but only through five different levels of difficulty (Brooks, 2017).

Among the most difficult problems he sees are: understanding the actions of pedestrians, including their abnormal behavior in poor weather conditions such as snow-covered roads, and the fact that a self-driving car can, in its own interest, cause harm to human-driven cars. Another reason for his skepticism is that if automation of mass transport such as trains is not yet fully automated to date, then how is it possible to automate the much more difficult car traffic?

According to Brooks, the five different levels of difficulty are:
1. People still do most of the car's functions, but some functions like steering or acceleration can be done automatically.
2. At least one driver assist system is automated, eliminating the need for the driver to physically control the car (hands off steering wheel and feet from pedals simultaneously).
3. The driver transfers some safety-critical functions to the cars in certain traffic or weather conditions.
4. Fully autonomous vehicles perform all safety critical functions in specified areas and in specified weather conditions.
5. Only at this level full human replacement in all circumstances may be achieved.

Ethical concerns have already emerged in a couple of recent events that will surely slow down the development and deploy-ment of technology, one of which has been the accident involving a Uber self-driving car in Arizona in which a pedestrian died (NTSB, 2018). The reason for this was a misinterpretation of the information related to the observations made by the radar and the Lidar radar: first, the pedestrian who was walking with a bicycle was identified as a car and then as a bicycle, thereby incorrectly predicting the pedestrian's path. The system would eventually have proposed emergency braking, but it was not allowed in autopilot, nor was it reported to the human operator in the car.

Other manufacturers, such as Nvidia, Tesla, and possibly Google, are also known to have encountered problems. One solution is remote control from control rooms in case of problems, but in this case it is no longer possible to speak of fully autonomous driving cars. Nissan is said to have opted for this and others are rumored to be considering it (Davies, 2018).

In fact, we believe it is realistic to reach level 4 mentioned by Brooks over time, i.e. fully autonomous operation in confined environments where pedestrians and cars are sufficiently separated or otherwise safe. Level 5 seems impossible for long in the future.

## 5.5 Big Data Analytics and Data Fusion

Due to the enormous increase in data through digitalization, big data has become a business-critical resource. A large amount of computing power and various tools for analyzing data are required to utilize the data. Businesses can use data to improve their core business, but also to create completely new business models.

Mass data analytics is already a well-established technology for processing and retrieving such information. Necessary functions include: editing and refining raw data for analysis, visualizing data for the user, and various statistical methods for extracting relevant data and making predictions based on the data collected. Artificial intelligence tools aim to automate this process and enable the analysis of massive amounts of data.

Data analysis methodologies can be divided into three types:

1. *Data analytics* refers to evaluating data based on past events,
2. *Predictive analytics* makes assumptions and tests based on historical data to predict future events, and
3. *Artificial intelligence / machine learning* analyzes data, makes assumptions and produces predictions on a scale that human analysis alone cannot achieve.

One advantage of computers over humans is the ability to retrieve and recall vast amounts of information that can be accessed quickly, for example, via the Internet. The machine does not get tired and can work 24 hours a day. If the data is of sufficient quality, the machine can outperform human performance in limited machine learning tasks.

Often too little attention is paid to the collection, naming, or annotation, and modification of good quality data for data analysis or machine learning methods, although the performance of the methods is largely dependent on it. You must be able to choose the right tools for each application. Most of the time that goes into data analysis takes time to edit and clean up the data for automatic analysis. Typically, this takes at least 80% of the time of the entire analysis process (Maikkola, 2019).

Business applications are an important application area for data analysis and artificial intelligence. The problem with applying machine learning to business databases is that most of them are too dirty or sparse in terms of machine learning. So, for the time being, there are few real machine learning applications compared to normal business analytics. Prediction is not required in many applications, as statistical reasoning and analysis of variances are sufficient in 80-90% of cases according to Dominic Ligot (Ligot, 2018).

In addition, data interpretability is usually more important in business applications than absolute accuracy. For this reason, traditional statistical methods are still dominant compared to black-box methods of neural networks. However, due to the hype of artificial intelligence, many methods implemented by conventional methods are referred in media as artificial intelligence, although they actually are not. The greatest benefit of artificial intelligence in business applications may therefore come from intelligent assistants who only help the user to make a decision.

Due to its speed and other capacity, the machine is also able to combine data from different sources in an unprecedented way. A good example is a recent article on a study published by Nature Human Behaviour, which found that human happiness is diminished by air pollution (Junttila, 2019). This conclusion was reached by analyzing the air quality of 144 cities in China and the number of emotion-related social media responses in these cities.

Sina Weibo's Twitter-like social media tweeds were used to help this research. The AI was used to analyze as many as 210 million social publications over a six-month period, in which the user had allowed the use of location information.

Daily human emotional expressions or likes are unreliable, but over a longer period of time they provide reliable information about the user's average emotional state. It is unclear to the reader what is artificial intelligence here and what is simply combining huge amounts of data from different sources using very fast computation and data analysis.

Countless applications of this kind can be found and the data used for either good or bad purposes. Internet giants in particular, such as Google, Facebook, Amazon and similar Chinese companies, collect huge amounts of information from their customers based on, for example, web, social media and shopping behavior, and can use this information to predict individual behavior, for personalized marketing of products etc. If combined information includes in addition to social media behavior, for example, a person's Internet searches, shopping and health information, it is possible to create dangerously accurate profiles of people.

Jaron Javier, a pioneer of the Internet, regards Facebook and Google's business models as shaping people's behavior, which goes much further than ordinary advertising (Ahlroth, 2019). “They are pushing our free will and our happiness as well as our ability to empathize. They are destroying the truth. In the end, they also take our soul.” The algorithms developed by these companies have not been published.

Computational models using advanced statistical mathematics and machine learning provide users with both “carrot” and sometimes “stick” - thus gradually changing their behavior to the benefit of businesses. (Ahlroth, 2019). Javier has written his thoughts in a recent book, "Ten Arguments For Deleting Your Social Media Accounts Right Now” (Javier, 2018).

## 5.6 Medical Applications

The application of artificial intelligence in medicine has aroused great interest and the potential applications are countless. A general trend in research in the field over the past 30 years has been the shift from knowledge-based approaches to data-driven methods (Peek et al., 2015).

Improving the quality of services and increasing security are key issues that artificial intelligence is hoped to help (Karlen, 2018). Professor Walter Karlen (ETH Zurich), in his article, gives an example of a Chinese TV show in which 25 cancer doctors competed against an artificial intelligence program for cancer diagnosis and prognosis based on brain analysis. Artificial Intelligence defeated the experts 2-0.

In his view, the Chinese have a clear lead in the use of artificial intelligence systems to support clinical decision-making, whether in the form of diagnostic tests or the implementation of large-scale hospital-wide systems.

According to Professor Karlen, doctors are bad at evaluating multidimensional data at the same time, that is, combining information from different sources - and they often tend to interpret things negatively. On the other hand, machines are bad for interpreting contextual situations, or for dealing with situations of uncertainty. Thus, at all times, machines are dependent on high quality and large amounts of data.

Both people and machines make mistakes! However, if well-functioning automated next generation artificial intelligence clinical decision support systems are to be put into practice, they must meet high medical standards and performance requirements, as well as local cultural, ethical and regulatory issues, and be cost effective and based on a sustainable business model (Karlen, 2018).

Interpretation of images generated by imaging equipment and other measurement data has become central to many medical diagnoses. For example, it may be the interpretation of images produced by microscopy, X-ray equipment, ultrasound camera or magnetic resonance imaging, or various physiological signals (ECG, EEG). For example, magnetic resonance imaging produces increasing numbers of clip images each year (Figure 5.6).

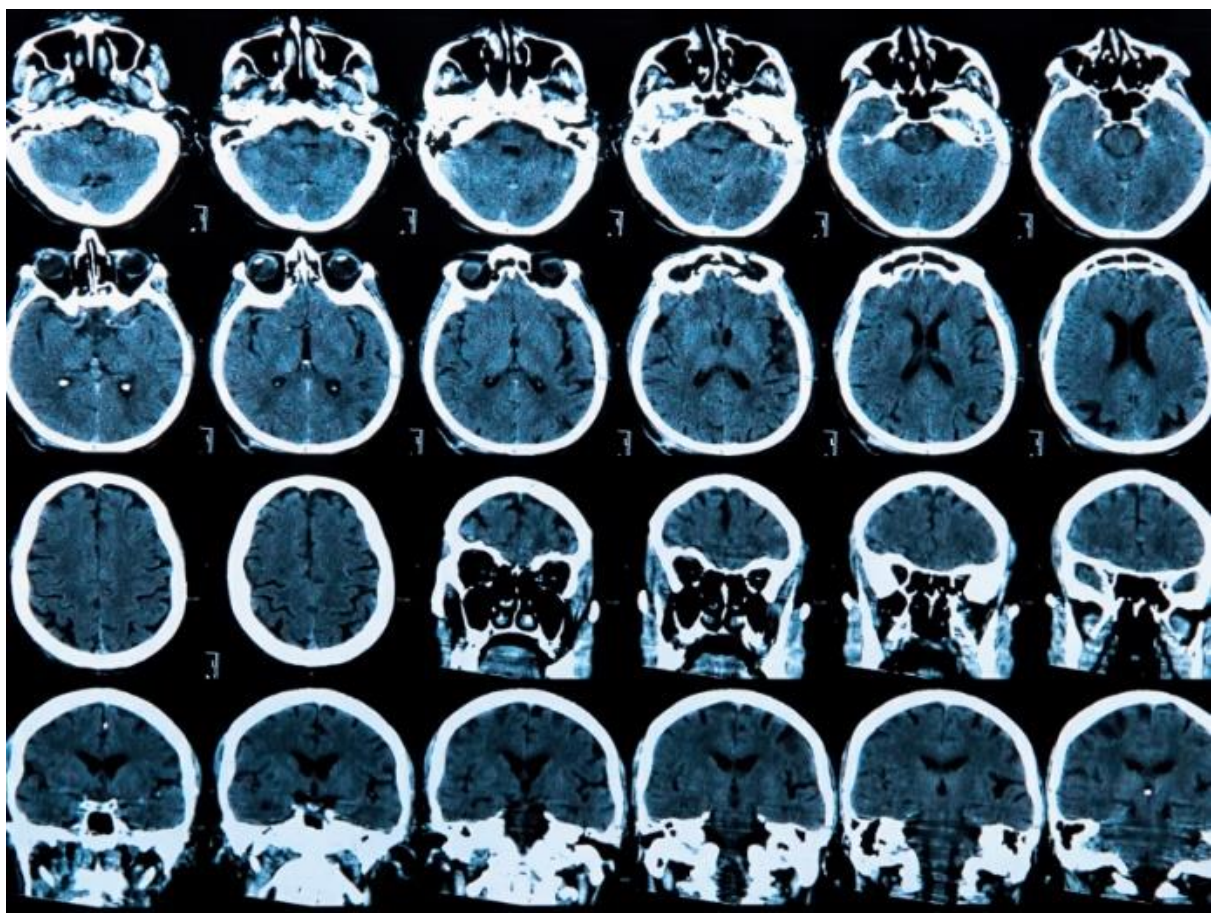

Figure 5.6. MRI scans of the human brain. (© 123RF)

Artificial intelligence and machine learning are hoped to assist physicians in interpreting such massive data - and perhaps even in the future - to automatically diagnose diseases and make drug recommendations based on them. The development and popularization of medical imaging techniques has led to a tremendous increase in the amount of data.

Section 8.5 outlines the problem of interpreting lung X-rays discussed in our study: How to identify healthy lungs with potentially ill lung images using the CNN deep neural network method. This information could be used to distinguish suspicious cases already in the health center, before a detailed examination by a very busy radiologist in the hospital. About 80% accuracy was achieved in the experiments, but raising it to an acceptable level would have required a much larger amount of training data.

This "simple" example demonstrated that the acquisition of annotated, i.e. named, training data is central to the application of deep learning methods to highly variable image data. As technology continues to evolve, imaging equipment is improving and old training data may not be very usable with new material.

On the other hand, creating annotated data is very laborious, as an example: which lung images are from a healthy person and which images may show symptoms of disease. There is not enough time for application experts such as radiologists in this case. Another major challenge is that it is difficult in medical applications to accept machine diagnoses that cannot be substantiated by a neural network. Thus, we believe that the interpretation of images based on machine vision and deep learning can only assist a physician in making diagnoses, not substitute a physician.

According to Chief Physician Päivi Ruokoniemi, artificial intelligence will bring added value to the automation of health care processes. Data mining in patient records can provide essential diagnostic information from hundreds of pages of patient records for use by the physician, as well as for other applications similar to normal physician practices (Ruokoniemi, 2018). She is very skeptical about the use of artificial intelligence based on deep neural networks in disease diagnosis and decision-making on drug therapies. "At its worst, artificial intelligence can recommend treatments that are unrelated to the patient's diagnostic status." This is related to the fact that current artificial intelligence can only find correlations, not causal relationships between, for example, a drug and its expected effects.

Her suspicions are supported by a publication in the prestigious Science journal looking at the vulnerability of machine learning methods to adversarial attacks containing false training data, and their relevance to medical applications (Finlayson, 2019).

A significant problem in the application of machine learning, especially to medical problems, is also seen in the fact that the solutions presented in research publications have mostly been tested with incomplete data and their results cannot be replicated in a real application environment (Scudellari, 2021). For example, according to an analysis by researchers at the University of Toronto, 85% of the publications on machine learning for COVID 19 detection from lung images did not meet reproducibility or quality requirements, and none of the presented machine learning models were ready for clinical use! Three ways are proposed to improve the situation: 1) to form joint teams of machine learning and medical experts, 2) to use high-quality data and know its origin, 3) to adopt standards for scientific conference communities that require the use of high-quality data and sufficiently comprehensive reporting in publications.

According to Tapio Seppänen, professor of medical information technology at the University of Oulu, "there are many products in the health industry that make data analysis and conclusions based on this". In practice, the development of such artificial intelligence is a long-term effort, as product compliance verification and medical validation take a long time. It is imperative to verify that the product is delivering what it promises and that it has medical benefits in patients' care.

Figure 5.7 shows the skin abnormality. It is ultimately necessary for an expert to determine whether it is a harmless mole or, for example, melanoma. However, the machine can make a preliminary assessment.

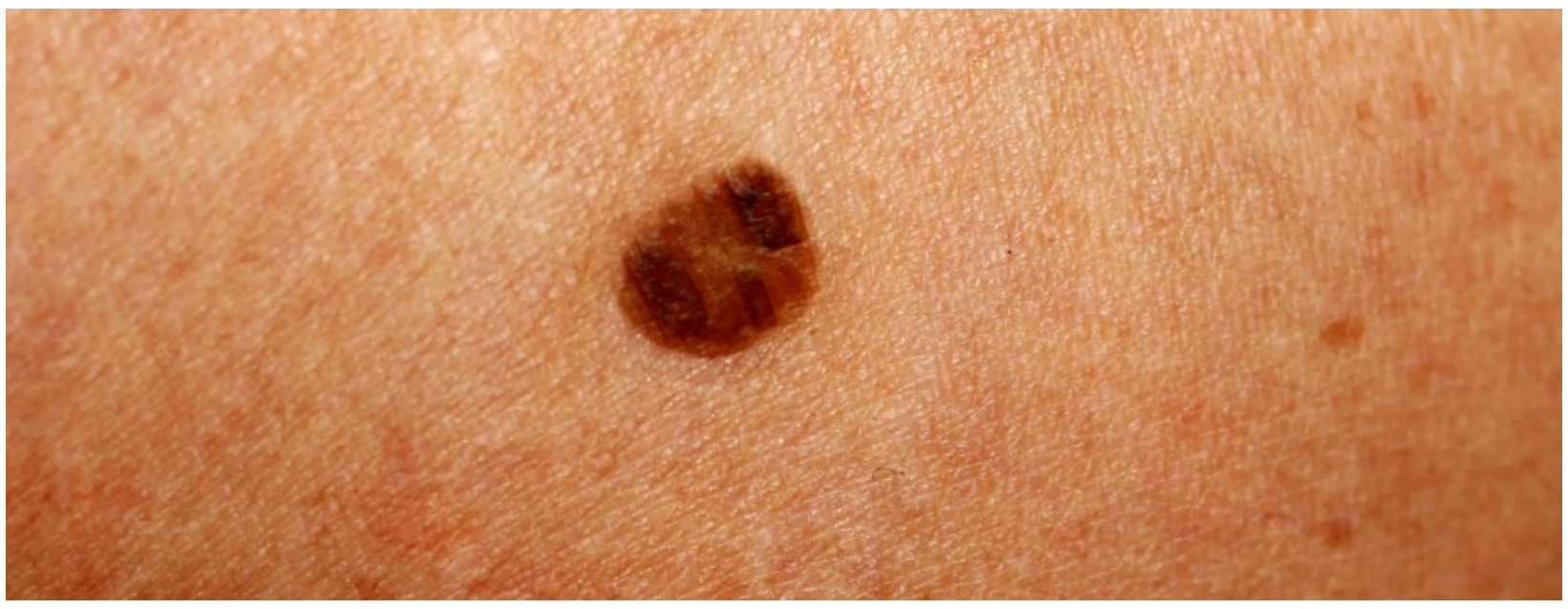

Figure 5.7. A dark area to be diagnosed on the skin. (© 123RF)

Because of liability issues, such solutions do not automatically make diagnostic decisions or suggest treatment measures, but serve as decision support systems for healthcare professionals. Their job is not (at least for a long time) to replace the doctor, but to act as his work tool. This also avoids the risk of making a lot of wrong decisions, because the machine is not perfect here, if not a human being.

Obviously, medical applications of artificial intelligence involve ethical issues, for which recommendations are discussed, among others, in the reference (Luxton, 2014).

## 5.7 Audiovisual Video Content Retrieval

Content-based image retrieval has been studied since the 1990s. The search system is given the desired query image. The system then searches the database for the most resembling images retrieved from the database and arranges them according to how close they are to the query image. For example, if a feed image shows Donald Trump, the system will search for images that resemble him the most.

The problem here is how well the content of the images can be represented by selected computationally simple features to allow accurate and fast retrieval. Typically, in the early days simple features related to color and texture and to the structure of objects were used. In a content-based video retrieval, a desired video clip is fed to the system and then the system searches the database for the most resembling ones. Instead of just showing images, one must also be able to model motion, for example, walking people, moving cars, etc. Deep learning solutions have made it possible to find a much more accurate search for both images and videos.

In many cases, audio provides significant additional information beyond just image and video information. For example, look for places in a movie where a certain person talks. This is a search for *audio-visual video content*. The rapid development of speech recognition technology in recent years (see Section 5.1) allows us to reliably recognize speech in different languages with video and convert it to text if desired.

An exciting EU-funded H2020 research project, MeMAD (https://memad.eu), was underway in 2018-20 to develop computationally effective, reusable and multi-purpose methods for analyzing multilingual audiovisual content. The aim was to revolutionize the processing of video collections and digital storytelling in the production of television programs and other media (MeMAD, 2018).

For example, the method developed in the project enables the audio-visual content of the video to be converted into text - directly into the desired language. This could revolutionize the way large video collections are handled by media companies, for example, and allow for the reprocessing of previously produced material and its application for new purposes. The solutions are planned to make it easier for the general public, and especially the hearing and sight-impaired, to find the material.

"The new discovery, accessibility and personalized service experience provided by artificial intelligence will continue to be vital success factors for the European media industry. Understanding content is one of the most promising areas of artificial intelligence at the moment," says Anssi Komulainen, Senior *Vice President, Innovation Strategy* at Yle (a Finnish public service media company) (MeMAD, 2018). The project involved four actual research partners and four companies or similar: Aalto University, University of Helsinki, University of Surrey, EURECOM, Yle, Lingsoft, Limecraft and INA. The project was coordinated by Professor Mikko Kurimo, Department of Signal Processing and Acoustics, Aalto University.

Audiovisual video analysis is also the focus in the Oulu-based company Valossa, which has developed a platform and methodology for advanced video content analytics (https://valossa.com).

## 5.8 References

Ahlroth J (2019) Addiktion algoritmi hallitsee meitä (An addiction algorithm is ruling us). Helsingin Sanomat 24.1.2019.

Aittokoski H (2018) Pian puhut koko ajan koneille (Soon you will speak all the time to machines). Helsingin Sanomat 23.9.2018.

Ake D (1986) Engineers at Martin Marietta put their $17-million experiment in UPI Archives 12.6.1986

Amidzic O, Riehle HJ, Fehr T, Wienbruch C & Elbert T (2001) Pattern of focal gamma-bursts in chess players. Nature 412(6847):603.

Brooks B (2017) The self-driving car's people problem. IEEE Spectrum, 32-35, 47-49.

Davies A (2018) Self-driving cars have a secret weapon: Remote control, Wired 2.1.2018.

Finlayson SG, Bowers JD, Ito J, Zittrain JL, Beam AL & Kohane IS (2019) Adversarial attacks on medical machine learning. Science 363 (6433):1287-1289.

Fohr D, Mellas O & Illina I (2017) New paradigm in speech recognition: Deep neural networks. Proc. IEEE Int. Conference on Information Systems and Economic Intelligence, 7 p.

Javier J (2018) Ten Arguments for Deleting Your Social Media Accounts Right Now. Bodley Head, 160 p.

Junttila J (2019) Puhdas ilma lisää onnellisuutta (Clean air will increase happiness). Helsingin Sanomat 24.1.2019.

Karlen W (2018) Is medicine ready for artificial intelligence? MedicalXpress, 6.9.2018.

Känsälä S (2019) Vaarallinen tarratemppu paljastui - voi ohjata auton vastaantulevien kaistalle (A dangerous sticker trick was found – it can drive a car to the oncoming lane). Iltalehti 3.4.2019.

Li H (2017) Deep learning for natural language processing: Advantages and challenges. National Science Review, 6 p.

Ligot D (2018) What is the brutal truth about machine learning? Quora Digest, September 2018.

Luxton DD (2014) Recommendations for the ethical use and design of artificial intelligent care providers. Artificial Intelligence in Medicine 62:1-10.

Maikkola M (2019) Datayhteiskunta on täällä (Data society is here). Kaleva 1.9.2019.

MeMAD (2018) Tutkijat opettavat tekoälyn kuvailemaan videoita (Researchers teach the AI to describe contents of videos). https://memad.eu

NTSB (2018) Preliminary Report Released for Crash Involving Pedestrian. National Transportation Safety Board.

Paukku T (2017) Googlen tekoäly AlphaZero opetteli hetkessä shakkineroksi (Google's AI AlphaZero learned quickly to become a genius in chess). Helsingin Sanomat 14.12.2017.

Paukku T (2018) Ihmisen go-lautapelissä voittaneen tekoälyn piti olla totta ehkä vasta vuonna 2035 (The AI that won a human in the go board game was not supposed be true before 2035). Helsingin Sanomat 18.10.2018.

Peek N, Combi C, Martin R & Bellazzi R (2015) Thirty years of artificial intelligence in medicine (AIME) conferences: A review of research themes. Artificial Intelligence in Medicine 65:61-73.

Riley T (2017) Artificial intelligence goes deep to beat humans at poker. Science news 3.3.2017.

Ruokoniemi P (2018) Tekoälyä on käytettävä harkiten lääketieteessä (AI should be used with care in medicine). Helsingin Sanomat 28.10.2018.

Russell S & Norvig P (2010) Artificial Intelligence: A Modern Approach, 3rd Edition. Pearson, 1152 p.

Schankland S (2019) IBM's AI loses to a human debater, but it's still persuasive technology. Cnet News 11.2.2019.

Scudellari M (2021) Machine learning faces a reckoning in health research. IEEE Spectrum 29.3.2021.

Tencent (2019) Experimental security research of Tesla autopilot. Tencent Keen Security Lab.

Wiki-Debater: Project debater

Wiki-Dickmanns: Ernst Dickmanns

# 6 Computer Vision - The Engine for AI Research

## 6.1 General about Computer Vision

The use of sensory information is central to the development of machines with similar intelligent properties as humans. Compared to other senses, vision is particularly important: man receives most of the information about his environment through visual perception. An estimated 30-50% of the brain is said to be used to process visual information.

Vision is influenced by a complex mechanism consisting of three main components: the environment (or object) to be observed, the lighting, and the observer. Three different schools of science study visual perception. Neurophysiologists seek to understand the function of the sensory and neural mechanisms of biological systems. Observational psychologists study psychological factors related to perception. Computer vision scientists are exploring computational, algorithmic, and technical problems associated with image acquisition, processing, and automatic interpretation.

The overall goal of computer vision is to make the machine see and understand what the view represented by the camera or other sensor contains and utilize this information in a variety of applications (Ikeuchi, 2021). For example, the machine must be able to identify objects and determine their positions and orientations, create a three-dimensional model of objects or imaged environment, detect changes in objects, and interpret the significance of different observations. The section on traditional machine vision found in Sections 6.1 and 6.2 is largely borrowed from Matti Pietikäinen's chapter in Encyclopedia of Artificial Intelligence (Tekoälyn ensyklopedia) or his other contributions (Pietikäinen, 1993).

Usually, automatic interpretation of the image is a very demanding task. Each application has its own specific requirements and universal methods have not been successfully developed. Changes in lighting, viewing direction and background, as well as other variations in the operating environment generally cause great difficulty in interpreting the described scene.

In the simple case, the images to be analyzed are two-dimensional, so that no depth information is needed to interpret them. Analyzing three-dimensional views is significantly more demanding because objects look different from different viewpoints. They may also partially overlap. In addition, objects or the camera, or both, can be in motion relative to one another, which not only causes additional problems but also facilitates the understanding of the content of the view.

As a discipline, computer vision is very challenging, interesting and multidisciplinary. Computer vision related problems are studied by computer experts, mathematicians, physicists, electronics and systems engineers, human vision psychology researchers and more.

In the scientific community, this field is commonly referred to as *computer vision*. Here mathematical theory, algorithms, and often also the connections to human perception play a central role. *Machine vision* is often referred to as vision for applications. A system engineering approach that takes into account the requirements and limitations of applications, imaging, algorithms, system architectures, hardware and software, interfaces, etc. plays an important role in this.

Computer vision issues are also being explored in many other areas. The closest related disciplines include digital image and signal processing, pattern recognition - and recently machine learning has played a very important role.

Imaging has a very important role for most applications. Figure 6.1 shows examples of different cameras in different applications. Cameras can operate like the human eye in visible light, but also in the dark. Instead of using a standard image sensor, ultrasound or X-rays familiar to medical applications - or radar for terrain and cloud imaging - can be used for image acquisition.

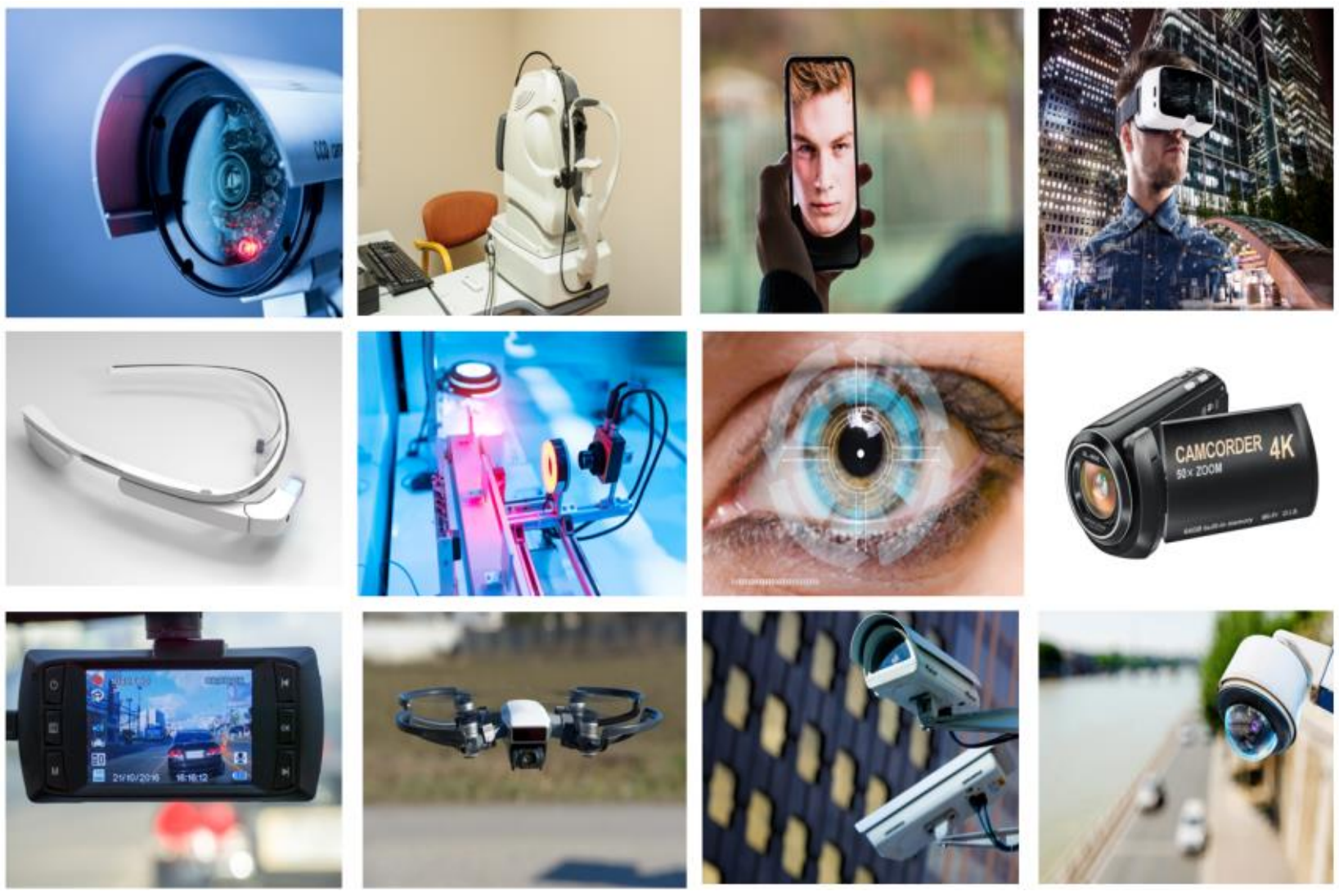

Figure 6.1. Examples of different cameras. (© 123RF)

## 6.2 Steps of the Traditional Computer Vision Process

### *6.2.1 Simplified image analysis process*

The following describes the different steps of a typical image analysis process used in the simplest machine vision applications. In this context, it is assumed that the view to be analyzed is a two-dimensional single image, so that there is no depth information or processing of video sequence to interpret the objects in the image. Figure 6.2 shows the main stages of the analysis in a block diagram (Pietikäinen 1993). When, for example, convolutional neural networks are used for the same purpose, the diagram becomes different (see Chapter 4).

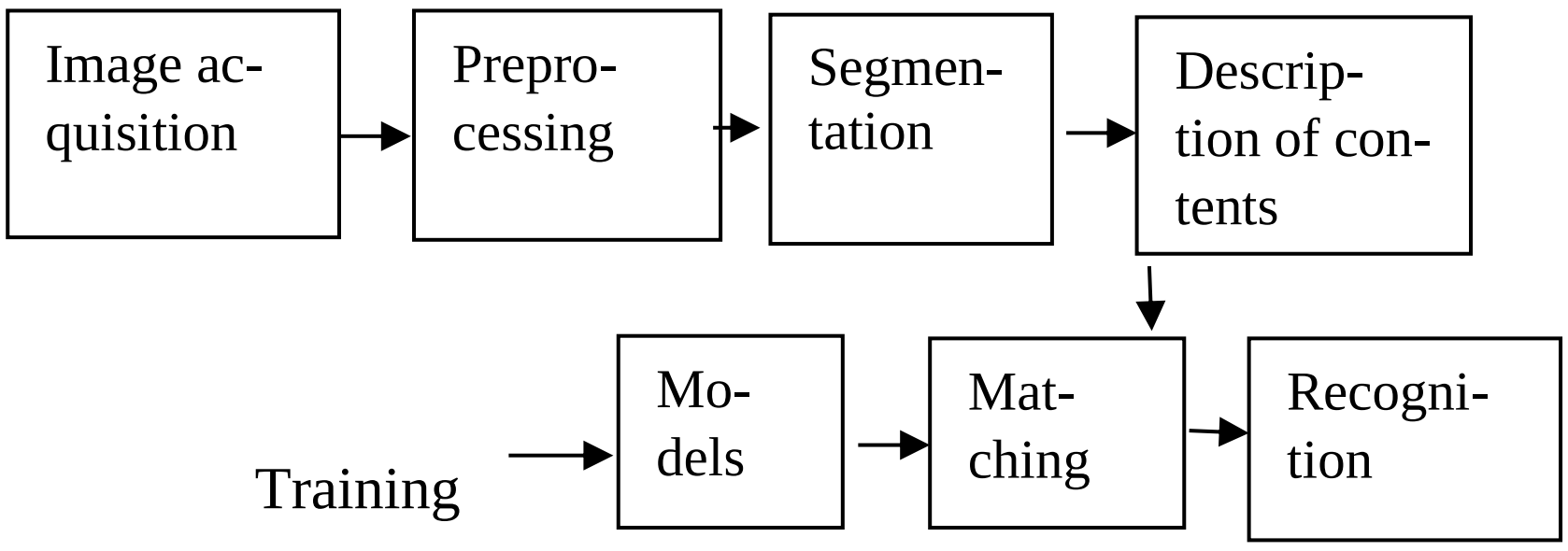

Figure 6.2. Simplified image analysis process.

The first step is to create an image with an image source, such as a standard digital camera or documentary scanner. In many applications, image analysis can be crucially facilitated if lighting and other imaging arrangements are carefully designed.

Depending on the application, a digitized image may have a very variable number of pixels. Typically, there are 512 x 512 to 2048 x 2048 pixels. Often grayscale images are quantized, for example, the tone scale is quantized to 256 levels. Similarly, a color image often uses 3 x 256 quantization levels, or 256 levels for each color component (e.g., RGB, red, green, and blue) to determine the pixel color.

A video image sequence consists of sequential images, i.e. frames, typically taken at 25 frames per second, but in many cases a higher speed may be required. For example, the micro-expression recognition in Chapter 9 required a speed of 100 frames per second for best results.

The next step is image preprocessing, whereby the image is converted into a format that is more advantageous for analysis by digital image processing. For example, the image can be normalized so that fluctuations in lighting and other environmental factors do not unduly influence the final result. Many times, the image is designed to filter out noise or other gray-level changes that interfere with the analysis, and to highlight features of interest. In addition, geometric correction of the image may be required, for example, due to distortion caused by the camera lens or to match images taken at different times.

Next, the image is segmented: the goal is to separate the objects and parts of objects from each other and their background. Traditionally, two alternative principles have been used for segmentation: region-based methods divide an image into homogeneous regions of gray scale, color, etc., and in edge detection sharp changes in color, i.e., the edges of regions, are detected.

Figure 6.3 shows an example of character segmentation by a method based on the use of adaptive grayscale threshold (Sauvola & Pietikäinen, 2000).

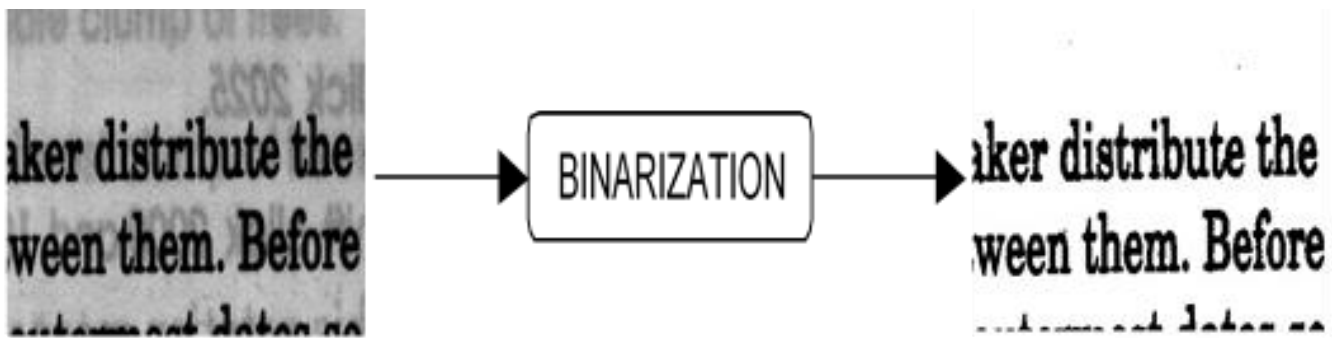

Figure 6.3. Character segmentation by adaptive threshold method. (© Elsevier)

Generally, segmentation is one of the most critical steps in image analysis, and segmentation methods vary depending on the application. Poor segmentation complicates the analysis and may even render it impossible, for example, adjacent characters might not have been separated in the image above.

After that, features describing the properties of the segmented areas, edges, etc., are computed, on the basis of which different objects can be distinguished. Such features include, e.g., shape, color, and texture describing the gray-level properties of regions.

Frequently, the object to be recognized consists of multiple segmented regions or edge segments. In this case, in addition to the properties of the regions (edges), information on the interrelationships between the regions (edges) is needed in describing the structure of the object. For example, networks (graphs) are used to describe a structure, whose nodes represent areas (edges) and links their interdependencies.

The generated representations are compared with the prototype target models taught to the system, thereby seeking to identify objects in the image or to detect deviations from the models. The objects in Figure 6.3 are typed letters, numbers, and other characters.

In the simplest case, the objects can be described with sufficient precision by global features describing the shape of the individual segmented regions or other properties (e.g., average gray level of regions, gray level variance, area, circumference, shape complexity). Different objects can then be identified using these features by pattern recognition methods or a neural network.

Generally, however, objects consist of multiple regions or edge segments, whereby, for example, they need to be grouped together to obtain features suitable for the classifier or to use so-called structural pattern recognition, such as a graph representation (see Chapter 3), to describe the properties of different regions and their location relative to each other.

### *6.2.2 Acquisition of three-dimensional information*

Analyzing three-dimensional views is considerably more difficult than two-dimensional cases. Feature detection and segmentation are complicated by, e.g., variations in illumination due to shadows and dependence of surface brightness on surface orientation.

It is difficult to calculate the properties used for recognition because objects look different in different directions and many things measured in the image are dependent on the viewing direction. It is also difficult to identify if there is only one side of the objects visible or the objects may partially overlap.

Analyzing three-dimensional surfaces usually requires information about the topography of the surfaces in the image. For example, information about the normal orientation of the surface corresponding to each visible point is useful. This kind of orientation information makes it possible to indicate important orientation edges, i.e., points at which the surfaces change direction rapidly. This then facilitates image segmentation so that the result is consistently smooth surfaces.

In order to determine the orientation of 3-D surfaces, so-called. shape from X technologies have been developed. They determine the shape of the surfaces (that is, the orientation of the surface at different pixels) by using different types of shape cues. These include, e.g., methods based on the use of shape from shading, shape from texture, and shape from contour / shape.

Generally, it is not possible to uniquely determine the orientation of surfaces using only one image. The problem can be alleviated by using two or more cameras in known locations, i.e., so-called stereo imaging. By finding matching points in different places in different images, the distances of these points from the camera system can be determined by the triangular measurement principle.

Depth information can also be obtained from motion by looking at the sequence of video images formed by several consecutive images as the camera moves, even if there is no accurate information about the locations of the camera at different times. Along with the depth, you get information about the motion that happened. Such methods based on the use of a normal camcorder are called passive because they do not transmit energy to the scene to be analyzed.

Active distance acquisition methods can often greatly facilitate analysis. Such solutions have already long been available, e.g., methods based on laser pulse travel time measurement and the use of structural light. It is possible to obtain depth information directly from objects by forming a so-called range image (depth map), where each visible point is valued by the distance of the point from the camera. An example of a widely used distance-measuring device in machine vision is a Kinect camera based on the use of active infrared light (Wiki-Kinect).

Figure 6.4 shows an example of a range image taken with a Kinect camera and a color image of the same subject (Herrera

Castro et al., 2011). In the range image, different distances from the camera are shown in different colors.

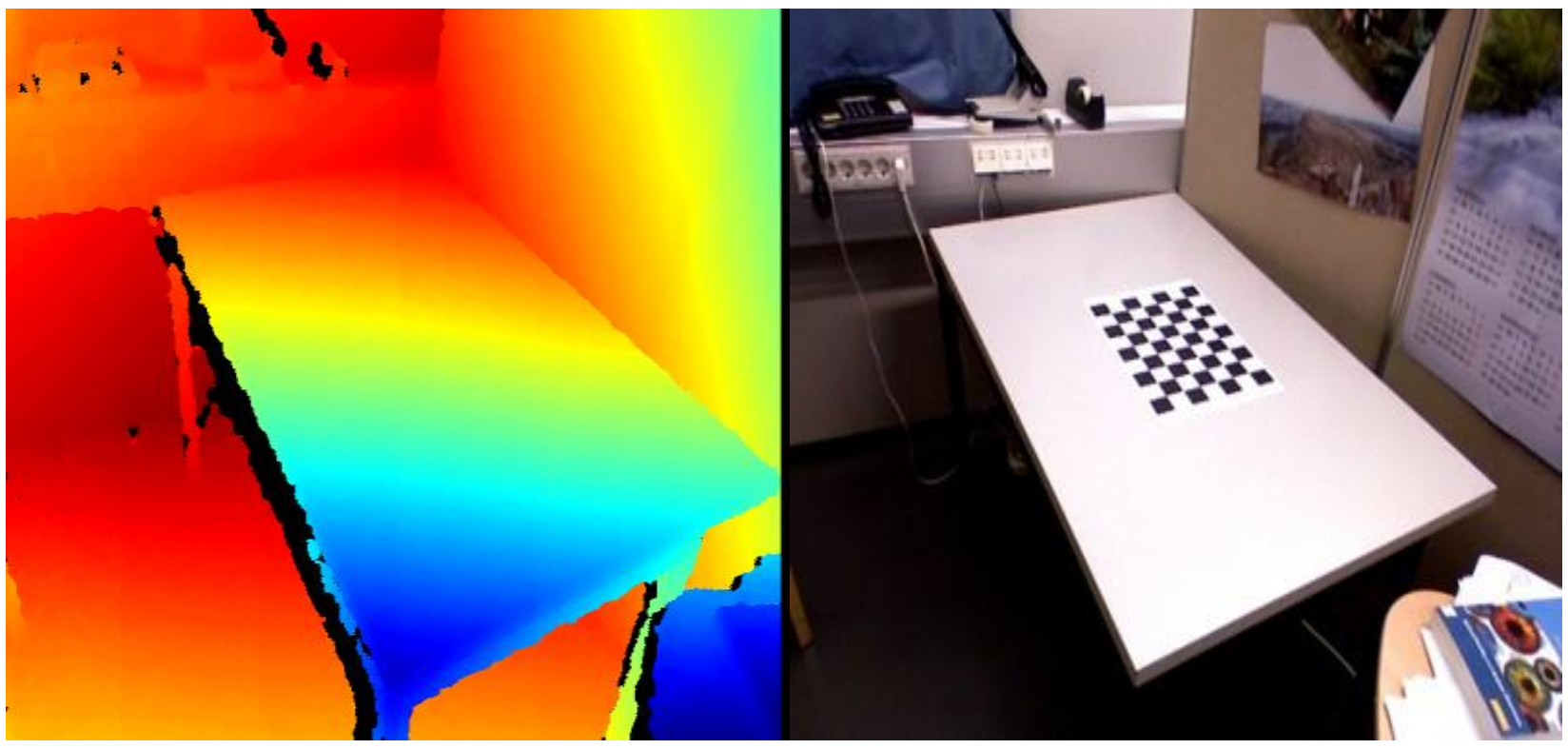

Figure 6.4. A range image and a normal color image. (© Springer)

Reprinted, with permission from Springer [Herrera Castro D, Kannala J & Heikkilä J (2011) Accurate and practical calibration of a depth and color camera pair. In: Computer Analysis of Images and Patterns, Proceedings, Lecture Notes in Computer Science 6855]

Depth cameras have greatly facilitated the analysis of 3-D images. They are now also coming to mobile devices. Google's Tango has been the first mobile platform to incorporate a range camera (Intel RealSense) and other sensors. It produces both depth maps and color images (RGB-D). It also has pretty good odometry, the ability to calculate distance based on information from motion sensors.

### *6.2.3 Analyzing 3-D views*

It is generally assumed that man organizes his knowledge of the environment as abstract concepts, for example, "house", "wall", "animal", "bear". There is an enormous gap in the interpretation of the image between such abstract concepts and the camera image's iconic presentation. The transition from image data to symbolic representation must be accomplished through intermediate steps.

Figure 6.5 shows a traditional approach consisting of a lower, intermediate and upper levels. This is partly based on Professor David Marr's conception of the three steps that human beings use when analyzing 3-D scene images (Marr, 1982), Section 3.4.

At the lower level, the image is preprocessed, segmented, and two-dimensional (2-D) features depicting target areas or edge segments are computed. This sub-system is called "low-level vision" or "early vision".

The primary function of the intermediate-level system is to detect and analyze three-dimensional (3-D) features. Three-dimensional information can be derived from the features computed at the lower level, or obtained directly by the methods developed for the acquisition of the 3-D information mentioned above.

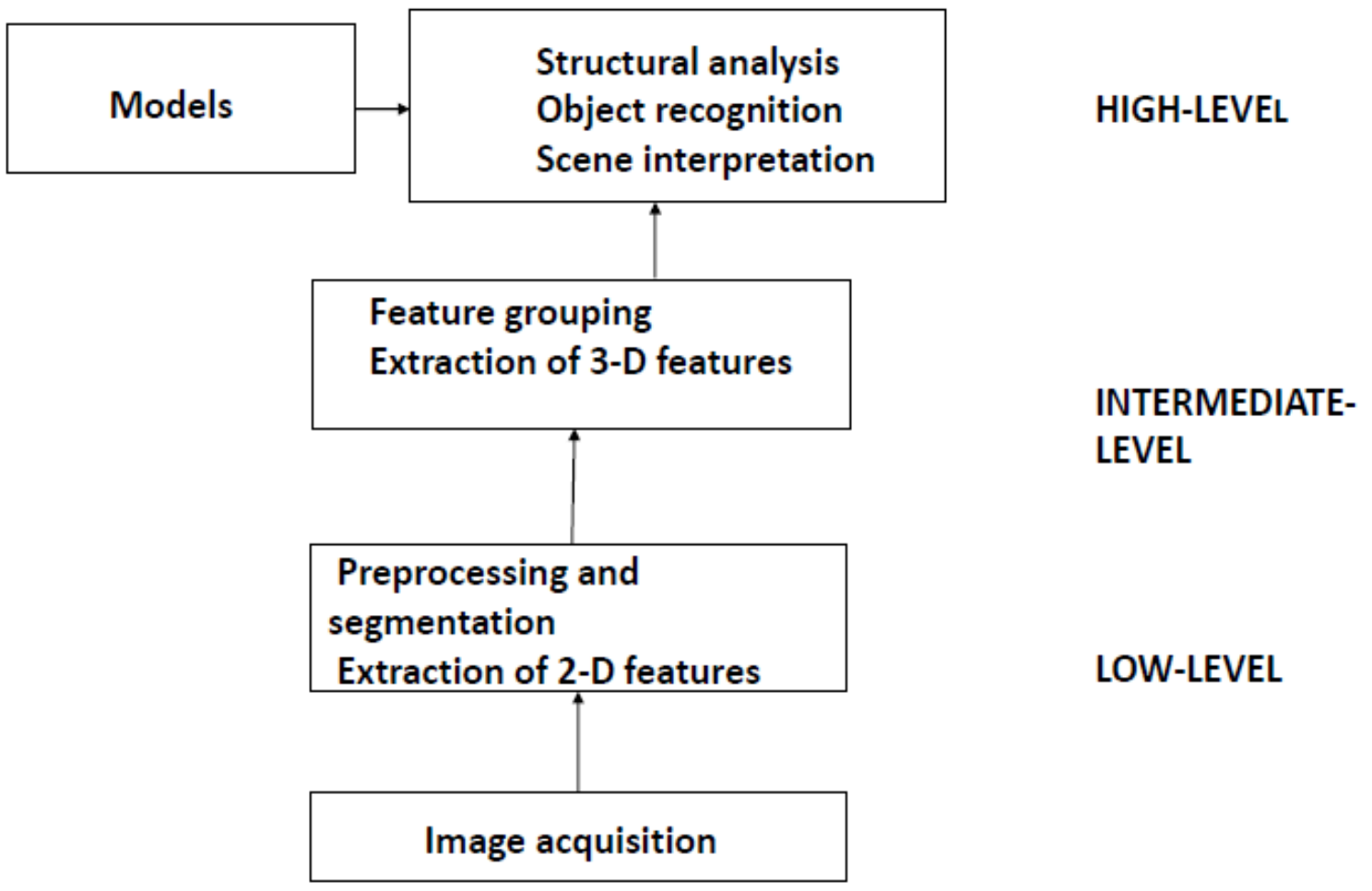

Figure 6.5. 3D view analysis process.

The task of a high-level vision system is often to find the models of previously known or learned objects that best fit with the structures represented in the image. Problems are caused by the appearance of the 3-D objects depending on the viewing location and the lack of two-dimensional image data, especially if the objects overlap. In addition, objects belonging to a particular category (e.g., "house") can vary greatly in appearance

## 6.3 Current Research Areas

Despite the numerous applications implemented, the field of computer vision is still research-focused. The solutions used in the applications have been largely customized. So far, there are only a few universal methods and solutions.

Mathematical modeling of visual information in real-world problems is often overwhelmingly difficult. It is particularly difficult to develop methods that work under changing conditions, for example, when lighting and camera positions vary. The high-speed requirements of applications also often impose special requirements on the methods being developed.

Figure 6.6 presents the major challenges associated with identifying objects in natural environments. The partial problems include (a) illumination, (b) deformation, (c) scale and viewpoint, (d) size and pose, (e) clutter and occlusion, (f) blur, (g) motion, (h) different instances of the same category, and (i) small interclass variations (Liu et al., 2020).

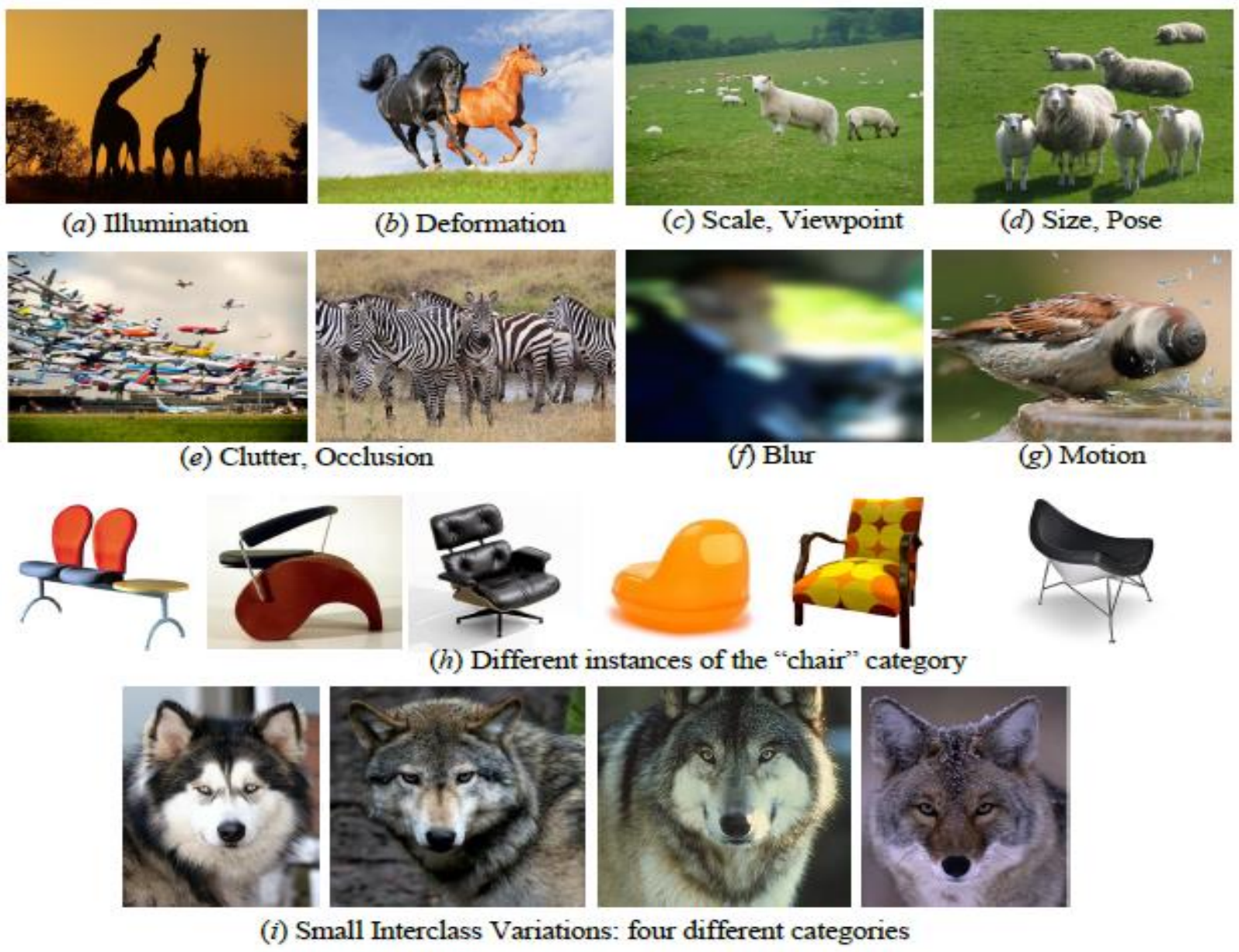

Figure 6.6. Challenges of object identification. (© CC BY 4.0)

From [Liu L, Ouyang W, Wang X, Fieguth P, Liu X & Pietikäinen M (2020) Deep learning for generic object detection: A survey. International Journal of Computer Vision, 58 p.], arXiv:1809.02165v4. This work is licensed under a CC BY 4.0. http://creativecommons.org/licenses/by/4.0/.

In order to achieve good classification accuracy, a huge amount of training samples is required to account for any variations that may occur. One way to reduce these fluctuations (and thus the number of training samples required) is to develop object representations that can withstand a wide range of variations, such as being reasonably invariant with respect to illumination and object size and position variations.

Since 2012, there has been a clear paradigm shift in computer vision research. The simplified solution models presented in Section 6.2 were often replaced, at least in part, by models based on deep neural networks.

Indeed, the application of deep convolutional neural networks has produced excellent results in many computer vision problems, such as character recognition, object detection and classification, face recognition, vehicle and pedestrian location, and so on. Solutions based on deep learning have been as the mainstream in scientific conferences - and due to the results obtained the interest on machine vision applications in industry has clearly grown.

However, the need for massive training data and computational complexity often limit the application of deep neural networks in practice. For example, for applications in industrial visual inspection, medical image interpretation, and micro-expression recognition it may be very difficult or impossible to find a sufficiently comprehensive range of training samples. Typically, the material available or readily available is unbalanced, with the most interesting rare deviations being underrepresented.

Interpreting three-dimensional information with deep neural networks is also particularly difficult, as obtaining sufficient training material is very difficult. This can be concluded from the many factors affecting interpretive data presented in Section 6.2.

Of special concern are applications that require very compact hardware implementation and low power consumption. These include smartphones, wearables, smart watches and glasses.

In scientific publications, classification accuracy or some related measure are often the most important and almost the only criterion for the performance of a new computer vision method. However, there are many other factors to consider when implementing applications and choosing methods.

For instance: What are the requirements for the given application? How many training samples are available and at what cost? What is the cost, size and power consumption of the system being developed? How user-friendly the system is? How can the system be updated and adapted to new environments?

Listed below are some timely key topics in a leading computer vision conference (ICCV 2017):

1. 3-D computer vision
2. Identification of actions
3. Big data and large scale methods
4. Biometrics: face and gestures
5. Analysis of biomedical images
6. Computational photography, photometry, object shape from various clues
7. Deep learning
8. Low level computer vision and image processing
9. Motion and tracking
10. Optimization methods
11. Identification: detection, categorization, indexing and matching
12. Robotic vision
13. Segmentation, grouping and representation of shapes
14. Statistical learning
15. Video: events, activities and video surveillance
16. Computer vision for something (X)

The following are typical problems that have recently been addressed in basic computer vision research.

Figure 6.7 illustrates the identification of objects (human, car) and segmentation of their instances by a convolutional network

.

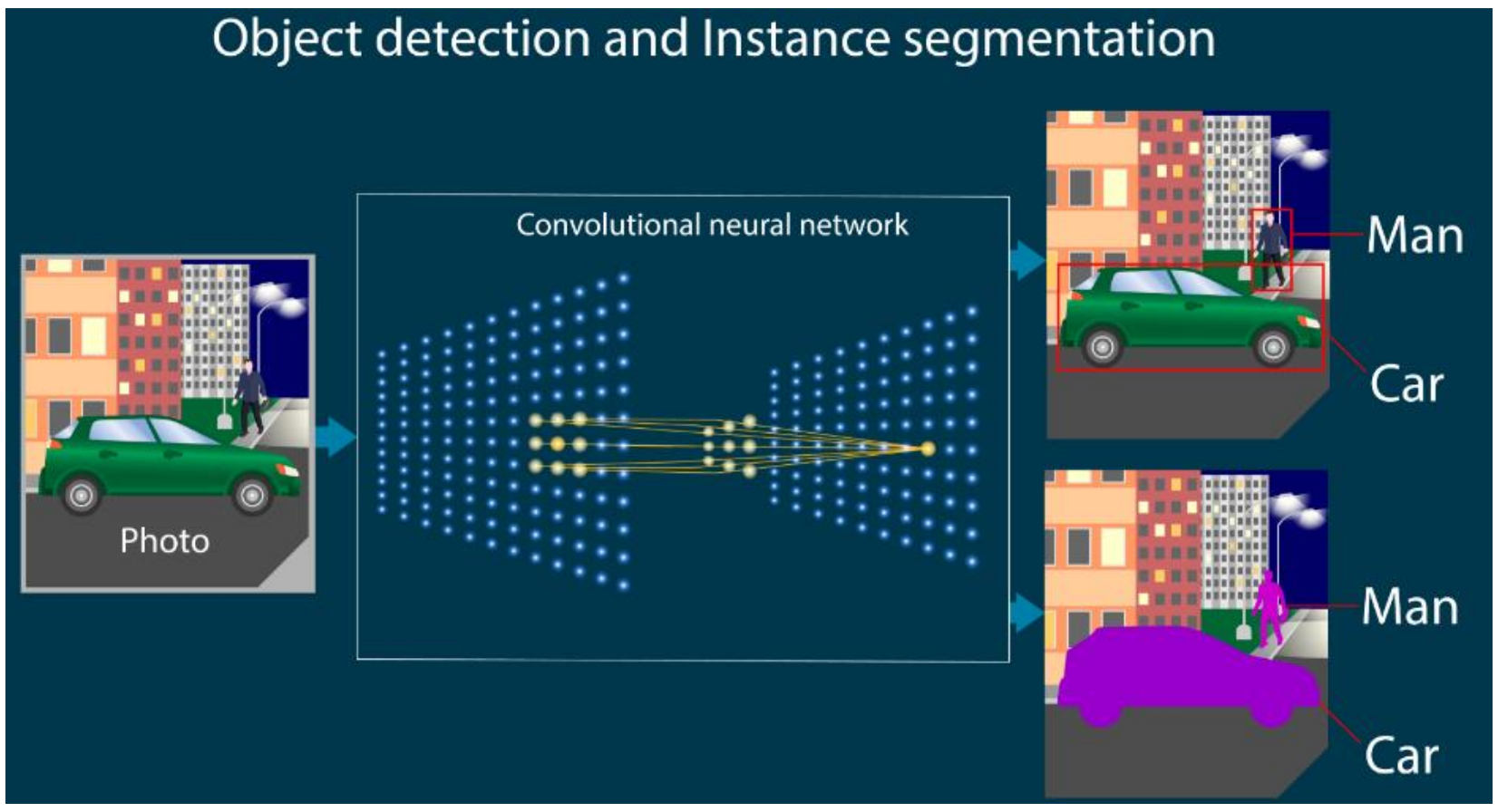

Figure 6.7. Identification of objects and segmentation of their instances. (© 123RF)

The following Figure 6.8 relates to computational photography (Pedone & Heikkilä, 2011). It removes the haze from the original photos above.

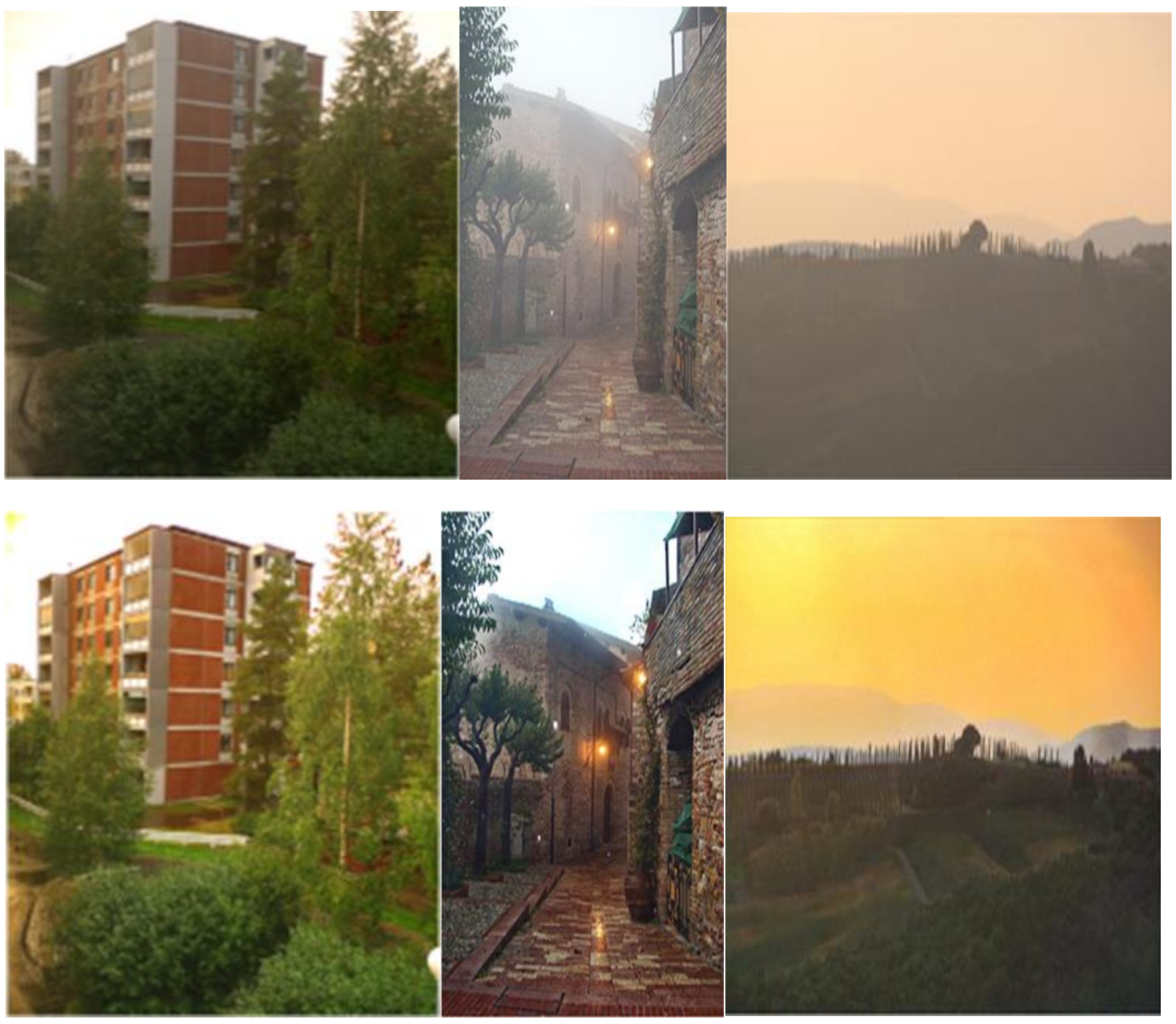

Figure 6.8. Computational removal of haze in an image. (© IEEE)

Figure 6.9 illustrates using a machine learning method to identify and locate cars and motorcycles in the image.

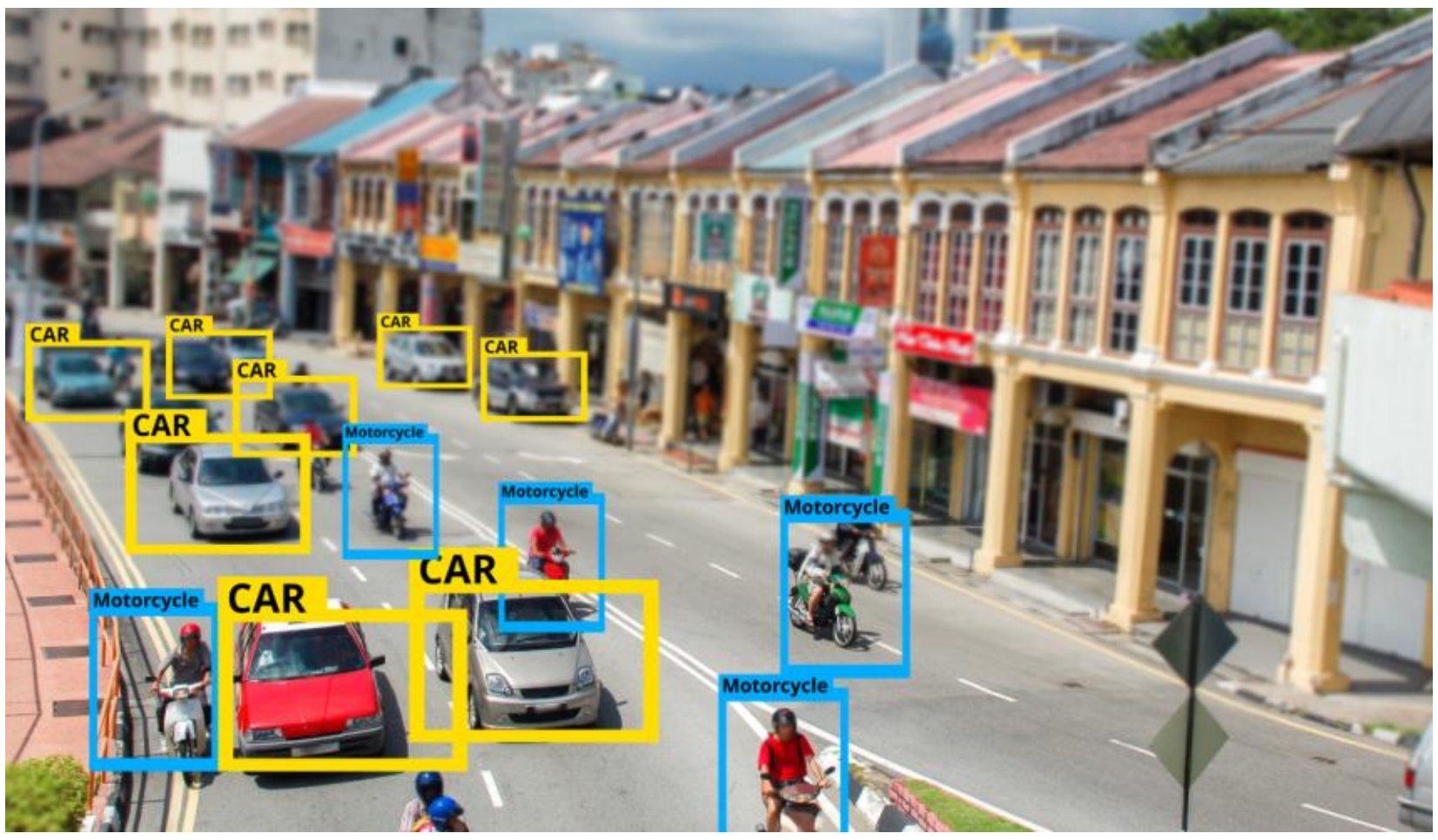

Figure 6.9. Object recognition using machine learning. (© 123RF)

Background removal and object tracking play a central role in various video surveillance applications. Figure 6.10 removes the motionless parts of the background from the video image and tracks the paths and activities of those persons who meet each other (Takala & Pietikäinen, 2007). However, removing the background may make it difficult to interpret human-environment interactions.

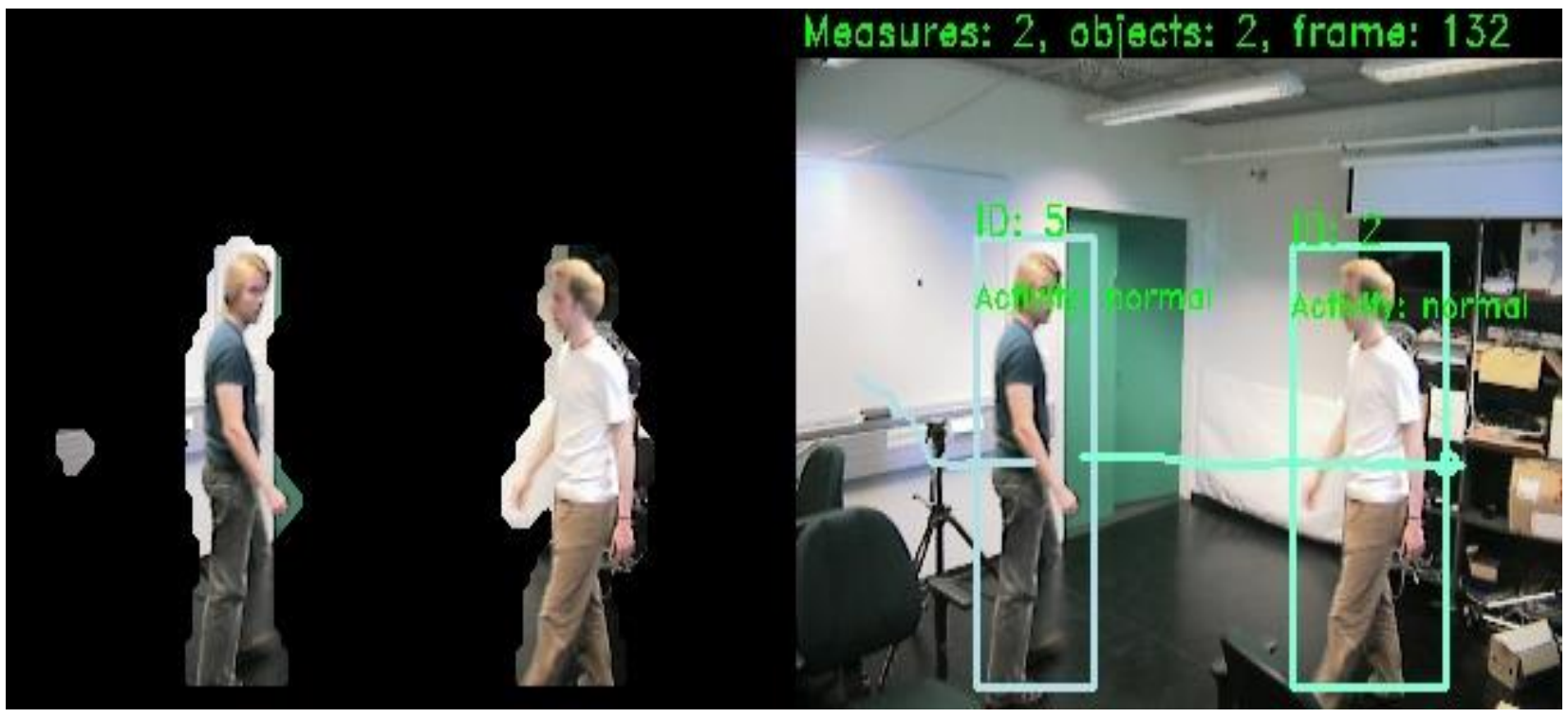

Figure 6.10. Background removal and tracking of moving objects. (© IEEE)

Biometric identification is a key area of research in machine vision, such as using face and iris (Figure 6.11 on the left). The human gait style is individual and can be used for biometric identification remotely. It might also be used to assess a person's emotional state (Figure 6.11 on the right).

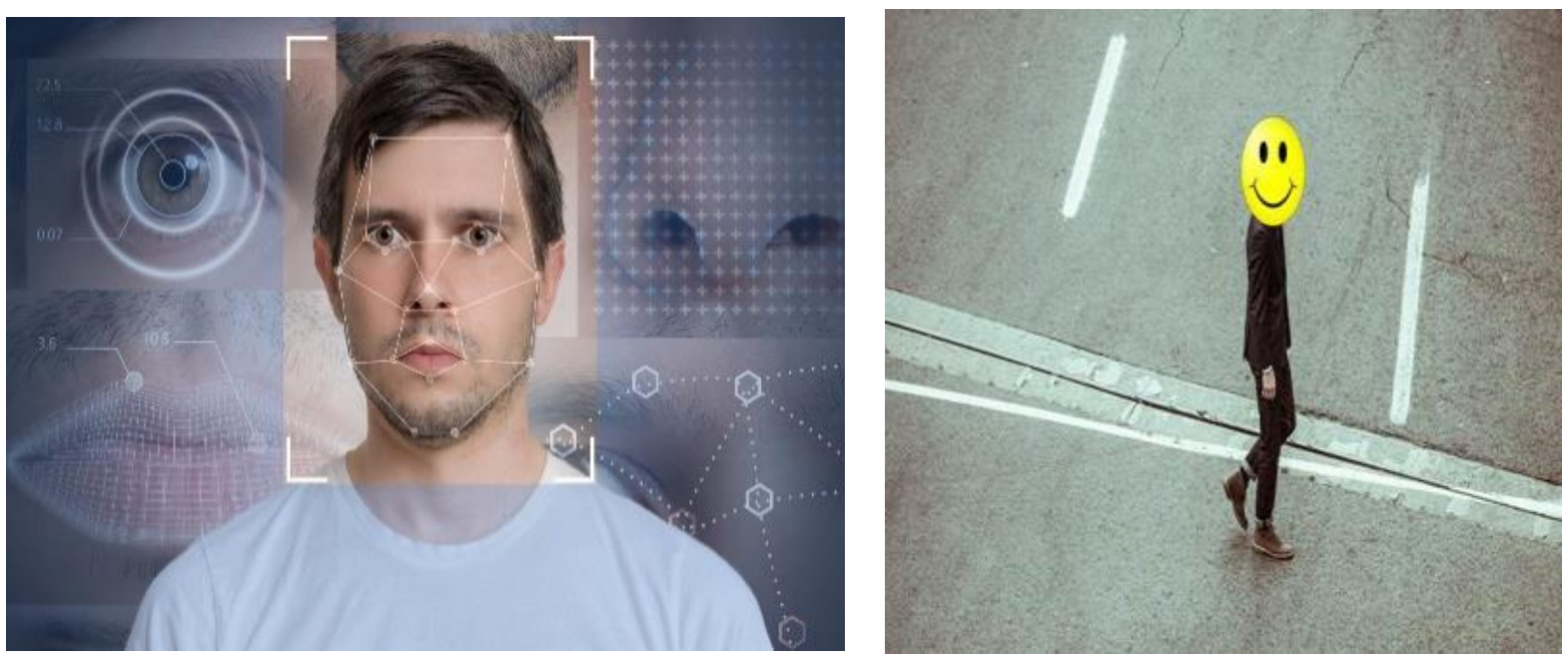

Figure 6.11. Biometric face recognition (© 123RF) and assessment of emotional state based on walking style. (© CMVS)

Below Figure 6.12 uses the distance and color image data generated by the Google Tango hardware mentioned in Section 6.2 to create a three-dimensional model of our laboratory space.

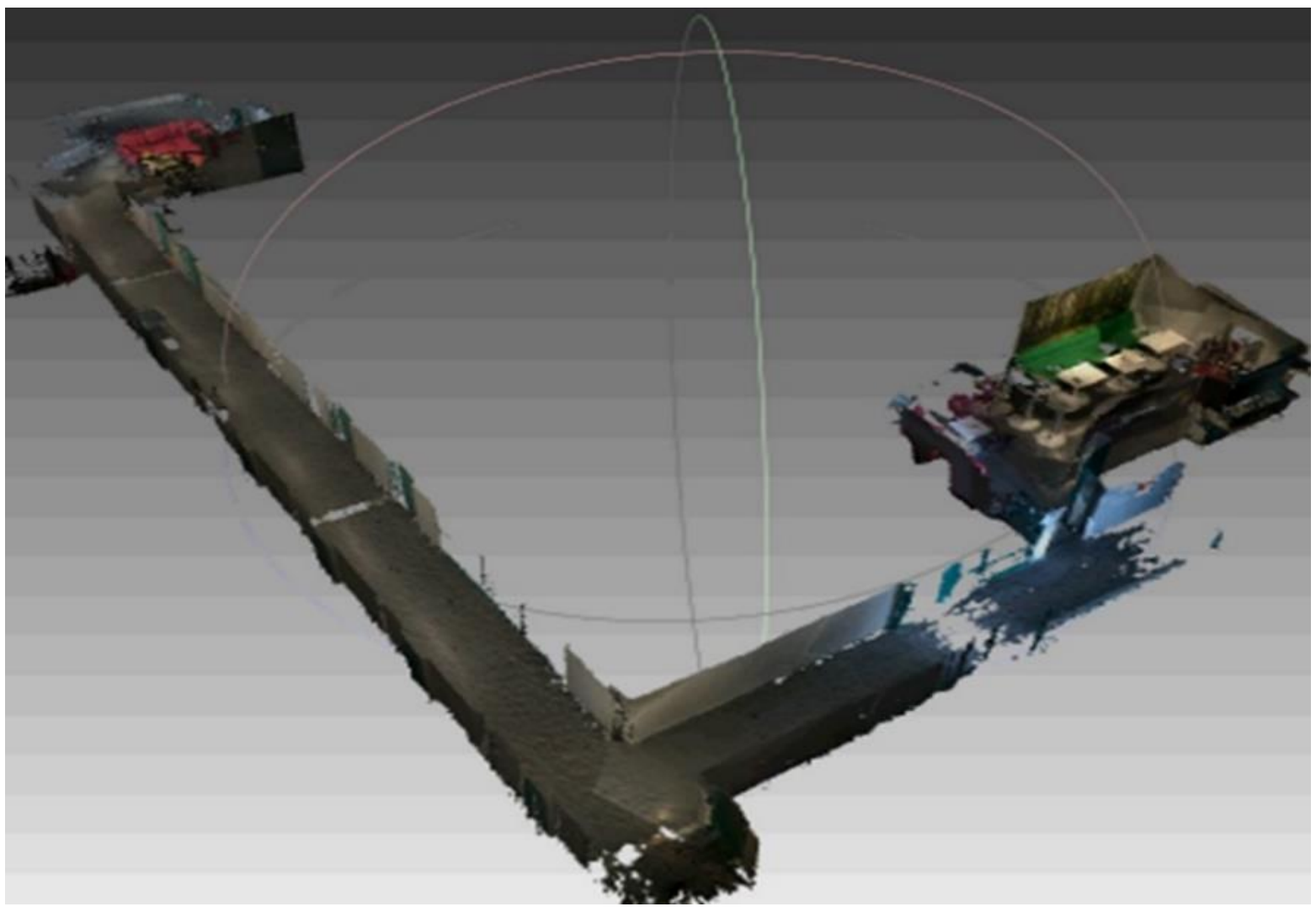

Figure 6.12. Created with Google Tango, a 3-D model of our research unit facilities (© CMVS).

## 6.4 Computer Vision Changing our Everyday Lives

Until recent years, machine vision has been largely in the background. It has been applied successfully in controlled conditions, such as visual quality control of industrial products, assembly and sorting tasks in industry, reading text from scanned documents, medical instruments such as cell counting, target detection and tracking in military applications, and environmental state analysis from satellite and aerial photographs. In stores, reverse vending machines for recycling cans and bottles have long used 3-dimensional machine vision.

In recent years, machine vision has become part of our daily lives. Biometric identification based on fingerprints, eye iris and face is already widely used, for example, at airports and for access control - and increasingly also on smartphones. They often include smile recognition as the first step in recognizing emotions. Internet searches based on image content have become much more accurate than before.

In order to prevent the growth of crime and terrorism, surveillance cameras have been installed everywhere in our environment. More and more efforts are being made to add automatic interpretation, for example, face or even behavioral recognition. The new cars have machine vision based lane guards and traffic sign detectors. Machine vision has also been used in art. For example, DeepArtEffects software can convert user-uploaded images to a selected painting style, https://www.deepar-teffects.com.

As we move through our environment in the near future, smartphones will be able to identify different objects, plants and birds and provide information to the user. While abroad, we are able to recognize foreign language text from ads and restaurant menus, as well as view customer reviews of the restaurant in question. Machine vision plays a key role in automatic shops, as already mentioned in Chapter 1 in relation to the system developed by Amazon.

We will see tremendous progress in technology for people with disabilities. Visually impaired people are able to read magazines and other documents. With the help of smart glasses, they are able to get information about their environment and even get information about the emotional state of their chat partner.

Machine vision plays a key role in autonomous or semi-autonomous vehicles, which are coming in confined and sufficiently safe environments. In medicine, imaging has come to play a central role in the diagnosis of diseases. Machine vision plays a significant role in helping doctors interpret images.

The development of imaging technologies opens up entirely new possibilities. For example, it is already possible today, even with printing technology, to make thin multi-lens cameras onto an image sensor. Figure 6.13 shows an image obtained with a thin multi-lens camera, from which it is also possible to make a three-dimensional interpretation based on the partial images.

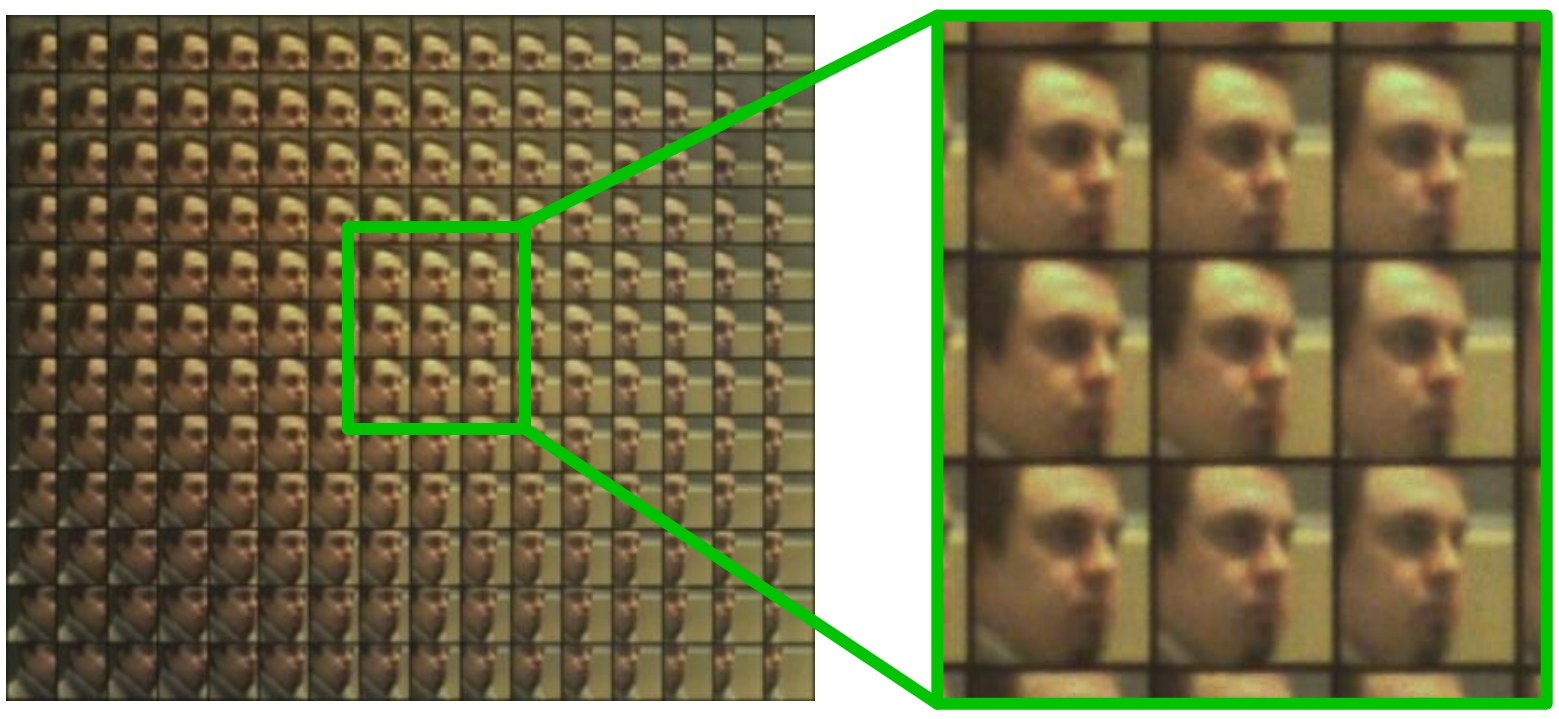

Figure 6.13. Picture taken with a lenslet camera. (© Sami Varjo)

In the future, these kinds of solutions can be embedded in the environment and, for example, on payment cards and, through them, implement novel human machine interfaces to intelligent environments with very little energy consumption (Varjo, 2016).

## 6.5 Center for Machine Vision and Signal Analysis

The **Machine Vision Group** of the University of Oulu was founded in late 1981 after Matti Pietikäinen returned from his research visit to the University of Maryland. Initially, the team's research was very application-oriented, with a focus on machine vision applications in visual quality control and robotics. Quality control, in particular, was already a very important field of application at that time, which made it possible to obtain external funding for research.

Since then, the group's research has expanded and deepened significantly, encompassing both internationally advanced basic research and applied research. The research and activities of the group for the first 25 years are discussed in the book "From Algorithms to Vision Systems - Machine Vision Group 25 Years" (Pietikäinen et al., 2006) and in the annual reports 1997-2015 (Pietikäinen et al., 2016). For 2016, see the Annual Report at http://www.oulu.fi/infotech/annual_report/2016/cmv. So far, machine vision research has led to the establishment of about 25 start-up companies, founded by earlier members of our group, and also new research teams have been spun off.

In the early 1990s the group returned to the texture analysis research started at the University of Maryland in 1980-81, which led to a very significant invention later: the Generalized Local Binary Pattern (LBP) and its application in face recognition (Chapter 7). In addition to texture and facial analysis (Section 8.2), the group has conducted world-renowned research in other areas, including: three-dimensional computer vision (Section 8.3), emotion recognition (Chapter 9), biometric identification (Section 8.2), intelligent human-machine interfaces (Section 8.4), industrial visual quality control (Section 8.1), and medical image analysis (Section 8.5).

The **Center for Machine Vision and Signal Analysis** (CMVS), which began operations in January 2016, combined the University's strong expertise in machine vision and biosignal analysis. In the Spring 2021 , the Center consisted of four professors (Olli Silvén, Guoying Zhao, Janne Heikkilä and Tapio Seppänen), one emeritus professor (Matti Pietikäinen), one associate professor (Mourad Oussalah) and three tenure track assistant professors (Li Liu, Miguel Bordallo Lopez, and Xiaobai Li). About 80% of the unit's staff of over 70 researchers and research assistants are from abroad. The Center is currently headed by Professor Olli Silvén.

The unit has played a key role in the development of artificial intelligence education at the University of Oulu since the 1980s. In the autumn of 2018, a new field of study in Artificial Intelligence was launched in the Information Technology Degree Program (Appendix L1).

CMVS's current research can be divided into five areas:

**Multimodal emotional user interfaces:** The overall goal of the study is to bring human-machine interaction closer to human-human interaction by considering the emotional state. A strong foundation is the group's research on passive imaging-based solutions, such as micro-expression recognition related to hidden emotions, dynamic facial expression and gesture recognition, analysis of physiological signals, and heart rate measurement from facial videos. Combining multi-modal data, masking a person's identity, and solutions that utilize machine learning are among ongoing research problems. Potential applications include e.g. human well-being monitoring, computer-assisted teaching, telemedicine and security technology.

**3-dimensional computer vision:** Advanced machine vision systems based on deep learning lack an ability to interpret 3-D views like humans do, making them difficult to use in many practical applications. The aim of the unit's research is to develop much more general solutions for the interpretation of 3-D views by combining data generated by multiple sensors (e.g. video and range cameras) and using modern machine learning methods to create a higher level semantic description to facilitate the interpretation of the given views. Potential application areas include autonomous robots and vehicles, augmented reality, and interpretation of views captured on mobile devices. The research is based on the group's solid experience in geometric calibration of cameras, 3-D reconstruction and machine learning.

**Toward resource-efficient AI:** Machine learning and statistical pattern recognition have been part of the unit's machine vision research since the 1980s. In recent years, deep learning has become central, but it has many weaknesses, such as the massive need for training samples, the need for high-performance computers, vulnerability to hostile attacks, and the inability to justify the reasons for classification decisions. The unit investigates computationally light, compact, and low-energy image, video, and 3-D data representations and classification methods that can be used in, for example, wearable devices, smart glasses, and embedded smart sensors. Efforts are being made to overcome the weaknesses of deep neural networks with both new representation methods and machine learning solutions.

**Medical signal analysis and biometrics:** The analysis of various physiological signals and biomedical images has been studied within the CMVS since the 1990s. The research focuses on the analysis of physiological signals in health assessment, the use of machine learning methods in biomedical applications, remote patient monitoring, and solutions related to privacy and information security. Recently, for example, the early detection of atrial fibrillation by remote measurement from facial videos has been studied (Section 8.2), as well as advanced brain signal analysis methods related to Alzheimer's disease research. Experts in both medicine and life sciences have been partners in the unit's research.

**Energy-efficient embedded machine vision systems:** CMVS has extensive experience in research and various applications of machine vision technology to support application development. The aim of the current research is to create knowledge for the design of distributed intelligent sensor systems and very low energy consumption high-performance computing solutions. The focus will be on the methods and technologies needed for large-scale, non-power-consuming sensing interfaces, using optical cameras for imaging or, for example, sensors operating on different radio waves. Potential applications include human-machine interfaces embedded in the environment and health monitoring devices.

## 6.6 References

Herrera Castro D, Kannala J & Heikkilä J (2011) Accurate and practical calibration of a depth and color camera pair. In: Computer Analysis of Images and Patterns, CAIP 2011 Proceedings, Lecture Notes in Computer Science, 6855:437-445.

Ikeuchi K, ed. (2021) Computer Vision: A Reference Guide, Second Edition. Springer, 1450 p.

Liu L, Ouyang W, Wang X, Fieguth P, Liu X & Pietikäinen M (2020) Deep learning for generic object detection: A survey. International Journal of Computer Vision 128: 261-318.

Marr D (1982) Vision: A Computational Investigation into the Human Representation and Processing of Visual Information. San Francisco: W. H. Freeman and Company.

Pedone M & Heikkilä J (2011) Robust airlight estimation for haze removal from a single image. Proc. CVPR 2011 Workshops, DOI: 10.1109/CVPRW.2011.5981822.

Pietikäinen M (1993) Konenäkö. Tekoälyn ensyklopedia (Machine Vision, Encyclopedia of Artificial Intelligence) (eds. Hyvönen E, Karanta I & Syrjänen M), 104-114.

Pietikäinen M, Aikio H & Karppinen K, eds. (2006) From Algorithms to Vision Systems – Machine Vision Group 25 years. University of Oulu, 254 p.

Pietikäinen M, Heikkilä J, Silvén O et al. (2016) Machine Vision Group – Annual Reports 1997-2015. University of Oulu.

Sauvola J & Pietikäinen M (2000) Adaptive document image binarization. Pattern Recognition 33:225-236.

Takala V & Pietikäinen M (2007) Multi-object tracking using color, texture and motion. Proc. IEEE Conference on Pattern Recognition and Computer Vision, 7 p.

Varjo S (2016) A Direct Microlens Array Imaging System for Microscopy. Acta Univ. Oul. C 588, 2016, 124 p.

Wiki-Kinect Kinect.

.

# 7 Local Binary Pattern – a Breakthrough

## 7.1 Local Binary Pattern Method

The main scientific achievement of the research group is the invention of the Local Binary Pattern (LBP) method based on the analysis of local binary patterns, and the development of its various variants and applications (Pietikäinen et al., 2011), (Pietikäinen, 2010).

Originally developed for the analysis of the surface textures in an image or portions thereof, it later developed into a general-purpose method for capturing the content of images and videos for identification. The textures are very diverse, with almost regular patterns or very irregular surface texture variations. It was later found that the method can also be applied to untraditional textures, such as images of human faces. Figure 7.1 (Liu et al., 2019) shows examples of images with very different textures.

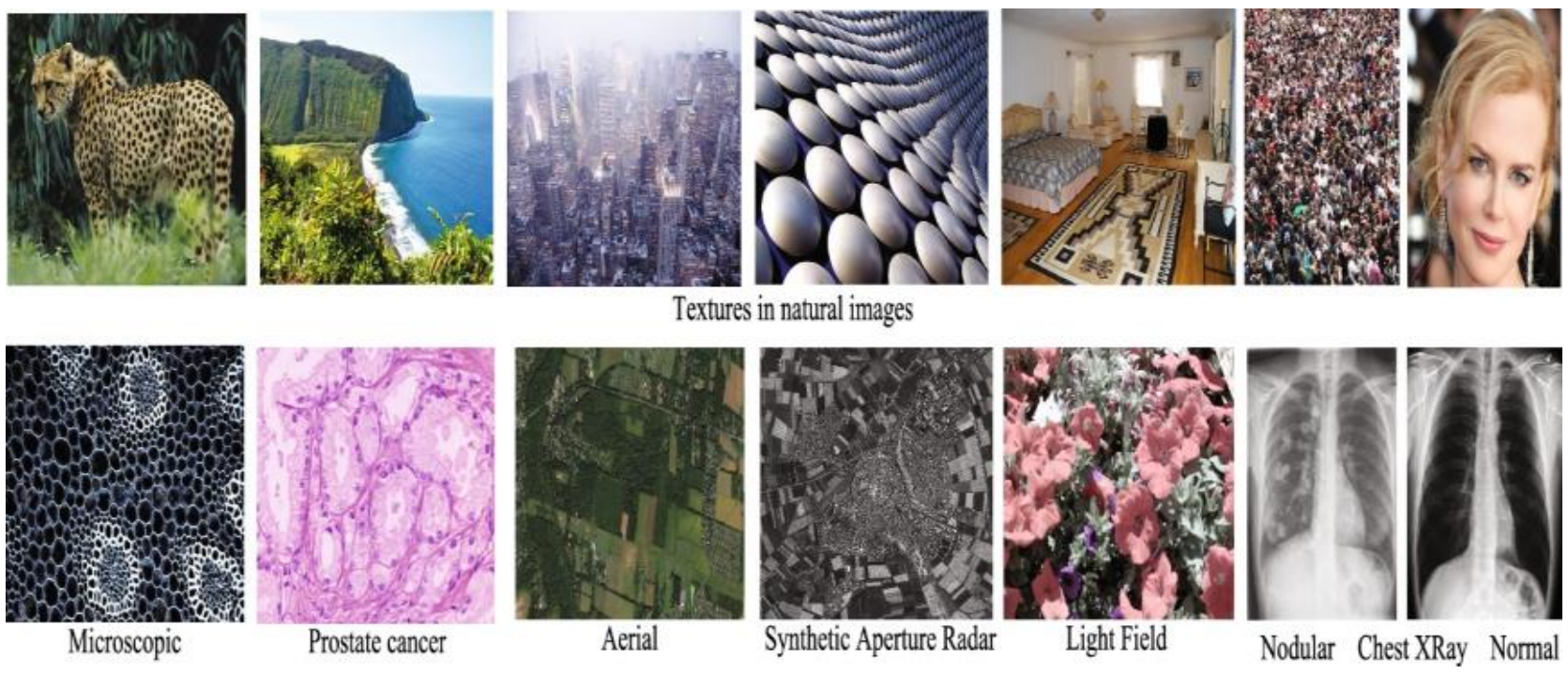

Figure 7.1. Examples of images with very different textures. (© CC BY 4.0)

The original basic LBP shown in Figure 7.2 was created as early as 1992-93 by co-operation of David Harwood, a visiting researcher from the University of Maryland, Matti Pietikäinen, and his postgraduate student Timo Ojala who is currently a professor. At that time and in the late 1990s LBP did not receive much attention in the scientific community.

In the early 2000s, a theoretical foundation was laid, which in turn led to the first real breakthrough (Figure 7.3). LBP was not of much interest to many top scientists before it was applied to face recognition (2004-2006), which can be used to identify a person's identity, facial expressions, and gender, for example (Figure 7.7).

The fourth major step was the generalization of the LBP method to video image sequences (Figure 7.6), whereby, for example,

changing facial expressions or words and phrases spoken in the video can be recognized from the video images.

LBP is an excellent example of the importance of long-term, well-focused research. The approach has deviated from the mainstream of research, and so it took ten years for the process to be truly breakthrough and accepted by the scientific community. Today, this highly cited method and its widespread application in various fields can be considered as one of the greatest success stories in Finnish computer science research.

The LBP method measures the texture content on the surfaces of objects in an image. The method is computationally simple and attractive to use as such.

An image converted to grayscale is processed pixel by pixel, for example, from top left to bottom right. For each pixel, its LBP code is computed in a selected size neighborhood, for example 3x3 or 5x5. Figure 7.2 shows an example of calculating the LBP code in a 3x3 pixel neighborhood (Ojala et al., 1996).

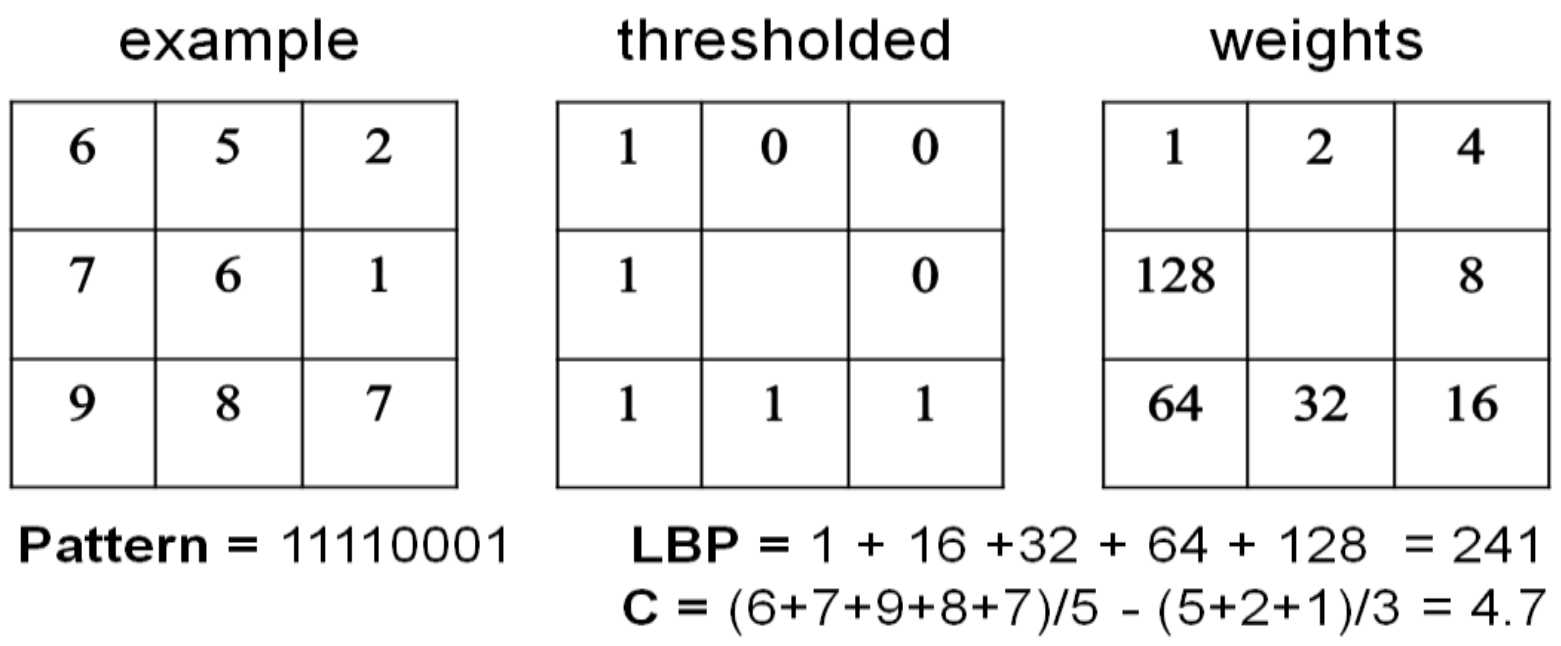

Figure 7.2. Example of calculating LBP code in a 3x3 neighborhood. (© Springer)

Reprinted, with permission, from Springer [Pietikäinen M, Hadid A, Zhao G & Ahonen T (2011) Computer Vision Using Local Binary Patterns]

In the left part of the figure, the numbers represent the gray scale, which is thresholded by the gray level of the center pixel in question (6). Neighbors with a tone greater than or equal to 6 are assigned a binary value of 1 and smaller values of 0, as shown in the center figure. Each location is assigned a weighting factor of 1 ... 128 (i.e. $2^0$ ... $2^7$) in a 3x3 environment, as shown at right. This allows the LBP code (241) to be calculated for the middle pixel under consideration. Using the eight neighbors of the midpoint, there are a total of 256 possible values, i.e., the length of the feature vector is 256. Similarly, the texture contrast (C) or strength (4.7) can be calculated by subtracting the sum of the shades of gray of pixels labeled with “0” from those labeled with “1”.

This same calculation is done for all pixels in the image or image area, for example from left to top to bottom right. By calculating the distribution of the number of times each LBP code occurs in a given region, the so-called LBP histogram is obtained to represent the texture of the region in question. When using contrast, a separate histogram is calculated for it.

Since then, the method has been improved and generalized. An important milestone in 2002 was the method published in the top-tier journal of computer vision and artificial intelligence, IEEE Transactions on Pattern Analysis and Machine Intelligence (PAMI). The computing neighborhood (3x3) was generalized to arbitrary distance (for example 5x5, 7x7, etc.) and a rotation tolerant version of the method was developed (Ojala et al., 2002). In contrast computing, the aforementioned generalized LBP method employs a statistical variance of the differences between the selected samples and the gray scale of the center under consideration.

Figure 7.3 shows an example in which different number of samples (P) are taken for the gray scale of an image at different distances (R) (Ojala et al., 2002). For each sample whose position in the horizontal and vertical directions is not an integer (1, 2, 3, etc.), a value is calculated by bilinear interpolation of the pixels of the original image in the neighborhood in question. As the number of sample pixels increases, the feature vector becomes longer, for example, in the case of eight neighbors (P), the feature vector has a length of 256, but with a value of 16, the length is already $2^{16}$, or 65,536.

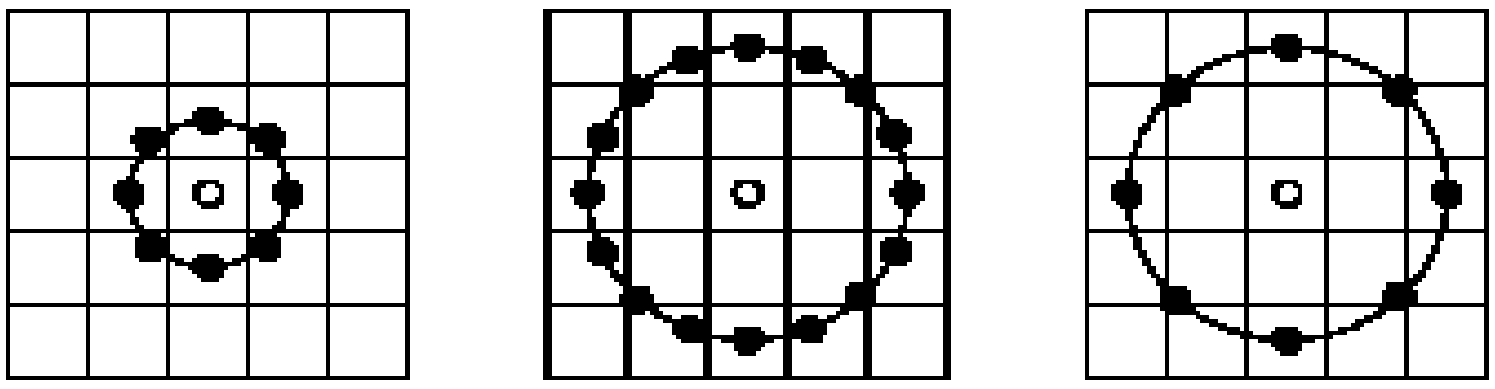

Figure 7.3. Generalized LBP. (© Springer)

However, our study found that the vast majority of information is in the so-called “uniform patterns”. Therefore, the number of codes generated by the method can be significantly reduced, i.e. to shorten the feature vector. This was an important observation, for example, when developing the face recognition method.

LBP can be interpreted as an operator analyzing the fine structure of an image, expressing in each pixel a micro-pattern, such as point, flat area, end of line, edge or corner (Figure 7.4) (Pietikäinen et al., 2011).

The histogram over the selected region depicts the number of times each pattern occurs in that region, i.e., gives information about the micro-texture of the region. This is a very simple version of the so-called Bag-of-Words (BoW) representation of the contents of an image area where the "words" correspond to different LBP codes (patterns) in the histogram.

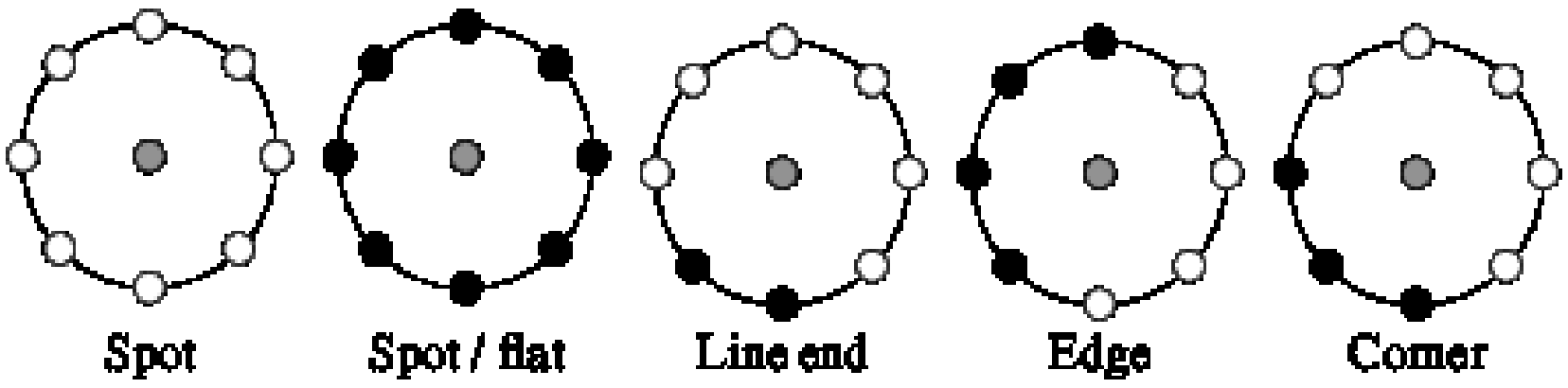

Figure 7.4. Micro-patterns detected by LBP operator. (© Springer)

The BoW principle has been widely used in computer vision in a wide variety of problems and applications, especially prior to the current AI using CNN deep neural networks. For a comprehensive overview of the various BoW and CNN methods for representing textures, see (Liu et al., 2019).

In Figure 7.5, the LBP method is applied to a face image. The value of each pixel in the center image has the LBP code calculated for it. It can be deduced from the figure that the method is quite insensitive to variations in illumination, since the LBP operator is independent of locally monotonically increasing or decreasing greyscales. The histogram to the right shows the number of occurrences of each LBP code over the entire image.

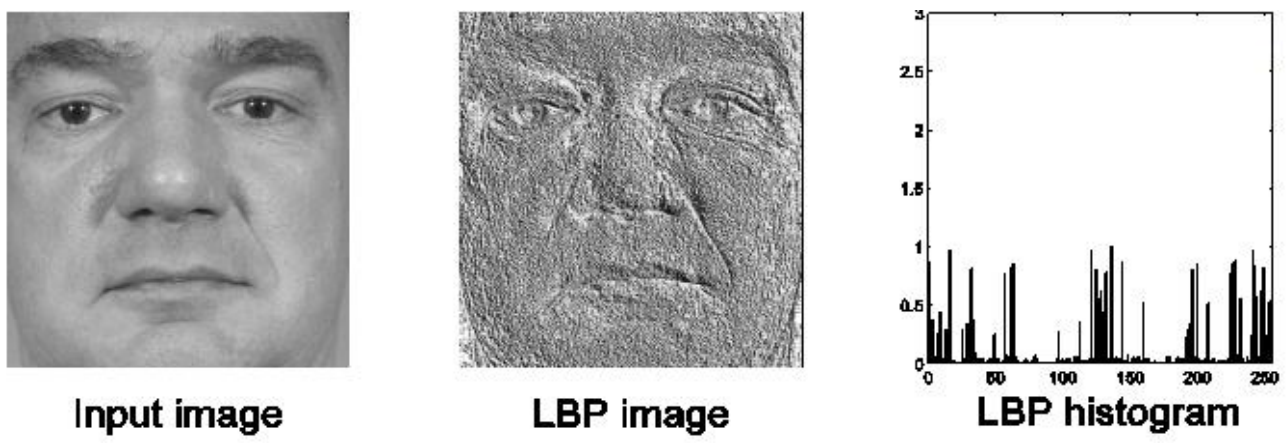

Figure 7.5. Application of LBP to a facial image. (© Springer)

Due to its computational simplicity, the LBP method is applicable to the real-time analysis and automatic interpretation of many types of digital image material. Because the operator is independent of monotonic (incremental or decremental) shades of gray in its computing neighborhood, the method is quite resistant to variations in lighting. For these reasons, the method has become very widespread in the scientific community and in a wide variety of applications (Pietikäinen et al., 2011).

A young doctor Guoying Zhao (now a professor) who came to Oulu from China in 2005, generalized the LBP method for dynamic textures, enabling the analysis of facial expressions, mov-ing objects and actions (Zhao & Pietikäinen, 2007). A major innovation in the LBP-TOP method is to do 2-D analysis in three orthogonal planes, one of which corresponds to the normal image plane (XY) and the two to the place-time planes (XT, YT). The distributions of the different directions are combined to represent the dynamic texture as shown in Figure 7.6.

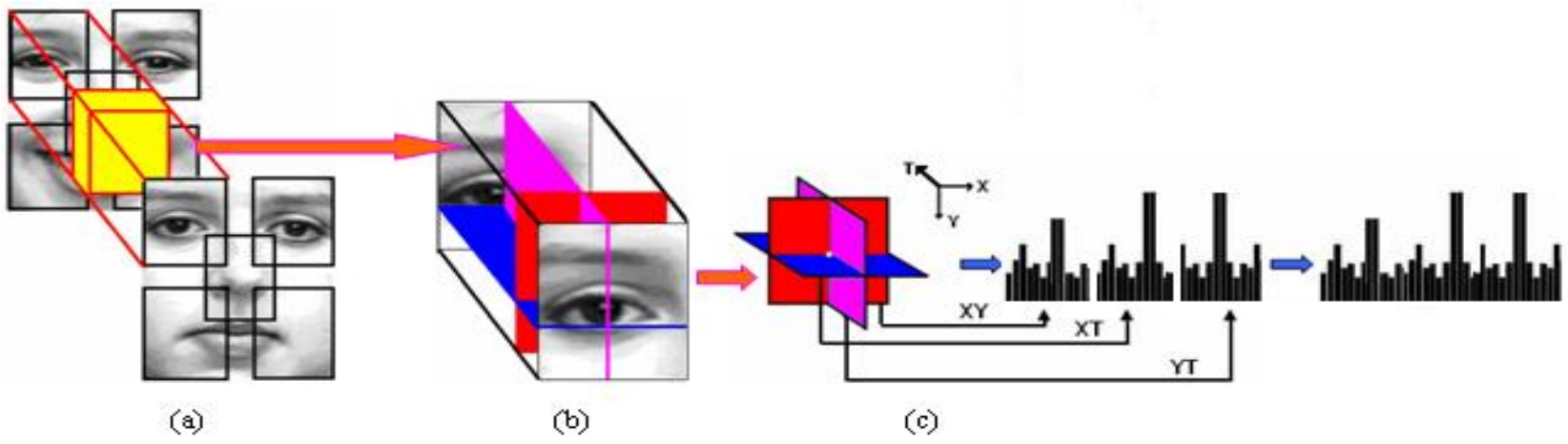

Figure 7.6. LBP-TOP in dynamic texture analysis. (© Guoying Zhao)

This method or its modifications have been extensively used in many problems, such as the recognition of ordinary facial expressions and micro-expressions (Chapter 9), visual speech analysis based on mouth movements (Section 8.2), anti-spoofing for face recognition (Section 8.2), human action recognition (Section 8.4), and biometric identification based on gait (walking style) analysis.

## 7.2 LBP in Face Recognition

Facial image analysis has been one of the major challenges in computer vision research in recent years, and our discovery of a suitable method has proven to be a major success (Ahonen et al., 2004, 2006). Figure 7.7 illustrates the principle of the method presented in simplified form.

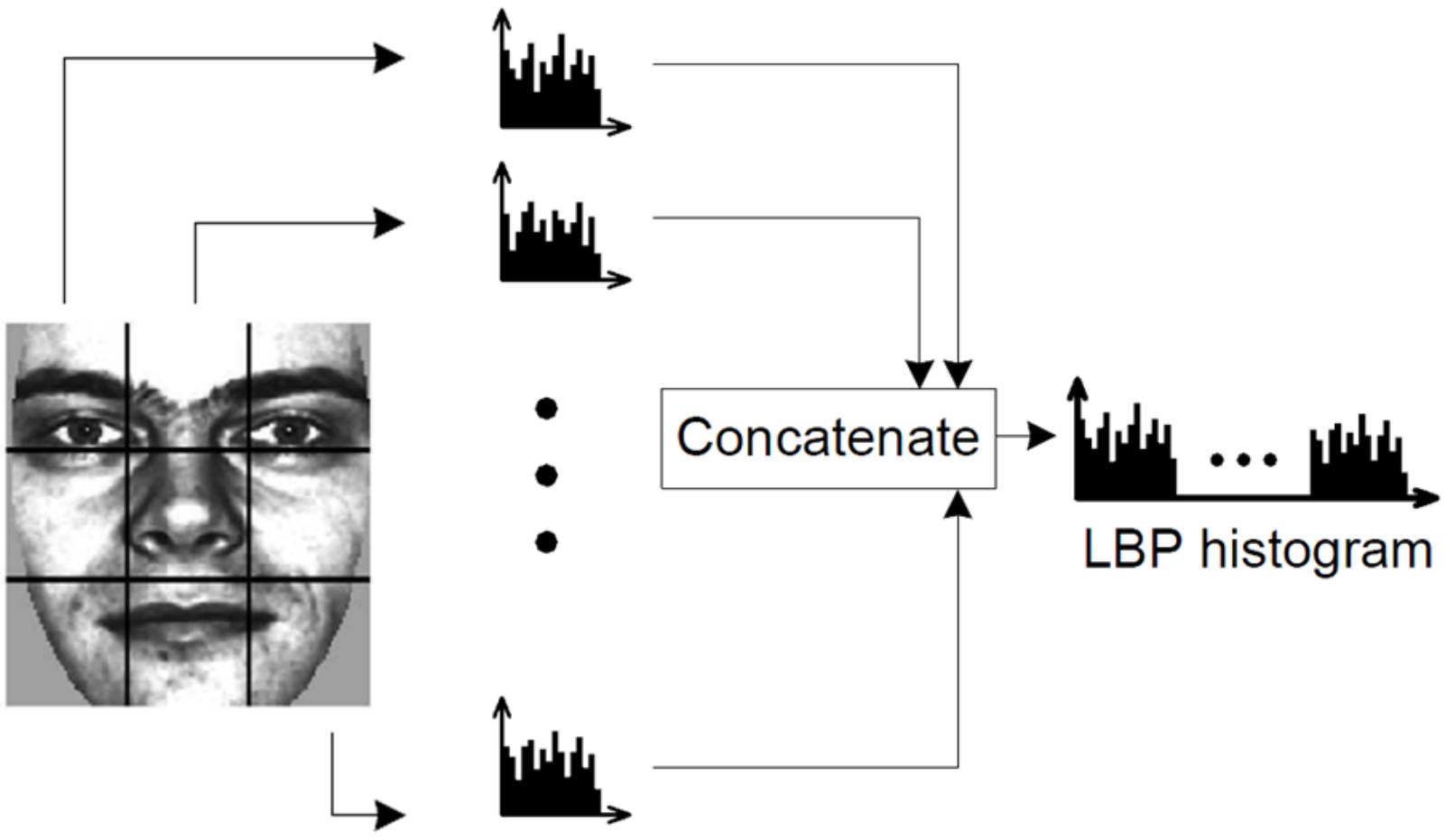

Figure 7.7. Face analysis with LBP. (© Springer)

Initially, the face is located with a face detection method, for example (Viola & Jones, 2004), and normalized to the same size and location by first searching key points, for example, the eyes and some other easily identifiable features in the area of the face. The LBP feature image is then calculated on the face area and the feature distributions for each window in the selected grid are calculated.

The LBP distribution represents the grayscale variations of the face, i.e. the texture, within the window in question. The location of the window gives information about the geometry of the face, i.e. where there is on the face texture and what kind of texture. By combining the distributions of all the windows into a common feature distribution over the entire face area, a LBP representation of the face content is obtained.

In the system training step, models are calculated for each individual to be recognized, generally using multiple facial images and the resulting feature distributions for each individual. Then, to identify who the unknown person is, compare their LBP description to the different person models taught and identify who it is if they find someone close enough. This case is a so-called *1:n recognition*, that is to say whose face is in question from the n persons trained for the system.

Face authentication, or *1:1 authentication*, is intended to verify that the person is one whose model is stored on the device during the training phase. Such a procedure based on the comparison of the described and the stored model is used, for example, to check the passport and identify the user when opening the smartphone. The third main method of face recognition is to use the so-called *watch list* to find the faces of people searched for in a large database, such as certain criminals.

In Section 8.2, the method has been used to detect spoofing attacks in a biometric identification application.

LBP and the resulting face recognition method became very popular in research and industry. References to publications are the most important measure of the scientific impact of research. Publications about LBP and the facial recognition method based on it are the most cited Finnish publications in the field of artificial intelligence in the 21st century. Several leading research groups have developed customized versions of our method.

At the European Conference on Computer Vision (ECCV 2014), the first LBP-based face recognition publication (Ahonen et al., 2004) presented at the same conference ten years earlier, was awarded a major award titled *Koenderink Prize for Fundamental Contributions in Computer Vision that has withstood the test of time.*

## 7.3 LBP: Present and Future

The above first versions of LBP have shortcomings, and research to improve them is still ongoing within the research community.

One of the biggest problems has been the sensitivity of the method to noise and other disturbances in the image. The LBP method has also encountered difficulties in dealing with so-called macro textures, where the elements of the texture pattern are large, several pixels in size.

Perhaps the best current method is the MRELBP (Median Robust Extended LBP), published in 2016, which has operator computation done in a larger local environment (from 9x9 - 17x17 pixels) and used to calculate median instead of average in local sample point neighborhoods (Liu et al., 2016).

In the calculation of the median, the pixels of the selected pixel neighborhood are in gray scale order and the middle one is selected. For example, the gray levels of the pixels in Figure 7.2 are (9 8 7 7 6 6 5 2 1) and their median is 6. Most LBP methods are such that the determination of a suitable LBP operator and its computing environment is largely application-specific.

Figure 7.8 summarizes the evolution of the LBP method over two decades, including the most well-known LBP variants (Liu et al., 2019).

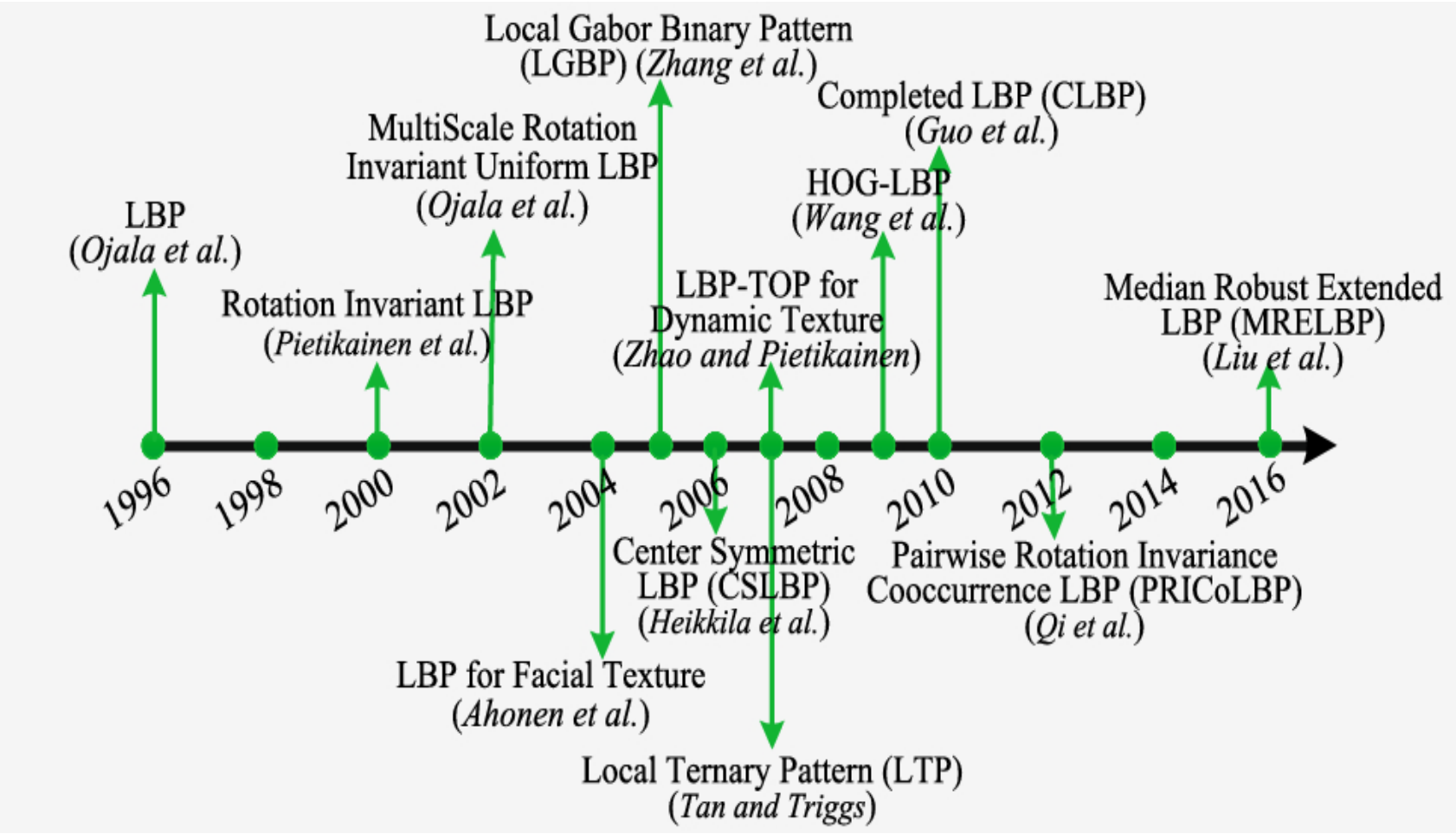

Kuva 7.8. Milestones in LBP research. (@ CC BY 4.0)

The use of the LBP method in face recognition has been further explored and very competitive results have been achieved. The focus has been e.g. better utilization of the facial structure and development of methods more tolerant to the effects of illumination and other variations than the above-mentioned method - while still preserving the computational simplicity of the original method. These new LBP methods are best suited for applications that require simple implementation and low power consumption, and massive amounts of training samples are not available.

However, methods based on sophisticated face positioning techniques, CNN neural networks, and enormous training material have been at the forefront of the latest breakthroughs in face recognition in natural environments (“in-the-wild”). Perhaps the most well-known example is DeepFace, developed by Facebook, which has used a three-dimensional face alignment method and a nine-layer neural network with 120 million weight factors (Taigman et al., 2014). Four million photos of Facebook users have been used for training. Another very significant development is the Google’s FaceNet, which directly learns description of faces in a compact, so-called Euclidean space, where distances correspond directly to a measure of the similarity of faces to be compared (Schroff et al., 2015).

Convolution-based operators have been used for years in texture analysis, including their application in face analysis. In this context, much of the recent use of the word "handcrafted" can be criticized because convolution-type filters, such as Gabor filters, found in the human visual system are often imitated in operator design. The operation of the LBP operator can also be seen as a combination of simple convolutions. For these reasons, the learning convolutional neural networks (CNN) presented in Chapter 4 can also be considered as texture methods (Liu et al., 2019).

Unlike basic LBP, CNN methods also perform very well in macro-textures with large variations. They are also able to learn directly from the training data a multi-scale representation, and enable them to learn the entire computing chain from convolutional filters to classification. The major problems for the applications are computational complexity, which requires special processors capable for high-performance computing, and the need for a very large number of training samples.

In our extensive comparisons, we also found that CNN-based methods are less resistant to noise, image blur, and small variations in viewing direction than the best modern LBP methods, especially the MRLBP operator (Liu et al., 2017). For these reasons, the applicability of CNN methods in most texture analysis applications is still very limited. In contrast, LBP operators are used in a very wide variety of applications.

Among the major challenges of coming years is to combine the best properties of LBP-type and CNN-based solutions, such as

1. compactness and computational lightness, allowing methods to be used in low power systems such as smartphones, smart glasses and clocks, embedded devices
2. tolerance to various image transformations and disturbances
3. ability to recognize macro textures and textures with large variations in appearance
4. ability to learn from data multiple-scale image representations

5. ability to learn like a human with a small amount of training data

The research community has already begun to tackle these problems, and a number of related publications have already been published. Our team has been active in promoting research in this field in the scientific community by organizing workshops at the conferences and by providing special issues to IEEE Transactions on Pattern Analysis and Machine Intelligence and the International Journal of Computer Vision.

## 7.4 References

Ahonen T, Hadid A & Pietikäinen M (2004) Face recognition with local binary patterns. Computer Vision, ECCV 2004 Proceedings, Lecture Notes in Computer Science, Springer, 3021:469-481.

Ahonen T, Hadid A & Pietikäinen M (2006) Face description with local binary patterns: Application to face recognition. IEEE Transactions on Pattern Analysis and Machine Intelligence 28(12):2037-2041.

Liu L, Chen J, Fieguth P, Zhao G, Chellappa R & Pietikäinen M (2019) From BoW to CNN: Two decades of texture representation for texture classification. International Journal of Computer Vision 127(1):74-109.

Liu L, Fieguth P, Guo Y, Wang X & Pietikäinen M (2017) Local binary features for texture classification: Taxonomy and experimental study. Pattern Recognition 62:135-160.

Liu L, Lao S, Fieguth P, Guo Y, Wang X. & Pietikäinen, M. (2016) Median robust extended local binary pattern for texture classification. IEEE Transactions on Image Processing 25(3):1368-1381.

Ojala T, Pietikäinen M & Harwood D (1996) A comparative study of texture measures with classification based on feature distributions. Pattern Recognition 29(1):51-59.

Ojala T, Pietikäinen M & Mäenpää T (2002) Multiresolution gray-scale and rotation invariant texture classification with Local Binary Patterns. IEEE Transactions on Pattern Analysis and Machine Intelligence 24(7):971-987.

Pietikäinen M (2010) Local binary patterns. Scholarpedia 5(3):9775.

Pietikäinen M, Hadid A, Zhao G & Ahonen T (2011) Computer Vision Using Local Binary Patterns, Springer, 207 p.

Schroff F, Kalenichenko D & Philbin J (2015) FaceNet: A unified embedding for face recognition and clustering. arXiv: 1503.03832v3.

Taigman Y, Yang M, Ranzato MA & Wolf L (2014) DeepFace: Closing the gap to human-level performance in face verification.

Proc. IEEE Conference on Computer Vision and Pattern Recognition (CVPR).

Viola P & Jones M (2004) Robust real-time face detection, International Journal on Computer Vision 57(2): 137-154.

Zhao G & Pietikäinen M (2007) Dynamic texture recognition using local binary patterns with an application to facial expressions. IEEE Transactions on Pattern Analysis and Machine Intelligence 29(6):915-928.

# 8 Towards Machine Vision Applications

Applied research on computer vision has gone hand in hand with advances in methodology and hardware. In the 1980s, the emphasis was on problems where the imaging conditions could be controlled, such as lighting and camera location to be kept constant. Because of the low computing capacity of computers, machine vision methods had to be computationally simple enough to access real-time operations. In our own group, in the 1980s, we focused on visual quality control of industrial products, as it had a large number of applications in Finnish industry and thus made it possible to obtain external funding for research (Section 8.1).

In the 1990s, methodological research and equipment had evolved so much that research began to focus on consumer-oriented applications, where people are often at the center, that is, the need to detect and identify people and their activities through images. The operating environment is changing and often requires analysis of video sequences rather than single images. The focus of the study was on facial image analysis: first, face detection based on skin color and, in the 2000s, facial recognition (Sections 7.2 and 8.2).

Biometric identification is a very important application, and our related research started as part of two large-scale European research projects. Later in the 2000s, research expanded to recognize facial expressions and emotions (Chapter 9) and speech recognition from mouth movements (Section 8.2). In face analysis, a study on heart rate measurement from videos was started in the 2010s (Section 8.2)

Our research into 3-D machine vision also began in the 1990s. The first milestone achieved was a method and tool for geometric calibration of the camera, receiving many citations and practical applications (Section 8.3). More recently, techniques related to 3-D modeling of the environment and methods to support augmented reality applications have been studied.

Our research into intelligent robots began in the 1980s (Section 8.4). Typical for applications is the three-dimensional information needed to analyze the robot environment, which can be obtained, e.g., by stereo vision systems, or 3-D cameras that provide direct distance information. In the 1990s, a major focus in our research was the so-called Machine of the Future project, with an application of a paper roll manipulator operating in a harbor environment.

In the 2000s, our research focused on the use of machine perception and machine vision methods in the control of moving robots. A demonstration system called *Minotaurus* was developed to op-erate a robot working in the laboratory environment. In this context, a study of human-robot / machine interaction based on the use of machine perception information was also initiated.

Medical image analysis research began in the Machine Vision Group in the 1990s, dealing, e.g., with magnetic resonance imaging (MRI) in surgery and analysis of skin images to identify melanoma. In the 2000s, use of functional MRI (fMRI) in brain imaging began to be explored. In recent years, major subjects have included the analysis of many types of microscope images and videos, chest X-ray, and fundus images (Section 8.5).

## 8.1 Visual Quality Control

Industrial and machine automation has been one of the most significant applications of machine vision since the 1970s. Visual quality control and sorting play an important role in applications in the electronics, metal, wood processing and food industries, for example. With automatic inspection it is possible to achieve better product quality and lower production costs. The machine is capable of performing visual inspection without tiring or taking breaks. Inspection and quality classification of wood surfaces and inspection of metal surfaces in industry are examples of key Finnish applications.

Visual quality control of industrial products was the most important application of machine vision in the 1980s. In Oulu we initially invested in the inspection of printed circuit boards and metal surfaces, but later the inspection of lumber became a key issue. In connection with the inspection of printed circuit boards, Olli Silvén, M.Sc. (who became later a professor), developed in his Ph.D. thesis a novel approach that compared the data measured from an image with the CAD model used to design that board, and recognized deviations from the model (e.g. breaks or too broad conducting wires) (Silvén et al., 1989). The approach was ahead of its time and did not lead to industrial exploitation at that stage. In connection with the inspection of metal surfaces, we investigated real-time defects detection from a moving metal strip in steel manufacturing (Piironen et al., 1990).

Lumber quality grading is a significant application where the use of image color and texture information clearly increases accuracy and product value (Silvén et al., 2003). Typical defects to be sought are knots, but the material may have only blue-tinged and spotty discoloration, as well as fully wood-colored knots, which can only be distinguished by their grain pattern.

The example in Figure 8.1 illustrates how faults in the same classes may vary greatly in appearance (Pietikäinen & Silvén, 2002). The so-called sound knot is firmly fixed to its surroundings and the dead knot is not. With the help of color and texture features and the developed unsupervised classification method, the rela-tive proportions of undetected faults and false alarms were reduced to less than 5%, which was less than half of the usual industrial level. The developed methodology was also applied to other industrial applications, e.g., sorting coffee beans.

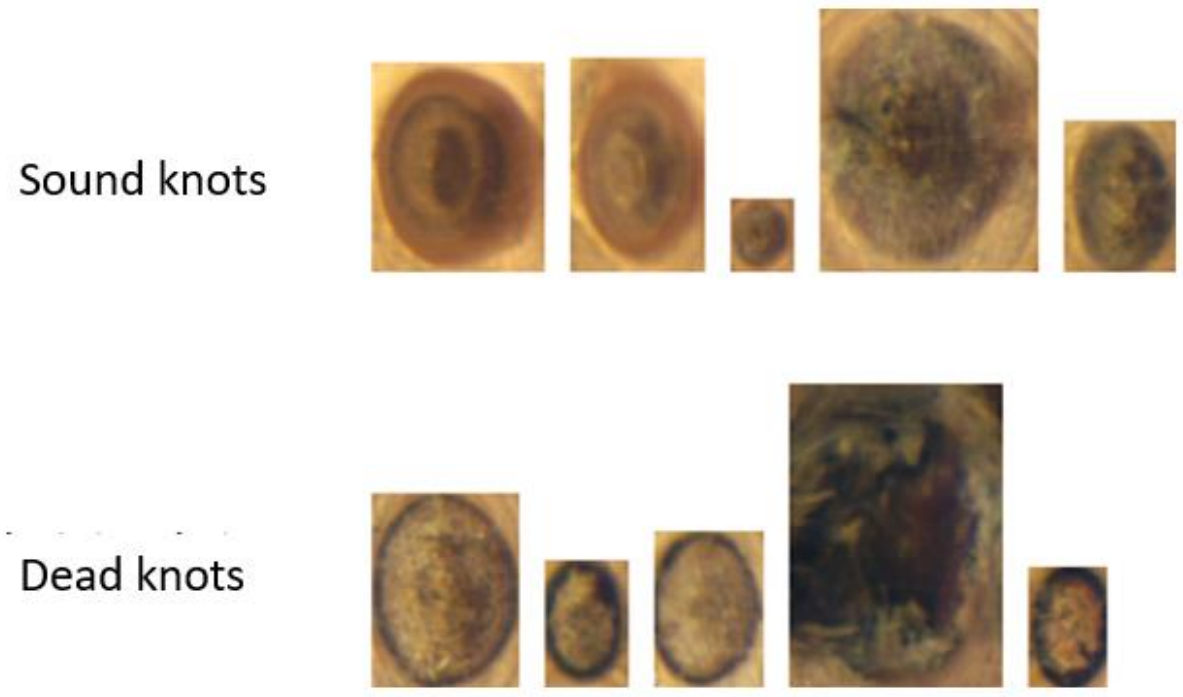

Figure 8.1. Variations in two different types of knots.

### *8.1.1 Training surface inspection*

There are two basic problems with surface quality checks: a distinction must be made between defects and defective areas, and then defects found must be classified for quality purposes. For example, defects in the steel surface must be able to be categorized at least into 2-D and 3-D categories, and the lumber inspection must be able to categorize the knots into different classes.

In many quality control tasks, there are even dozens of fault classes, and distinguishing them from one another requires even human sophistication. Often there are deviations from the ideal quality on the surface, but only cases that violate the given criteria are to be interpreted as defects.

Defect detection and imaging solutions for inspection equipment are generally designed according to the typical failure characteristics of each application. On this basis, the system can be trained for defect detection and recognition. For example, the appearance of sawn timber depends on the site and the quality control equipment must be adapted to these natural changes.

Traditionally, pattern recognition systems are trained through human-selected samples of defects and flawless material as shown in Figure 8.2 (Pietikäinen & Silvén, 2002).

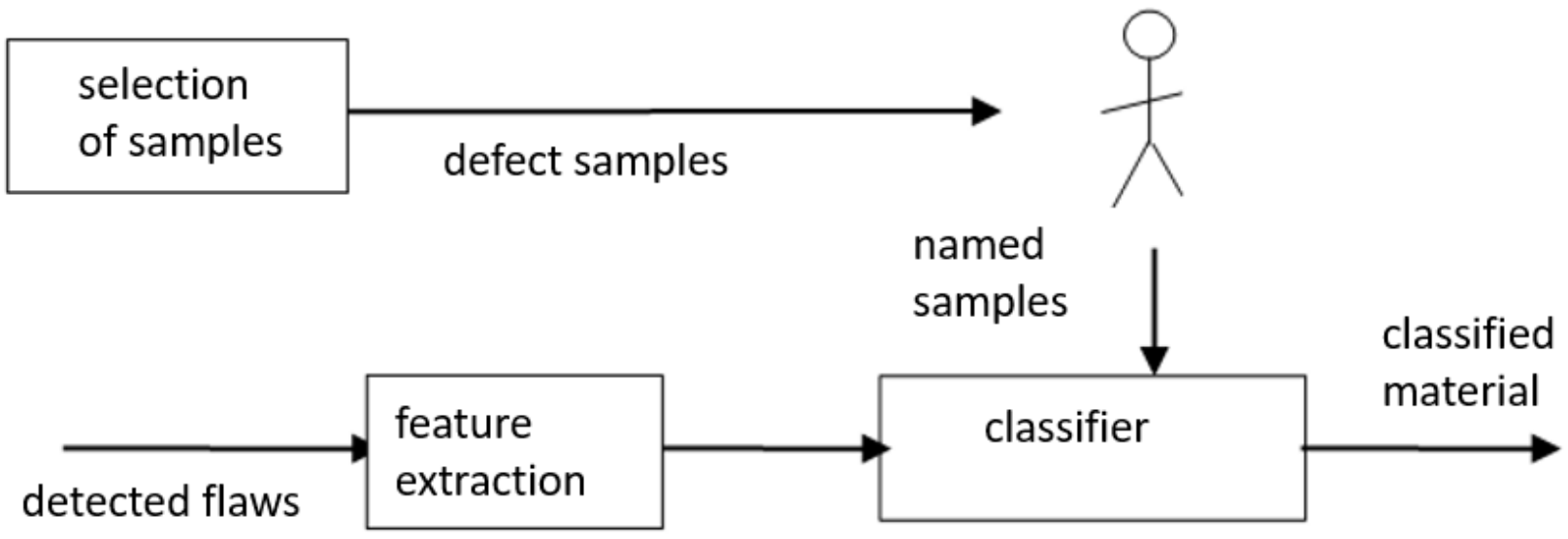

Figure 8.2. Training of classifier with human classified samples.

However, human selection is prone to error, so many of the samples are labeled incorrectly. In addition, exhibiting the training with samples is laborious and the training has to be repeated as the appearance of the material varies. This procedure is therefore rare in industrial systems.

Typically, industrial quality control systems are not really trained, but the classification parameters used by them are adjusted based on the result of the inspection. Figure 8.3 illustrates this principle, which addresses the function of a rule-based classifier. The problem is that the adjuster has to be familiar with the classifier's solutions, because the parameters do not always have an easily understood connection with the test result.

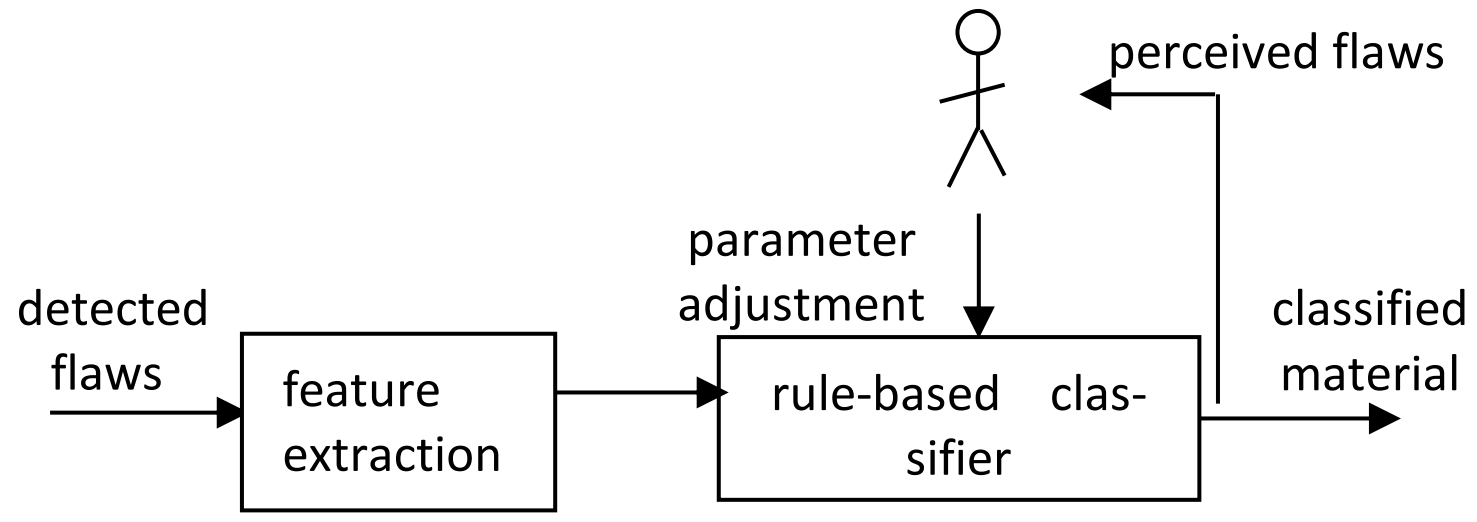

Figure 8.3. Rule-based classifier tuning by adjusting parameters.

The third option is the use of unsupervised training, whereby individual samples need not be named. The idea is to cluster samples isolated from the teaching material and visualize the resulting groups to the operator of the inspection system to determine class boundaries, as shown by the diagram in Figure 8.4.

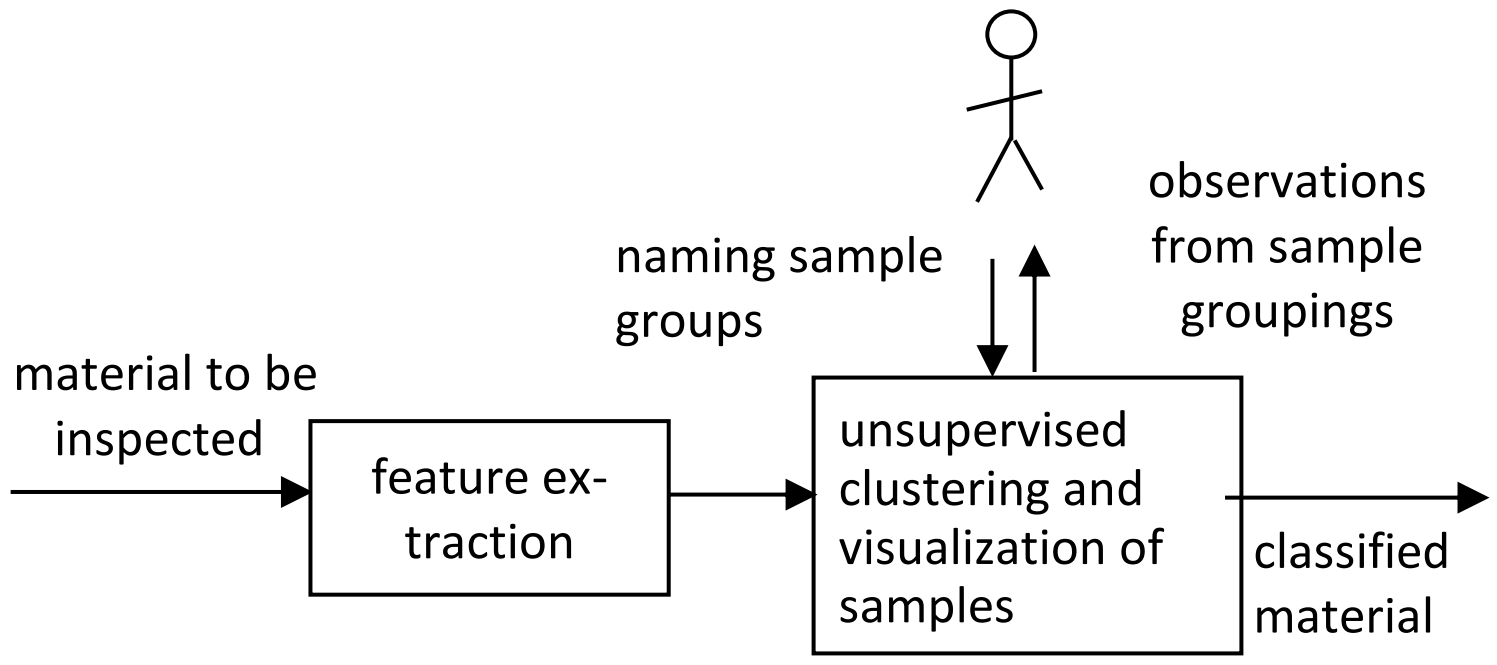

Figure 8.4. Training based on visualized clustering.

*8.1.2 Focus on visualization*

For example, a self-organizing map can be used for visualization. Re-training is easy, as is changing class boundaries. This solution has been found to significantly improve the accuracy of sawn lumber and steel inspection.

Figure 8.5 shows a grouping of different types of knots using a self-organizing SOM map developed by late Professor Teuvo Kohonen (Niskanen, 2003), (Kohonen, 2001). The solution laid the foundations for systems of leading wood surface inspection equipment developers and has worked well for Nordic wood species. Thus, the interest of Finnish system suppliers focused on machine vision based wood strength assessment techniques, for which our group has also developed new solutions (Hietaniemi et al., 2014).

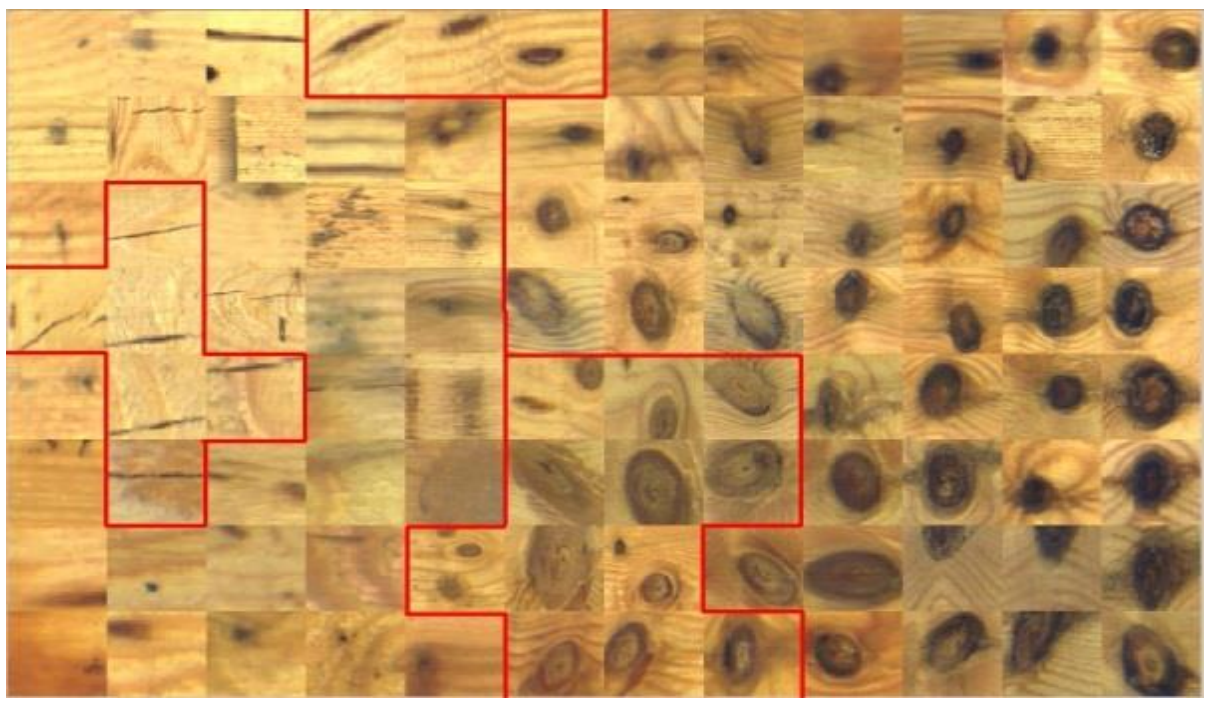

Figure 8.5. Grouping of knots with the SOM method. (© Acta Univ. Oul.)

Niskanen M (2003) A Visual Training Based Approach to Surface Inspection. Acta Univ. Oul. C 186, 125 p. http://jultika.oulu.fi/files/isbn9514270673.pdf

However, our recent collaboration with Brazilian scientists has shown that wood species in the south are different from those in the Nordic countries, sawmills handle more species and contain more habitat variations, which means that the SOM-based solution does not work.

Therefore, the selection of training samples must be made smarter and adequately matched to the wood material to be taught with a smaller number of training samples. Deep neural networks are clearly not suited to this problem as such. Partially supervised and active learning methods seem to be a good starting point. Newer tools, better than SOM, must be used to group and visualize non-linear unbalanced data (see Chapter 3)

## 8.2 Face Analysis and Biometric Identification

Face recognition and analysis is one of the most important abilities of a person in everyday life. Face analysis is used for different tasks. We can recognize other people in their faces, whether they are old or young, male or female, European or Asian. We can also see from the face whether a person is happy or sad, tired or nervous, whether he or she is speaking true or possibly lying.

Faces are very important in social communication between people. Emotional expressions communicate both emotional state and intentions related to behavior. Mouth movements are important clues in recognizing speech in distracted conditions and hearing impaired people can even read from lips. The position of the head and the direction of the gaze tell what the person is paying attention to at each moment.

Faces also tell about human health. For example, people with autism have difficulty understanding and expressing emotions, or pain can often be recognized by facial expressions. Facial information can also be used to find information that cannot be seen by humans, such as involuntary, very rapid micro-expressions and the measurement of heart rate from color video based on small variations in the color of the face.

Our own research on facial analysis began as early as the late 1990s. The first step in face analysis is usually so-called face detection, that is, to first find where in the image there is a face or faces. Thereafter, the processing can be applied to that image area and, for example, attempt to identify that person.

We developed a method for detecting skin pixels from a video image, using so-called skin locus. The solution takes advantage of the knowledge that the pixels on the skin are close to each other in a suitable color coordinate system. For this and other facial analysis research, we collected a Physics-based Face Database of 111 individuals under different lighting conditions. This database has since been used by numerous research groups around the world.

The actual breakthrough was presented in the previous section, i.e. an LBP-based method for facial representation and its applications to various problems, such face recognition, face detection, facial expression recognition, gender identification, age estimation, identification of spoken phrases from mouth movements, face anti-spoofing, and micro-expression recognition.

The following describes the application of the LBP method for fraud prevention (anti-spoofing) in biometric identification. In addition, we look at heart rate measurements from facial color variations, face analysis in human-machine interfaces, and visual speech recognition from mouth movements. Identification of micro-expressions by the LBP method is discussed in Chapter 9.

### *8.2.1 Biometric identification and spoofing prevention*

Biometric identity verification and recognition are important applications in artificial intelligence. Verification or authentication refers to whether the person in question is who he or she claims to be. Recognition means who or if any of the many individuals stored in the database are concerned. We want to get rid of user names and hard-to-remember codes.

Biometric identification offers solutions to this. Identification using physical properties may be based, for example, on a person's face, iris, fingerprints, blood vessels, or DNA. Person-based behavioral solutions use, for example, a person's way of interacting with a device, voice, or walking style (gait).

People recognize each other mainly on the basis of their faces. Indeed, face recognition is the most natural and widely used method for biometric authentication and identification, for example in checking passports at airports or controlling access to confined spaces or computers. In recent years, it has proven to be a major problem, both in face recognition and in other solutions, that biometric systems can be fooled in different ways.

Face recognition can be spoofed by showing to the system other people's pictures, video clips, or natural-looking masks. Figure 8.6 shows examples of the use of a photograph and a mask to gain unauthorized access to our group's research laboratory. A video demonstration of presentation attack detection is available (Web-Antispoofing).

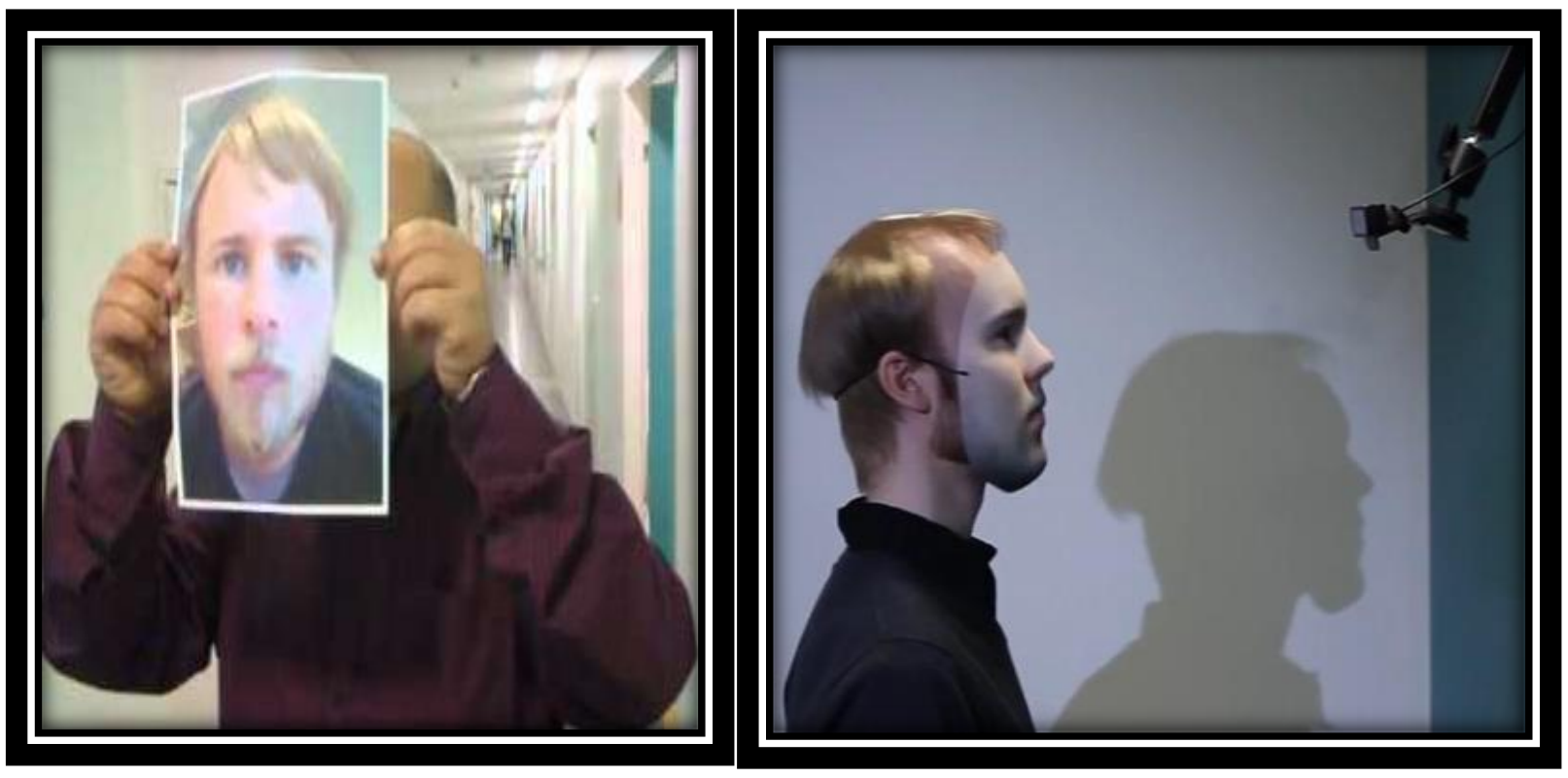

Figure 8.6. Cheating on identity verification. (© CMVS)

In 2011, we introduced an LBP-based solution for blocking face-based spoofing attempts. The rationale for this solution is that the faces captured by the camera are images of three-dimensional subjects, whereas photographs are two-dimensional subjects. The photo used for the scam features glosses, variations in the texture of the image, etc., which are not seen directly in the face. Figure 8.7 shows our original method (Määttä et al., 2011).

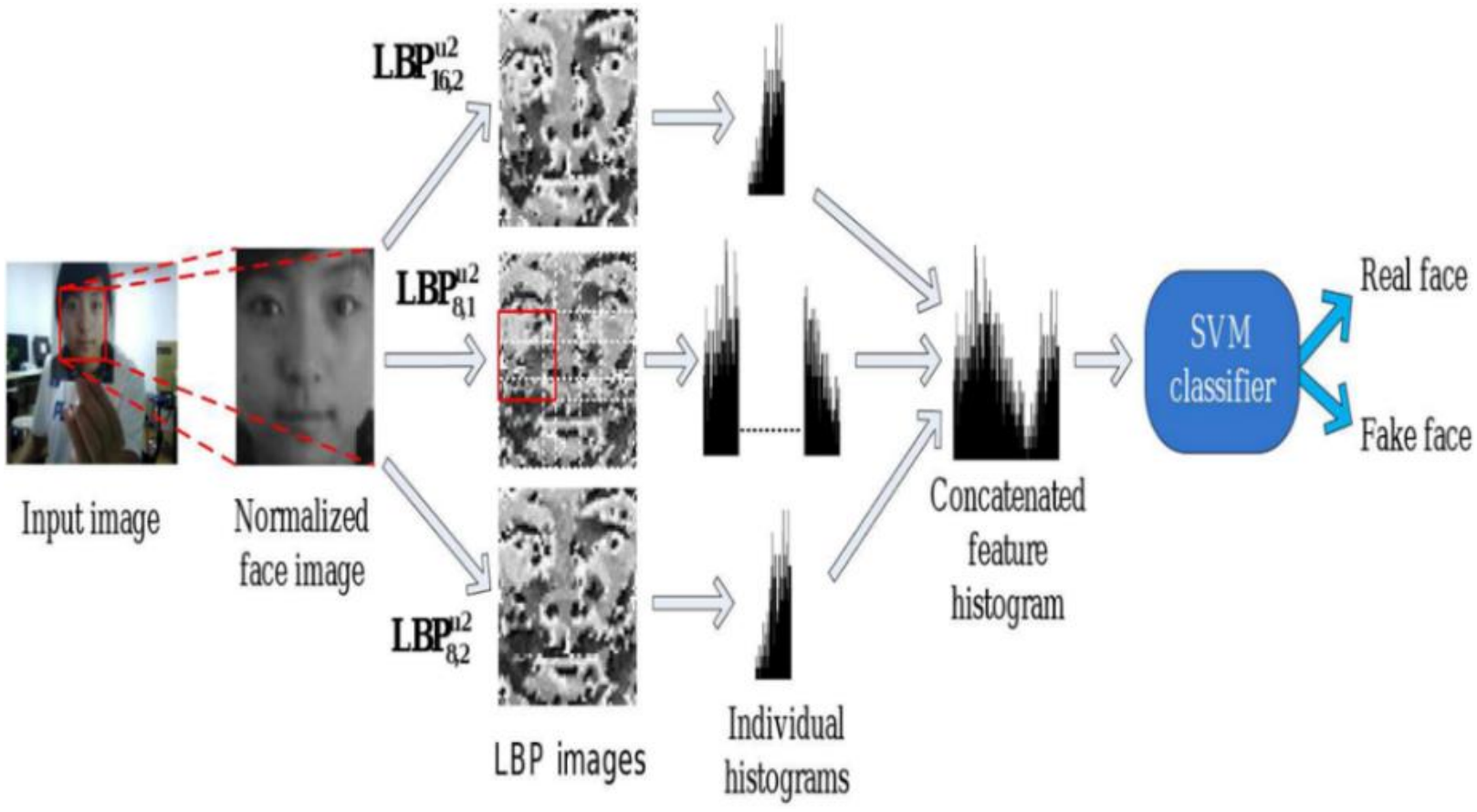

Figure 8.7. Spoofing detection with LBP method. (© IEEE)

The facial image under consideration is processed by LBP operators using different scaling and sampling modes, and their feature distributions are then combined. The system is taught beforehand with samples of true and false faces. The recognition uses a Support Vector Machine (SVM) to determine whether it is a true or a false face.

The method proved to be a major breakthrough. Several other research groups began to apply modifications to the aforementioned LBP method for both fraud prevention in photographs and video. Our publications presenting the method received two awards: the IET Biometrics Premium Award 2013 (Määttä et al., 2012) and the BTAS 2016 Five Year Highest Impact Award (Määttä et al., 2011). The method we have presented also motivated the use of LBP for anti-spoofing in connection with other biometric identifiers such as eye iris, fingerprints, gait, and even audio speech.

The use of masks has proven to be a very difficult problem, and the previous solutions suitable for photos and videos are not sufficient. For this problem, in 2016, we presented a method based on the application of a method based on applying the heart rate measurement method to be described below. A heart rate pulse can be detected from the real color variations of the face, but it cannot be detected through masks (Li et al., 2016).

### *8.2.2 Detection of heart rate from face video*

Traditionally, heart rate measurement requires devices that come into contact with a person. When measuring the cardiac electrical curve, electrodes, usually ten, are attached to human skin to help measure electrical potentials produced by the heart. Heart rate can be measured with sensors mounted on the chest or other parts of the body such as the wrist, finger or earlobe.

The interest in measuring heart rate remotely using a video camera started in the research community around 2010. The motivation was to develop a device that allows people to conveniently see their heart rate at home from the “health mirror” alongside their daily routine or, for example, when working at a terminal.

The first solutions were based on detecting very slight color variations of the face or on muscle movements (Li et al., 2014). The methods are based on the fact that the heart pumps blood at a rate of heart rate to the head, causing minor color changes and muscle movements on the face. A similar principle can also be applied to measure the respiratory rate through small color changes caused by variations in blood oxygen levels.

The problem with these solutions was that the methods developed worked well only under near ideal conditions. The illumination variations and the subject's movements hampered the measurement.

In 2014, we introduced at the IEEE Conference on Pattern Recognition and Computer Vision (CVPR), the leading computer vision conference, a method based on detecting tiny color changes which largely eliminated the shortcomings of the earlier attempts and was thus much closer to practical applications (Li et al., 2014). The principle of the method is shown in Figure 8.8.

Initially, in step 1, the Viola-Jones face detection method finds a face in the video image of a face (the area with yellow box). There are 66 landmarks highlighted in red, nine of which are used to find a region of interest (ROI) colored in blue. The area is well suited for detecting minor color variations. As faces move, landmarks are followed to monitor changes in the area of interest.

The RGB (Red-Green-Blue) camera's green channel is used to detect color changes, as it was found to be the best for this purpose in experiments. Step 2 compensates for ambient light variations by comparing face color variations with average background color variations. In step 3, elastic movements within the region of interest, such as those caused by changes in facial expressions, are eliminated. In step 4, e.g., the frequencies outside the normal heart rate range are filtered out from the signal (in our system, the heart rate is assumed to be between 42 and 240 beats per minute). The heart rate value is obtained by calculating the so-called power spectrum of the signal using Fourier transform. In the example of Figure 8.8, the heart rate is 67.2 beats per minute (Li et al., 2014).

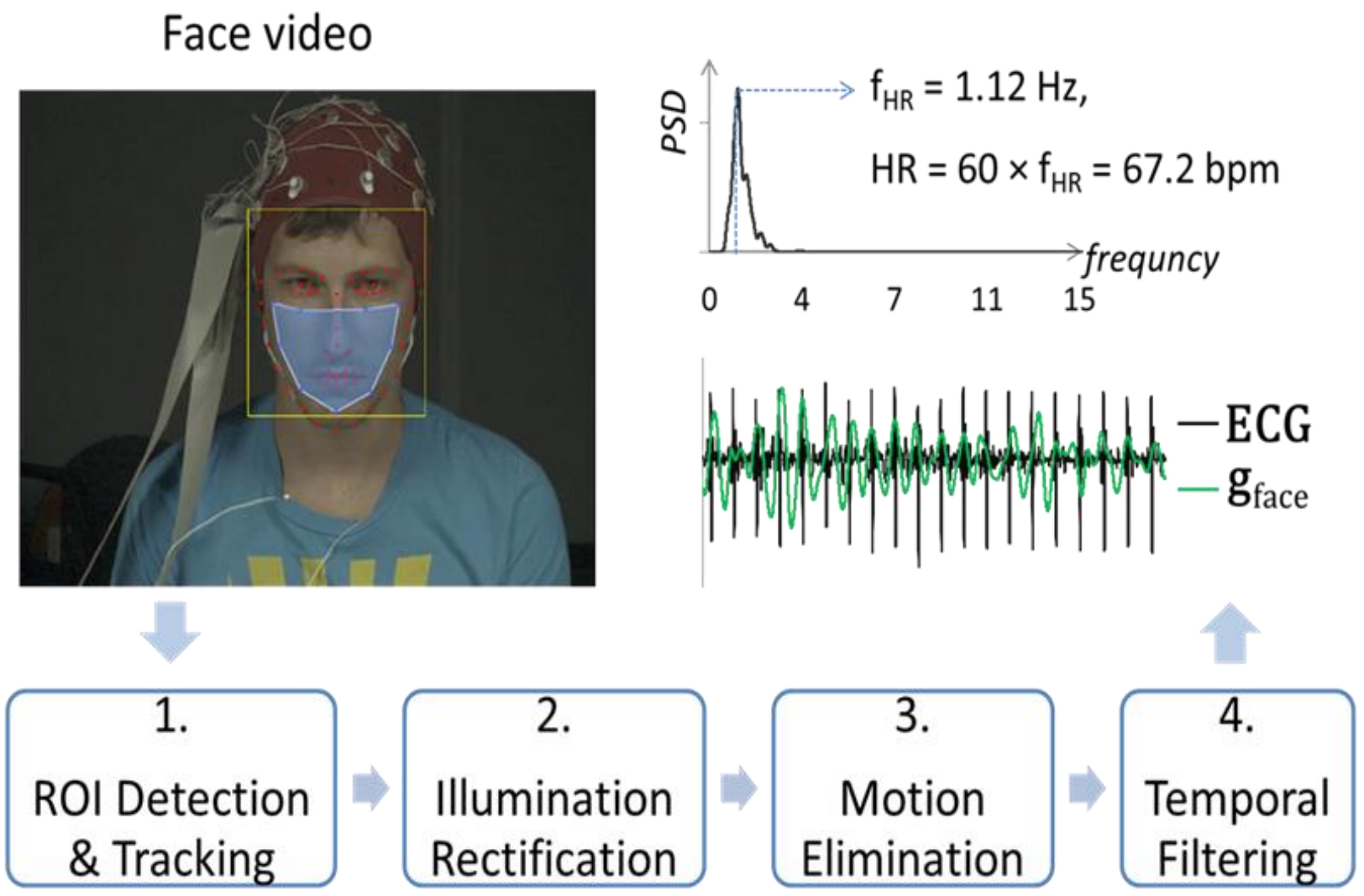

Figure 8.8. Heart rate measurement from face video. (© IEEE)

The developed method worked better in the tests performed than the corresponding previous methods. This was particularly noticeable with the use of the multimodal MAHNOB-HCI database, images under conditions of varying ambient lighting and some movement of the faces of the subjects.

The performance of the reference methods collapsed, but our own method still worked quite well. We also conducted interesting experiments for future applications, where we monitored the heart rate change of a test person playing a game on a tablet computer and compared the result with the estimates of a commercial heart rate monitor.

In addition to the aforementioned face mask scam prevention method, our unit is investigating the use of our method in telemedicine, with an area of application in early detection of atrial fibrillation. For this study, the Oulu Bio-Face Test Database (OBF Database) has been collected in collaboration with experts from the University of Oulu's Medical Research Center and Oulu University Hospital (Figure 8.9) (Li et al., 2018).

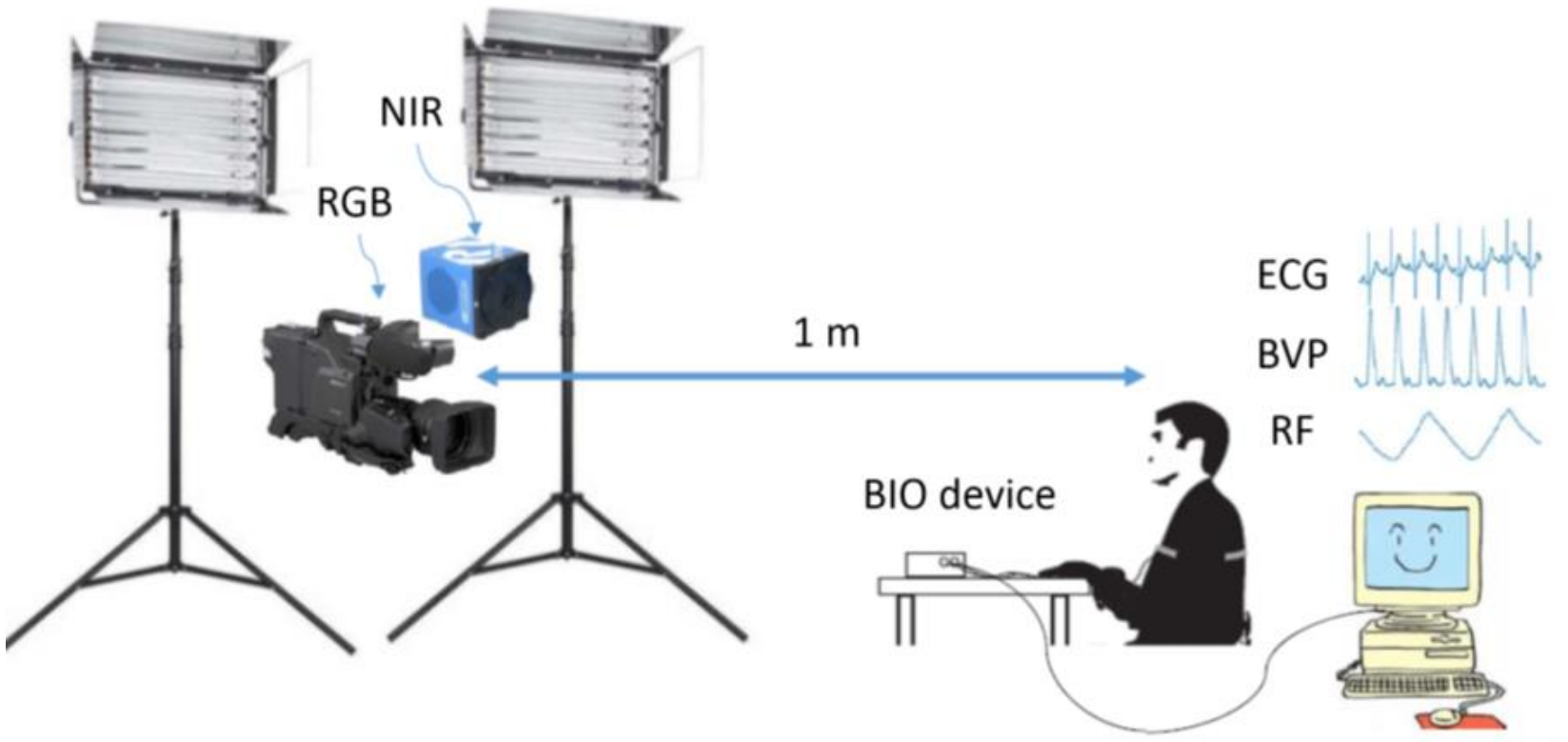

Kuva 8.9. Collection of OBF test database. (© IEEE)

From videos we are interested in measuring the heart rate (HR), heart rate variations (HRV), and respiratory rate (HF) of the subjects images (Li et al., 2018). In addition, reference measurements of these physiological signals have been made by conventional methods. Preliminary tests have shown promising results on video with measured HRV features - for the first time in the world. The above physiological signals are also important in analyzing the human emotional state. Emotional intelligence is discussed in Chapter 9.

### *8.2.3 Face analysis in human-machine interfaces*

In Figure 8.10, face analysis methods have been applied to the real-time human-machine interface developed by us (with Jukka

Holappa as the lead implementer). The system recognizes persons and their gender by the LBP method.

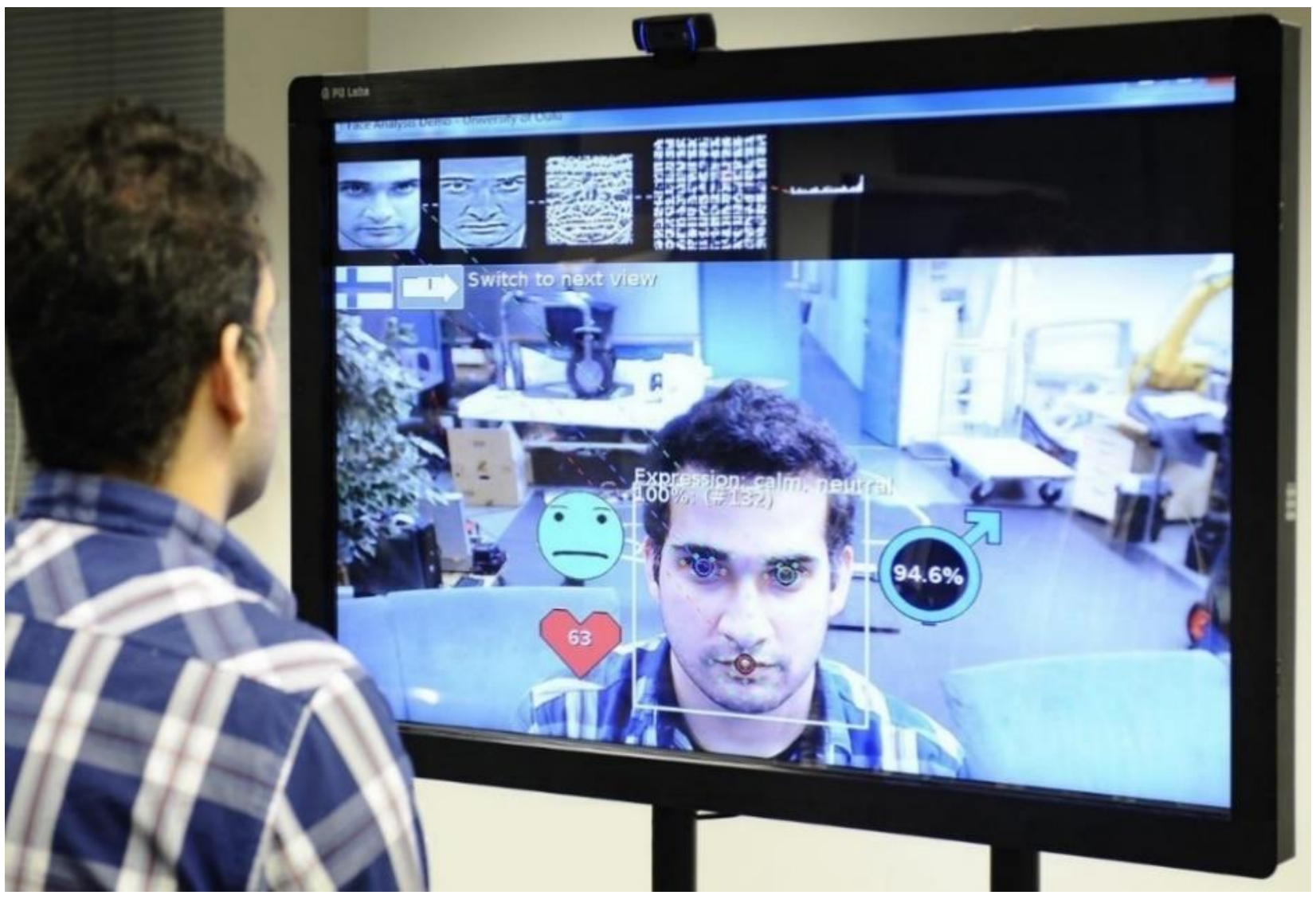

Figure 8.10. Facial analysis for human-machine interface.

In the picture, the machine is 94.6% sure a male is in the view. His heart rate is 63 beats per second, and the expressions detected from the movements of the landmarks on his face reveal that he has a neutral mind.

The system has been in continuous operation in the entry lobby of the research group, and was presented at various exhibitions and during so-called Abi-days for high-school students of Northern Finland. A real-time demonstration of the system being applied to Donald Trump and Hillary Clinton's election debate video can be found in the reference (Web-TrumpClinton).

The use of face analysis in smart glasses is also an interesting application. For example, a doctor could use facial information to aid in the diagnosis. Devices for aiding blind and other visually impaired people is another major application area. Intelligent glasses can be used to identify nearby people and gain insight into their emotions (Section 9.3). In Figure 8.11, the user (Jukka Holappa) wears Vuzix M100 smart glasses and is shown a measured facial view of what he sees.These glasses feature an Android operating system familiar to smartphones, a 1.2-gigahertz processor, a camera, a microphone, a speaker, a display, and a gesture interface. We implemented in the device the basic operations of facial image analysis, including face detection, tracking of a moving face, face recognition, gender identification, facial landmark detection, and facial expression recognition.

All computing was done on the smart glass's own processor, without wirelessly transferring images to external devices. Low image resolution achieved real-time operation, but due to the high CPU usage, the battery lasted only about an hour. This demonstrates the importance of developing machine vision or other methods of low power consumption in portable devices.

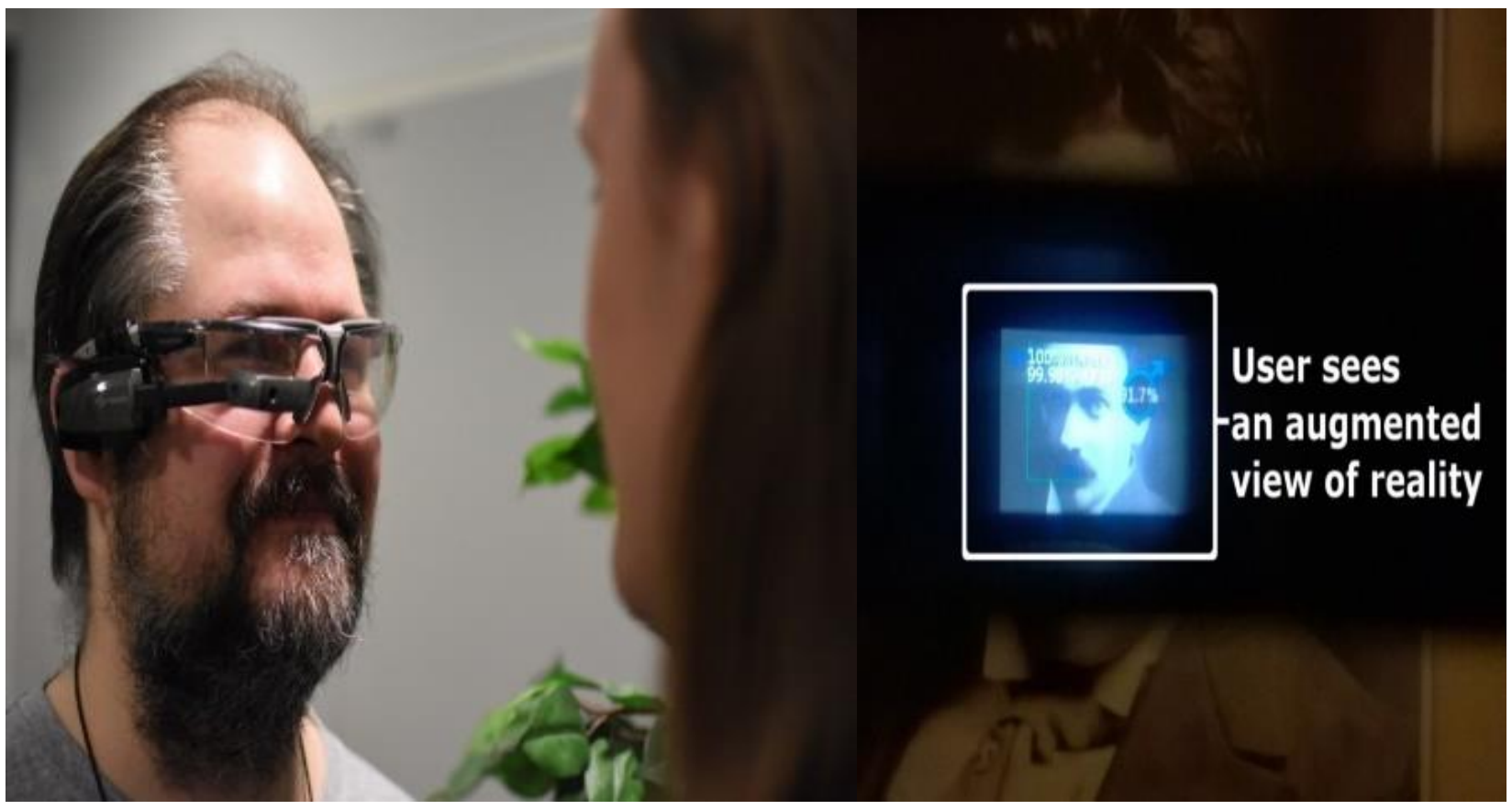

Figure 8.11. Face analysis with smart glasses. (© CMVS)

### *8.2.4 Visual speech recognition from mouth movements*

Human speech can also be read from the movements of the mouth and lips. This is often referred to as lip reading. For those who are hearing impaired, reading from the lips is a very important skill, and even under very disturbing conditions, it can improve audio-only recognition.

In the renowned science fiction novel and movie "2001: A Space Odyssey", the HAL 9000 robot was able to read from the lips of people on a spaceship. This subject has also been studied in computer vision. Initially, the goal has been to identify individual words or phrases.

Motivated by the excellent success of LBP-based face recognition, in 2006 we began to investigate visual speech recognition by applying the LBP-TOP method developed at that time to this problem. Figure 8.12 is a block diagram of our first system (Zhao et al., 2009).

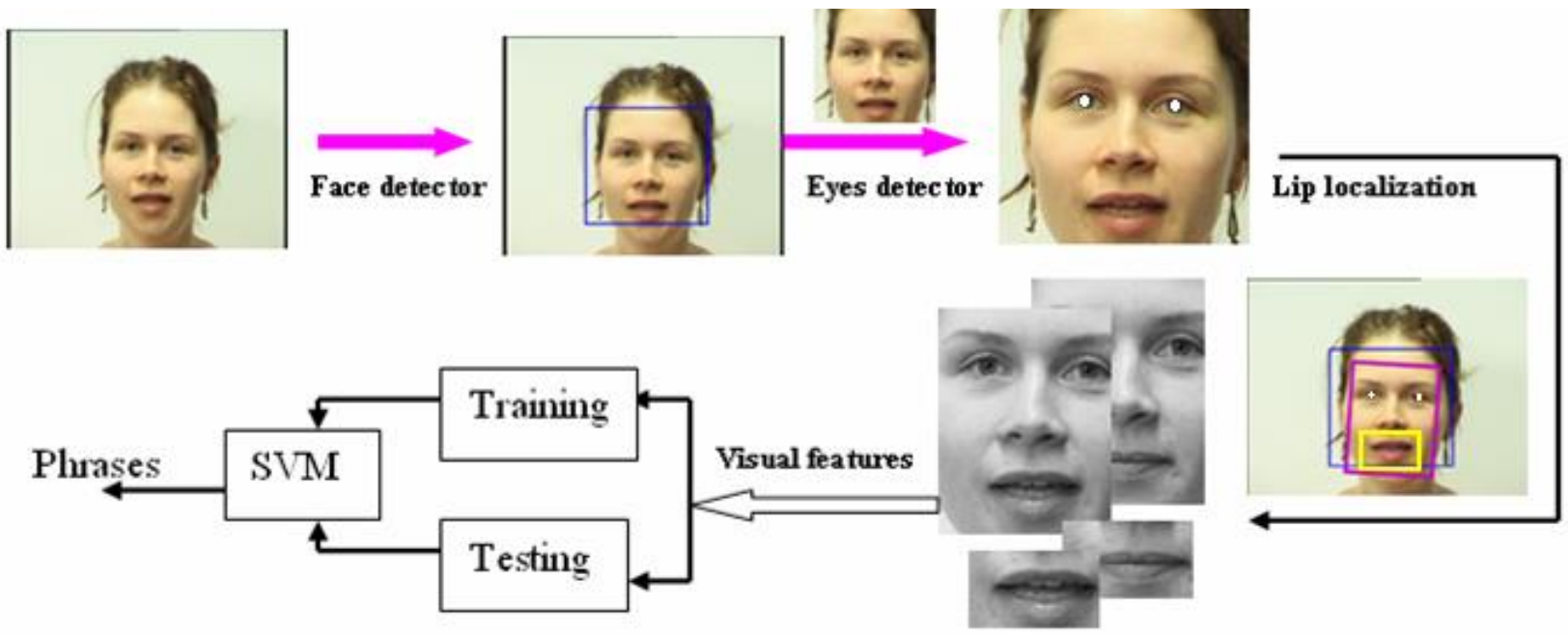

Figure 8.12. Visual speech recognition with LBP method. (© IEEE)

At first, the face is detected in the image and then the eyes in the face. Using this information, it is estimated where the mouth is located, assuming a close-up front view of the face. From the mouth area, the LBP-TOP feature distribution is calculated from the selected length of speech, Figure 8.13.

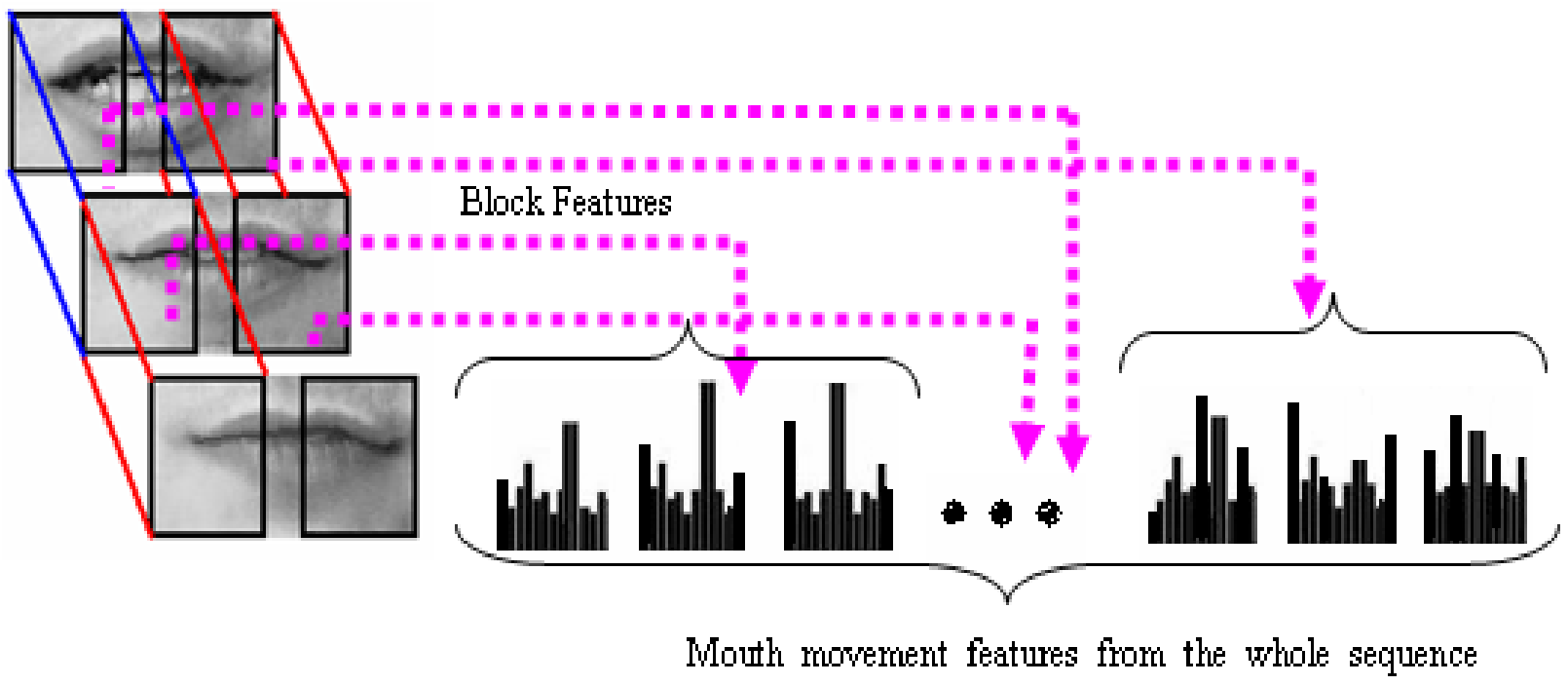

Figure 8.13. Description of mouth movements with the LBP-TOP method. (© IEEE)

The system is taught by phrases spoken by multiple speakers, for example, our original OuluVS database had a total of 817 video clips out of the phrases in Table 8.1, with 20 speakers pronouncing them one to five times.

Table 8.1. Phrases to be recognized (© IEEE)

| | | | |
|---|---|---|---|
| C1 | "Excuse me" | C6 | "See you" |
| C2 | "Good bye" | C7 | "I am sorry" |
| C3 | "Hello" | C8 | "Thank you" |
| C4 | "How are you" | C9 | "Have a good time" |
| C5 | "Nice to meet you" | C10 | "You are welcome" |

The original method achieved 60% recognition accuracy with this database (Zhao et al., 2009). Later, it was investigated how speech variations between phrase lengths and between different speakers can be better unified, resulting in more than 20% better results for this database and the best results published by then for other test databases. The results of our study were published in 2014 in the top-tier journal IEEE Transactions on Pattern Analysis and Machine Intelligence (Zhou et al., 2014).

Recently, motivated by deep learning research, recognition of complete sentences has been successful, for example through collaboration between Oxford University and Google Deep-Mind. One of the reasons for this success is that, due to a significant improvement in the accuracy of speech recognition, the massive collection of training material from mouth movements is straightforward from video footage of audiovisual faces.

Training end-to-end phrases in the same way as in acoustic speech recognition (see Chapter 5) has also greatly helped (Fernandez-Lopez & Sukno, 2018). However, there is still a long way to recognize continuous speech from mouth movements.

## 8.3 3-D Computer Vision and Augmented Reality

Much of the computer vision research has focused on analyzing two-dimensional images. However, we live in a three-dimensional world. Thus, machine vision must be able to analyze and interpret three-dimensional information: how far the obstacles are from a moving robot, what is the robot's location on the map, how to recognize objects from different angles, what is the three-dimensional structure of the object or environment.

Augmented reality represents real and artificial objects in the same 3-D environment. Wearable computers, such as smart glasses, have created a great need for such technology.

Our 3-D computer vision research began in the 1990s and was highlighted by the development of a new method and Matlab tool for precise geometric calibration of the camera. One of the most important subjects of the research led by Professor Janne Heikkilä is precise location of the user in relation to the operating environment. Other key research problems include estimating the structure of the 3-D view and creating a dense 3-D model using multiple images.

### *8.3.1 Camera calibration*

Geometry plays an important role in machine vision. The laws of geometry and optics determine how a three-dimensional environment is mapped to a camera's two-dimensional sensor. Understanding imaging geometry is thus important in developing image analysis methods. Geometric camera calibration is important because it is a prerequisite for accurate 3-D image measurements.

Calibration eliminates lens distortion in the image and provides a geometric match between the environment and the image (Figure 8.14). For this purpose, our group developed a highly precise camera calibration method and implemented a tool freely available on the Internet (Heikkilä & Silvén, 1997), (Heikkilä, 2000). The method has been widely cited and used by both the research community and industry.

The general principle is that the camera to be calibrated takes images from a test pattern or object whose pattern positions are well known. The information obtained can be used to determine a method that corrects geometric distortion in each pixel of an image.

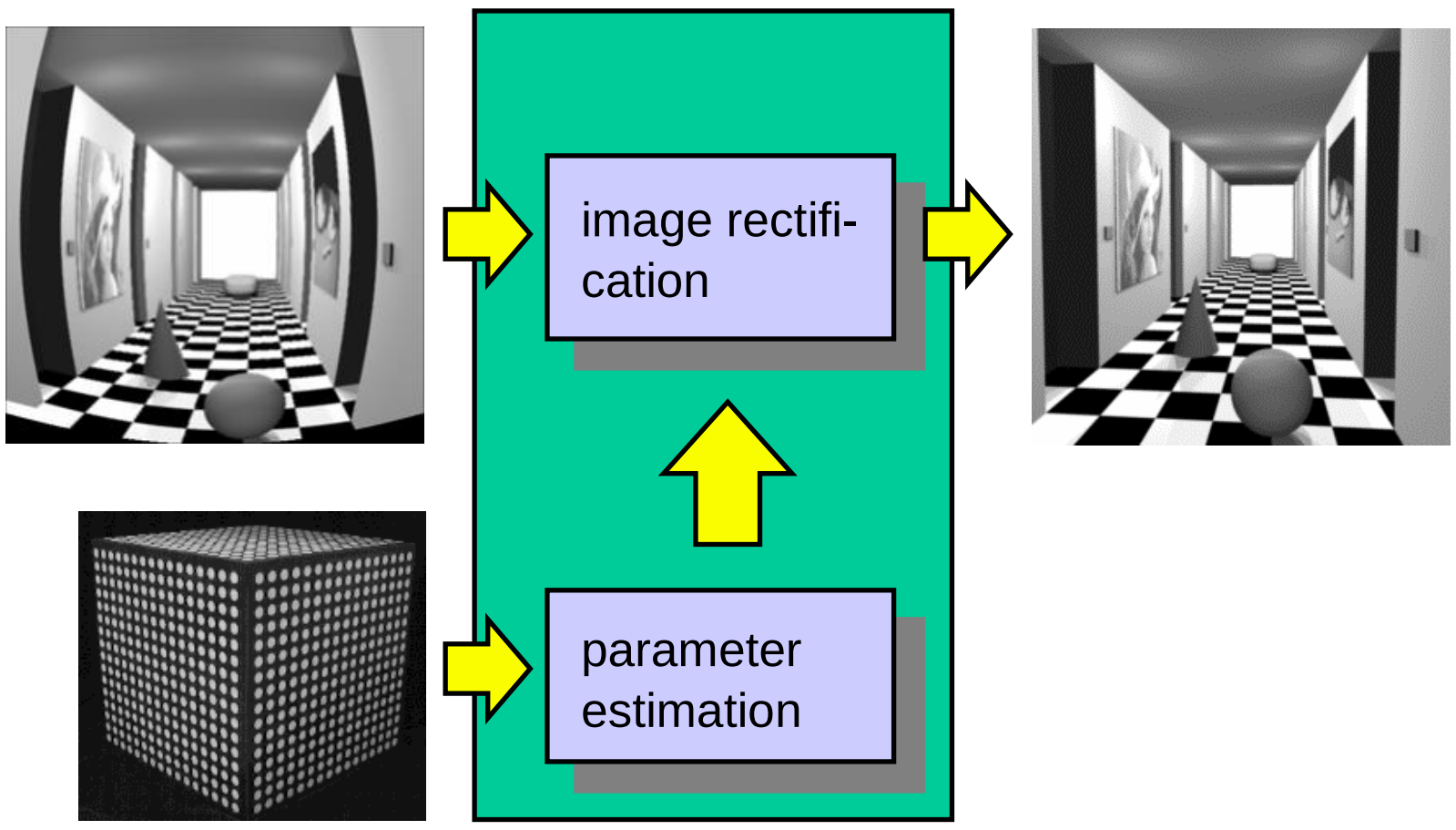

Figure 8.14. Geometric camera calibration. (© Janne Heikkilä)

Later, the group developed other new and widely used methods for calibrating cameras, the best known example being the method for calibrating a three-dimensional Kinect camera (Herrera et al., 2012). Kinect became known as the game controller for Microsoft's Xbox 360 system, and since then it and its later versions have been widely used in 3-D machine vision research.

*8.3.2 Creating a 3-D model*

Creating a three-dimensional model of multiple-camera images or a moving-camera image sequence has been one of the major problems with 3-D computer vision. Figure 8.15 takes standard images of the subject from different viewing points and creates a three-dimensional model of the subject using multi-camera stereo techniques (Ylimäki et al., 2015).

A well-known example using this type of technology is the publication "Building Rome in one day". It describes how to create a three-dimensional model of a city in a single day, with the help of numerous photographs taken by travelers from different locations.

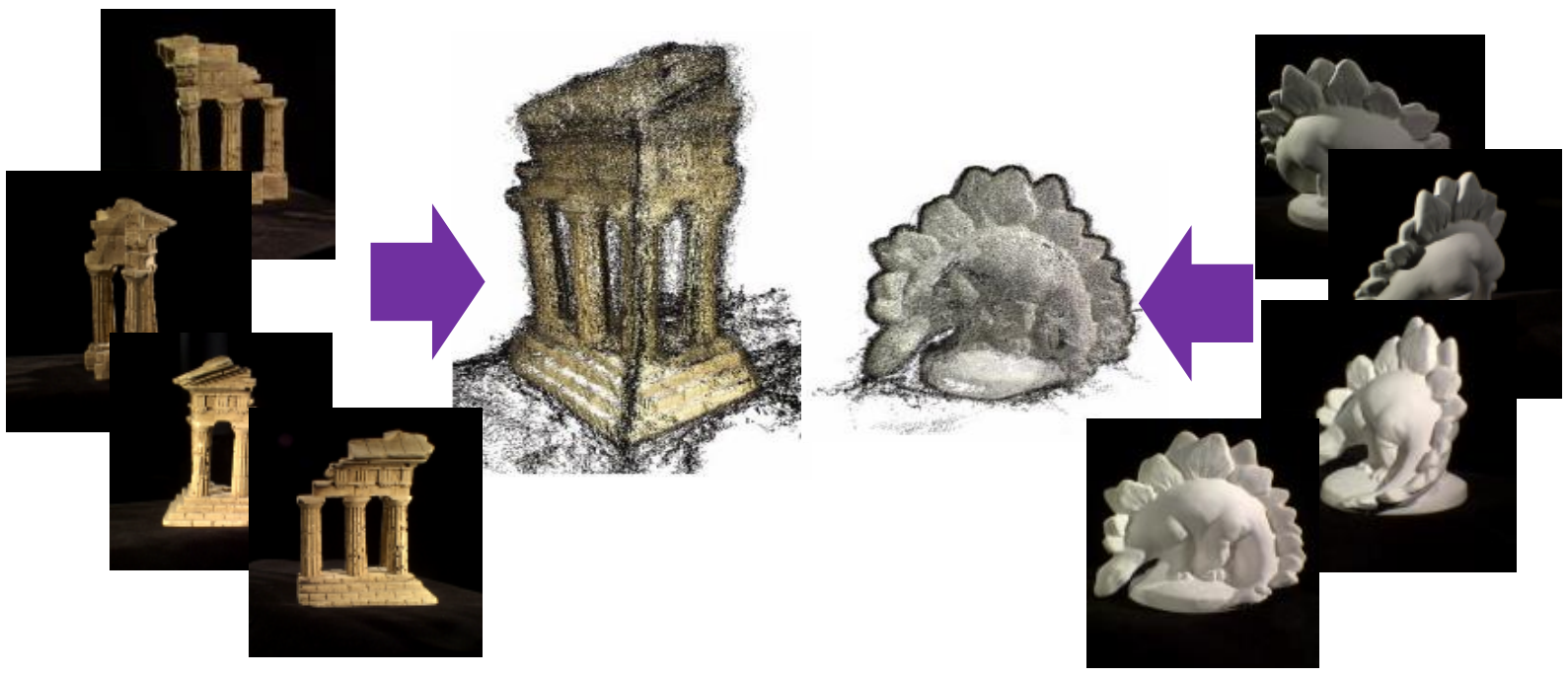

Figure 8.15. Creating a 3-D model using the multi-camera stereo method. (© Janne Heikkilä)

By searching for similar features in sequential images of a video sequence, it is also possible to create a three-dimensional model of the subject to be imaged, referred to herein as "structure from motion". However, using similar points, it is not possible to get a dense 3-D model, but a sparse model.

One of Professor Heikkilä's recent research interests has been the use of a sparse model produced by a moving camcorder and a distance-based range camera to create an accurate 3-D model. An example of using Google's Tango system for this was shown in Figure 6.12.

In addition to the traditional geometric machine vision, methods based on deep learning have recently been introduced for determining the position of the camera and for 3-D reconstruction. A key benefit is their greater reliability in situations where the view has little point-to-point correspondences as required by conventional methods. These new types of methods are also actively explored in our unit.

*8.3.3 Towards augmented reality applications*

Augmented reality is a significant application area for 3-D vision. Figure 8.16 shows an example of adding textual information about the environment the viewer sees in the scene captured by the camcorder.

Augmented reality combines artificial, or virtual, reality with a real-life camera view. Our team has in recent years focused on 3-D machine vision techniques that provide even better features for augmented reality applications.

A well-known example of augmented reality is Pokemon Go. However, so far, application development in games has been limited by the lack of user experience and the limitations of display technology.

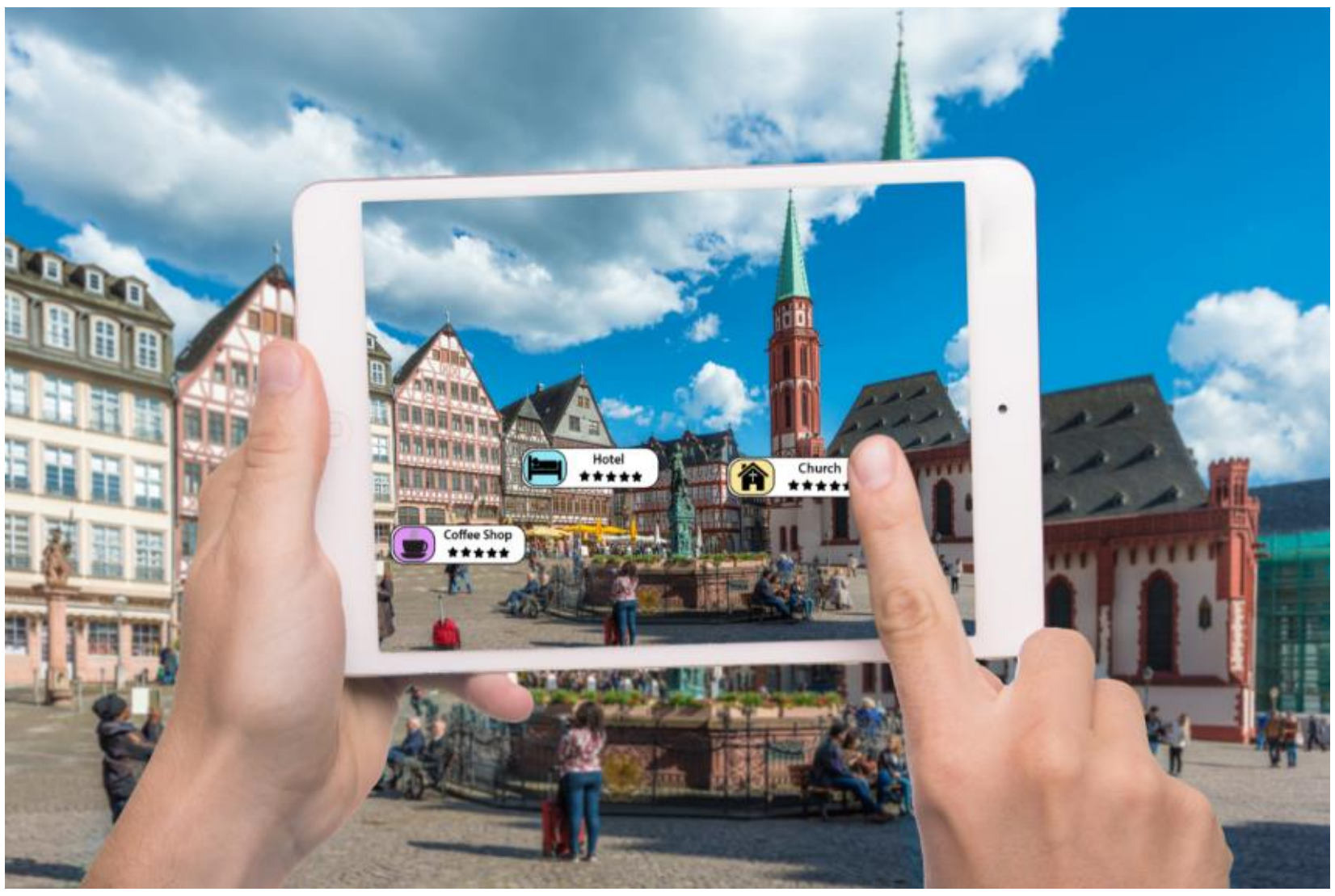

Figure 8.16. An example of augmented reality. (© 123RF)

The HoloLens glasses introduced by Microsoft are a new step forward, enabling the inclusion of three-dimensional holograms in the user's field of vision. According to Janne Heikkilä, such technology can be utilized in, for example, maintenance. If there is an exact three-dimensional model of the item being serviced, such as a building, the location of the hidden structures, such as pipes or wires, can be visualized by the repairman.

As another augmented reality application in the near future, he mentions in-vehicle aids for adding virtual objects to the windshield, such as signage and weather or driving information.

## 8.4 Machine Perception in Human-robot Interaction

In robotics, computer vision can be applied to various materials handling, sorting and assembly tasks, for example in the automotive industry. An ordinary robot must be carefully programmed in advance for selected tasks, as unexpected changes in the working environment cause malfunction. Visual and other senses make robots more adaptable, resilient and more independent. There are many such applications that utilize relatively simple machine vision.

The use of machine vision in the navigation of mobile machines and robots has also been the subject of much research. Potential applications include, for example, mining and forestry machines, as well as machines that are usually used in hazardous or unpleasant conditions (underground, underwater or in space). Future cars will also increasingly use machine vision to help the driver.

The development of more intelligent robots requires the combination of many artificial intelligence technologies. The machine must be able to percept its environment using, for example, visual and tactile senses, interpret sensory information, design free paths for movement of the hand or entire machine, perform the necessary mechanical movements, react quickly to unexpected situations, etc.

Our research group has been doing research on intelligent and autonomous robotics since the 1980s. Initially, the focus was on the development of robot platforms and solutions utilizing simple machine vision and other sensory information. In the 2000s, the focus shifted to human-robot interaction.

### *8.4.1 Intelligent robots*

The development of autonomous mobile robots began in the 1960s. The first independent mobile robot was the "Shakey" robot developed at SRI's Artificial Intelligence Center in 1966-72, which was able to sense its environment and make decisions based on sensed information. Significantly influencing on modern robotics, this robot was able to perform tasks in a laboratory environment that required planning, route searching, and repositioning simple objects in the environment.

In our group, in the early 1980s there were enthusiastic graduate students who developed as their "hobbies" a micro mouse, a mechanical miniature robot looking for its path in the middle of the maze. They used a simple ultrasonic radar and light sensors to sense the environment. National and international competitions were held on the topic, with the current professors Olli Silvén and Juha Röning and their teams doing well.

Based on this hobby-based background, we began to develop autonomously moving vehicles as a part of Juha Röning's dissertation work. The project developed a Controlled Test Vehicle (CAT) robot that was able to move and avoid obstacles in a laboratory environment using the three-camera stereo method developed by us at the University of Maryland to acquire three-dimensional information about the environment (Figure 8.17).

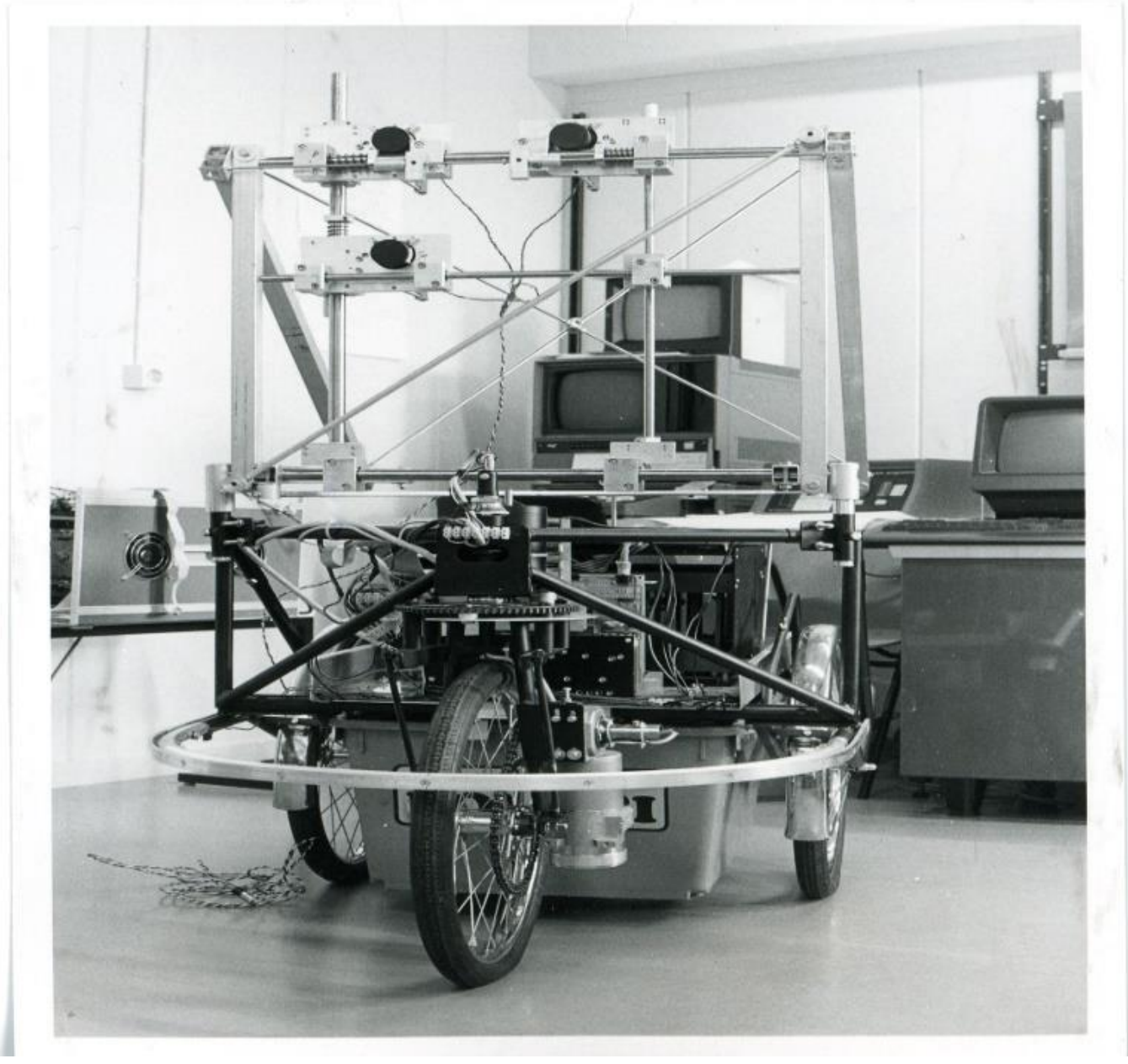

Figure 8.17. CAT mobile robot.

Significant applied research in the field of intelligent machines was carried out in Oulu already in the early 1990s in connection with the VTT-University co-operation project “Machine of the Future”, funded by Tekes and the industry. It sought to increase autonomy by applying advanced senses and reasoning techniques.

The industrial pilot was to increase the autonomy of a paper roll loading manipulator operating in the harbor environment. The system utilized 3-D modeling of the environment based on advanced sensors, such as laser depth measurements, and a design-execution-monitoring principle to control the system. The intention was not to replace the manipulator's supervisor, but to make his or her work easier and more efficient. One of the significant results of the project was education. The research related to this project involved several young researchers who later made a significant career: current professors Visa Koivunen, Jukka Riekki and Juha Röning, as well as docents (adjunct professors) Tapio Heikkilä and Kari Pulli. Later Dr. Pulli created an excellent career in Silicon Valley.

Today, a paper roll manipulator could be much more advanced than nearly 30 years ago. Machine senses are now much better and more reliable, providing a much more accurate description of the 3-D environment.

The machine could now navigate autonomously in its operating environment, avoid obstacles, and handle the loading task independently. However, people would need to make sure that everything goes according to plan and deal with problem situations - possibly with multiple manipulators operating simultaneously.

### *8.4.2 Human-robot interaction*

Developing intelligent machines and robots that interact with humans has been one of the great challenges of artificial intelligence research. Inspiration for this research came from the 1968 science fiction film “2001: A Space Odyssey”, directed by Stanley Kubrick, based on the book by Arthur C. Clarke. It introduced the HAL 9000 computer, which was able to hear, speak, design activities, recognize faces, see, evaluate facial expressions, give artistic judgments, and even read speech from lips.

Machine vision plays a key role in the development of natural human-machine interfaces resembling this (Figure 8.18). One important application is the various service robots that are coming into our daily lives. Such robots would cope with many of the routine tasks previously performed by humans. Because they operate in the same environment as humans, humans and robots need to be able to interact in the same way that humans interact with each other - this  is often called natural human-machine interaction. For example, a service robot could guide visitors to the museum or help the elderly at home.

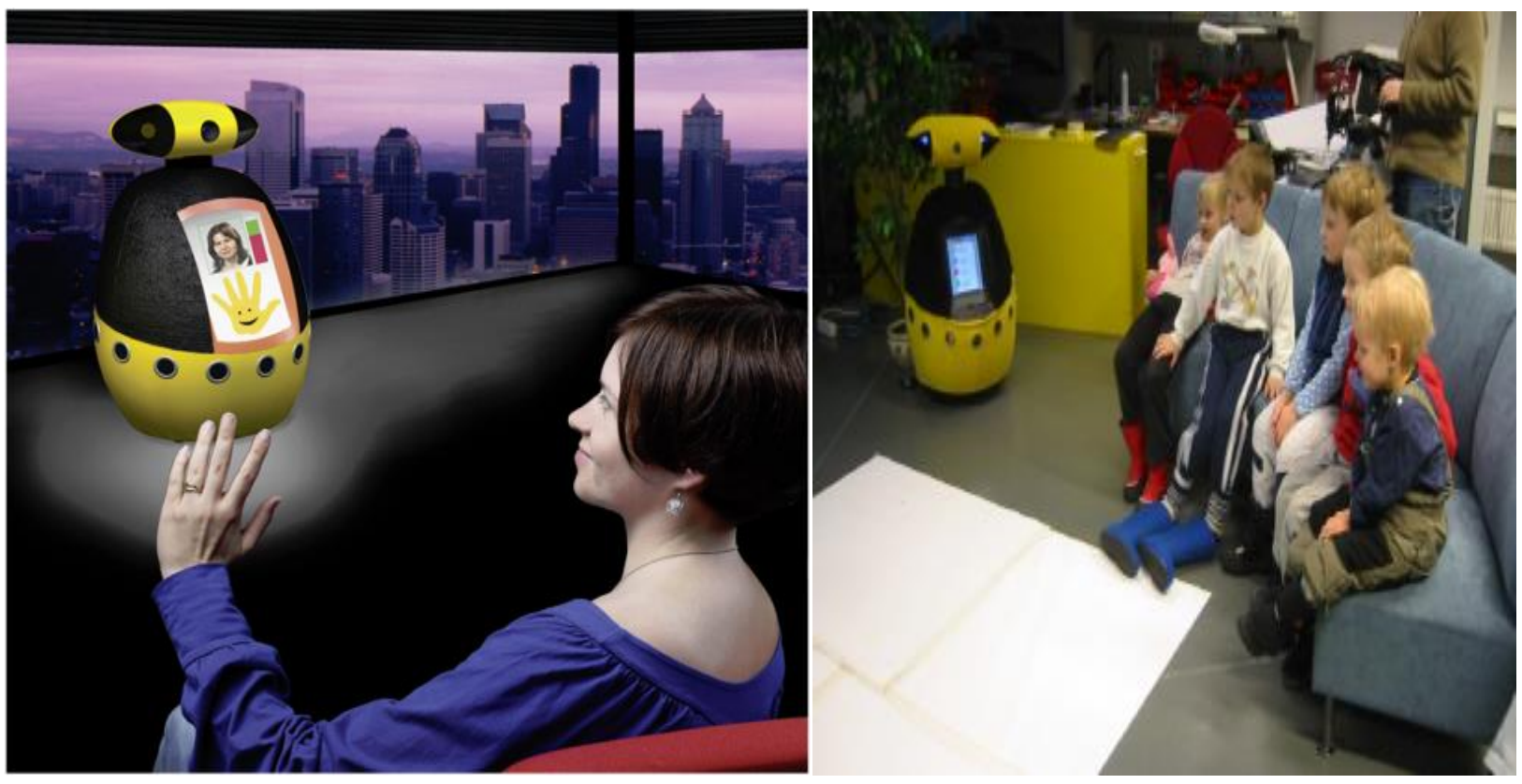

Figure 8.18. Human-robot interaction should be easy and natural. (© Jukka Kontinen)

Such a service robot must be able to find and preferably also identify its user. In this way, it could safely provide personalized service to its user and should recognize the user's emotional state in order to be able to have an affective, emotionally responsive interaction.

The robot should also be able to communicate with its user and understand the given commands by recognizing speech and gestures. Simple gestures can also be used to communicate remotely.

The robot must also maintain a natural response based on the information provided by its observations, for example by means of an avatar display resembling a human face. The robot must be able to learn its behavior and the tasks it is supposed to do.

In line with these principles, in the early 2010s, with the support of the Academy of Finland and the European Regional Fund, we realized a mobile robot called Minotaurus (Röning et al., 2014). We collaborated with the robotics group led by Professor Juha Röning.

Figure 8.19 shows the sensory system of the Minotaurus, which consists of a standard camcorder, Kinect cameras providing distance information, a microphone system, and an audio-visual display. The robot was connected to the sensory network system of our laboratory in order to observe its movements in different locations of our workspaces.

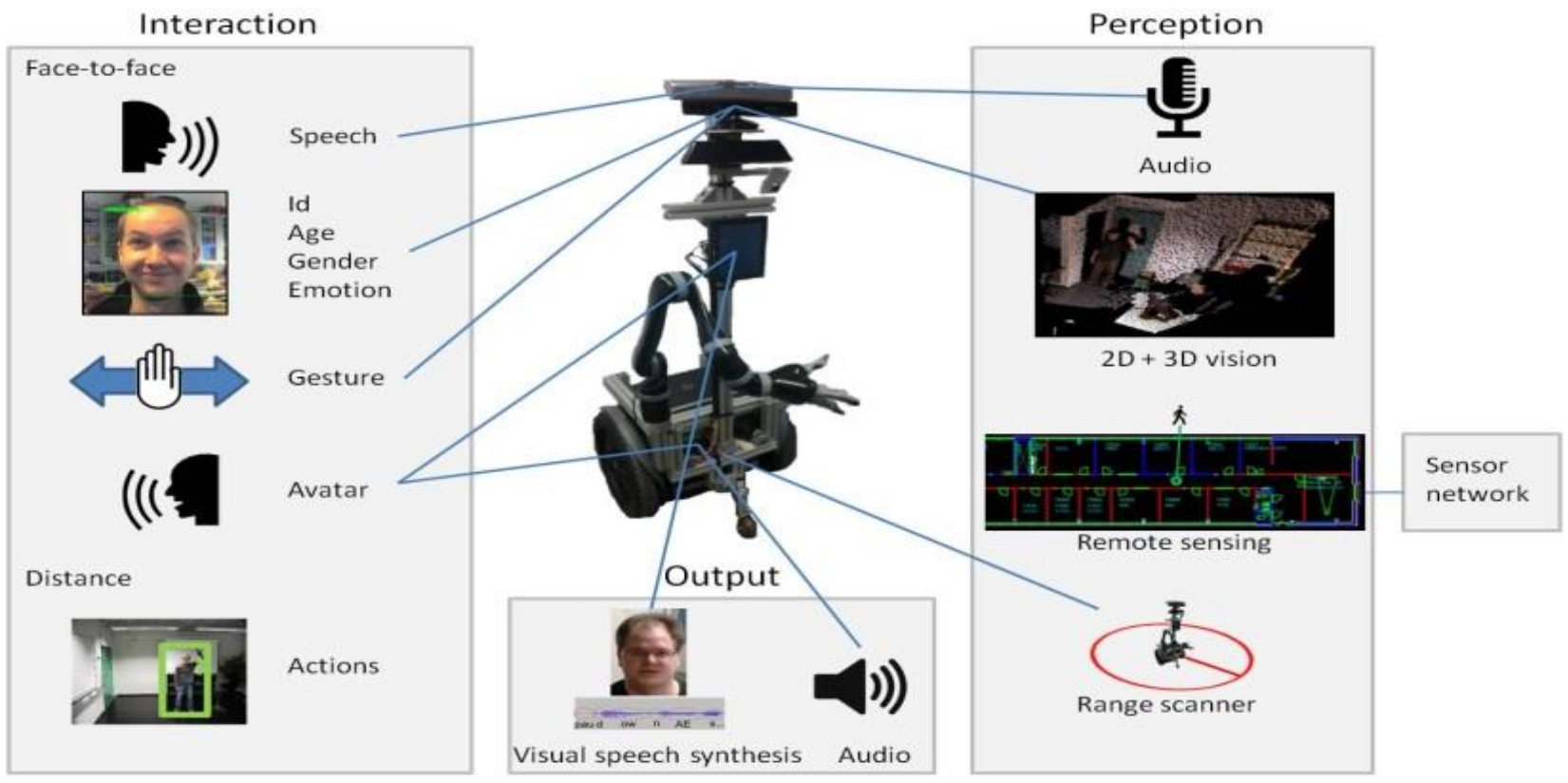

Figure 8.19. Minotaurus robot for investigating human-robot interaction. (© Springer)

Reprinted, with permission from Springer [Röning J, Holappa J, Kellokumpu V, Tikanmäki A & Pietikäinen M (2014) Minotaurus: A system for affective human-robot interaction in smart environments. Cognitive Computation]

Figure 8.20 depicts communication with the Minotaurus in our laboratory premises. The Kinect 3-D camera is located above the screen. If possible, the identity of the person communicating with the robot will be recognized from the face. This can be used to personalize the robot's services to the needs of that user. If the person is unknown, in which case identification fails, it is still possible to measure from a facial image so-called soft biometrics such as gender, whether he or she is a child, young, middle-aged or old, whether he or she is European or Asian, etc. Such information can also be used to personalize the services of the robot according to the user. For example, the response of a robot to a child should be different from that of an adult.

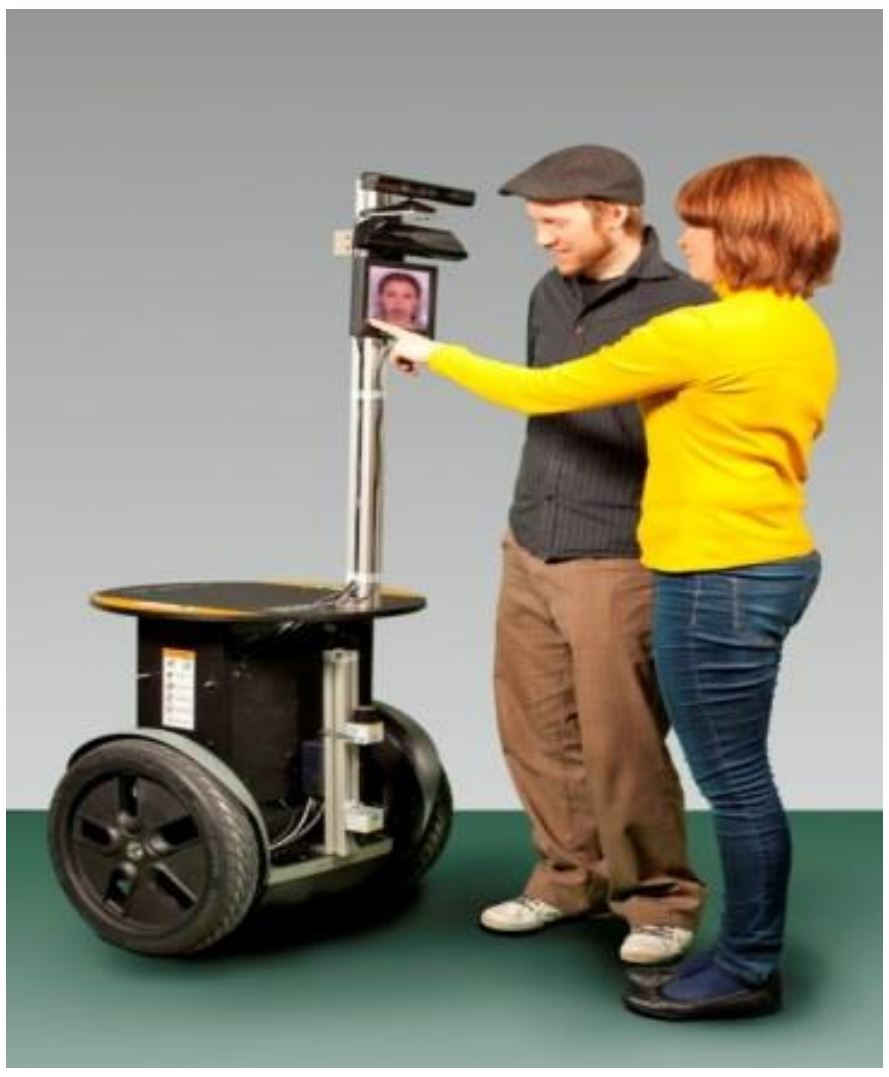

Figure 8.20. Testing human-robot interaction. (© Jukka Kontinen)

In order to engage in an emotive interaction, the machine must be capable of the same type of conversation as human interaction. For example, the robot should prefer to approach people who are at good mood rather than those who do not seem to be interested.

The Minotaurus used facial expressions for simple, albeit unreliable emotion recognition, able to recognize, alongside a neutral look, clearly stated prototypic expressions of "happy", "sad", "surprised", "disgust", "angry", or "fear". Better results would have been achieved with the latest methods developed for spontaneous facial expressions and also taking into account the user's speech and gestures.

It was also possible to talk to the robot with speech, gestures and actions. The first two were used in close-range and the third from a distance. For speech recognition, we used the freely available PocketSphinx (https://github.com/cmusphinx/pocketsphinx) system, with support for both Finnish and English. Small microphones arranged in a matrix provided information on the direction of the speaker. That way, it was known who the robot was talking to, in front of the camcorder or the Kinect depth camera familiar from Microsoft's Xbox 360 game controller.

Gesture recognition is a wide area of research in computer vision, with significant application areas such as sign language recognition for hearing impaired people. There is no common body language and different gestures have different meanings in different cultures. Our robot used simple shapes or movements as gestures, which were taught to the system beforehand by "showing" examples of each gesture. The user was able to select various gestures for the system by displaying them to the Kinect camera during the training phase. Each of the taught gestures could then be used as a simple control command for the robot.

People use their movements for a variety of communication in a very natural way, for example, a stranger can be greeted remotely by waving hands. A similar type of function was developed for Minotaurus, which allowed, for example, human-robot interaction to be initiated remotely.

Because the image produced by the camcorder was too inaccurate for further detailed action analysis, we used a very simple solution based on the LBP method we had previously developed (Kellokumpu et al., 2011). Figure 8.21 clarifies this (Röning et al., 2014).

The person seen by the robot is searched for in the image and LBP-TOP histograms of motion are computed in four areas. These can be used to identify the system's pre-taught actions, such as *don't come* or *come* here.

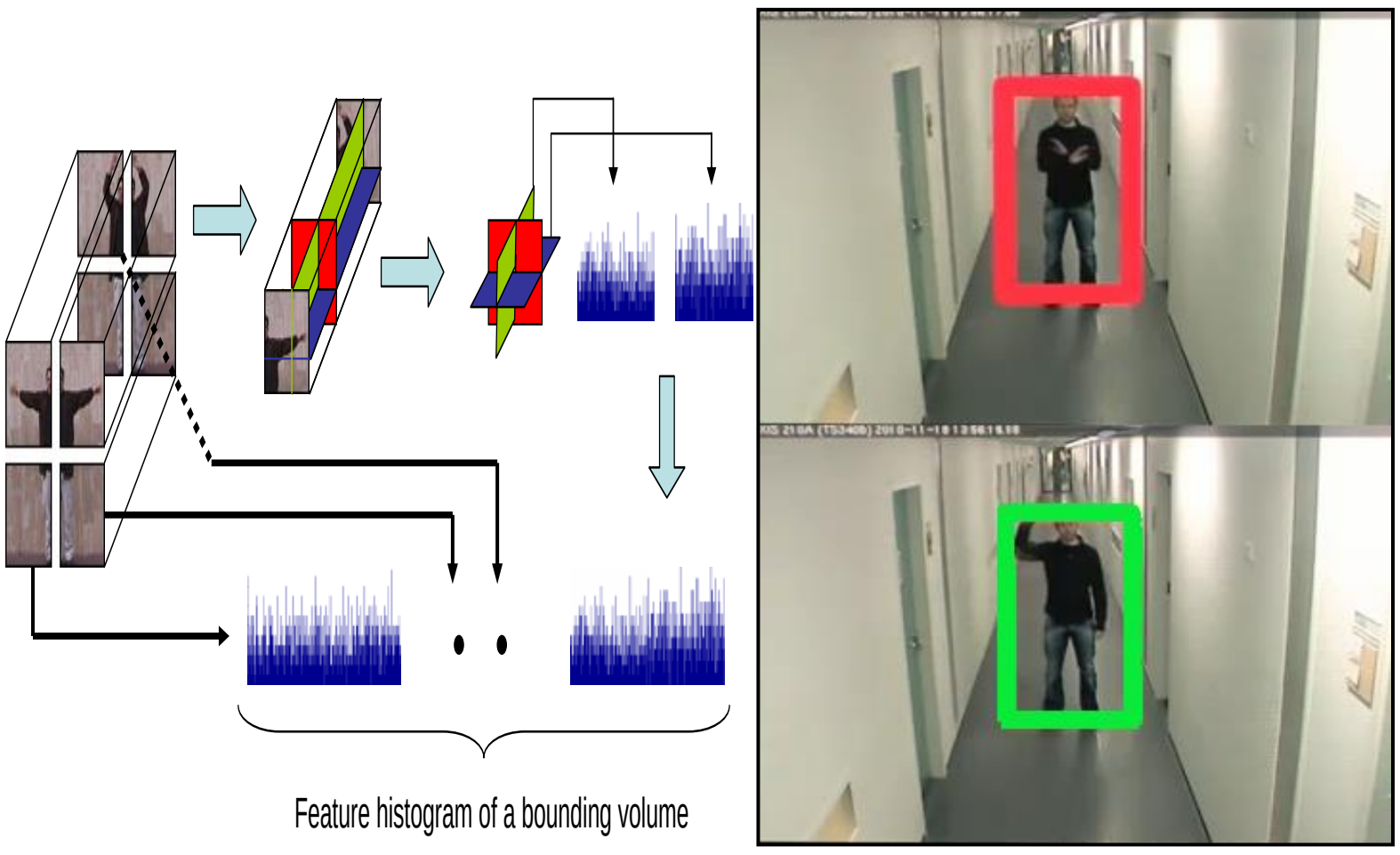

Figure 8.21. Recognition of simple human actions with the LBP-TOP method. (© Springer)

The robot responds to the user with a talking avatar. Such approach has been found to facilitate human-machine voice interaction. We used a simple system developed by our research team that associates the talking mouth region with a constant face background (Zhou et al., 2012).

The system, trained using a publicly available audiovisual video database, is capable of producing normal-looking speech visualization in the mouth area, be it Finnish or English (Figure 8.22).

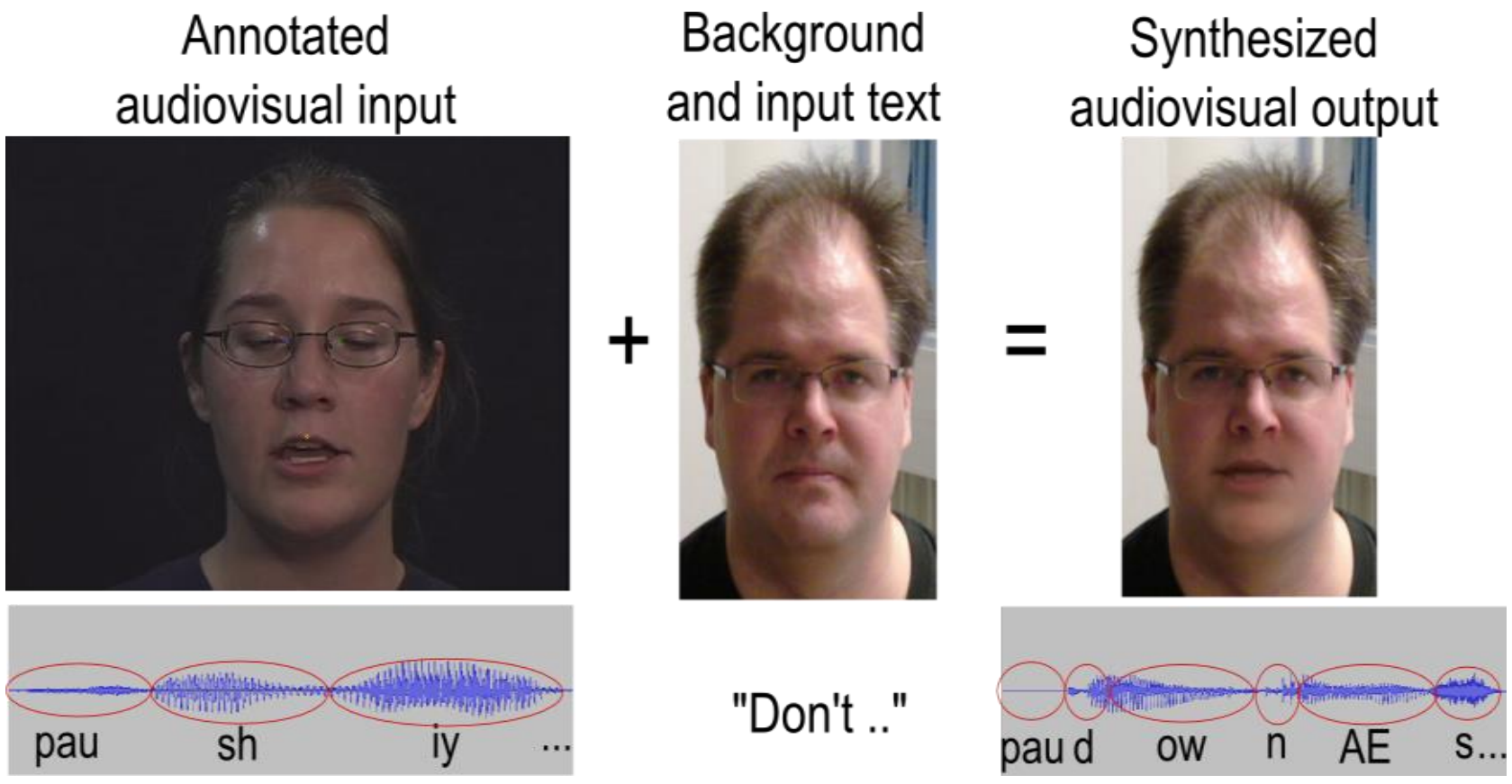

Figure 8.22. Audiovisual face animation. (© Springer)

Reprinted, with permission from Springer [Röning J, Holappa J, Kellokumpu V, Tikanmäki A & Pietikäinen M (2014) Minotaurus: A system for affective human-robot interaction in smart environments. Cognitive Computation]

We made various demonstrations with the system. Below is an example of a conversation between a human and the Minotaurus robot. A video demonstration of the communication can be found at (Web-Minotaurus).

<Human> Hello!
<Minotaurus> Hi!
<Human> Are there any news?
<Minotaurus> Yes, there are totally twenty feeds.
Give the number you want to hear.
<Human> Nineteen
<Minotaurus> OK, the topic of the news number nineteen goes like:
Saudi Arabia to import Finnish education expertise.
<Human> What is the weather like?
<Minotaurus> Oulu Finland weather: minus seven Celsius, overcast
<Human> Bye!
<Minotaurus> Bye! Have a nice day!

The system we developed contained most of the basic components that could be designed, for example, in a service robot operating in an intelligent environment. It would have the ability to move around in an indoor environment. It could use its senses to recognize people, their expression, their roughly defined age, their gender, gestures and actions. It could also recognize speech and have an ability to retrieve information from the Internet.

The latest version also had a hand that allows the robot to grab objects and, for example, move them from one place to another. The robot receives information about its environment through a sensory network. The environment could have embedded display devices that could be used to request services when the robot is not nearby. Using this, the robot can provide a variety of services. It can serve as a guide and take guests to meet the people they have agreed to meet.

The robot can also act as a messenger to share some of the messages it knows with people it knows. The robot may also provide the ability to retrieve various information through discussion, such as on-line data or contact information of people working in that environment. All of the above functions must be performed in the most natural way possible, utilizing the robot's ability to recognize human facial expressions and then change its behavior based on emotional state.

## 8.5 Machine Vision in Medical Image Analysis

Interpretation of images obtained with various imaging equipment plays a central role in medicine. This may be, for example, examination of the results of microscopy, X-ray equipment, ultrasound camera or magnetic resonance imaging.

Already in the 1990s, the Machine Vision Group studied use of magnetic resonance imaging (MRI) in surgery, analyzing skin images to identify melanoma, and measuring the area covered by psoriasis rash. In the early 2000s, the use of functional MRI (fMRI) in brain imaging was investigated.

In recent years, the main areas of focus have been microscopic image and video analysis and pulmonary X-ray analysis. Figure 8.23 shows an example of a microscopic image analysis study with the University of Helsinki (docent Johan Lund), in which the actual tissue of interest (stroma) was segmented from the epithelium by the LBP / C texture method and the support vector classifier (Linder et al., 2012). The main result of this collaboration may be the development of a practical tool for the diagnosis of malaria (Linder et al., 2014).

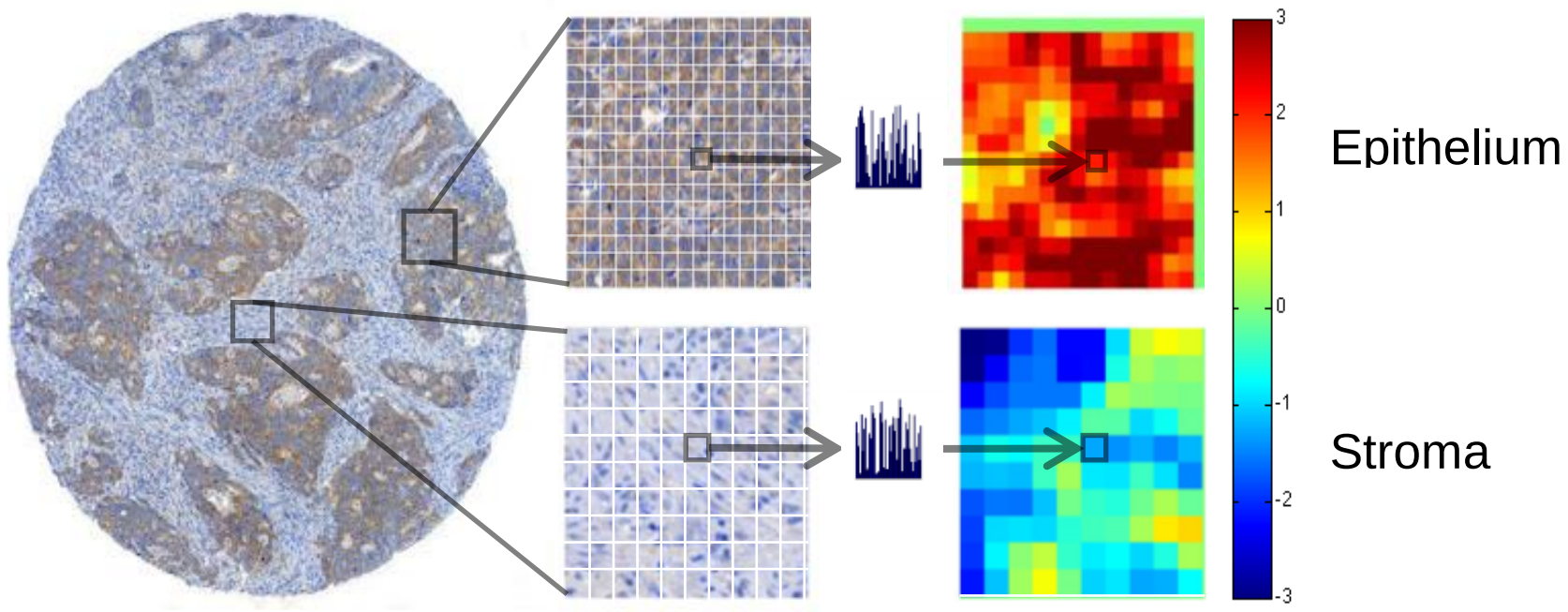

Figure 8.23. Segmentation of a virtual microscope image.

From [Linder N, Konsti J, Turkki R, Rahtu E, Lundin M, Nordling S, Ahonen T, Pietikäinen M & Lundin J (2012) Identification of tumor epithelium and stroma in tissue microarrays using texture analysis. Diagnostic Pathology].

### *8.5.1. Deep neural networks for classification of chest X-ray images*

There is a great need for automatic interpretation of chest X-ray images, as they are produced in large quantities in hospitals every day, and it is not possible for physicians specialized in X-ray interpretation to go through all the details. It would be advantageous if a preliminary interpretation of the images could be made already in health centers.

Most of the pictures taken have lungs healthy. If hospital radiologists could only focus on cases where the lungs may have disease-related changes, their precious time for other tasks would be saved.

The aim of our study was to be able to distinguish between images of healthy and potentially ill lungs, classifying of the image material into two categories. Figures 8.24a and 8.24b show examples of healthy lungs, and Figure 8.24c shows a patient with pneumonia (Chen et al., 2016).

However, the automatic interpretation of the images is difficult because they exhibit considerable variations due to the imaging conditions, such as illumination, lung position and scale. Images may include images taken by different hospitals, devices, and different operators. In addition, it may be difficult for sick patients to be in the best position during imaging.

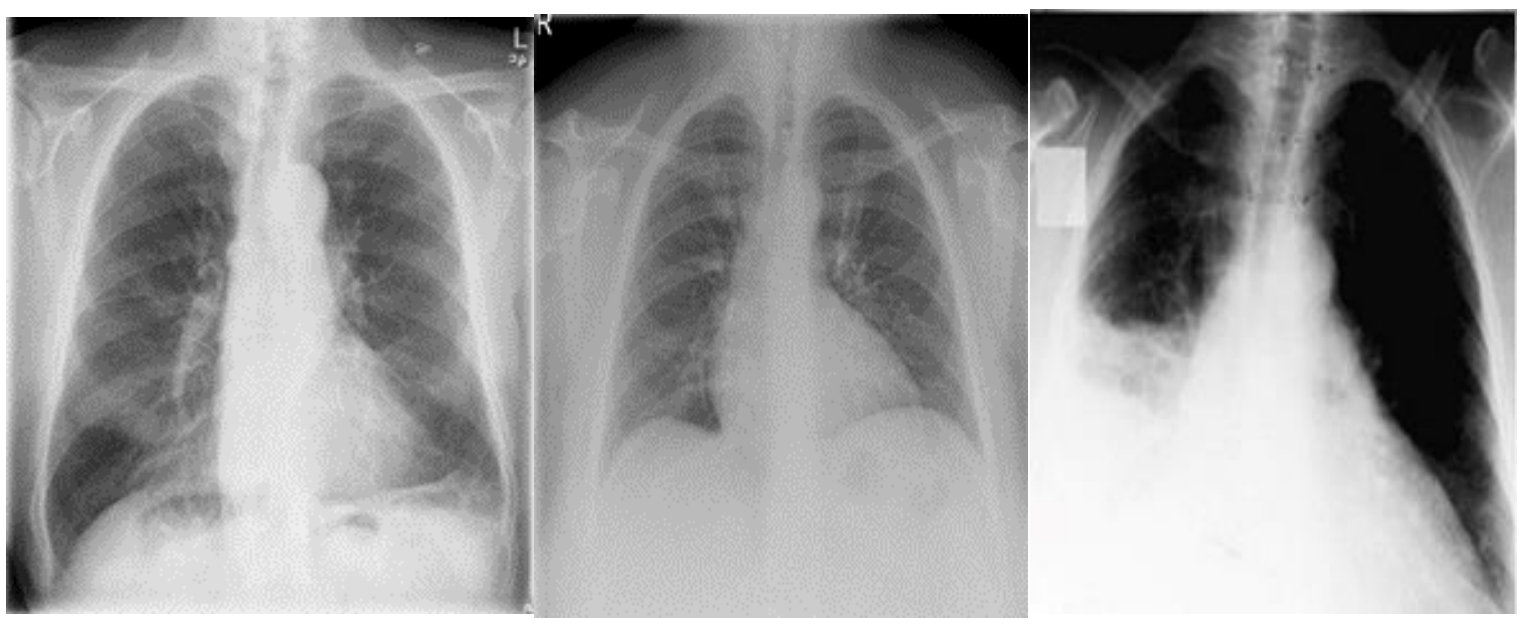

(a) (b) (c)

Figure 8.24. X-ray images. (a) and (b) normal, (c) pneumonia. (© IEEE)

In the developed method, the alignment of the images is first performed (Figure 8.25). The set of images is then digitally augmented for CNN deep network training with sample versions of three different scales as well as cropping, mirroring, and rotating. Finally, a fine-tuned CNN model is used to classify unknown test images.

Two image databases were collected for the study by the Oulu University Hospital. The first consisted of 755,969 magnetic resonance imaging (MRI) images of 512 x 512 pixels taken from the brains of 1000 patients used for training. Each image was annotated by radiologists, either healthy or possibly diseased. Both of these were nearly as many. The second set consisted of 4,000 X-ray chest images taken from 2,000 patients at a resolution of 2688 x 2688 pixels. Half of the pictures were of a healthy person and half of them were possibly ill.

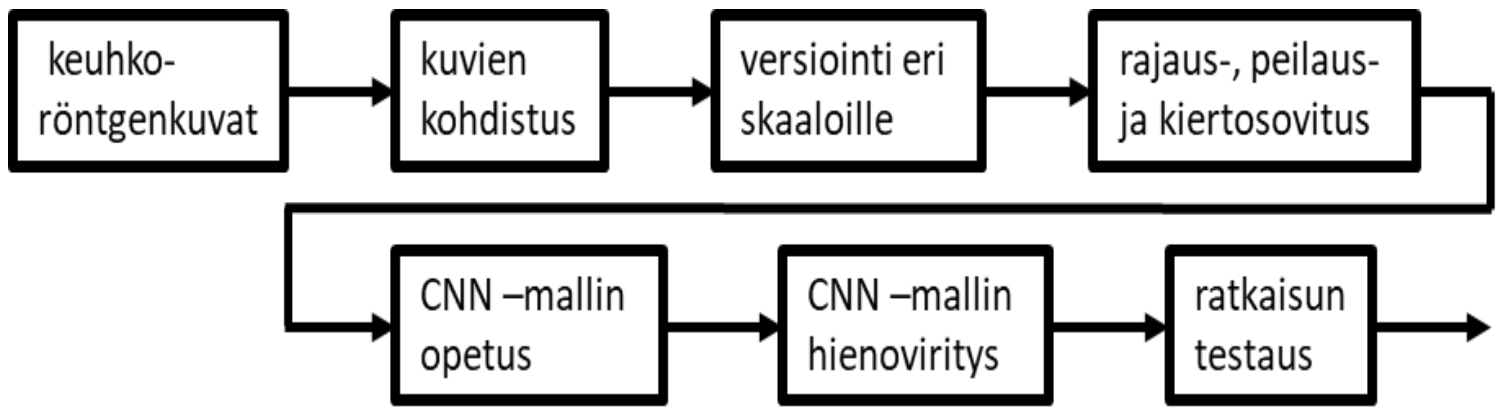

Figure 8.25. Classification of X-ray images with a CNN neural network method.

The MRI image set was used to teach the CNN neural network by first supplementing it with three different scales. Half of the lung images were used to fine-tune the CNN model by first aligning, complementing with three different scales and adding cut, rotated, and rotated images. The other half of the lung images were used for testing with the fine-tuned CNN model by first performing only an alignment operation.

The developed method achieved over 80% classification accuracy, clearly better than the conventional methods. However, this result is not good enough for clinical use, as pictures of sick people should not be included among the healthy pictures. Significantly more training samples would be required to create a good enough CNN model, but creating such a set of samples is a very laborious and expensive operation.

## 8.6 References

Chen J, Qi X, Tervonen O, Silvén O, Zhao G & Pietikäinen M (2016) Thorax disease diagnosis using deep convolutional neural network. Proc. 38th Annual International Conference of the IEEE Engineering in Medicine and Biology Society (EMBC 2016), 2287-2290.

Fernandez-Lopez A & Sukno FM (2018) Survey on automatic lip-reading in the era of deep learning. Image and Vision Computing 78:53-72.

Heikkilä J & Silvén O (1997) A four-step camera calibration procedure with implicit image correction. Proc. IEEE Conference on Computer Vision and Pattern Recognition, June 17-19, San Juan, Puerto Rico, 1:1106-1112.

Heikkilä J (2000) Geometric camera calibration using circular control points. IEEE Transactions on Pattern Analysis and Machine Intelligence 22(10):1066-1077.

Herrera Castro D, Kannala J & Heikkilä J (2012) Joint depth and color camera calibration with distortion correction. IEEE Transactions on Pattern Analysis and Machine Intelligence 34(10):2058-2064.

Hietaniemi R, Bordallo Lopez M, Hannuksela J & Silvén O (2014) Real-time imaging system for lumber strength prediction. Forest Products Journal 64(3-4):126-133.

Kellokumpu V, Zhao G & Pietikäinen M (2011) Recognition of human actions using texture descriptors. Machine Vision and Applications 22(5):767-780.

Kohonen T (2001) Self-Organizing Maps, Third Edition, Springer.

Li X, Alikhani I, Shi J, Seppänen T, Junttila J, Majamaa-Voltti K, Tulppo M & Zhao G (2018) The OBF database: A large face video database for remote physiological signal measurement and for Atrial Fibrillation detection. Proc. IEEE International Conference on Face and Gesture (FG 2018).

Li X, Chen J, Zhao G & Pietikäinen M (2014) Remote heart rate measurement from face videos under realistic situations. Proc. IEEE Conference on Computer Vision and Pattern Recognition (CVPR 2014), Columbus, Ohio, 4265-4271.

Li X, Komulainen J, Zhao G, Yuen PC & Pietikäinen M (2016) Generalized face anti-spoofing by detecting pulse from face videos. Proc. International Conference on Pattern Recognition (ICPR 2016), 4244-4249.

Linder N, Konsti J, Turkki R, Rahtu E, Lundin M, Nordling S, Ahonen T, Pietikäinen M & Lundin J (2012) Identification of tumor epithelium and stroma in tissue microarrays using texture analysis. Diagnostic Pathology 2012, 7:22.

Linder N, Turkki R, Walliander M, Mårtensson A, Diwan V, Rahtu E, Pietikäinen M, Lundin M & Lundin M (2014) A malaria diagnostic tool based on computer vision screening and visualization of plasmodium falciparum candidate areas in digitized blood smears. PLoS ONE 9(8):e104855.

Määttä J, Hadid A & Pietikäinen M (2011) Face spoofing detection from single images using micro-texture analysis. Proc. International Joint Conference on Biometrics (IJCB 2011), Washington, D.C., USA, 7 p.

Määttä J, Hadid A & Pietikäinen M (2012) Face spoofing detection from single images using texture and local shape analysis. IET Biometrics 1(1):3-10.

Niskanen M (2003) A Visual Training Based Approach to Surface Inspection. Acta Univ. Oul. C 186, 125 p.

Pietikäinen M & Silven O (2002) Konenäkö (Computer vision). 25th Anniversary Book of Pattern Recognition Research in Finland, Eds. Iivarinen J, Kaski S & Oja E, Pattern Recognition Society of Finland, 74-85.

Piironen T, Silvén O, Pietikäinen M, Laitinen T & Strömmer E (1990) Automated visual inspection of rolled metal surfaces. Machine Vision and Applications 3(4):247-254.

Röning J, Holappa J, Kellokumpu V, Tikanmäki A & Pietikäinen M (2014) Minotaurus: A system for affective human-robot interaction in smart environments. Cognitive Computation, 6(4):940-953.

Silvén O, Niskanen M & Kauppinen H (2003) Wood inspection with non-supervised clustering. Machine Vision and Applications 13(5-6):275-285.

Silvén O, Virtanen I, Westman T, Piironen T & Pietikäinen M (1989) A design data-based visual inspection system for printed wiring. In: Advances in Machine Vision, ed. JLC Sanz, Springer-Verlag.

Zhao G, Barnard M & Pietikäinen M (2009) Lipreading with local spatiotemporal descriptors. IEEE Transactions on Multimedia 11(7):1254-1265.

Zhou Z, Hong X, Zhao G & Pietikäinen M (2014) A compact representation of visual speech data using latent variables. IEEE Transactions on Pattern Analysis and Machine Intelligence 36(1):181-187.

Zhou Z, Zhao G, Guo Y & Pietikäinen M (2012) An image-based visual speech animation system. IEEE Transactions on Circuits and Systems for Video Technology 22(10):1420-1432

Ylimäki M, Kannala J, Holappa J, Brandt SS, Heikkilä J (2015) Fast and accurate multi-view reconstruction by multi-stage prioritized matching. IET Computer Vision 9(4):576-587.

Web-Antispoofing: Anti-spoofing demonstration

Web-Minotaurus: HRI demonstration

Web-TrumpClinton: Trump-Clinton debate

# 9 Emotion AI – the Next Breakthrough?

## 9.1 Introduction

The purpose of emotions is to try to increase things that promote human well-being and avoid things that endanger life (Wiki-Tunne), (Puttonen & Heikkinen, 2018). According to Lauri Nummenmaa, Director of the Emotion Laboratory at University of Turku, the feelings can be compared to a thermostat, which keeps the room temperature normal and regulates the heating when deviating from the set value. The universal, culture-independent basic feelings include happiness, sadness, disgust, anger, fear and surprise.

According to research by Nummenmaa's group, the human body experiences strong emotions, and the above-mentioned basic emotions can be located in the same places, regardless of cultural background. (Nummenmaa et al., 2014). After the pictures, videos and stories were presented to the test subjects, they were asked to color the drawing where and how intense the feelings of the material presented felt. Only the feeling of "happiness" spreads throughout the body (Figure 9.1).

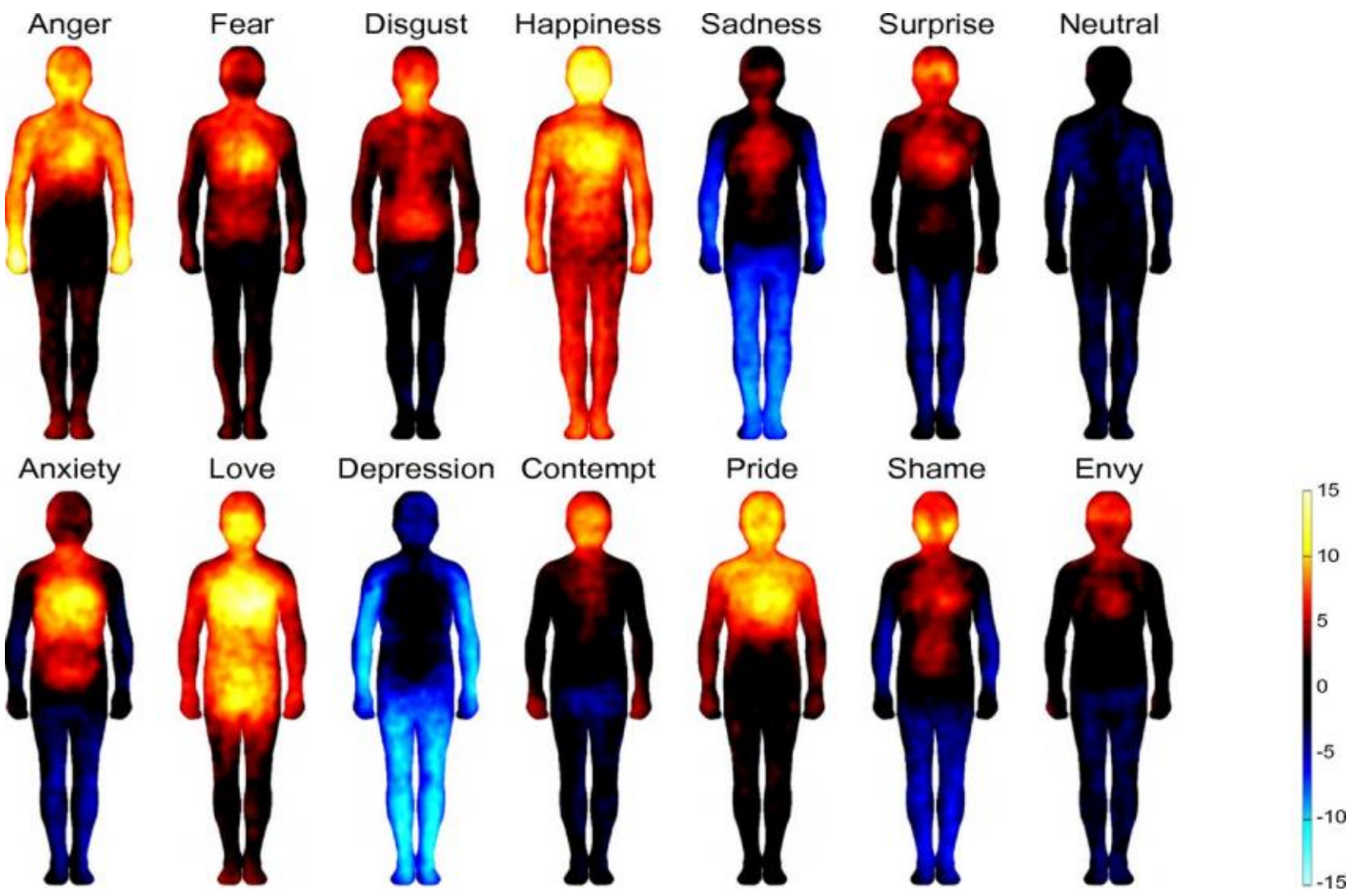

Figure 9.1. Locations of different emotions on the body map. (© PNAS)

However, caution should be exercised in generalizing the results of this body map to most real-world situations. The experiments have been conducted in a typical laboratory environment and the influence of many key factors on the results has not been elucidated, for example with regard to the design of the experimental setup to produce different emotional states. Section 9.7 introduces the challenges of emotional intelligence that need to be considered when designing practical applications. For example, variations in imaging conditions will cause serious problems when applying machine vision to automated emotion analysis.

Facial expressions have also been found to be similar throughout the world, but due to cultural differences learned and individual differences, the use of expressions may differ. For example, Americans smile and gesture more than East Asians or Finns, but that doesn't mean they are happier.

Emotions and intelligence are closely related. Emotions guide human activities alongside motives, goals, and data processing in the brain. They promote human adaptation to the environment and different situations. Emotions affect a person in many ways. They may appear in facial expressions, speech tone, body movements and gestures. Emotional state affects a variety of physiological signals describing the internal state of the human being, which can be obtained, for example, from the brain's electrocardiogram, heart rate, skin temperature, skin conductivity, blood pressure or respiratory rate.

Emotions and empathy play a central role in human-to-human interaction. Most of the communication between people is non-verbal communication: facial expressions, gaze, gestures, body movements and postures, variations in voice tone, touch, etc.. Estimates of non-verbal communication vary. Albert Mehrabian, in his much quoted study, stated that 38 % of face-to-face communication involves voice and tone, 55 % non-verbal behavior (e.g., facial expressions), and only 7% words (Mehrabian, 1972). However, these results have been criticized. Indeed, in research, they relate those to more communicative attitudes and feelings.

Only a small part of the human brain deals with verbal communication. It is estimated that at least 60% of the brain is related to vision. However, perhaps only 20% of this is devoted solely to vision, the remainder deals with the sharing of more modalities, e.g., vision and touch, vision and motor functions, sight and attention, vision and meaning. As babies, people learn non-verbal communication in their social and emotional interaction, with face as the dominant channel of communication instead of voice. After learning to speak, people use non-verbal communication, such as facial expressions, voice tone, and other non-verbal communication, at least partly unconsciously.

Human **emotional intelligence** refers to the ability to recognize our own emotions and those of others and to use this knowledge to guide our own behavior and achieve our goals. The importance of emotional intelligence for different applications is clear. For example, when making shopping decisions, feeling of a particular product is often more important than the details of the product. Studies have shown that when it comes to deciding between several options, it is very difficult for a person to do so by mere logical reasoning. Salespeople who can read the customer's feelings are better salespeople. When a customer views a product in a store or shop window, their feelings tell us what they like and what they don't.

According to Jussi Toivanen, Managing Director of Microsoft Finland, more empathy would be needed (Nalbantoglu, 2018). He said people should become more human as technology advances and machines become more intelligent.

In understanding emotions and social skills, man is superior to machines, and this competitive advantage should be exploited. According to a study conducted by Microsoft, Finnish companies are in the forefront of exploiting artificial intelligence technology in Europe, but lagging behind in understanding the meaning of emotional intelligence and empathy. In Finland, the focus is on process optimization and cost savings, but not on solutions that understand people and their needs.

Nowadays, when interconnected intelligent devices control our communication and our daily lives, attention to emotions plays a central role in artificial intelligence - we can talk about **Emotion AI** or **Affective Computing** (Picard, 1997). Communication with technology is becoming more interactive, resembling the interaction between people. Intelligent devices in our environment are able to capture people's emotions and moods and create more personalized user experiences. Emotional intelligence is needed in systems being developed to read user emotions and use them to adapt actions.

According to Rana el Kaliouby, director of Affectiva, a spin-off company launched by Rosalind Picard’s affective computing research at MIT, emotional intelligence will grow into a billion-dollar business in the next few years. It will revolutionize many industries, such as market research, the automotive industry and health care. Many technology companies have begun to invest in the emotional intelligence. Alongside Affectiva, Apple, Google, Microsoft and Baidu, for example, see emotional intelligence as a major component of their systems. Also in Finnish companies the interest in recognizing emotions has clearly increased lately.

However, automatic recognition of emotions is a very difficult task. It is known that emotions depend on the context in which they are attempted to be recognized. Factors affecting the evaluation of the emotion state include: operating environment, body posture, sounds and words, and the cultural background of the individual being assessed (Feldman Barret et al., 2011). In addition to facial expressions, a number of different factors are needed, so-called multi-modal approaches. Many times the tone and body movements already tell a lot about emotional states.

## 9.2 Cognition and Emotions Go Hand in Hand

Cognition refers to those phenomena of the human mind that can be described or explained as information processing. Cognition and emotion are closely interrelated (Wiki-Emotion).

According to cognitive explanations, humans use emotions to process information from both their bodies and their environment. The human mind is thought to function in that it constantly processes and directs its emotions. Feelings arise as a result of judgments and interpretations. Feeling creates motivation for action as well as repeating our behavior. Although cognitive explanations focus on exploring information processing, they also accept that physiological state can alter emotion.

Studies have shown the effect of the physiological states of the body on the experience of feeling. In one study, subjects were asked to feel anger. One group was told to squeeze their hands at the same time, the other just "felt angry". The results showed that people holding their hands in fists felt more angry than others (Wiki-Tunne).

A cognitive explanation for the emergence of emotions is given by the so-called *appraisal theory*. According to it, situational factors and previous experiences create a feeling (Figure 9.2) (Scherer et al., 2001, Wiki-Appraisal). The person's own assessment of the situation is important in feeling the feeling. According to the theory, emotions arise from evaluations of events that cause certain reactions, or emotions, in different people.

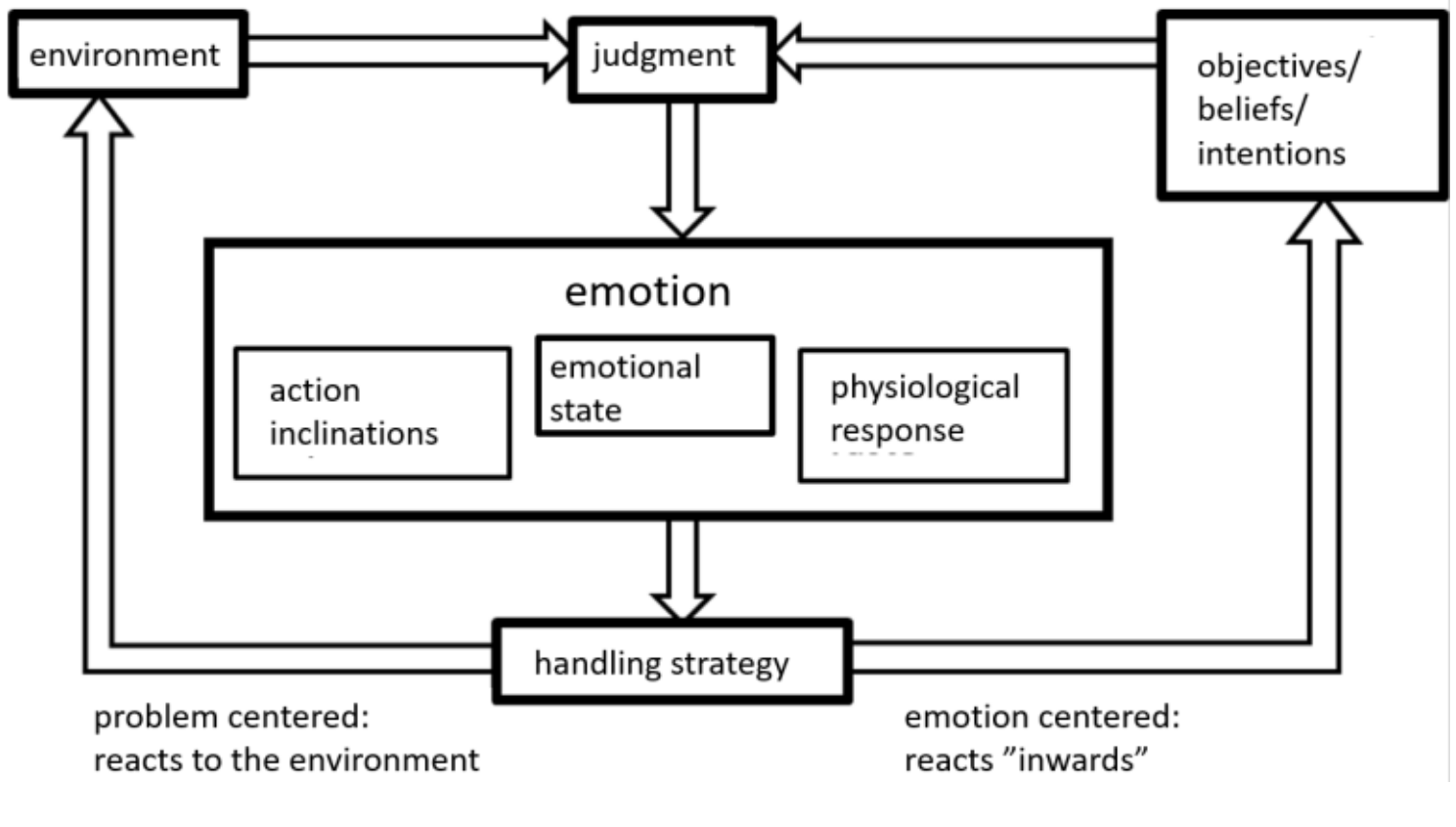

Figure 9.2. Cognitive appraisal theory. (e.g. Scherer et al., 2001).

Wikipedia, for example, describes going to a first date. If dating is successful, the person may feel happy, glad, excited, moved and/or be eagerly awaiting because this assessment can be expected to have positive long-term effects - starting a new relationship, getting engaged or even getting married.

Conversely, when dating is negative, emotions can become depressed, sad, empty or frightened (Wiki-Appraisal). The reason-ing and understanding of these emotional reactions will also influence future evaluations. Note that such an estimation theory, or variations thereof, allows for individual variations in the same event in different people.

## 9.3 Applications of Emotion AI

Emotions play a central role in human-to-human interaction. When talking to a robot or chatbot, one finds "machine-like", insensitive speech disgusting. In particular, empathy that is the ability to understand the feelings of the interlocutor, is seen to play a significant role in good interpersonal communication.

The same flaw more generally applies to interaction in digital media, where we often do not see our interlocutor. Solutions to problems have been sought, e.g. , in a project funded by Tekes and industry in 2018-19, “Quantifying Human Experience for Increased Intelligence within Work Teams and in the Customer Interface”, coordinated by cognitive scientist Katri Saarikivi from the Cognitive Brain Research Unit, University of Helsinki (Web-Humex, 2017).

The role of our group was measuring the emotional state using machine vision. Other participants were Aalto University research groups on Complex Systems and Learning Environments. The companies involved in the project were interested in: utilization of emotional intelligence in computer-mediated interaction with clients, health services, and digital marketing.

A good example of using today's emotional intelligence is the sales laboratory acquired for teaching purposes at Haaga-Helia University of Applied Sciences, based on emotional analysis software developed by MIT University's spin-off company Affectiva (Mielonen, 2018). It observes more than thirty points of motion on its face, apparently using action units as defined by Paul Ekman. The system has been trained with over seven million faces. It classifies a person's face based on measured information into one of seven prototypic expressions. The Haaga-Helia system describes how customers respond to a salesperson, and the aim is to get salespeople to change their behavior based on the customer's emotional state.

Emotional intelligence can be used in many ways in computer-assisted teaching. For example, if the subject to be taught has not been understood or the student is not attentive enough, it is readily apparent from the student's expressions and changes in emotional state.  Professor Sanna Järvelä, investigating computer-assisted learning at the University of Oulu, explains that emotional intelligence can be utilized more widely in the future digital learning environments, which should understand students' goals, assess performance, and provide active guidance to enable each student to achieve their learning goals in a natural way of interacting with the machine.

In particular, basic education - from pre-school to university - and lifelong learning have become increasingly important teaching and learning environments. According to Järvelä, the basic problem in current technology-assisted and/or virtual learning environments is that they do not provide the same social and emotional presence as face-to-face interaction. Emotional intelligence can significantly improve this issue and make invisible issues related to emotions and cognitive learning processes visible.

Interpreting emotions is also important in intercultural communication. It is often difficult for unfamiliar Europeans to infer the face of Asian people and vice versa. In this way, emotional intelligence can support both parties in making judgments.

The lack of verbal communication with people is a major problem for the visually impaired. The machine can help them identify objects and people in the environment. Identifying a partner's emotion state would be very important in natural interaction. Smart glasses like those in Section 8.2 could be helpful. The emotion state can be transmitted to the visually impaired by means of acoustic signals, artificial speech, or touch feedback with  a device attached on the body.

Emotions also matter to our health. Very strong negative or positive emotion can cause a heart attack or cerebral hemorrhage. Depression and other types of long-term negative emotion can increase the risk of heart disease and diabetes. On the other hand, long-term positive emotional state is known to have a positive effect on psychological and somatic, organic diseases.

In health and medicine, emotional intelligence can help monitor our mental state, identify pain conditions, or even help make a proactive diagnosis of Parkinson's disease or coronary heart disease. Emotional information can also be used in research and therapy for such as depression, autism, traumatic stress disorder and bipolar disorder.

It is possible to see up to 30 medically relevant symptoms on people's faces by machine vision, such as getting hints of pain or states of mind (fatigue, depression, anxiety, mental strain, stress, etc.) or detect facial changes and asymmetries, for example, when a cerebral hemorrhage might start (Thevenot et al., 2018).

By combining data from sensors that measure physiological signals, it is possible to develop better devices, for example, on the wrist, for continuous monitoring of human health and well-being. By using intelligent glasses with machine vision capabilities, the physician could obtain useful information about the patient's face for diagnosis.

Emotions also tell when a person is tired, frustrated, or nervous while driving a car - or when he or she is properly alert. The disclosure of such information has been of increasing interest to car manufacturers in improving both safety and the driver's user experience. An example of a great need is the Kuopio bus accident in late summer 2018, where detecting and alerting the driver of a change in his driving condition early enough could have prevented the accident.

Important potential applications for emotion recognition include analyzing people's behavior and facial expressions during airport security checks and revealing lies during interrogations. The advantage of machine vision is that the person to be investigated does not need to be fitted with any measuring equipment that could change his or her behavior. However, lie detection with a machine is a very difficult task and cannot be reliably performed on the basis of perceptual information alone. Emotional status analysis may also be useful, for example, in job interviews (Figure 9.3) or in preparation for such.

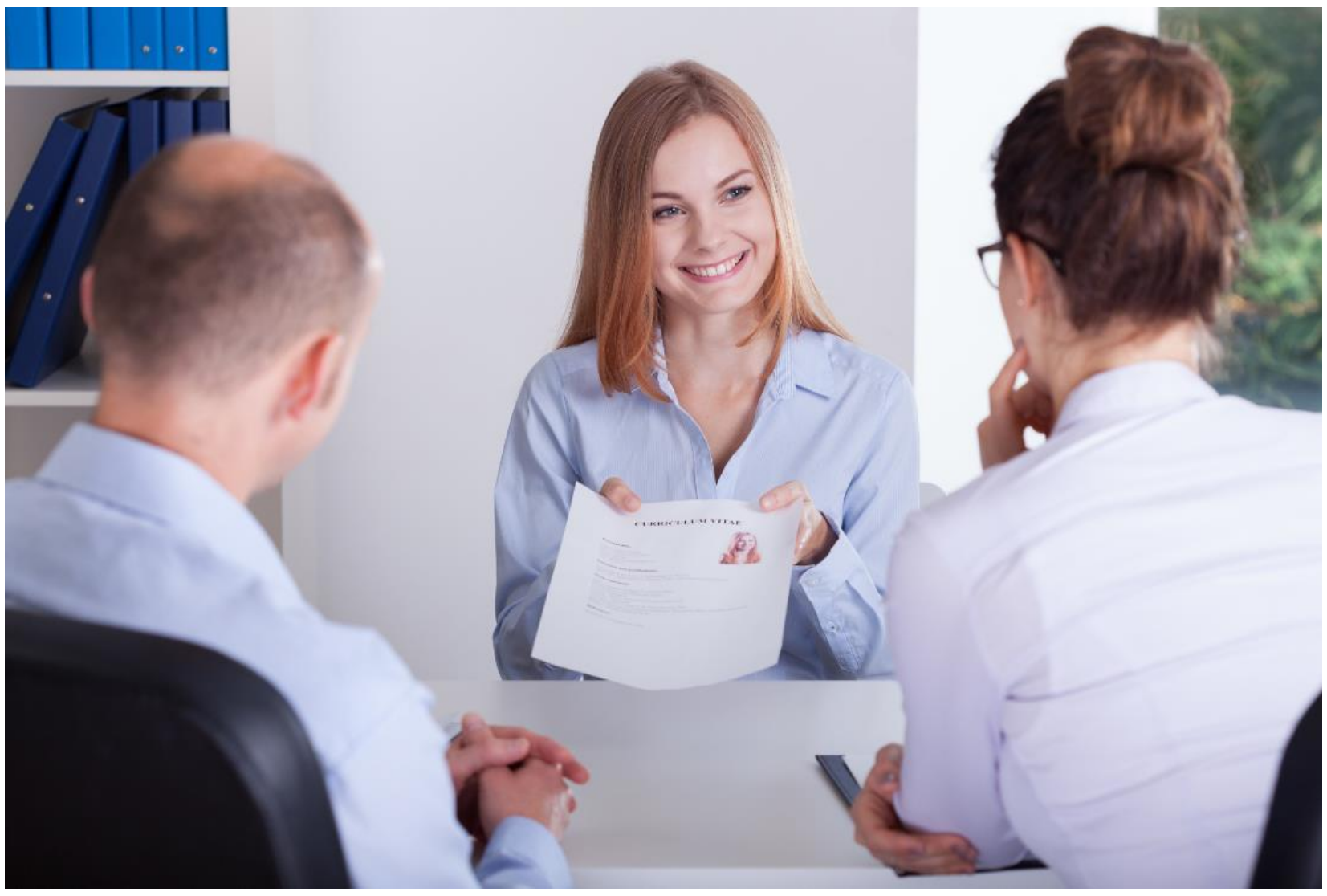

Figure 9.3. Emotions in a job interview. (© 123RF)

Customer profiling and user experience analysis are important commercial applications. Facial expressions tell us how interested customers seem to be in certain products, etc. These types of applications are becoming more common among the first because they usually do not require very high precision and can be very important commercially.

Emotional toys are much more natural playmates for kids than non-emotional toys. Simple toys recognizing child's expression already exist or are being developed (Web-Toys). A puppet recognizing an emotion state would be much better able to "communicate" with a child. Communicating with toy robots would be much more natural, in the same style as the other human-machine interfaces mentioned earlier. Emotion recognition can also be used to teach children to stay alert and, more generally, to use emotions and empathy in their interaction.

Emotions can also be utilized in the game industry. Affectiva has introduced features that utilize emotion mode in the *Nevermind* video game (Web-Affectiva). The system recognizes the player's emotion state and changes its behavior and user experience accordingly. For example, if a player is under stress, the gaming world will become darker and more distorted, but will calm down as stress decreases.

Nevermind uses in its analysis multimodal data from a heart rate monitor, Apple watch and eye movement measurement device, and in its latest version Affectiva's emotion recognition software, which analyzes players' actions on the video data produced by their web cameras.

## 9.4 Facial Expressions in Emotion Recognition

Charles Darwin was the first to study facial expressions and their relationship to emotions, and in his 1872 book, “The Expression of Emotions in Man and Animals”, concluded that expressions can be described as discrete categories of emotions (Ekman, 2009). In addition, the set of expressions and persons he had observed and studied were universal, that is, independent of the person.

The six most common so called prototypic expressions on the face are happy, sad, disgust, angry, fear, and surprise. Contempt is also often included in basic expressions. In addition, the face may look neutral. Depending on the application, there may also be other interesting facial expressions, such as arrogance, driving fatigue, pain when experiencing pain, or nervousness when tense.

This approach is based on imagining emotions as separate classes, as Darwin suggests, because of their fundamentally different structures. Figure 9.4 shows examples of seven commonly used prototype expressions and neutral expression taken from the widely used Cohn-Kanade database.

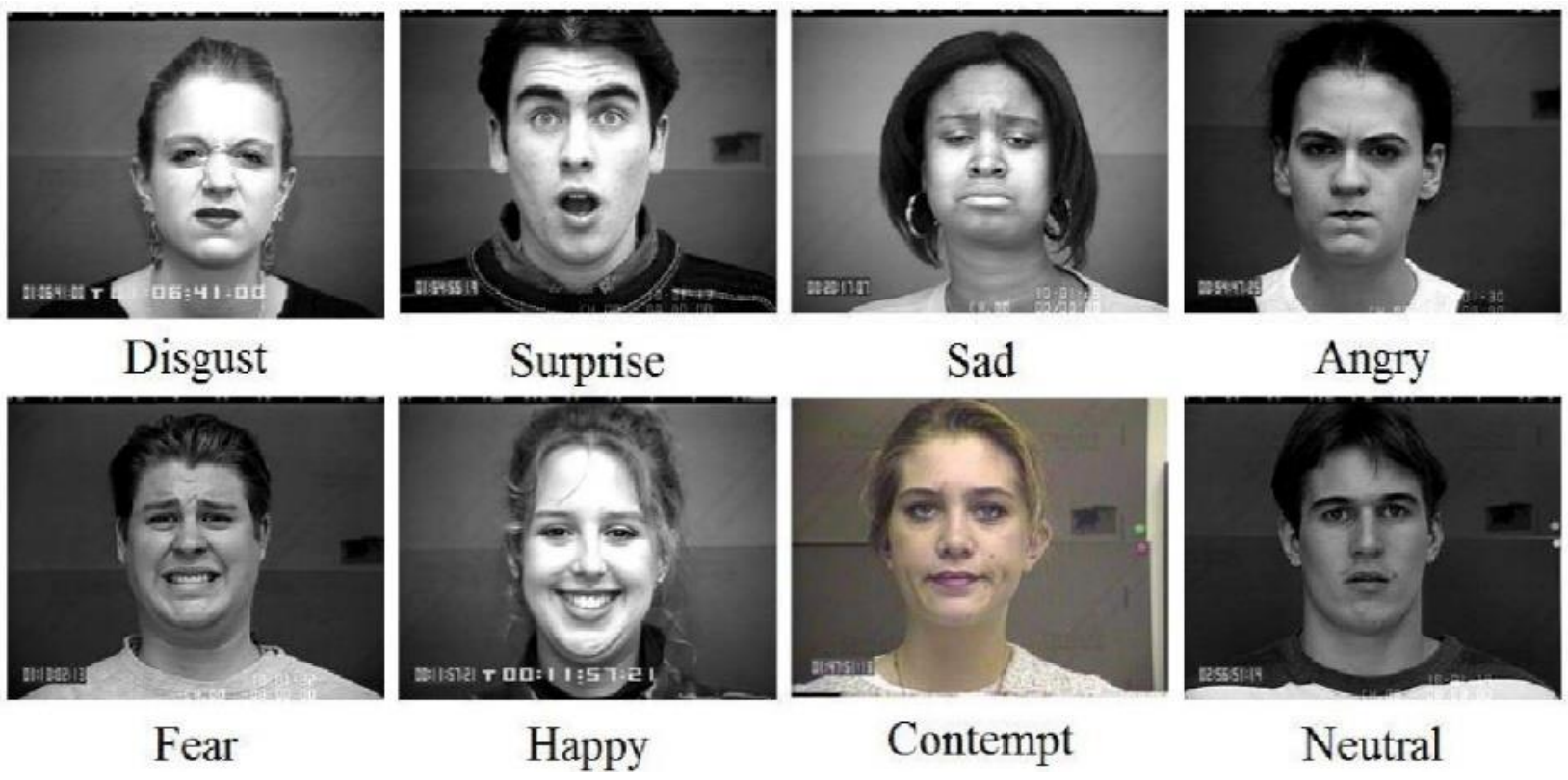

Figure 9.4. Prototypic facial expressions.

The facial expressions can be created by the facial action coding system (FACS). It describes facial muscle movements to produce various expressions (Ekman & Friesen, 1978). Figure 9.5 shows examples of action units (Martinez et al., 2017).

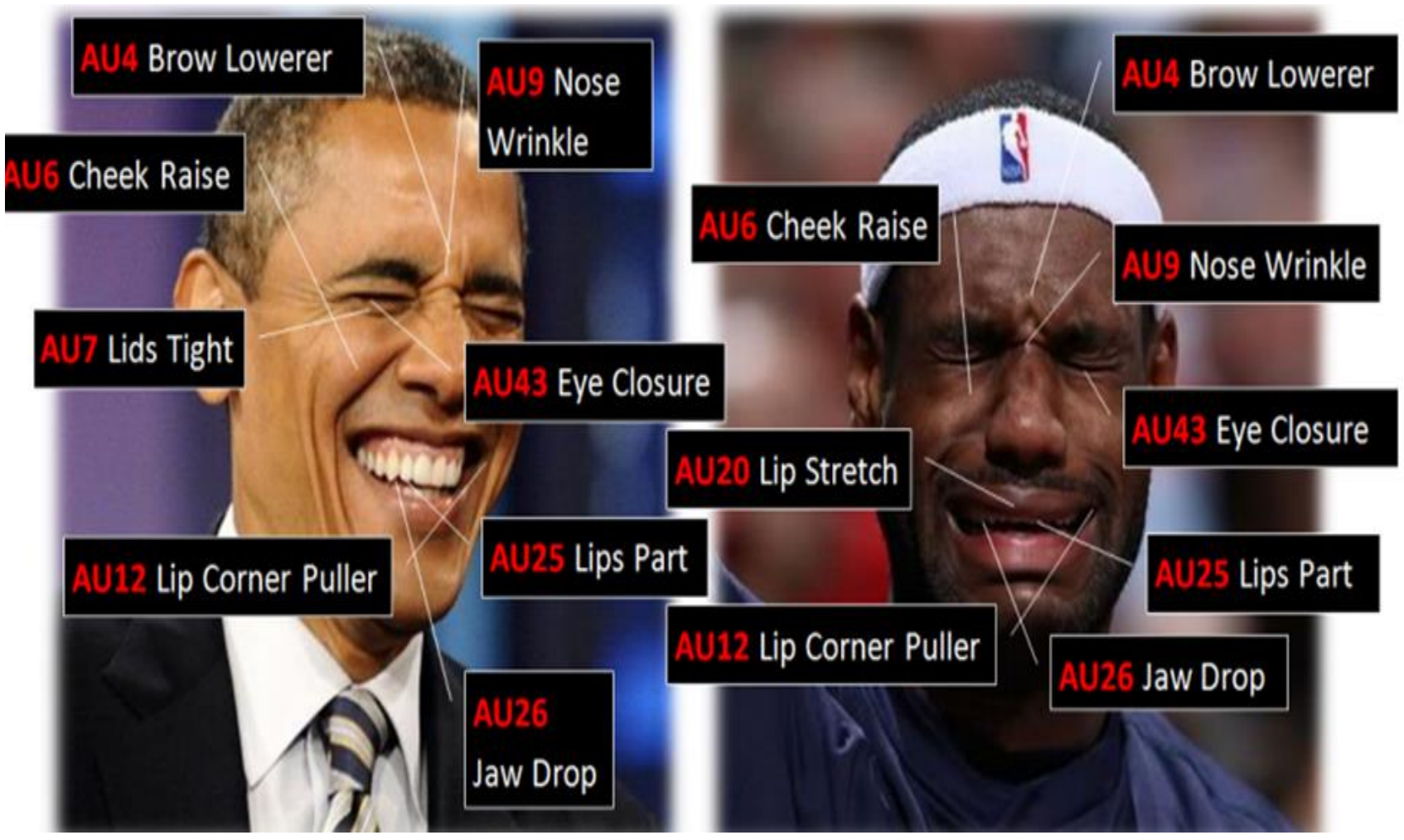

Figure 9.5. Examples of action units. (© IEEE)

The basic unit is the Action Unit (AU). FACS, the latest version of which was developed by the psychologist Paul Ekman in 2002, is a commonly used standard for categorizing emotion expressions (Ekman et al., 2002). It is used in computer vision as well as in psychological research and facial animation.

The advantage of FACS coding is that they can be used to describe facial muscle movements for the majority of potential expressions. The big challenge for machine vision is to be able to detect the various action units automatically with sufficient reliability. Laboratory conditions have provided quite good results for most of them, but the problem is greatly exacerbated by natural variations in lighting, head pose changes, lower image resolution, and so on. Figure 9.6 shows how the expressions "joy" and "anger" can be composed.

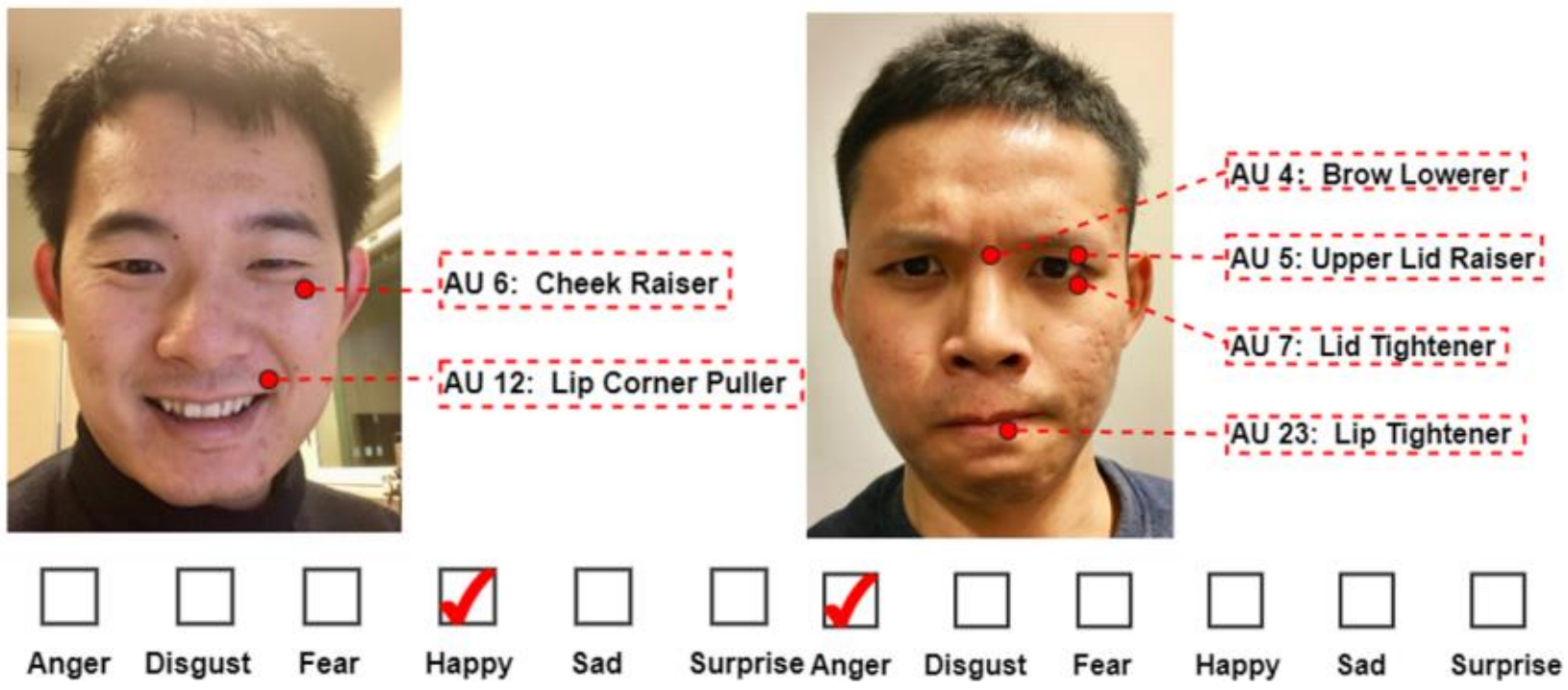

Figure 9.6. The expressions “joy” and “anger” were born as a combination of movements. (© Henglin Shi and Zitong Yu)

In the above examples, two emotion AI researchers at the Center for Machine Vision and Signal Analysis are present. Notice the combined effect of the movements of different action units.

Another way to describe emotions is to group them by dimension. The much used two-dimensional model assumes that there are always values for activation state (arousal) and emotional value (valence) associated with different emotional states as shown in Figure 9.7. For example, happy emotion is seen as both a positive arousal and valence.

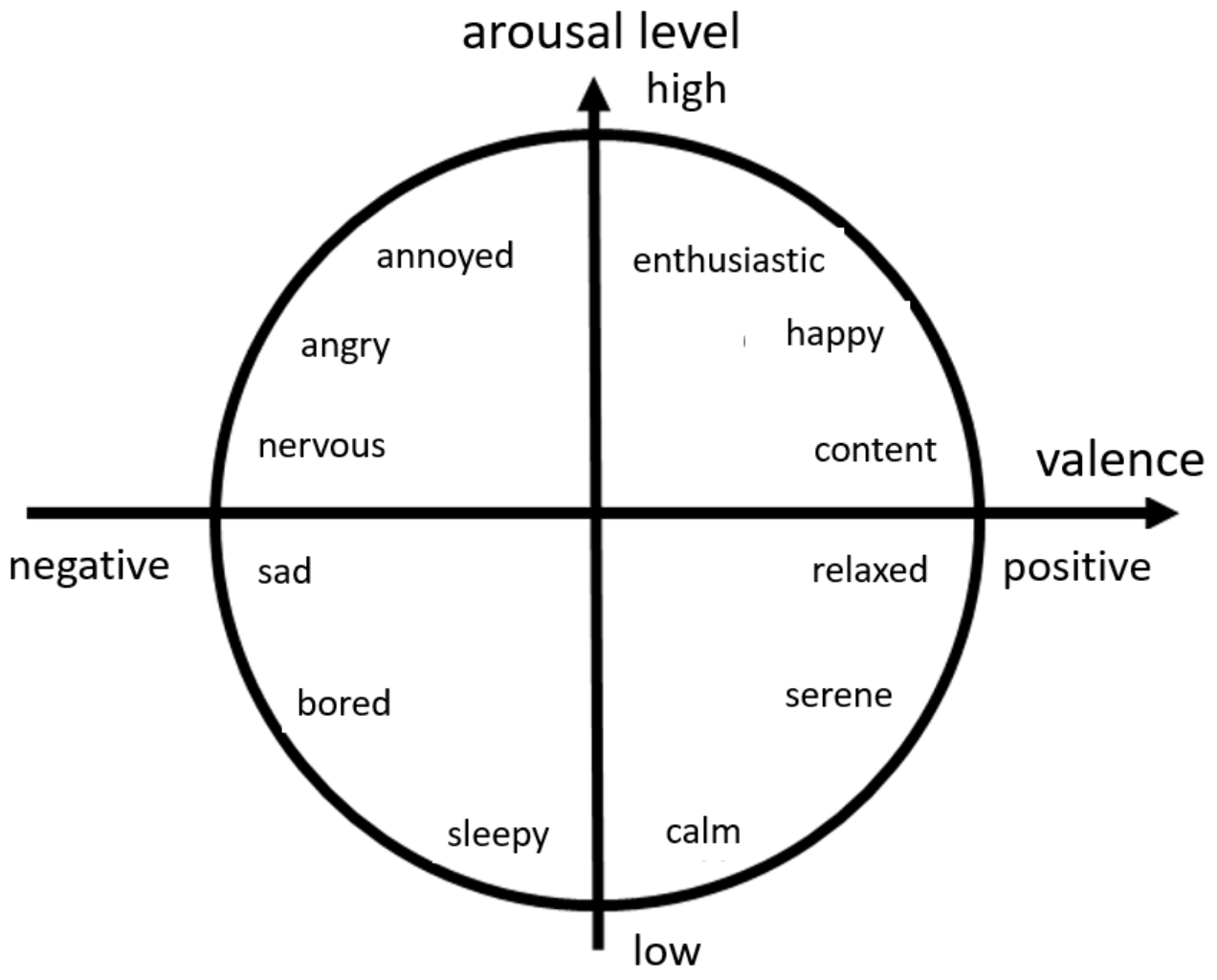

Figure 9.7. Two-dimensional emotion model.

Finding the valence reliably is much easier than arousal, so often it is only used, dividing emotions into positive ones on the right (like happiness, peacefulness), negative ones on the left (anger, sadness, nervousness) and almost neutral in the middle.

Facial expressions can also be divided into two main categories based on their speed and intensity:

1) Macro-expressions are mostly clearly visible and have a duration of about 0.5 to 5 seconds and
2) Micro-expressions very fast (0.03-0.5 seconds) and low in intensity.

Macro-expressions are voluntary expressions, meaning we can even pretend to smile or other expressions. Micro-expressions, on the other hand, are involuntary. In Figure 9.8, the horizontal axis represents the duration of the image and the vertical axis the intensity.

In addition, there is often interest in the types and durations of the individual expressions or their action units. For example, in the initial phase (Onset) the expression may change from neutral towards the smile, at the peak (Peak) the smile will be maximal and in the final phase the expression may return to neutral again.

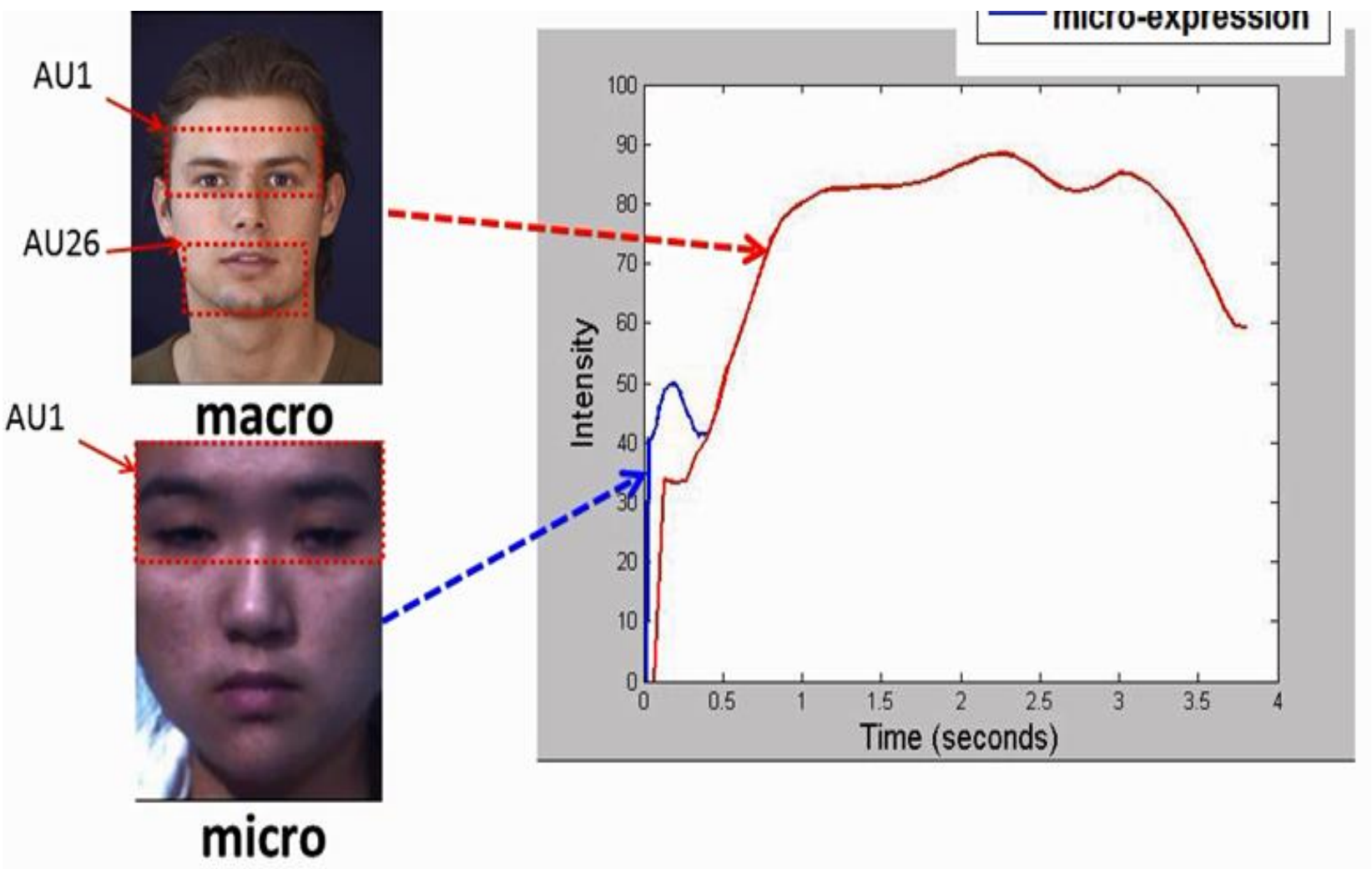

Figure 9.8. Micro- and macro-expressions. (© CMVS)

## 9.5 Micro-expressions and their Identification

Micro expressions are very fast unconscious facial movements. They also reveal emotions that the person does not want to express (Figure 9.9).

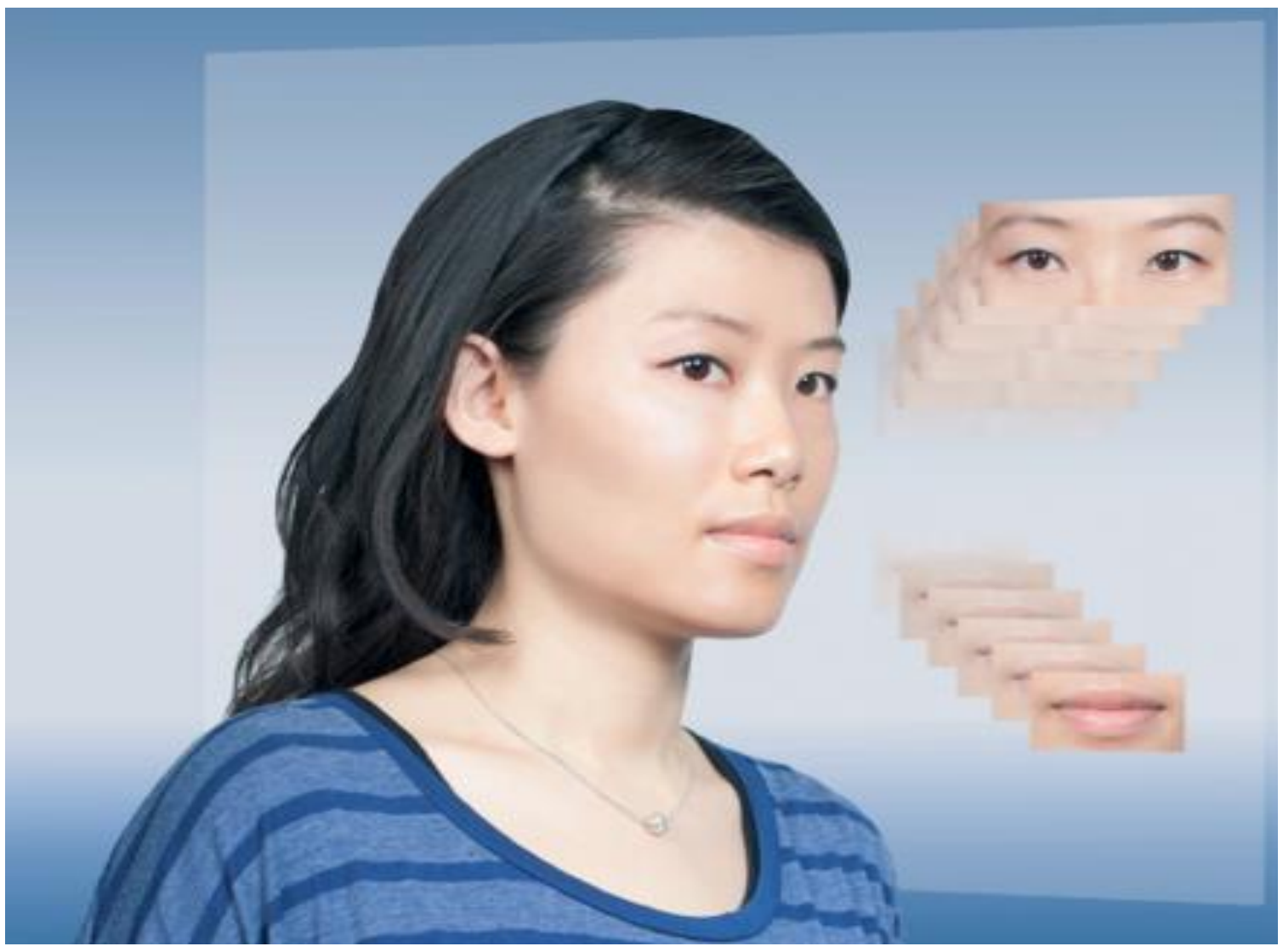

Kuva 9.9. Micro-expressions usually occur in the eyes and mouth regions. (© Jukka Kontinen).

As mentioned above, not only speed, but also the very low intensity of micro-expressions can be distinguished from ordinary expressions. The micro-expression can also be one-sided so that it starts slower, but ends very quickly, because the whole expression does not have time to come to the face.

According to Dr. Xiaobai Li (STT October 14, 2017), for example, in a happy normal feeling, corners of the mouth will rise clearly, but in the micro-expression, only one may rise slightly, or sometimes both. In a micro-expression corresponding to a negative feeling, the inner corners of the eyebrows are pressed down for a short time, and with surprised faces both eyebrows, or just one, may rise slightly. This interpretation is not entirely incontrovertible, since there is no unambiguous knowledge of how micro-expressions appear.

Micro-expressions can occur, for example, when lying in border control or police interrogations, hiding one's opinion in business negotiations, computer-aided teaching when a student has not understood what he or she is teaching, in some psychiatric illnesses or in poker. It is very difficult to see them with the human eyes. It has been found that even a well-trained person can only recognize less than 50% of micro-expressions.

In 1966, researchers Ernest Haggard and Kenneth Isaacs were the first to detect such facial expressions when examining films taken from therapy sessions of couples, noting that these "micro-momentary expressions" flash so quickly that they are difficult to see except in slow motion filming (Li et al., 2018).

Such expressions later became known from the work of psychologist Paul Ekman, who also called them micro-expression. Particularly noteworthy was the case where Ekman and his colleague were studying a video of a psychiatric patient. Later, the patient admitted that she had been about to commit suicide, even though she seemed to be happy with the video all the time. However, as the video was examined more picture-by-picture, a hidden anxiety was found on the face that lasted for a period of two consecutive frames (1/12 s).

Micro-expressions are often confused with other fast expressions in public because of their apparently good “market value”. Even in Finland, a human personality test based mainly on micro-expressions has been marketed. However, an article in the Helsingin Sanomat newspaper dealing with it stated that these expressions could not be true micro-expressions, and that it is not possible to determine a person's personality mainly through video clips taken from the face (Tiainen, 2018). One of the reasons for the market value is probably the television series "Lie to Me", presented in 2009-2011, in which the main character was one of the world's best lie detection experts, able to analyze for example micro-expressions.

However, with over 40 years of practical experience in interpreting human speechless communication, Joe Navarro (Navarro, 2011) has warned against giving micro-expressions too much importance. For example, lying cannot be revealed solely on the basis of micro-expressions or other individual behaviors. They are indicators of stress, psychological distress, anxiety, aversion or tension - but not lying. In his view, more generally, more attention should be paid to everything the human body communicates, not just the face and the micro-expressions.

In the United States, TSA has a number of people trained at passenger airports to examine the behavior of travelers, including micro-expressions. Launched in 2007, the SPOT (Screening Passengers by Observation Techniques) program aims to see if there are potential terrorists among travelers.

However, recent leaks indicate that the result has been quite poor, with only 1% of the 30,000 checkpoint passengers being arrested related to cases involving drug use or unreported baggage imports. No terrorists found! (Web-TSA)

Training tools have been developed to improve the detection of micro-expressions, for example by Paul Ekman. In Finland, too, it might have been considered to look at micro-expressions and other behaviors in border control, as mentor, magician and trainer Jose Ahonen said he has been training Finavia security inspectors (Myynti & Markkinointi, 8.6.2016). According to him, identifying micro-expressions is also useful in negotiations, sales situations or recruitment interviews. The article also refers to the famous Bill Clinton's claim that he had no sexual relationship with Monica Lewinsky. Leaving in front of the camera, Clinton is said to have a micro-expression related to arrogance, which revealed him lying. As an interesting detail, Ahonen says that when he follows discussion programs for learning how to find micro-expressions, he finds the contradictions between speech, expression and gestures of particular interest.

Another Finnish mentalist specializing in micro-expressions is Mr. Pete Poskiparta. In an article presented in YLE news (https://yle.fi/news/3-9912709, 12.11.2017), he says that many car dealers are particularly good readers of micro-expressions. Rapidly flashing expression of arrogance, a one-sided smile, can mean that the buyer is over-bidding and promising a very good deal for the seller. In the micro-expression of a liar, he says that there is an expression of supernaturalism and complacency, that is to say the person's face flashes one-sided smile immediately after the lie.

According to Professor Matsumoto, there are two nerve pathways that originate from different parts of the brain (Matsumoto & Hwang, 2011). The pyramidal pathway controls voluntary facial movements, while the extrapyramidal pathway directs involuntary emotional activities. When people are in intense emotional situations but need to control their expressions, both of these pathways become activated, causing a "controversy" in facial regulation that allows for instantaneous leakage of micro-expressions.

### *9.5.1 Automatic recognition*

Automatic real-time recognition of micro-expressions has long been considered a difficult task due to the extremely short duration and low intensity of the expressions. The first studies were about acted, not real spontaneous expressions. In 2010, we were the first in the world to investigate natural or spontaneous micro-expressions, and in 2011 we published the results at the International Conference on Computer Vision (Pfister et al., 2011).

A test database had to be created for the study, because such a study had not been done before. The principle behind our creation of the SMIC (Spontaneous MICro-expression) database was that viewers were shown video clips evoking heavily various emotional states and were asked to remain motionless and keep their faces expressionless.

Examples of videos shown include Youtube cat videos (happy micro-expression) and the movies *Lion King* (sadness), *Shining* (fear), *Hellraiser* (disgust) and *Capricorn* (surprised) (Li et al., 2018). A total of 164 samples of micro-expressions were found by hand-crafting from the high-speed video footage taken with a fast 100 frames per second camera. Figure 9.10 shows the principle of an improved version of the micro-expression recognition method developed (Li et al., 2018).

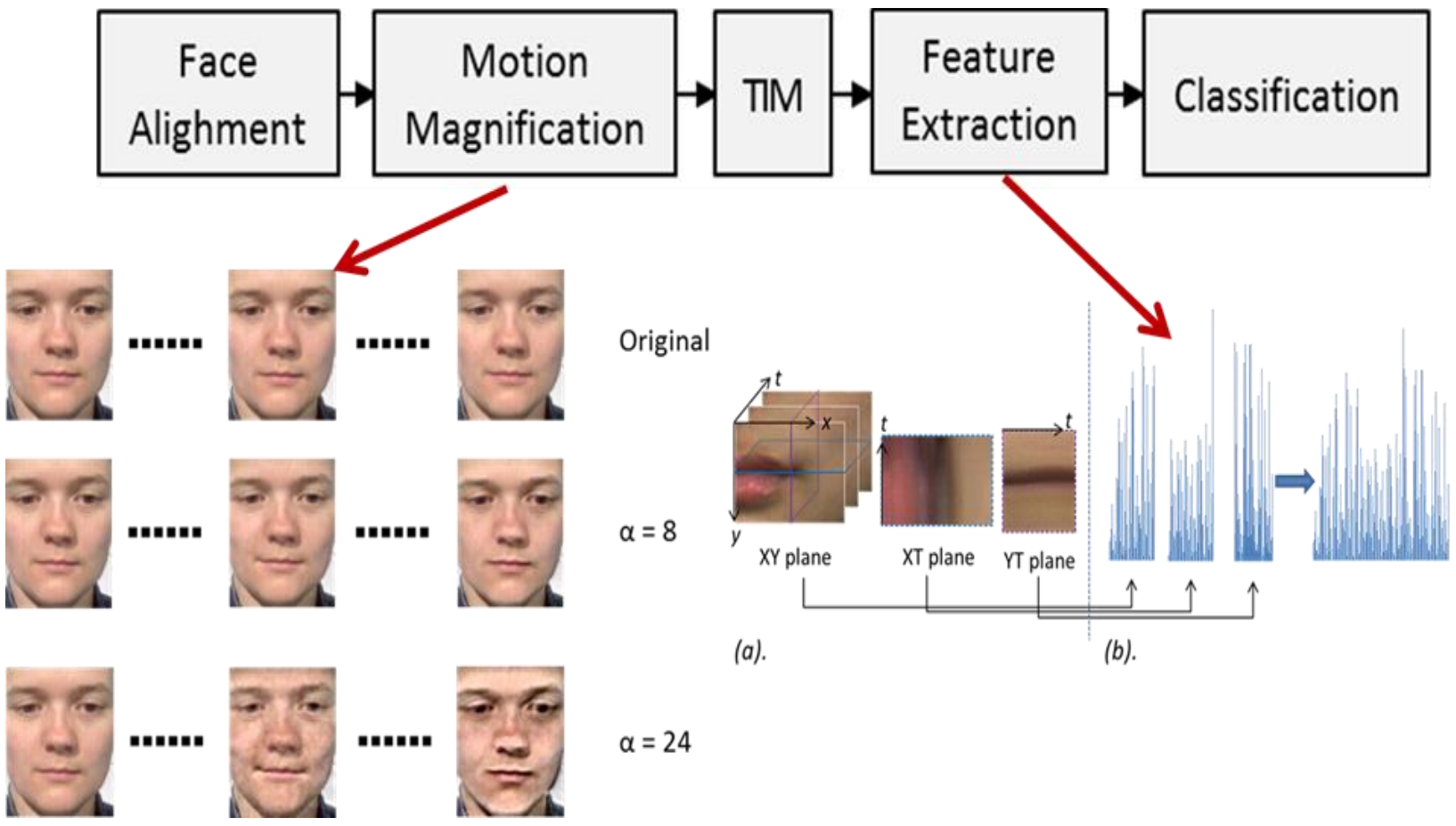

Figure 9.10. Improved version of the micro-expression recognition method. (© IEEE)

Initially, faces found on different videos are aligned at the same location in the face area using face alignment based on detected landmarks, and then motion magnification utilizing the motion of a sequence of images, is used to bring out the details. Subsequently, extra images are generated between two consecutive images by a temporal interpolation (TIM) method to obtain more images for ultra-fast micro-expressions and to make expressions of different lengths uniform in the feature extraction method. Then, the micro-expression features are described, for example, by the LBP-TOP method (Zhao & Pietikäinen, 2007) (see Section 7.1). The classification algorithm uses the computed features to identify what the micro-expression is.

In order to speed up the discovery of micro-expressions, a spotting method was developed which, prior to using the above-mentioned method, searches the video for locations that exhibit extremely rapid changes between successive images. Rapid changes can be caused, for example, by micro-expressions or eye blinks. The spotting method allows to use computationally much more laborious expression recognition only in those parts of the video sequence that may have a micro-expression.

Micro-expressions have recently been studied in many groups, but there are still many open research challenges as discussed by Zhao & Li (2019). Among these are lack of comprehensive datasets, detecting action units for micro-expressions, working in realistic uncontrolled situations, handling both micro- and macro-expressions, using context clues and multi-modal learning, and analyzing multiple persons in interactions.

## 9.6 Threats and Practical Problems

Publicly, there are often threats that technology will allow automatic recognition of the emotional state of any street person - the same style as face recognition can already be used to identify one. However, emotions are a much more vague concept than identity, and face images alone are not enough to recognize them.

There are still many practical problems associated with emotional state recognition. Right now, we need good, almost direct front-view photos with moderate resolution. Emotional expression is partly cultural and perhaps also genetic in nature, which may make it difficult to obtain a universal interpretation of the emotional features measured in each person's face and speech. Managing this problem involves a great deal of psycho-physiological knowledge acquisition.

It is certain that countermeasures are also under development. Technology has always been a race and it certainly is also here. For example, if in video conferencing there is a suspicion that the other party is trying to interpret emotions, solutions resembling the real-time beauty algorithms available for taking selfies on some mobile devices may in the future hide emotions without the recipient's knowledge. Now, video encoders hide reception errors, so why not hide half the emotion in pictures and speech? And, in such cases, the detection of heart rate and respiratory rate from videos can be significantly impaired even by active illumination, as long as such an analysis of physiological features and the like is at the transmission end.

But does emotional intelligence technology lead to face-to-face meetings for reasons of trust? Then the analysis or manipulation of the "channel" will not always be possible in real time by technical means.

And yet, all technology has both positive and negative sides. By providing a person with constant feedback on emotional information that he unknowingly conveys, is it not possible for a person to be able to avoid at least the transmission of negative feelings through practice? In job interviews, that can be good. Technology will probably have to provide the tools for such conditioning, for example, by telling the practitioner about the emotions that he or she is unknowingly communicating.

## 9.7 Challenges and Future Perspectives

Emotional artificial intelligence research is still largely in its infancy. Much of the research, as well as the basic theories of emotion recognition, have been made for a small number people with acted, non-natural expressions. The spontaneous natural expressions in our daily lives are very varied and different. The state-of-the-art technology is therefore most suitable for applications where emotion recognition accuracy is not critical.

The context greatly influences to expressions. Are we dressing up for weddings or funerals, are we resting or do some physical activity, do we watch a pet or something more boring? It has also been found that our cultural background affects facial expressions. According to a large population survey, Americans have the most smiles, Chinese and Japanese have the least smiles (McDuff, 2018). This does not mean that Americans are happiest.

The result is believed to be influenced by the degree of individuality of the people and, above all, by the heterogeneity of the population. In a multicultural environment, people need to gesture more to get their message across. And a smile does not necessarily mean that one is happy. Even the Finns are very low-profile when discussing, southern Europeans express a lot and use gestures. Social norms also affect feelings. Women generally smile more than men and children look differently than adults. In the morning, people's emotional state is at its most intense and falls towards evening.

All of these types of factors affect facial expressions. In addition, most current facial expression analysis methods work well only with almost front-facing videos, head position should not vary greatly, for example when speaking or gesturing, lighting should not vary widely, and the face should not be covered too much.

Further research is needed to achieve reliable operations without the limitations mentioned above (referred to as "in the wild"). Recognition of facial expressions during speech is also problematic and we should be able to utilize both together, since speech and mouth movements also contain significant information about the emotional state.

These issues are particularly difficult for recognizing very fast and hard-to-detect micro-expressions. Current methods only work well under almost studio conditions. In addition, current test databases (SMIC, CASME) have been imaged under such conditions (Li et al., 2018). This hampers the development of recognition methods that are suitable for demanding conditions. On the other hand, creating new databases is very tedious: micro-expressions are rarely found, and searching for video for annotation and system training is largely manual.

Facial expressions alone are not enough for most emotional AI applications. In addition, other modalities are needed, such as gestures, body movements, speech and gaze changes. Various physiological signals provide intrinsic information about emotions such as heart rate and heart rate variations, respiratory rate, electroencephalogram (EEG), skin conductivity. The role of speech and language is particularly important in human-machine interaction.

Better connections to human cognitive functions would also be important. Our thinking is related to emotions. In a futuristic brain-to-vehicle project Nissan is investigating the use of a simple brain electricity headgear (EEG) to adapt to driver's driving style and predict movement patterns (Thubron, 2018). The purpose is, for example, to predict the need for braking or turning and to perform the action 0.2-0.5 seconds faster than the driver, thereby preventing incidents with a small improvement in reaction speed.

By combining different modalities, it is possible to achieve better results in emotional state assessment. Figure 9.11 (partially edited from Vinciarelli et al., 2009, Fig. 1) is an example of the use of different modalities. A research problem of its own is how best to combine and learn multiple modality data.

As an example of multimodal recognition, we have investigated the use of facial expressions and EEG signals in emotion recognition (valence/arousal) for virtually expressionless face videos from the multimodal MAHNOB test database (Huang et al., 2016). Of the individual modalities, the EEG was clearly better than the facial expressions due to the low expressions of the faces, but when properly combined, the result continued to improve - well beyond what a person could rate based on those videos.

In addition, we investigated the determination of a group's "average" emotion state in a video by combining the facial expression information of each individual with information obtained from the upper body and the imaged background scene (Huang et al., 2018). Again, multimodal data yielded significantly better results than data obtained from individual modalities.

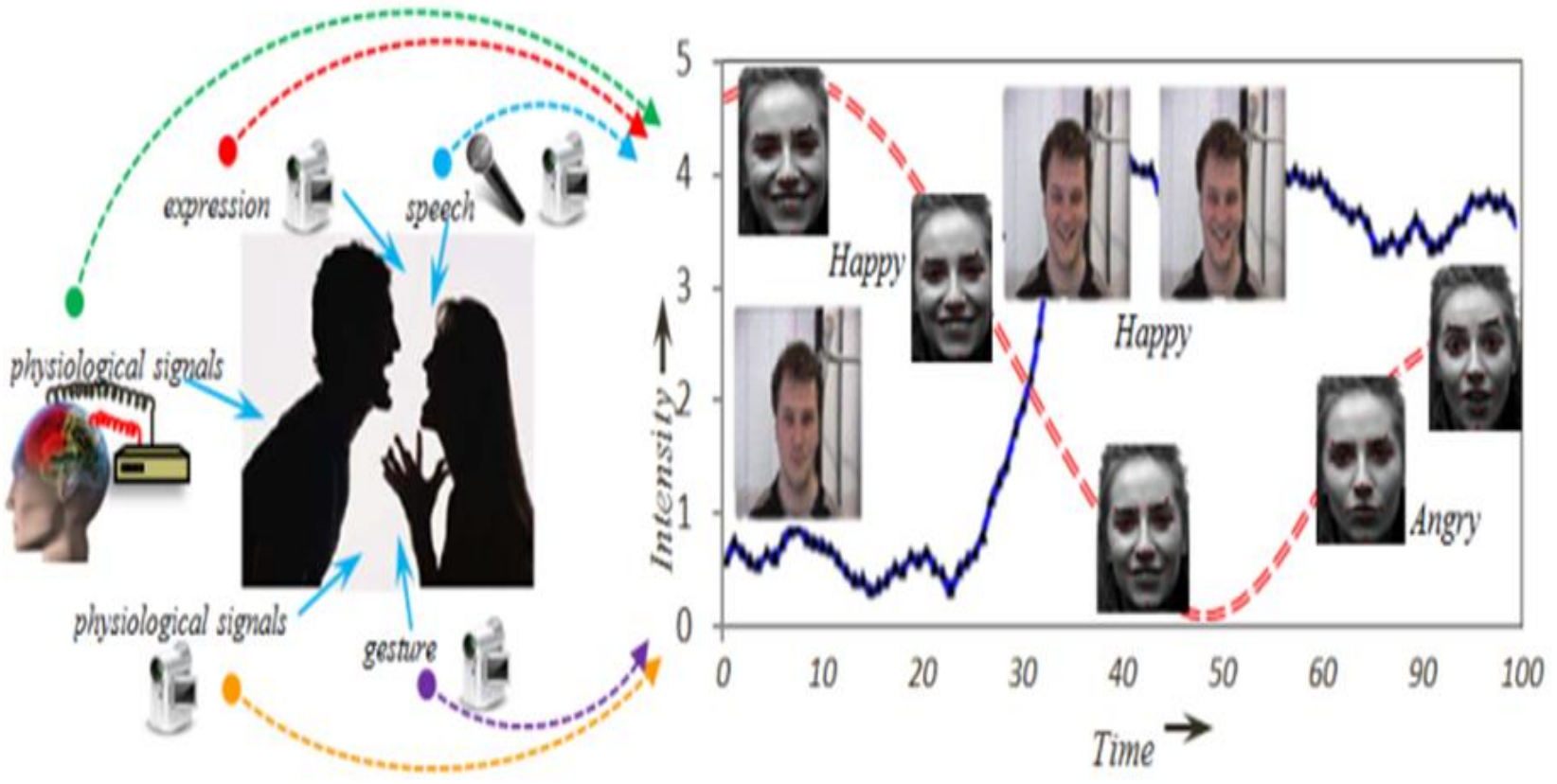

Figure 9.11. Multimodal emotion recognition.

The best success in recognizing emotions can be achieved by learning, if possible, personal emotion models for each person. This is similar to what was the case with the much easier problem of speech recognition. Speaker-dependent, that is, speech recognition system trained individually by each person's speech samples is much easier than speaker-independent - and only after significant improvements through deep learning and massive training data can speaker-independent recognition be successful nowadays.

The ways in which people express their emotions vary enormously, so it is not possible to achieve sufficiently good results for many applications using a common model. Focusing on only one application can improve recognition. In fact, this has been typical for most machine vision applications. Generally, in a given application, variations in the imaging conditions can be minimized and application-specific test data collected to the widest possible range of people, and these techniques facilitate analysis to obtain sufficiently high reliability.

## 9.8 References

Ekman P (2009) Darwin's contributions to our understanding of emotional expressions. Philos Trans R Soc Lond B Biol Sci 364(1535):3449-3451.

Ekman P & Friesen W (1978) Facial Action Coding System: A Technique for the Measurement of Facial Movement. Consulting Psychologists Press, Palo Alto.

Ekman P, Friesen WV & Hager JC (2002) Facial Action Coding System: The Manual on CD ROM. A Human Face, Salt Lake City.

Feldman Barret L, Mesquita B & Gendron M (2011) Context in emotion perception. Current Directions in Psychological Science 20(5):286-290.

Huang X, Dhall A, Goecke R, Pietikäinen M & Zhao G (2018) Multimodal framework for analyzing the affect of a group of people. IEEE Transactions on Multimedia 20(10):2706-2721.

Huang X, Kortelainen J, Zhao G, Li X, Moilanen A, Seppänen T & Pietikäinen M (2016) Multi-modal emotion analysis from facial expressions and electroencephalogram. Computer Vision and Image Understanding 147:114-124.

Li X, Hong X, Moilanen A, Huang X, Pfister T, Zhao G & Pietikäinen M (2018) Towards reading hidden emotions: A comparative study of spontaneous micro-expression spotting and recognition methods. IEEE Transactions on Affective Computing 9(4):563-577.

Martinez B, Valstar MF, Jiang B & Pantic M (2017) Automatic analysis of facial actions: A survey. IEEE Transactions on Affective Computing 10(3):325-347.

Matsumoto D & Hwang HS (2011) Evidence for training the ability to read microexpressions of emotion. Motivation and Emotion 35(2):181-191.

Mehrabian A (1972) Nonverbal Communication. Aldine-Atherton, Illinois: Chicago.

McDuff D (2018) Large-scale and longitudinal emotion analysis. Keynote speech, Workshop on Large Scale Emotion Recognition and Analysis, Xi'an, China.

Mielonen M (2018) Kaupan myyjä saattaa kohta tunnistaa, oletko iloinen vai nyrpeä (A salesperson may soon recognize whether you are happy or not). Helsingin Sanomat 29.10.2018.

Nalbantoglu M (2018) Tulevaisuus tarvitsee lisää empatiaa (More empathy is needed in the future). Helsingin Sanomat 6.11.2018.

Navarro J (2011) Body language vs. micro-expressions. Psychology Today Dec. 24.

Nummenmaa L, Glerean E, Hari R & Hietanen JK (2014) Bodily maps of emotions. PNAS 111(2):646-651.

Pfister T, Li X, Zhao G & Pietikäinen M (2011) Recognising spontaneous micro-expressions. In Proc. International Conference on Computer Vision, Barcelona, Spain.

Picard R (1997) Affective Computing. MIT Press.

Puttonen M & Heikkinen K (2018) Tunteet tarttuvat herkästi (Feelings are contagious easily). Tiede (science) Magazine, 2018.

Scherer KR, Shorr A & Johnstone T (Eds.) (2001) Appraisal Processes in Emotion: Theory, Methods, Research. Oxford University Press.

Thevenot J, Bordallo López M, Hadid A (2018) A survey on computer vision for assistive medical diagnosis from faces. IEEE J. Biomedical and Health Informatics 22(5):1497-1511.

Thubron R (2018) Nissan's mind-reading cars can predict drivers' actions. Techspot News 4.1.2018.

Tiainen A (2018) Mikroilmeet paljastavat tunteitamme, vaikka yrittäisimme peittää ne (Micro-expressions reveal our feelings, even if we try to hide them). Helsingin Sanomat 22.2.2018.

Vinciarelli A, Salamin H & Pantic M (2009). Social signal processing: Understanding social interactions through nonverbal behavior analysis. IEEE Conference on Computer Vision and Pattern Recognition, CVPR 2009.

Zhao G & Li X (2019) Automatic micro-expression analysis: Open challenges. Frontiers in Psychology, 07 August 2019.

Zhao G & Pietikäinen M (2007) Dynamic texture recognition using local binary patterns with an application to facial expressions. IEEE Transactions on Pattern Analysis and Machine Intelligence 29(6):915-928.

Web-Affectiva: This innovative video game can sense your emotions and respond accordingly

Web-Humex (2017) Uusi projekti tutkii tunteita ja vuorovaikutusta työelämässä (A new project investigates emotions and interactions in work life). Helsinki University 5.7.2017.

Web-Toys: Emotions at play: the potential for emotion enabled toys, Affectiva

Web-TSA: Yes, the TSA is probably profiling you and it's scientifically bogus. Business Insider, 6.5.2015.

Wiki-Appraisal:Appraisal Theory

Wiki-Emotion: Emotional Intelligence

Wiki-Tunne: Tunne (Emotion)

# 10 Is Super-intelligence a Threat?

## 10.1 Super-intelligence and Risks of AI

Super-intelligence refers to an imaginary “agent” that transcends the intelligence of the sharpest and most talented people (Figure 10.1). It can arise in the context of so-called explosion of intelligence or similar technological singularity. At that time, the development of intelligence accelerates beyond the reach of human understanding (Wiki-Singularity).

Some researchers believe that singularity will soon follow the development of generic, human-like artificial intelligence, which in turn will develop new artificial intelligence solutions. Depending on the proposer, the prerequisites for this are some or all of the following:

1. Computers become fast enough, in particular, quantum computers could contribute to achieving singularity. This thinking prevailed when artificial intelligence was seen primarily as a combinatorial search problem, and this assumption is still not rare. The central justification is the huge parallel processing capacity of the brain.
2. The memory capacity of computers becomes adequate, whereby singularity occurs when a sufficient number of neurons can be realized. It is debatable whether one billion, or one hundred billion, like in the human central nervous system, or over thousands of billions of neurons is needed. At the same time, of course, the senses with neurons and motor abilities should be realized in order for the artificial intelligence to be able to independently acquire knowledge and experience.
3. Sufficient data and knowledge has been collected, after which artificial intelligence starts to learn more independently. The Cyc project (Wiki-Cyc), which has compiled a huge set of rules for everyday life, represents this direction. This has been the longest project in the history of artificial intelligence.
4. Significant breakthroughs in artificial intelligence theory are achieved for which researchers and developers are needed. This is the "give us enough money" justification disliked by the decision-makers. The challenge is that the research should investigate significantly different approaches. The current mainstream is to explore the potential of deep convolutional networks, where deadend can lead to funders' frustrations with oversized promises and a "back-winter" of research.

Professor Nick Bostrom of Oxford University (Wiki-Boström) has written a bestselling book on super-intelligence called "Superintelligence: Paths, Dangers, Strategies" (Bostrom, 2014). In his book, he wants to replace the sci-fi singularity term with an explosion of intelligence.

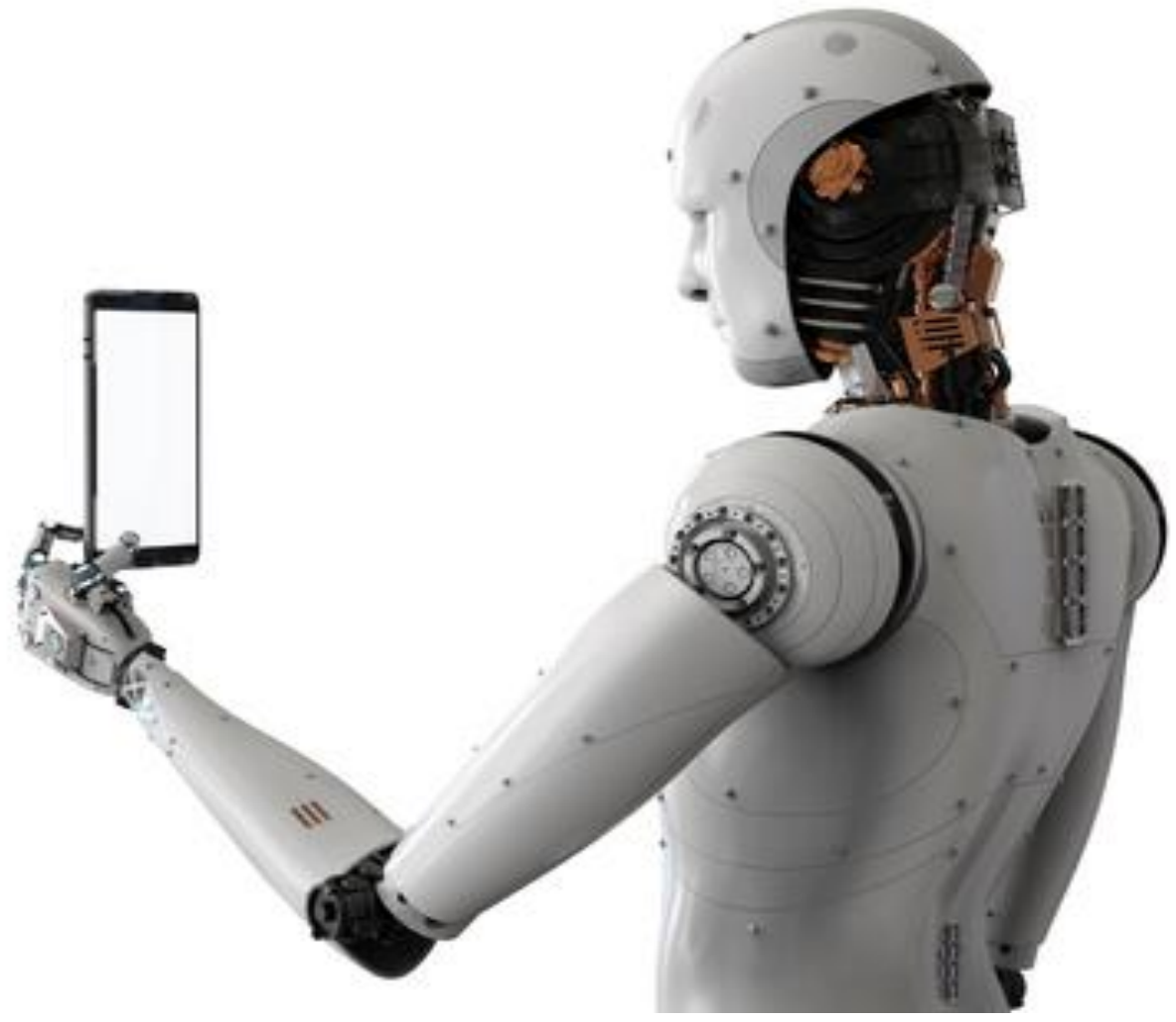

Figure 10.1. Do I sometimes get super-smart? (© 123RF)

However, both terms represent a crossing point beyond which society can no longer be identified due to overwhelming machine intelligence. Bostrom, as well as previous singularity predictors, led by writer, entrepreneur, scientist, and futurist Raymond Kurzweil (Wiki-Kurz), Google's technical director, and Vernor Vinge (Wiki-Vinge), a mathematician, computer expert and science writer, see the destiny of mankind to be in the hands of smarter machines.

One of the key justifications for the predictions has been the rapid development of computer components and computer systems. Biological neurons operate at speeds of up to 200 Hz, tremendously slower than modern microcomputers, let alone modern supercomputers. In addition, human neurons transmit impulse-like signals at a maximum rate of 120 meters per second, while electronic systems can communicate optically at the speed of light.

Computer memory and computing capacity can be continuously increased as technology advances. The latest quantum computers under development are believed to lead to the emergence of super-intelligence. However, along with super-intelligence, there are strong doubts about the introduction of quantum computers in real life (Dyakonov, 2018).

Belief in super-intelligence has strengthened with recent advances in machine learning. Deep neural networks using massive data have achieved better results in many tasks than with humans, as discussed in Chapters 4-5.

The dangers of artificial intelligence have been the subject of much debate lately - from threat paintings by physicist Stephen Hawking and entrepreneur Elon Musk. Bill Gates, too, has been concerned about the result of the super-intelligence.

The late Stephen Hawking has argued (2017) that artificial intelligence can replace humans after someone develops technology that is capable of continuously improving themselves. The result is a new kind of life. He has argued that artificial intelligence could even be the worst thing that has happened to our civilization throughout its history if we do not learn how to prepare for it and avoid potential risks.

Artificial intelligence could bring threats such as powerful autonomous weapons or new ways for few to oppress many. It could bring a major disruption to our economy. To prevent this, “best practices and efficient management” should be required from the creators of artificial intelligence.

Elon Musk has argued (2018) that artificial intelligence is more dangerous than nuclear weapons and that there should be a controlling body watching the development of it. Previously, he has also argued that artificial intelligence is much more dangerous than North Korea and that developments in the industry should be monitored.

The authors of this book suspect that the greatest risks of artificial intelligence culminate in influencing human behavior. There is already evidence of the effectiveness of social media manipulation using artificial intelligence.

## 10.2 Significant Limitations with Current AI

However, it should be noted that most of the strongest critics are not artificial intelligence researchers, meaning that they may not be fully aware of how difficult it is to develop strong artificial intelligence in particular.

In 1956, it was predicted that after 25 years, people would be able to concentrate on leisure activities, as machines do much of the work. However, the development of intelligent machines has proven to be much more difficult than we thought: success has been achieved only in well-defined tasks in weak artificial intelligence - through a major investment in research and development. The development of systems with general human-like intelligence is, according to most researchers, impossible.

Some critics of artificial intelligence, led by Professor Hubert Dreyfus already in the 1970s, concluded that it was generally impossible to develop machine intelligence. Human reasoning is largely based on intuition: the inexplicable realization of what coding into a computer program is too difficult to do. It is difficult for a machine to cope with tasks that require creativity, extensive general knowledge or aesthetic evaluation, knowledge, causal evaluation, emotion. He has also argued that because computers do not have a body, childhood, and cultural practices, they cannot become intelligent (Moreno, 2021).

Professor Roger Schank, one of the pioneers of artificial intelligence research (related to understanding natural language and case-based reasoning), said in October 2018 that artificial intelligence does not even exist at this time. Recent developments in his mind are only about very fast computing (CNN interview Oct 18, 2018). Only when the machine is able to talk to people in the same way as he talked to a journalist can artificial intelligence be talked about. So, in his opinion, only the so-called strong artificial intelligence is real AI.

According to Helsingin Sanomat (Paukku, 2017), another pioneer of artificial intelligence, Patrick Henry Winston (1943-2019), who was long the director of the artificial intelligence laboratory at MIT, has said. "When people talk about artificial intelligence, they actually talk about the computer's raw computing power. A machine is not smart if it looks smart. Instead of artificial intelligence, we should talk about tricks that can be done with a computer."

Judea Pearl, the winner of the Turing Prize for Information Technology, a pioneer of probabilistic artificial intelligence and a pioneer in causality research, believes that current artificial intelligence is only a sharpened version of old AI, being able to do what it has done already a generation ago, that is, to find hidden regularities in a large amount of data (Hartnett, 2018).

"All of the convincing achievements of deep learning add up to curve fitting," Pearl said. However, he has been impressed by how many problems can be solved simply by fitting curves to the data.

The human brain is known to have about $10^{11}$ (= 100 billion) neurons. Each of these can connect to up to 10,000 other neurons, carrying messages through up to $10^{15}$ synaptic interfaces (Web-Memory).

One key factor for unmatched brain performance is their massive parallel processing of data with low power consumption (Dicarld, 2018). The average adult power consumption is about 100 watts, and the brain accounts for about 20%, or 20 watts. The IBM Watson computer that won the man in the Jeopardy quiz, a very limited issue, claimed 20,000 watts.

Achieving brain performance as a whole in human intelligence in similar tasks to the same power use would require that we be able to mimic the structure of the brain well enough with back modeling, through reverse engineering. The development of such a computer is far in the future, if at all possible.

Namely, the rest of the human body has a huge amount of neurons (Figure 10.2) in the senses transmitting intracellular signals to the brain. This calls into question the idea of creating super-intelligence solely on the central nervous system. How to create an entity that can function in its environment?

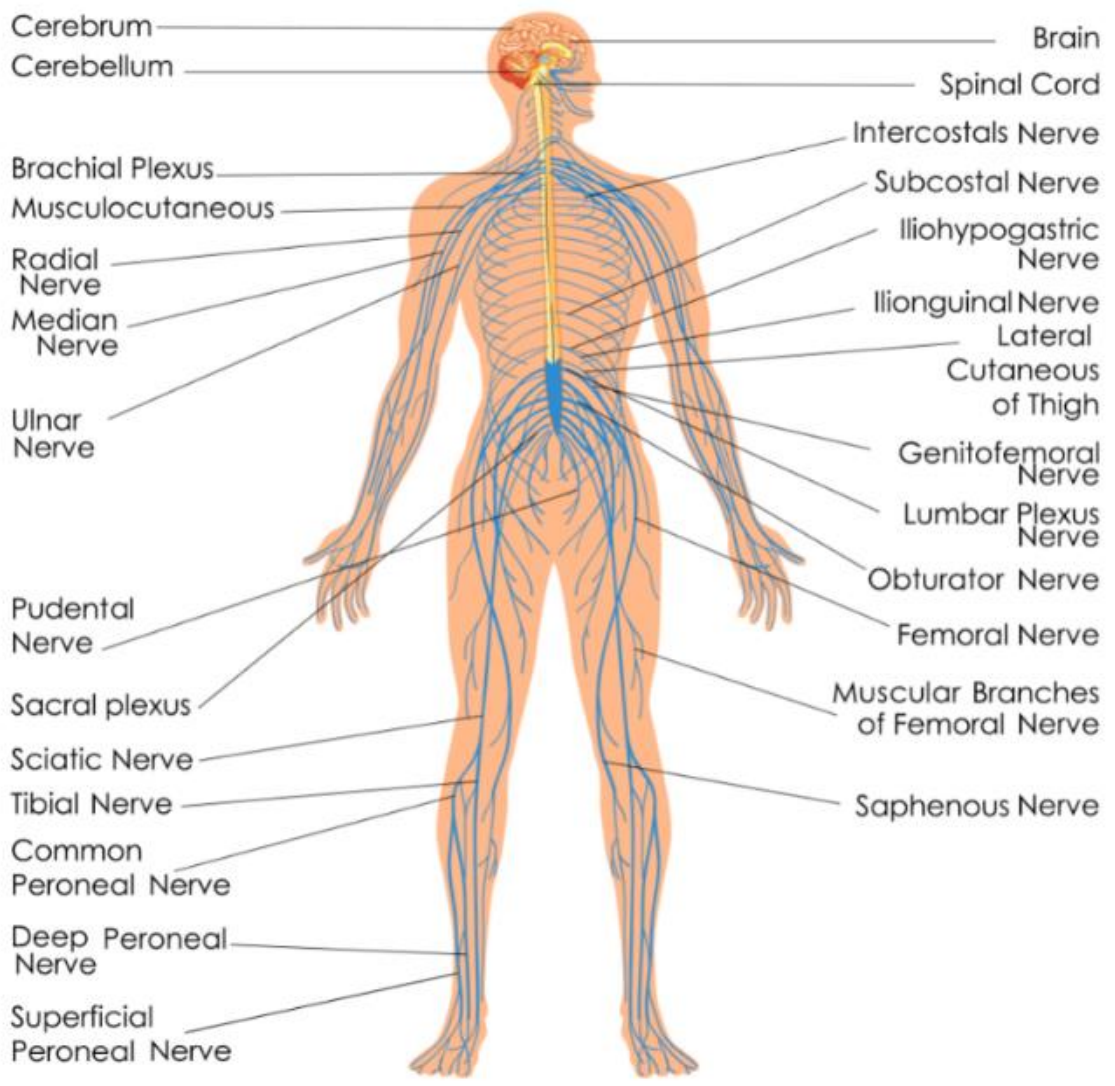

Figure 10.2. Human nervous system. (© 123RF)

Currently, artificial intelligence is mainly based on one algorithm, i.e. deep learning (Section 4.5) and its variations. In Figure 10.3. is an example of a typical deep learning convolutional neural network.

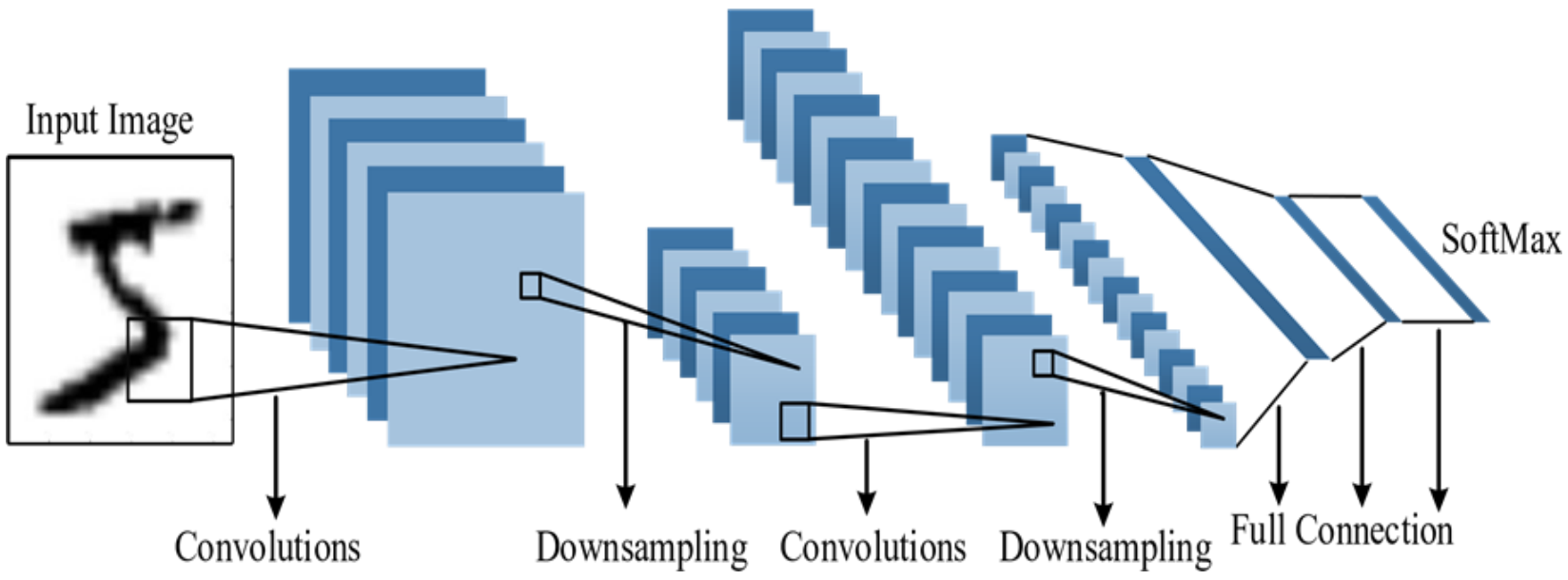

Figure 10.3. Learning with a convolutional neural network. (© Li Liu)

Professor Gary Marcus (New York University) has recently published a very interesting and eye-catching article, "Deep Learning: A Critical Appraisal," on the limitations of deep learning and future challenges of AI (Marcus, 2018). He says deep learning has led to an undisputed development in many problems, such as speech recognition, image recognition, and game playing - and has aroused enormous interest in artificial intelligence in a variety of media.

However, we are very far from artificial intelligence like human intelligence. Expectations have risen too high, and there may be a "back winter" of research again if the over-expectations are not met. In his article, Professor Marcus presents ten concerns about current deep learning methods (Marcus, 2018):

1) Deep learning requires a lot of data. People are able to learn abstract relationships between objects with a few trials. For example, the word "teen" refers to a person between the ages of 13 and 19. You can then immediately determine whether there are any teens in your circle of friends.

   This learning does not require numerous examples, but the ability to represent abstract relations with algebra-like variables. Humans are able to learn such abstractions as early as seven months of age, in a few minutes, from a few examples. Deep learning methods lack the ability to learn abstractions based on verbal definition. They work best when there is a huge amount of teaching samples available.

2) Deep learning is actually low learning and has a limited ability to move from one problem to another. The word "deep" primarily refers to the structure of the neural network, that is, how many hidden layers are used instead of the previously used one. Deep learning is in no way capable of handling abstract concepts such as "right" or "democracy", and even more specific concepts such as "ball" and "opponent" can be difficult to interpret. Transferring a trained deep learning method to a slightly different problem can be extremely difficult.

3) Deep learning lacks a natural way of dealing with hierarchical structures. For this reason, for example, the understanding of natural language by these methods has not yet reached very advanced levels. Most deep learning language models represent sentences as mere word sequences, whereas in reality, language is largely a hierarchical structure, where larger structures can be recursively constructed from smaller components. At the lowest level are phonemes and then morphology or combinations of letters or phonemes. Next are single words and then combinations of words, or syntax. Semantics describes the meaning of an expression already spoken or written. At the highest level is pragmatism, which is related to the limitations of how to use the words used and how to interpret the language in different situations.

   Problems similar to those of automatic language interpretation are encountered, for example, in routing, scheduling, etc. tasks, and in robotics.

4) Deep learning has difficulty with open-ended reasoning problems. For example, Marcus exemplifies the inability of a machine to distinguish between nuances of, say, "John promised Mary to leave" and "John promised to leave Mary".

As a result, the machine is unable to determine who is leaving anyone or what happens next. Some progress on such problems has been made recently, but is not even close to human ability. Reflections on these problems can be seen when examining, for example, machine translation problems from Finnish to English or vice versa.

5) Deep learning is not transparent enough. The methods based on neural networks work to a large extent as the so-called "black boxes". Systems contain millions or even billions of parameters, the importance of which cannot be well analyzed by developers. If deep learning systems work well enough and independently in the application, the issue is not very problematic. But for example, in medical diagnoses or business applications, the user wants to understand why the system came to a particular solution, or why a self-driving vehicle crashed.

6) Deep learning is not integrated with prior knowledge. Current deep learning methods work independently without utilizing other potentially useful information to solve the problem. On the other hand, people use a variety of prior information to solve problems. For example, the use of training samples alone does not adequately solve problems where knowledge of the laws of physics is required.

7) Deep learning lacks the natural ability to distinguish cause and effect relationships from correlation. The method is capable of learning the complex interdependencies between its inputs and outputs, but it has no natural way of presenting cause and effect relationships, for example, in medical diagnoses of a disease and its symptoms.

8) Deep learning assumes that the surrounding world is largely constant. The method works best in stable ground conditions, such as table games where the rules of the game do not change. On the other hand, for example, in politics and economics, the rules are constantly changing.

9) Deep learning works well as an approximation, but the answers it gives cannot always be fully trusted. A number of cases have been reported, particularly in machine vision, where deep learning has yielded completely false results. A famous example is where a speed limit traffic sign and a three-dimensional printed turtle are identified as rifles.

10) Deep learning is difficult to put into practice. It is relatively easy to make systems that operate under limited conditions, but it is difficult to guarantee that they will operate under different conditions with new data, which may not be reminiscent of prior teaching data. Training, for its part, requires examples where different cases are even in balance.

As a result of these problems, Professor Marcus suggests that deep learning must be complemented by other techniques in order to access general artificial intelligence that resembles human intelligence.

Deep neural networks and their variations are currently being researched extensively, and at least partial solutions to many of these shortcomings can be found. In any case, it is clear that current networks are very far from being used for deeper, strong artificial intelligence. Another major challenge of related research is how to solve computational problems related to deep learning, leading to the need for massive training data, the use of ultra-fast dedicated processors, and high power consumption. These significantly limit the use of deep learning in, for example, mobile devices or smart sensors embedded in the environment, where the energy consumption must be very low.

Deep learning works very well with limited problems when there is a very large and sufficiently comprehensive supply of natural training data. There are many such applications as well. However, in many practical problems, huge amounts of training data are not available. However, deep learning systems should be able to generalize their activities to cases not included in the training data. Even in infancy, people can learn what they see from single or a few images without the need for massive amounts of training data.

In addition to these, current artificial intelligence is largely tied to a specific task or problem: to identify objects, to recognize speech, to make stock price predictions, etc. The human brain is capable of handling a huge variety of tasks, many of them simultaneously. Again, the artificial intelligence system must be taught to each task individually with a massive amount of training data. A huge number of separate but cooperative deep learning networks would be needed to be able to perform tasks that resemble everyday human activities.

## 10.3 Artificial vs. Human Intelligence

In order for a machine to have strong artificial intelligence like human intelligence, the following capabilities would be needed:

- How do you make your machine think? Current artificial intelligence is at its best in machine sensing, but the machine cannot think.
- How do you get the machine to do many different tasks like many people do, often at the same time?
- How do you make your machine learn and improve through experience as people do?
- How do you get the machine to justify its decisions, which are necessary in many applications, such as medicine, business applications - and self-driving vehicles?

- How do I make a machine work in a changing operating environment without re-training it? How to make the machine work even in simple everyday tasks (Figure 10.4)?

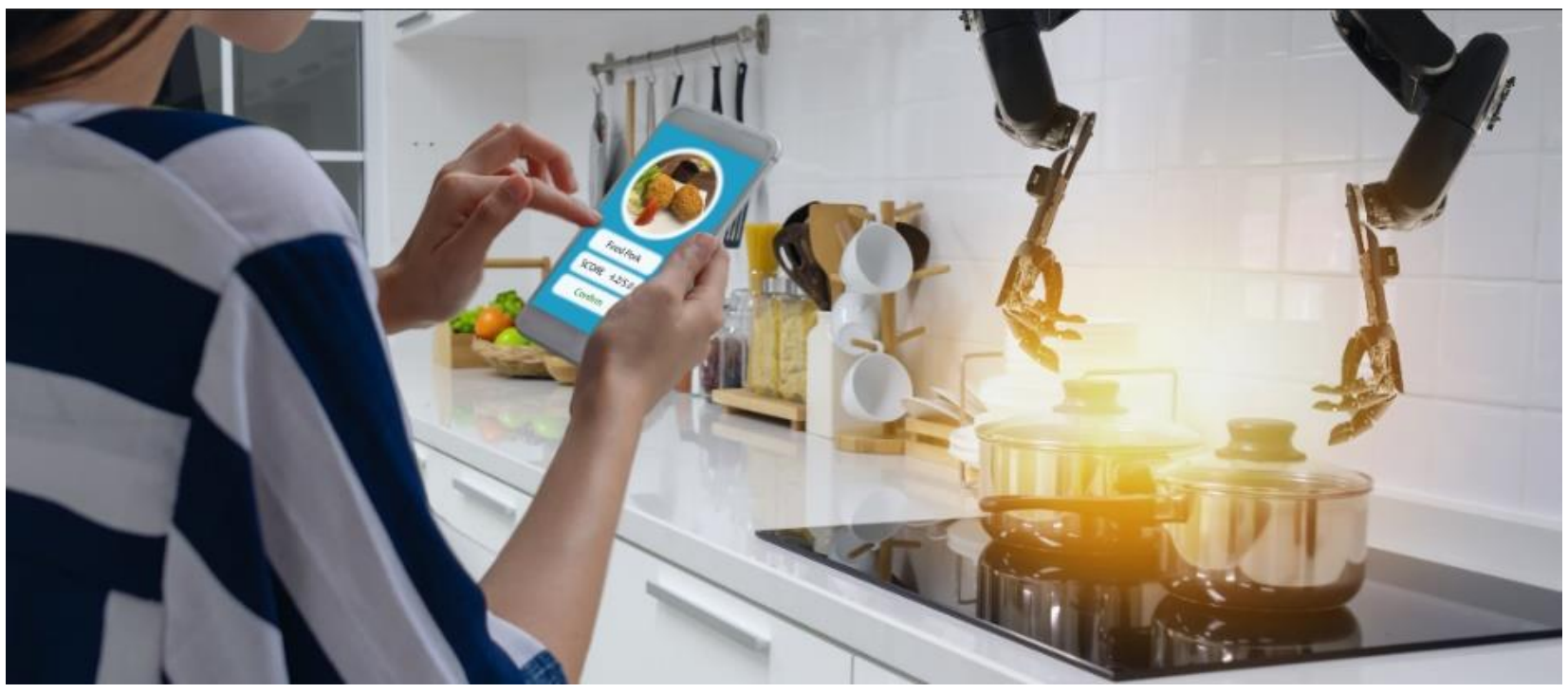

Figure 10.4. Kitchen robots have been awaited. (© 123RF)

- How can the machine understand spoken and written language and learn with their help as people learn?
- How can you make the machine learn how to recognize objects from one or a few sample images, as little children can do? The brain also uses cognitive functions of more general form than deep nets to interpret images.
- How does the machine gain the ability to understand causality, i.e. cause-effect relations? For example, why did the medical diagnosis program end up with the reported result? Why is the economy projected to evolve as predicted by an artificial intelligence program?
- How do I get the machine to design complex tasks and change plans during execution? For example, to change the intended route of the car when the situation changes abruptly and prevent you from colliding with a reindeer that appears in front of the car.
- How to bring to the machine consciousness, which refers to the totality of the senses, experiences, feelings, memories and thoughts that an individual experiences at any given moment, or the individual's awareness of themselves and their environment (Wiki-Consciousness).
- How could you get an original human-like creativity on the machine?
- How do you get the machine to feel? People are very sensitive and emotional, they see, hear, think and feel - and emotions guide their thoughts.
- How can you make a machine evaluate moral issues, for example, what is right and wrong?
- How does the machine get the intuitive ability typical of the human brain? Intuition means knowing or understanding where information comes directly from an object or event - not by reasoning. For example, for some reason, the path of the walk is changed from normal and then something surprising happens, or think of a person and he or she will call. It is often referred to as the "sixth sense".
- How do you implement computers that use massive parallel computing with low power consumption like the human brain?

It would be very expensive to develop and maintain systems with the above features and sufficient reliability. From infancy, a person can identify new objects he or she has seen on the basis of just a few examples, learn the interdependencies between objects and things, learn to speak, read and understand language, walk, interact with other people, etc. Artificial intelligence is far from this!

In conclusion, the imagined super-intelligent systems in Figure 10.5 seem impossible to implement based on current knowledge. We believe that artificial intelligence is above all a human helper, does not control him. Semi-autonomous vehicles facilitate driving and traffic, search for suitable driving routes and prevent accidents. Artificial intelligence helps doctors diagnose diseases, not replace them. It helps people in their daily lives and helps them to interpret the massive amount of data, for example in corporate decision-making.

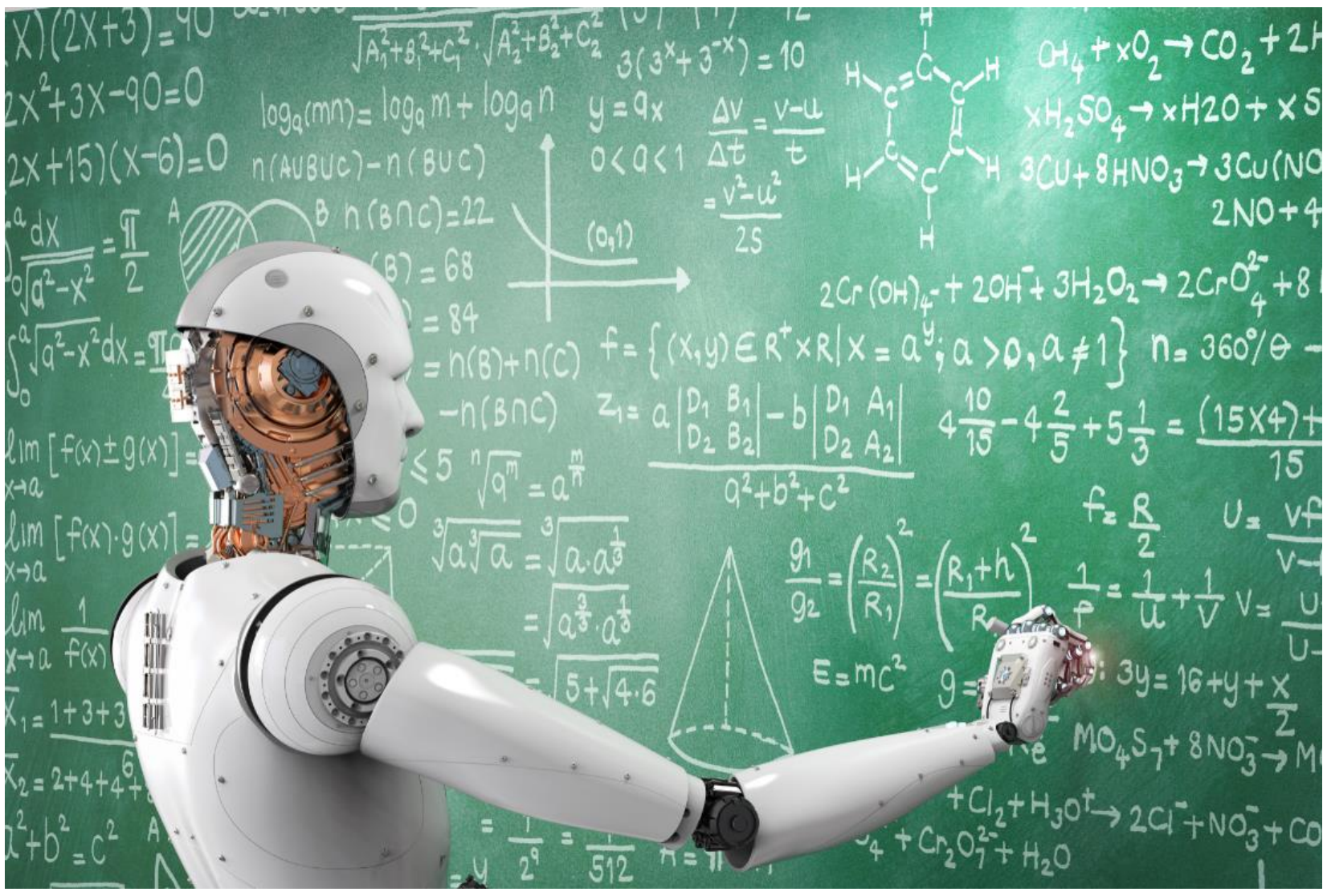

Figure 10.5. Imagined super-intelligence. (© 123RF)

Machines can also show primitive creativity, such as giving artists new tools to help them work. Emotions play a central role in human-to-human communication. We will soon be able to utilize elementary “emotional intelligence” in human-machine interaction, and research on this is also being carried out in our own research unit.

In the words of Patrick Henry Winston, as mentioned earlier: “A machine is not intelligent if it looks intelligent. Instead of talking about AI, we should talk about the tricks that a computer can do." You can teach a machine good or bad things. If a person displays images of violence on the machine, the machine learns to recognize them. Weapons have used machine vision and other sensory information, for example, in the search for and tracking of targets long ago. Artificial intelligence techniques can also be applied to decision-making in "smarter" weapon systems, if desired. As technology advances, smart features will be added to a wide variety of applications and systems, but humans are their developers, not artificial intelligence.

## 10.4 How from Now on Towards Strong AI?

Due to the shortcomings of current deep learning methods, there has recently been a growing debate about what to do after deep learning (Vorhies, 2018). In this regard, limitations of the back-propagation algorithm, for example, in relation to its requirement for differentiation, have been disclosed. This means that the partial derivatives of the function being modeled are assumed to be continuous at each point.

Even Geoffrey Hinton, a pioneer in deep learning, has taken part in this discussion and presented a solution called Capsule Networks (CapsNet), for such as image classification. The aim is, for example, to reduce the enormous need for additional training material for variations in the position, size, distortion, color, texture, etc. of objects in the image, particularly in the case of three-dimensional machine vision problems (see Chapter 6). The method adds new layers inside individual network layers to increase invariance to the above image transformations and thereby significantly reduce the amount of training material needed. By comparison, a person is often able to recognize an object even if he has only seen it from one angle.

An interesting solution is the gcForest technology introduced by professor Zhi-Hua Zhou (Zhou & Feng, 2017). In his research, he has shown that so-called gcForest, a method based on forest search, regularly wins CNN and RNN in both text and image categorization tasks and has many significant advantages over the above methods:

- The complexity of the model is adaptively determined by the data,
- requires only a fraction of the training data needed by CNN and RNN,
- can be run only on a central processing unit (CPU) without GPU high-performance processors,
- teaching is as fast or often faster and well adapted to distributed computing,
- contains much less so-called hyperparameters and works well with default settings,
- is based on easy-to-understand random trees, as opposed to completely opaque neural networks.

In summary, gcForest (multigrained cascade forest) is a decision tree based ensemble classifier that preserves the sequential structure of deep networks but replaces opaque neurons with groups of random forests combined with fully random decision trees (see Section 4.3).

How, then, can the artificial intelligence system gain more human-like features, expand its application capabilities, and move toward strong artificial intelligence? You can get some guidance on this by looking at some of the ideas of the artificial intelligence experts.

As early as 1990, late *Marvin Minsky*, one of the most prominent researchers in the history of artificial intelligence, argued that in order to develop truly intelligent systems, it would be necessary to integrate neural or other massive parallel computing and symbolic data processing traditionally used in artificial intelligence (Minsky, 1990).

Neural computing can cope well with, for example, pattern recognition and knowledge gathering, but so-called symbolic systems are needed for high-level thinking, such as goal-based reasoning, structuring things, and exploring cause and effect relationships. How to implement such an integrated system was, in Minsky's opinion, the greatest challenge for research. The conclusion of the first author of this book, based on his experience, was largely similar to that in his professorship's inaugural presentation "Artificial intelligence developers are facing great challenges" (Pietikäinen, 1992).

*Zeeshan Zia*, an artificial intelligence researcher working at Microsoft , has largely agreed with Minsky in his comments. In his view, the three main research themes for the coming decades are:

1. Combining paradigms based on symbolic computation, probabilities and deep learning. At the highest level of awareness, we think with symbols. How would you describe a food recipe, for example? How do you communicate math? How do you decide how to move from one place to another? Symbol processing is central to any general intelligence that resembles human intelligence.
2. Sensors produce noisy data and our world models are noisy. Graphical likelihood models have been very useful in modeling many practical problems, especially if there are few random variables or a large amount of domain knowledge is available. In addition, such models are much easier to understand than deep learning models. Current causal modeling tools aiming for cause estimation are based on probabilities.
3. Deep neural networks are particularly useful in lower level pattern recognition problems, such as machine vision, speech recognition and, increasingly, limited interpretation of natural language. Professor Andrew Ng, one of the leading researchers in deep learning, has suggested that the intellectual tasks that a person can do in less than a second are ideal for neural networks.

However, this is not entirely true. Deep learning works well when there is a very large and sufficiently wide range of natural training data available. In many practical problems, a huge amount of instructional data is not available. Already in infancy, people are able to learn what they see from single or a few pictures

*Sridhar Mahadevan*, who heads the Data Science Lab at the Adobe Research Center, says that artificial intelligence is now dominated by the data science / machine learning / deep learning paradigm. Everything starts with data and we should now get off it to the next paradigm. The data cannot provide an explanation of the events that would be necessary, especially in many critical applications, such as self-driving accidents or incorrect X-ray diagnosis of cancer. Why did the system malfunction?

In his view, the necessary explanation cannot be obtained from machine learning systems, but

1. A combination of statistical and symbolic reasoning is needed.
2. Nor can statistical reasoning produce causal explanations, as has been shown many years ago. Causal reasoning explores "what-if" type questions. For example, what happens if the US raises steel tariffs? Data science cannot answer this, but needs to know what the consequences of the operation are and the countermeasures of other countries.
3. Data represents what has happened, that is old news. Many applications do not yet have all or sufficient data for learning, such as automatic car steering in all circumstances. Figure 10.6 depicts multidimensional data, from which explanations are searched for phenomena.

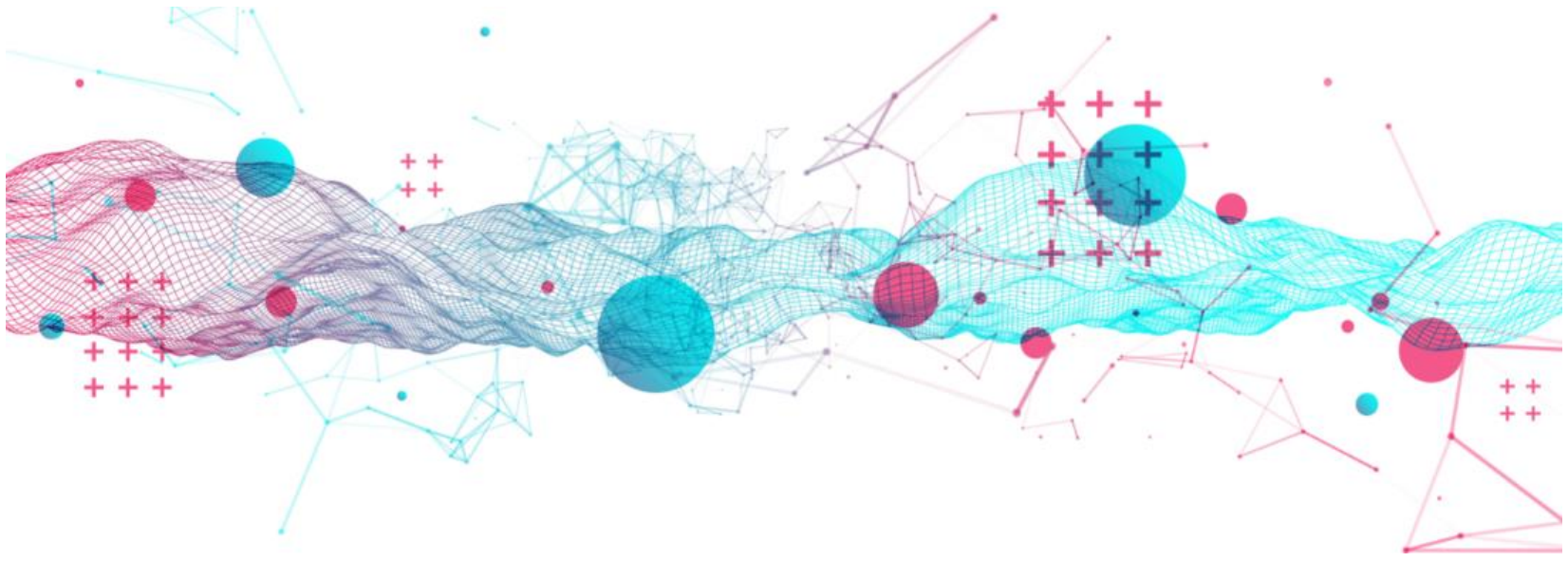

Figure 10.6. Visualized data structures. (© 123RF)

Professor *Yoshua Bengio*, a pioneer in deep learning, wants to build artificial intelligence at the heart of deep learning, but it should be expanded to include reasoning, causal learning, and exploring the surrounding world for machine learning and intelligence (Knight, 2018). "We would need to face the tough challenges of artificial intelligence and not be content with short-term incremental development. If you really want to approach the level of human intelligence, this is a different game, and academic research is better off carrying that torch."

*Fei-Fei Li*, a Stanford University professor and also with Google, known as the mother of the vast ImageNet image database that provided a breakthrough in deep learning, has suggested that humans should now be brought more in the center of artificial intelligence (Knight, 2017). "The excellence of today's artificial intelligence is largely pattern recognition, which is very task-oriented. It lacks awareness of contextuality (the connections between things) and the flexible way people have to learn.

We want to create technology that makes people's lives better, our world safer, and our lives more productive and better. To achieve this requires understanding of contexts like human intelligence, abstracting knowledge and reasoning."

Professor *Judea Pearl*, a pioneer in causal cause-effect research, has said that artificial intelligence has completely forgotten what intelligence really is (Hartnett, 2018). Only when the machine is capable of analyzing cause-effect relationships and answering the question Why?, can real artificial intelligence be talked about. "Instead of finding only a correlation between malaria and fever, the machine should be able to conclude that malaria is causing fever."

Causal reasoning is at the heart of human intelligence and provides the basis for scientific thinking and reasoning, but in science it is largely forgotten - even if compared to statistics and probabilities. Pearl has recently written a book entitled "The Book of Why: The New Science of Cause and Effect" (Pearl & Mackenzie, 2018).

*Brent Oster*, who has been for four years at Nvidia, a company specializing in high-performance artificial intelligence hardware, has worked on hardware architectures and methodologies related to deep learning implementations (more recently he has been heading ORBAI - Artificial Intelligence and Robotics company). He believes that we can get closer to strong AI only by implementing systems much more resembling the structure and function of the human brain (Oster, 2018).

Oster has potential long-term solutions to this issue. Current machine learning, often also referred to as artificial intelligence, is, in his view, merely a matter of adapting very large functions to even larger sets of data so that these functions can predict future data. In his view, the word "neural" and even less the word "artificial intelligence" should not be used in this context. Existing "neural networks" and "neurons" are in structure and function simply a small subset of a much broader and richer family of synthetic neurons, neural networks, and methods.

### 10.5 References

Bostrom N (2014) Superintelligence: Paths, Dangers, Strategies. Oxford University Press, 352 p.

Dicarld JJ (2018) To advance artificial intelligence, reverse-engineer the brain. Wired, 2.3.2018.

Dyakonov M (2018) The case against quantum computing. IEEE Spectrum, November 2018.

Hartnett K (2018) To build truly intelligent machines, teach them cause and effect. Quanta Magazine.

Knight W (2017) Put humans at the center of AI. MIT Technology Review, October 9, 2017.

Knight W (2018) One of the fathers of AI is worried about its future. MIT Technology Review, November 17, 2018.

Marcus G (2018) Deep learning: A critical appraisal. arXiv:1801.00631

Minsky ML (1990) Logical vs. analogical or symbolic vs. connectionist or neat vs. scruffy, In Artificial Intelligence at MIT, Expanding Frontiers, Patrick H. Winston (Ed.), Vol.1. MIT Press, 1990. Reprinted in AI Magazine, Summer 1991.

Moreno A (2021) 5 reasons why I left the AI industry. Towards Data Science, April 5.

Oster B (2018) Is machine learning anything more than an automated statistician? Quora blog, October 4, 2018.

Paukku T (2017) Tekoälyn viisasten kivi on superäly (The philosopher's stone of AI is super-intelligence). Helsingin Sanomat 20.6.2017.

Pearl J & Mackenzie D (2018) The Book of Why: The New Science of Cause and Effect. Basic Books, 432 p.

Pietikäinen M (1992) Tekoälyn kehittäjillä mittavia haasteita (The developers of AI face major challenges). AKTUUMI, University of Oulu 4:23-28.

Vorhies V (2018) What comes after deep learning. Data Science Central blog.

Zhou Z-H & Feng J (2018) Deep forest. arXiv:1702.08835v3.

Web-Memory: Human memory

Wiki-Boström: Nick Bostrom

Wiki-Consciousness: Consciousness

Wiki-Cyc: Cyc

Wiki-Kurz: Raymond Kurzweil

Wiki-Singularity: Singularity

Wiki-Vinge: Vernor Vinge

# 11 Summary - Does the AI Hype Continue?

## 11.1 Artificial Intelligence Today

Artificial intelligence research has been promising big for over 60 years. Numerous simulated and real-world toy demonstrations have been implemented in laboratories to demonstrate automatic environment modeling and behavioral learning, but the demonstrations have been scaled weakly to a larger scale, and robots have been left unused.

In recent years, machine learning and artificial intelligence applications have finally begun to come to the surface of everyday life. The development is largely due to advances in computer technology. The Internet has provided the opportunity to acquire massive data sets and thus supported the development of methodology, away from limited laboratory environments. At the same time, there has been tremendous progress in the development of hardware support for machine learning methods and the use of masses of data. Chapters 5 and 6 included examples of applications, including real success stories. All related to the so-called weak artificial intelligence, that is, limited application problems.

One advantage of computers over humans is the ability to retrieve and recall vast amounts of information that can be accessed quickly, for example, via the Internet. The machine does not get tired and can work 24 hours a day. If the data is of sufficient quality, the machine can outperform human performance in limited machine learning tasks. However, too little attention is often paid to gathering, naming, and modifying good quality data for machine learning, although the performance of these methods is largely dependent on it. Typically, such data processing requires up to 80% of the workload required to apply machine learning.

Due to its speed and other capacity, the machine is also able to combine data from different sources in an unprecedented way. Section 5.5 provided an example of a study which found that human happiness is reduced due to air pollution (Junttila, 2019). Countless applications of this kind can be found, and the data used for either good or bad purposes. If the information to be combined is not only somebody's behavior, but also, for example, a person's Internet searches, shopping purchases, health records and bank information, it is possible to create dangerously accurate profiles of people. The biggest threat posed by artificial intelligence is probably people's own behavior, that is, how much we give our private information to everyone without protection, for example through social media.

Machine vision is an area for which numerous industrial applications have been developed for a long time, beginning with the recognition of machine-typed characters and the visual quality control of industrial products in the 1980s or earlier. Image or video search from databases, biometric identification, and video surveillance are examples of major applications in recent years. New and more demanding applications are constantly being found, for example in intelligent human-machine interfaces.

Robotics and automation are widespread applications where intelligent functions have been introduced for a long time, such as automotive assembly tasks, which were already used in the 1980s. With the recent advances in technology, artificial intelligence can be applied to more everyday tasks to replace routine work or to create entirely new opportunities, such as improving the quality of life of the elderly or disabled.

Evolving technology, artificial intelligence, and robots cannot, of course, substitute for immediate human contact, but can bring tremendous benefits to the elderly or ill living alone at home, whether through constant face-to-face communication with friends and relatives, searching the Internet for other entertainment, shopping, on-line health monitoring and telemedicine services, housekeeping and other assistive services, or virtual travel around the world. Artificial intelligence, machine senses and robotics help the visually impaired see, the hearing impaired hear, wheelchair users to move semi-autonomously, and each person to constantly monitor their fitness and health.

An interesting example of the near future of autonomous transport is the self-propelled container ship designed by Norwegian fertilizer giant Yara for fjord transport (Aittokoski, 2018a). The loading system planned in Finland is also becoming autonomous. The same article also mentions, e.g., an autonomous train in Australia to rail ore transportation. These are good examples of limited application areas where even full self-steering can be achieved, unlike unlimited road transport. The role of artificial intelligence in these is not the key, but part of automation solutions that have evolved over the years.

Particularly significant progress has been made in speech recognition during the past decade. Until a few years ago, recognition was too unreliable for most applications and sensitive to environmental noise. Today, the point has been reached that speech recognition is gaining a very central role in human-computer interfaces. One of the first examples of this development is already having a reliable Google search by voice, for example in Finnish or English. Leading artificial intelligence companies have invested tremendously in voice recognition technology and have launched personalized assistants based on it. An example of a new Finnish application is the recognition of dictations of doctors directly to the text in the Hospital District of Helsinki and Uusimaa (Tammi, 2018). Considerable progress has also been made in language translation, with Google Translate as a significant example.

Success has also been achieved in other applications where mass data is inherently used. For example, information on a large population of symptoms of various diseases can be used to teach the system to make fairly good predictions based on symptoms. Google is said to have been able to make predictions based on search data for years, with reasonable accuracy, including the spread of the flu and other viruses.

Artificial intelligence has also been successfully applied to real-life music and poetry, in which there is a wealth of prior example data. The article "Sibelius or Homo Deus?" shows readers how to make a computer compose like Jean Sibelius (Sirén, 2019). However, composing on the basis of artificial intelligence just imitates the style of the composers, said Esa-Pekka Salonen in an interview with Ykkösaamu on May 18, 2019. Independent, frontier and offensive compositions that humans have not been able to do, machines have failed to create.

At its best, artificial intelligence is used in applications that do not require accurate results and the machine does not have to make an actual decision, create a new one or diagnose it. On the other hand, "at its worst, artificial intelligence can recommend, for example, treatments that are unrelated to the patient's medical condition in any way," says Senior Physician Päivi Ruokoniemi in an article in Helsingin Sanomat (Ruokoniemi, 2018).

Artificial intelligence has the ability to adapt machines to humans more than humans to learn new technology. Impact could even extend to the evolution of the provision and delivery of social and health services. This requires a growing role for users in developing and adapting systems. Thus, for example in the field of public software procurement, learning artificial intelligence can be one way of transforming non-recurring, delayed and outdated solutions into continuous development.

Like other technologies, artificial intelligence can also be intentionally misused by its practitioners. Machine learning works according to what kind of data is being fed to it. If learning is based on false and untruthful data, the results can be misused. Another worrying example is the use of Facebook data for profiling people by Cambridge Analytica, selling the information to outsiders, and thereby influencing the election with negative profile-specific campaigns such as anti-immigration messages or undermining the political opponent.

It is also problematic to make false pictures and videos using computer vision and machine learning techniques. For example, the face of a certain politician is made to speak naturally in a speech by another person, such as an actor. In the Fall of 2018, there was an incident leading to a temporary ban on a White House reporter for the CNN television station, whose video was artificially speeded up so that the reporter appeared to be grabbing the person taking the microphone.

Facial recognition is one of the most advanced artificial intelligence applications today. Due to the large number of applications, a great deal of effort has been invested in research in this field. Deep learning and easy access to mass data for developers of various services (e.g. Google, Facebook, Baidu) have significantly accelerated development in recent years.

Section 8.2 looked at face analysis and its application to biometric identification from the perspective of our own research. Like many other technologies, face recognition can also be misused. For example, large numbers of CCTV cameras have been installed across urban areas, and the misuse of the information they provide clearly endangers privacy. This is also exacerbated by the fact that face recognition is not reliable enough in changing environmental and imaging conditions, so there is a high potential for misinterpretation. The greatest risk of error is in applications that try to find criminals or other persons considered to be dangerous for the society in a watch list on videos taken by a surveillance camera (Laperruque, 2018).

International technology leaders in artificial intelligence have an enormous amount of data that they collect from their clients. They also have huge computing resources and cloud services under their control. With their very large computing resources, these companies are able to demonstrate impressive application examples, which, however, may still be far from real real-world applicability.

Examples include Go board game AlphaGo (Paukku, 2018a) developed by DeepMind, winner of the human world champion, and Nvidia’s system generating unknown face images (Paukku, 2018b). These achievements receive extraordinary attention in the media, and thus increase the hype of artificial intelligence - and the market for companies that have developed them. The face-generating method is currently based on GANs (Generative Adversarial Networks) of high interest, which have two competing networks (see Section 4.9): one responsible for generating (images that look natural), and the other rating the content (does the generated image look natural?) (Goodfellow et al., 2014).

A significant part of the development of machine learning applications currently being carried out in Finland relies on the services of technology leaders. However, their use may be relatively expensive and may involve security risks. Nor are they always available. The best chances for Finnish companies to succeed are in their own products, in new niche applications that aren't of interest enough to AI technology leaders, and in various services.

However, artificial intelligence and machine learning are in themselves massive problems: millions of years of evolution are being sought to be quickly automated. Sometimes disappointments will inevitably arise, as research is working on single-goal developments, even though it is a multi-objective entity, which may require evolutionary methods to combine, select, and understand the features of solutions as humans become overwhelming. Technology developers usually have a horizon-biased style of "total solution is around the corner" and this is also true for machine learning and artificial intelligence.

Thus, it would be better to focus on limited application problems with weak artificial intelligence, where artificial intelligence is more like a human assistant. not a ruler. The term *augmented intelligence* is often used in this context. The hype surrounding artificial intelligence could also be reduced by using a more application-oriented term *machine intelligence* instead of artificial intelligence, without even having to compare human intelligence. It can be expected that such machine intelligence will be more and more normal in the future integrated with other technology, improving the adaptability of applications to new situations, functionality and usability.

In this sense, this is a normal phase of technology development, with new solutions being introduced and no longer even talking about artificial intelligence. With the advancement of research and technology, we can better respond to the great challenges of artificial intelligence, etc., and provide systems with more human-like features.

## 11.2 Prospects and Challenges for AI

By integrating various technologies related to artificial intelligence, such as masses of data, machine learning, speech, natural language interpretation, machine vision, and robotics, very new types of applications can be created. A futuristic example is the social humanoid robots which imitate humans, receiving considerable attention from the media, such as the Sophia robot developed by a Hong Kong company (Wiki-Sophia) (Figure 11.1).

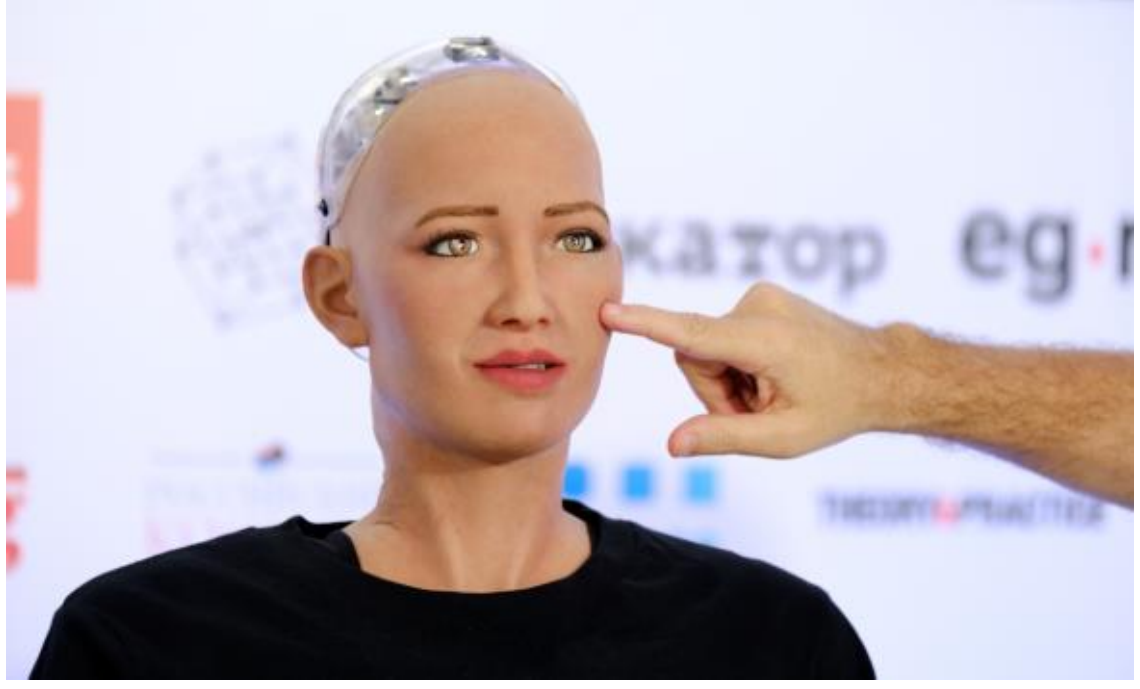

Figure 11.1. Sophia robot. (© 123RF)

Combining several types of weak artificial intelligence features (for example, vision, hearing, touch, speech, face animation) into the same system and application may be an area where astonishing results can be achieved as technology advances in the future.

Our networked society could also allow to use so-called *collective intelligence* provided by multiple weak artificial intelligence computers, possibly at different location, to solve highly "complex" AI problems (Wiki-Collective).

Finland's success in artificial intelligence requires not only investing in applications but also investing heavily in long-term research in the field in order to produce sufficiently original solutions. Education also plays a key role in success. There is a need for both improving citizens' data literacy and training for artificial intelligence specialists. Those specializing in artificial intelligence should take a broad approach to the field, ranging from mathematics, programming skills, to both traditional and new methods of artificial intelligence and data analysis.

The next big step in artificial intelligence requires a reassessment of existing methodological and implementation solutions. In many applications, massive data is either unavailable or too expensive to obtain. Artificial intelligence focused on machine learning, and deep learning in particular, is vulnerable to adversarial attacks in which the system is fed with incorrect teaching data that does not resemble the right, unnoticed by the eyes or other senses (Heaven, 2019).

This is a major threat in many key applications, such as medicine, self-driving vehicles, and AI based weapon systems (Finlayson et al., 2019). Too much reliance on the services of technology leaders is expensive and can put security at risk. There are also opportunities for small internationally networked economies for new types of artificial intelligence based business models.

The widespread adoption of artificial intelligence in societies requires people to have confidence in this technology. The ethical issues of artificial intelligence related to this are beginning to receive increasing attention in various communities (Wiki-Ethics), (Web-ECStrategy). Confidence is diminished, for example, by the fact that if a machine learning method is trained with incomplete “biased” material, the results it gives may be incorrectly weighted. It has been observed, among other things, that some facial recognition methods have not worked as reliably for the colored population, as their share of the teaching material has been too small. Similar examples have also been found in medical applications, for example.

According to leading researcher Ilkka Tuomi (Tuomi, 2019), increasing energy consumption means that only a few developers of artificial intelligence models can keep up with the development. The energy consumed by the best (and deepest) deep learning models has increased tenfold each year in the 2010s. Google, Facebook and other leading artificial intelligence companies are already the world's largest users of renewable energy

- and as the trend continues, wind and solar power may not be enough for others in the near future.

Sweden has long been a trendsetter in the provision of global Internet services in the Nordic countries, not least thanks to Skype, Spotify and Klarna. The background is probably a well-established culture of international trade. Similarly, in Finland, should we seek the foundation for artificial intelligence in our strong communications and information technology skills? Advances in information and communications technology and electronics are, by and large, leading to ultra-dense networks. Such developmental pathways include, for example, the increasing use of proximity electronics such as activity and health bracelets and rings. Similarly, we see networked wireless technology increasingly invading applications in buildings, machinery, equipment, industrial production and pets, and so on.

As a result, the amount of data collected alone is increasing tremendously, with personal data protection issues being challenged by the concentration of data and analytics on data centers. On the other hand, information from technical sources is rarely categorized, as opposed to, for example, images entered by people into Internet services. This challenges the development of artificial intelligence applications based on current machine learning.

According to many estimates, artificial intelligence and machine learning will increasingly shift to so-called edge and fog computing (Figure 11.2). At that time, services that require heavier computing and analytics are provided at wireless base stations or even in a decentralized collaboration between measurement nodes, mobile phones, vehicles, and other terminals.

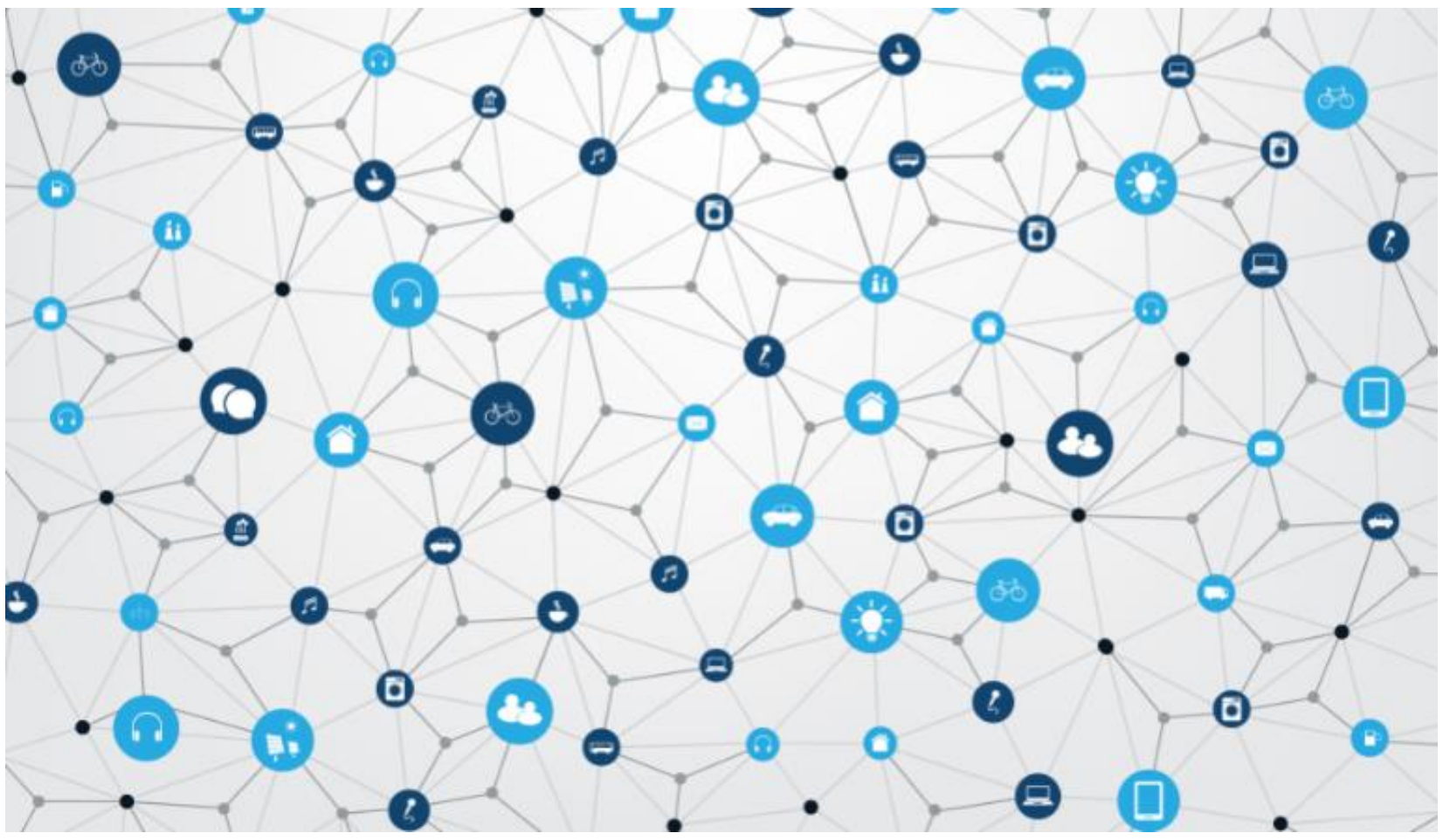

Figure 11.2. Fog computing is highly distributed. (© 123RF)

This improves reactivity, robustness and data protection by dropping the role of centralized resources, but challenges machine learning: there is less data available, and procedures based on deep neural networks are clearly in difficulty.

Thus, instead of using methods that require large classified data, researchers should aim for solutions that learn from limited sample sizes. As a practical result, we can, for example, obtain distributed dynamic data search engines from which we can ask about information changes.

On the other hand, ultra-dense network terminals will increasingly operate with help of their environment by collecting energy without needing batteries or external power supply. It should also be possible to reduce energy consumption to curb climate change. Thus, both computing and memory use must be energy efficient. Excessive power consumption prevents the development of many applications, such as smart sensors embedded in mobile devices, smart glasses, smart watches, or intelligent sensors embedded in the environment.

Achieving fast-learning and low-energy artificial intelligence requires investment in research into effective presentations. For example, LBP-like or other neural network methods using binary-formatted data could bring significant efficiency improvements. Indeed, research funders need to understand the need to support both the development of energy-efficient learning algorithms using traditional information technology and solutions based on new methods.

A very hot area of near future research is machine learning with one or at most a few training samples (so-called one or few shot learning), as people learn many things. A simple example is the Bayesian program learning mentioned in Section 3.4 (Lake et al., 2015). With such methods, new wireless communication technologies support the embedding of artificial intelligence in the human immediate environment instead of Internet giant data centers. Indeed, the IT pendulum may swing once again from a focus on decentralization.

Continuous learning of the new without forgetting previously learned knowledge is central to human intelligence, but artificial intelligence lacks such an ability. This topic is also a new, very important area of research in artificial intelligence. In this context, the terms “continual learning” or  life-long learning” are used (Wiki-ContinualAI) (Jha 2020).

Artificial intelligence systems must be able to justify to the user the reasons for their decision, i.e. why, for example, it ends up diagnosing a particular disease on the basis of the information taught to it. Rationale for decisions is particularly important in, for example, medical applications, self-driving vehicles and military applications. However, systems based on machine learning lack the ability to justify - for the user (or other programs) they appear as “black boxes”. Explainable AI is a very topical research area in this regard. For more information, see (Wiki-Explainable AI) and (Web-DarpaXAI).

The ability to analyze cause-and-effect relationships (i.e., causality) is also seen as one of the key elements in creating stronger artificial intelligence. Section 10.4 already mentions the statements of the pioneer, Professor Judea Pearl. An introduction to the topic from the perspective of artificial intelligence and applications can be found, for example, in references (Gontalonieri, 2020) and (Dickinson, 2021).

Strong artificial intelligence resembling human intelligence is still a long way off. Today's artificial intelligence is seen as a sharpened version of what artificial intelligence was long ago.

The next step requires a re-evaluation of the fundamentals of artificial intelligence and a combination of different approaches - for example, combining neural network methods that use connectionist pattern recognition and methods that handle symbolic information. The Neuro-Symbol Concept Learner program, announced by MIT, IBM and DeepMind in the spring of 2019, is an interesting step in this direction. It learns a greatly simplified version of its surroundings in a child-like way, looking around and talking (Knight, 2019). Artificial intelligence similar to human intelligence would require much more brain-like systems in structure and function to overcome enormous methodological, computational, and power consumption challenges (Ideami J, 2021), (Oster, 2018).

The reference (Romero, 2021) examines the views of several AI experts on how deep learning could be taken to the next level, towards truly intelligent systems: Convolution networks (CNNs) and their limitations, as well as pre-labeled teaching data, should be eliminated using unsupervised training instead of supervised; move to hybrid models using symbolic processing and deep learning; incorporate cognitive characteristics into systems; and leverage ideas and the latest findings from neuroscience and human brain research.

## 11.3 References

Aittokoski H (2018a) Norja seilaa itseohjautuvaan tulevaisuuteen (Norway is sailing towards a self-guiding future). Helsingin Sanomat 19.9.2018.

Dickinson D (2021) Why machine learning struggles with causality? TechTalks Blog 15.3. 2021.

Finlayson SG, Bowers JD, Ito J, Zittrain JL, Beam AL & Kohane IS (2019) Adversarial attacks on medical machine learning. Science 363 (6433): 1287-1289.

Goodfellow I, Pouget-Abadie J, Mirza M, Xu B, Warde-Farley D, Ozair S, Courville A & Bengio Y (2014). Generative adversarial networks. Proceedings of the International Conference on Neural Information Processing Systems (NIPS 2014), 2672-2680.

Gontalonieri (2020) Introduction to causality in machine learning. Towards Data Science, July 9, 2020.

Heaven D (2019) Deep trouble for deep learning. Nature 574:163-166, 10.10.2019.

Ideami J (2021) Towards the end of deep learning and the beginning of AGI. Towards Data Science. March 2021.

Jha S (2020) Continual learning - where are we? Towards Data Science 17.9.2020.

Junttila J (2019) Puhdas ilma lisää onnellisuutta (Clean air increases happiness). Helsingin Sanomat 24.1.2019.

Knight W (2019) Two rival AI approaches combine to let machines learn about the world like a child. MIT Technology Review, 8.4.2019.

Lake BM, Salakhutdinov R & Tenenbaum JB (2015) Human-level concept learning through probabilistic program induction. Science 350(6266):1332-1338 1332.

Laperruque J (2018) Unmasking the realities of facial recognition. Project On Government Oversight POGO 5.12.2018.

Oster B (2018) Is machine learning anything more than an automated statistician? Quora blog, October 4, 2018.

Paukku T (2018a) Ihmisen go-lautapelissä voittaneen tekoälyn piti olla totta ehkä vasta vuonna 2035 (The AI that won a human in the go board game was not supposed be true before 2035). Helsingin Sanomat 18.10.2018.

Paukku T (2018b) Kuvan ihmistä ei ole olemassa (The person shown in the picture does not exist). Helsingin Sanomat 27.6.2018.

Romero A (2021) 5 deep learning trends leading artificial intelligence to the next stage. Towards DataScience, April 26, 2021.

Ruokoniemi P (2018) Tekoälyä on käytettävä harkiten lääketieteessä (AI should be used with care in medicine). Helsingin Sanomat 28.10.2018.

Sirén V (2019) Sibelius vai Homo Deus (Sibelius or Homo Deus). Helsingin Sanomat 28.4.2019.

Tammi S (2018) Lääkärin sanelut suoraan tekstiksi (Physician's dictations direcly to text) . Helsingin Sanomat 25.6.2018

Tuomi I (2019) Sähkönkulutus on tekoälyn kompastuskivi (Electricity consumption is an obstacle for AI). Helsingin Sanomat 31.8.2019.

Web-ECStrategy: High-level expert group on artificial intelligence 10.3.2021

Web-DarpaXAI: Explainable artificial intelligence (XAI)

Wiki-ContinualAI: Continual AI

Wiki-Ethics: Ethics of artificial intelligence
Wiki-ExplainableAI: Explainable artificial intelligence
Wiki-Collective: Collective intelligence
Wiki-Sophia: Sophia robot

# Appendix L1: What should be Taught about AI?

## L1.1 General

The revolution brought by artificial intelligence poses major challenges for the development of skills and the education system. Universities are facing problems, as many companies in Oulu, for example, have started investing in artificial intelligence and recruited more than 15 doctoral researchers in two years (2016-2018). It is more and more difficult to recruit new top-level students and researchers anymore (Pietikäinen et al., 2017). Students of information technology are mostly employed in companies from the very beginning of their studies.

On the other hand, the fragmentation and short-term nature of research funding is making it difficult for top-level research leading to original new innovations and "deep know-how" in artificial intelligence. Finland's strength has always been innovative solutions, even various niche applications with a large global market.

Now even areas with a long Finnish tradition of key technologies in artificial intelligence, such as machine learning, computer vision, and automatic speech and language understanding, have been struggling. Fortunately, thanks to recent artificial intelligence hype, there has been some improvement - mainly in machine learning.

An increasing number of doctoral students and researchers specializing in artificial intelligence are coming from abroad. For example, our university's machine vision research group has a dozen or so hard-working Chinese researchers and students, many of them with Chinese grants. It is clear that paid international master's programs at universities should not be seen as an income item, but that they can provide motivated top-level students for the needs of Finnish research and business.

The profound knowledge of artificial intelligence can only come from long-term research. In recent years, access to research funding in Finland has become more difficult significantly, and the Academy of Finland is increasingly funding projects of only two years' duration. The focus of Business Finland (formerly Tekes) has now shifted, as its name suggests, to more and more business-oriented research. While this kind of funding can achieve short-term gains, original results and innovations that generate new knowledge are created through basic research. Indeed, many start-up companies related to artificial intelligence have emerged as a result of university research.

Mathematics and programming skills form a central basis for expertise in artificial intelligence. The motivation to study mathematics in schools has undoubtedly decreased, and the number of applicants for university studies in technical fields has dropped.

The problem begins already in elementary school, where motivated students cannot keep up with their abilities. The attitude towards mathematics in homes and social institutions is also often disparaging. Is the high school subject structure correct? Are there too many different topics in the long mathematics curriculum, and more profound knowledge and the learning skills suffer? However, we see recent efforts to strengthen the role of mathematics in high schools as a positive step.

The need for knowledge in mathematics and mathematical thinking is not limited to “hard” science and technology. Future medical doctors will have to deal more and more with the immense masses of images and information. They need to be able to understand what this information can bring, to use new advanced technologies and tools in their daily work - and to collaborate with experts in various fields to research and develop new technology for monitoring human well-being, diagnosing and treating diseases.

The use and interpretation of large masses of data play a central role in modern economics. One must, for example, be able to analyze the current situation and forecast future trends based on current and past knowledge. As early as the early 1990s, it was said that Wall Street was the largest employer of artificial intelligence doctors in the United States. The state of the environment can be monitored by collecting massive amounts of measurement data from various influencing factors and interpreting this data. New technologies incorporating artificial intelligence can help prevent climate change and its effects.

Even human scientists, far from mathematics and technology, are increasingly using a variety of measurement data, as well as artificial intelligence related devices and tools, in their research. Regardless of the industry, most decision-makers should be able to understand what artificial intelligence is and does not, and what it allows to prevent them from being taken over by outside "marketers”.

Future teachers need to know what digitalization, automation and artificial intelligence are all about, and thus be able to help their students to succeed and to be employed in the future. Automation and artificial intelligence will undoubtedly eradicate routine tasks, but the need for tasks that require analytical thinking, creativity, craftsmanship, and the ability to adapt to ever-changing circumstances will surely increase. Giving a positive image of mathematics and technology at an early stage, including for girls and women, is important to help create a more equal society.

The ability to function in the modern information society is relevant to all citizens and jobs. Data literacy, and also knowing what the capabilities and limits of artificial intelligence are, are important. Fear of technological threats doesn't help, because artificial intelligence is a human helper, not a ruler. The greatest danger to the individual is probably the inability to utilize the ever-increasing and renewing tools of artificial intelligence and how to protect his or her private information in information networks and social media. Figure L1.1 illustrates various perspectives that may need to be explored in research and application of artificial intelligence.

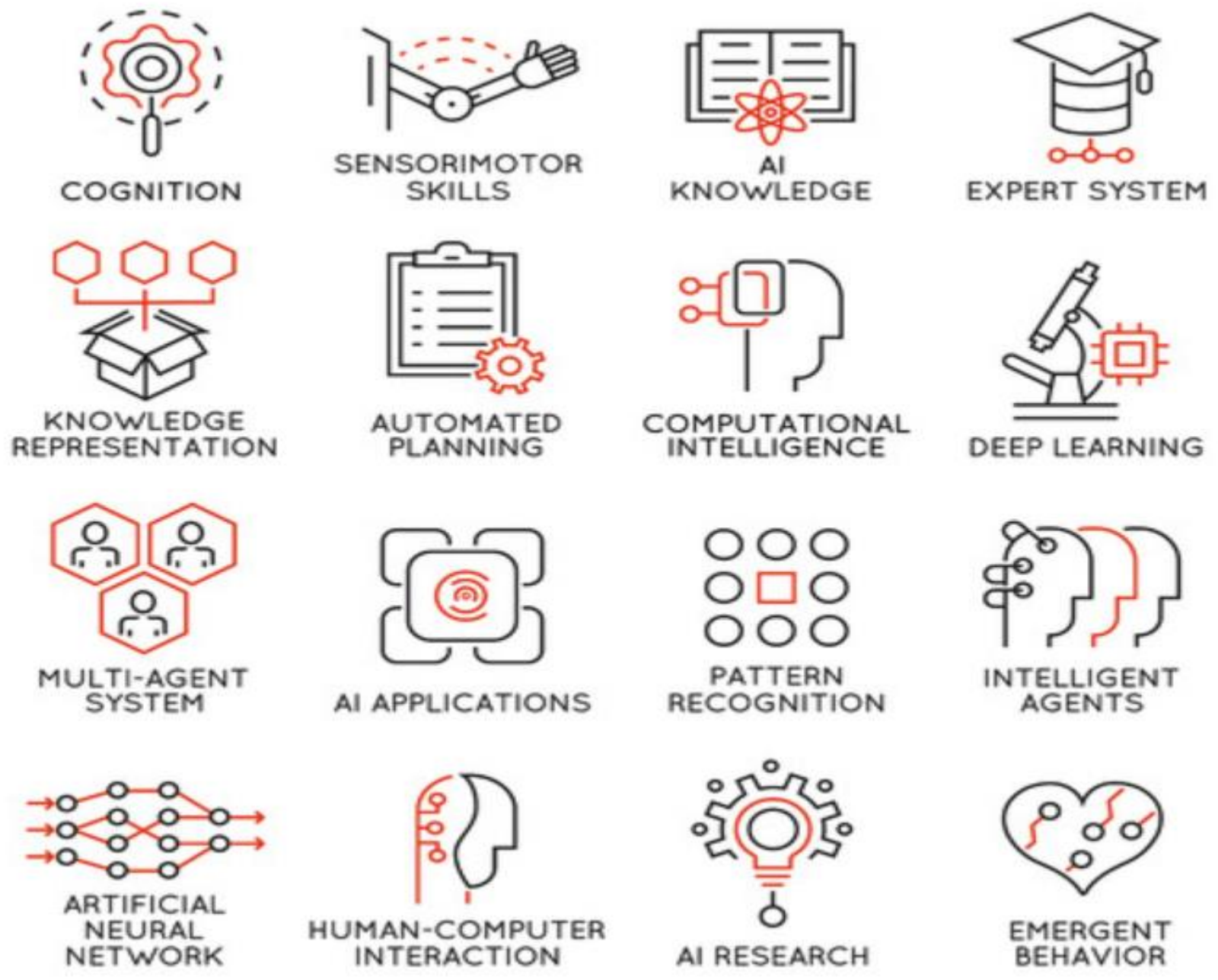

Figure L1.1. Various perspectives to AI. (© 123RF)

This appendix focuses on the core of artificial intelligence, knowledge of which is required in most applications. The teaching of artificial intelligence will be examined from the perspective of online teaching available through the Internet and the training of university-level artificial intelligence experts. The key to developing good mathematics and mathematics skills at primary and secondary level is that of all students. Knowledge of the basics of computer programming at this stage is also important.

## L1.2 Online Learning in Artificial Intelligence

The teaching of artificial intelligence has begun to be delivered as online courses. Bernard Marr, on his Forbes site, presents ten of the best (mostly American) free online courses, see (Web-Bernard).

The courses deal with e.g. machine learning basics, applications and practical implementations. Some of the courses are already designed for people with some technical skills to implement their own neural networks. Some of the courses are intended for the general public, that is, for those who want to know how to apply this technology to solving a variety of real problems.

However, it should be noted that on-line courses provide a rather narrow view of artificial intelligence, mostly focusing on deep learning or its applications. Section L1.3 introduces a broad field of study designed for the training of artificial intelligence experts for graduate students in Information Technology. The online courses give the general public an appropriate introduction to the basics of machine learning and can also complement the training of artificial intelligence specialists.

Bernard Marr considered the following to be the best free online courses in 2020: Elements of AI - University of Helsinki; Learn with Google AI; Intro to Artificial Intelligence - Udacity; Machine Learning - Stanford University (Coursera); AI for Everyone - Andrew Ng (Coursera); Data Science and Machine Learning Essentials - Microsoft (EdX); Machine Learning Crash Course - Google; Learning from Data (Introductory Machine Learning) - Caltech (EdX); Artificial Intelligence A-Z: Learn How to Build an AI - Udemy; Creative Applications of Deep Learning with Tensorflow - Kadenze (Class Central).

For Finns, perhaps the best-known of these, in addition to the course of University of Helsinki (Web-Helsinki), is Coursera's Stanford University machine learning courses (Web-Coursera), which are lectured by Andrew Ng (Wiki-Ng), a world-renowned lecturer in the field. He has been founder of the Google Deep Learning Group (Deep Brain) and leader of artificial intelligence research at Baidu.

Online education provided by Coursera, founded by Ng, is free. Only a certificate of completion of the course is required. The course covers a wide range of implementations of various machine learning applications, such as speech recognition and more efficient Internet searches. Advanced courses go into the statistical basics of deep learning, deep neural networks, and some development tools.

The lecturers of the extensive Udacity Intro to Artifiical Intelligence course are well-known artificial intelligence researchers Peter Norvig and Sebastian Thrun. The former is another author of the classic book on artificial intelligence (Russel & Norvig, 2010) and has recently served as Google's director of research (Wiki-Norvig).

The Elements of AI online course for the general public offered by the University of Helsinki and Reaktor company has been widely used in Finland and recently also elsewhere. Originally an English course is now also available in Finnish, cf. (Web-Helsinki). The course consists of six chapters: 1) What is artificial intelligence? 2) Problem solving with artificial intelligence, 3) Practical applications of artificial intelligence, 4) Machine learning, 5) Neural networks, and 6) Effects of artificial intelligence. More detailed information on the contents of these chapters can be found on the course website.

This book, and especially Chapters 3-4, is designed to serve a fairly wide readership, helping to understand the basics, possibilities, and limitations of artificial intelligence.

### L1.3 Degree Program in Artificial Intelligence at the University of Oulu

The University of Oulu has been teaching computer science courses since the early 1980s. Previous teaching included basic courses in both artificial intelligence and machine learning (neural networks and pattern recognition). In addition, the basics of digital image processing and machine vision were covered. From the outset, research and teaching have been closely intertwined: new courses have largely been created to meet the needs of research and have been taught by researchers and research assistants. On this basis, over the years, new research groups focused on various fields of information technology have emerged alongside machine vision, all of whom have strong activities in the field of artificial intelligence. These include research on medical signal analysis, ubiquitous computing, data mining, and intelligent systems and robotics. This research has made a significant contribution to curriculum development.

In 2016, a systematic renewal and expansion of the Artificial Intelligence curriculum was started in the Computer Science and Engineering Degree Program. The starting point was that diploma engineers (MSc) who are deepening into artificial intelligence must have a solid basic mathematical training, good programming skills, and a sufficiently broad vision of artificial intelligence, because now so hot topic machine learning is only one part of artificial intelligence. A new field of study in artificial intelligence started in the Fall of 2018, but the new courses included in it were completed and held before that. The latest content of the study guide for Computer Science and Engineering MSc students and course descriptions can be found in (Web-Oulu).

Graduate engineers specializing in artificial intelligence are given a strong basic education in mathematics, programming, computer engineering, human-computer interaction, digital signal and image processing, and computer graphics. Basic studies in mathematics and computer science include: vector algebra, differential and integral calculus, linear algebra (matrix calculus), complex analysis, statistical mathematics, computer mathematics, and data structures and algorithms. The normal training of diploma engineers in mathematics and mathematics is supported by additional courses on regression and variance analysis, and optimization, closely related to AI and machine learning.

The recent core courses Artificial Intelligence have been Introduction to Artificial Intelligence, Artificial Intelligence, Machine Learning, Deep Learning, Computer Vision, Affective Computing, Natural Language Processing and Text Mining, and Multimodal Data Fusion. Courses related to massive data processing, or big data processing, are Journey to Data Mining and Big Data Processing and Applications. A modern course dealing with information networks is an Introduction to Social Network Analysis. In addition, it is possible to choose e.g. a course related to virtual reality VR systems and people; and Fundamentals of Sensing, Tracking and Autonomy related to machine sensing.

In addition to the above, at least audio and speech processing focused teaching would be needed, but no teaching resources for this have been found so far. Chapters 3 and 4 of this book are largely based on material developed by Olli Silvén for the Introduction to Artificial Intelligence course.

The following is a brief introduction to the main courses in Artificial Intelligence at the University of Oulu, based on course presentations. Our intent is to give the reader an idea of what kind of artificial intelligence studies are needed by knowledgeable IT professionals. Courses in the Bachelor of Science (Engineering) degree are marked with **P** (basic studies) or **A** (subject studies) in the title of the course and advanced courses for the Master's degree are marked with **S**. The code of each course is given in parenthesis.

*Introduction to Artificial Intelligence (521160P)*

The course is designed for students of all disciplines. It does not require theoretical knowledge of artificial intelligence, but focuses on identifying and solving different types of artificial intelligence problems with existing tools. The course consists of lectures and exercises in groups of students from many different fields. Each group must have at least one student with programming skills.

After completing the course, the student will be able to identify any artificial intelligence techniques that may be applicable to problem solving, able to distinguish between search, regression, classification, and clustering problems, able to explain the use of supervised and unsupervised learning, and performance measurement. This differs from the online course of the University of Helsinki and Reaktor, especially in that our own course is clearly more data-driven.

Course contents: 1) Introduction: importance of artificial intelligence, 2) Search methods: AI of games, 3) Regression methods: learning the causality, 4) Classification methods: identifying categories, 5) Clustering methods: identifying class structures, 6) Supervised learning, 7) Unsupervised learning.

*Artificial Intelligence (521495A)*

Course based on Artificial Intelligence: A Modern Approach (Russell & Norvig, 2010), which is probably the most widely used textbook in the world for artificial intelligence teaching.

After completing the course, the student will be able to identify the types of problems that can be solved with artificial intelligence, know the basic concepts of intelligent agents, know the most common search methods and principles of inference based on logic. He or she can also apply simple methods for reasoning involving uncertainty and for machine learning based on observations. In addition, the student will be able to program the most common search methods. The course includes two programming exercises.

Content of the course: 1) Introduction, 2) Rational (intelligent) agents and uninformed search, 3) Informed search, 4) Adversarial search (games), 5) Uncertainty, 6) Markov decision processes, 7) Reinforcement learning, 8) Bayesian networks, 9) Learning from samples, 10) Advanced applications.

*Machine Learning (21289S)*

After completing the course, the student will be able to design simple optimal classifiers based on basic theory and evaluate their performance, be able to explain Bayesian decision theory and apply it to minimal error and minimum cost classifiers, apply basic gradient search to linear nonlinear discriminant function, and apply regression methods to practical machine learning problems. The course includes supervised laboratory work and independent work.

Content of the course: 1) Introduction, 2) Bayesian decision theory, 3) Discriminant functions, 4) Parametric and non-parametric classifications, 5) Extraction of features, 6) Classifier design, 7) Example classifiers, 8) Statistical regression methods.

*Deep Learning (521153S)*

After completing the course, students will become familiar with deep learning and its basic methods and how to use them in a variety of problems. The course consists of lectures, five assignments and a final project.

Content of the course: 1) Introduction to deep learning and course content, TensorFlow tutorial, 2) Fundamentals of deep learning: linear regression, logistic regression, loss function, stochastic gradient search, simple practices for training basic models, 3) Neural networks, deep networks, auto-encoders, 4) Convolutional neural networks (CNN), 5) Current CNN applications in computer vision, 6) Deep models for text and sequences (RNN and LSTM).

*Computer Vision (521466S)*

After completing the course, the student understands the basics of image formation, presentation and modeling. He or she is able to use basic machine vision techniques for image recognition problems. In addition, the student is able to use two-dimensional transformations for model fitting and image registration, and can explain the basics of three-dimensional imaging and reconstruction. The course includes homework. Prerequisites are required to complete the Digital Image Processing course or equivalent.

Content of the course: 1) Introduction, 2) Imaging and representation, 3) Color and shading, 4) Image features, 5) Identification, 6) Texture, 7) Motion from 2-D image sequences, 8) 2-D models and transforms, 9) 3-D sensing of 2-D images, 10) 3-D transformations and reconstruction.

*Affective Computing (521285S)*

After completing the course, the student will be able to explain the theory and modeling of emotions, will be able to implement algorithms related to the recognition of emotions from visual and audio signals or the combination of multiple modalities. He or she also has a general view of applications of affective computing (emotion AI).

Content of the course: 1) History and development of affective computing, 2) Psychology of emotion theory and modeling, 3) Emotion recognition from different modalities: facial expressions, speech, EEG, 4) Crowdsourcing, 5) Synthesis of emotive behaviors, 6) Emotion recognition applications.

*Natural Language Processing and Text Mining (521158S)*

After completing the course, students will be able to understand, design and implement basic systems for searching and querying text (on-line), taking into account linguistic factors and performing clarification of word meaning, performing (statistical) inferences using corpus (language database), and editing (statistical) language modeling tool kits, lexical on-line databases and various natural language processing tools.

Content of the course: 1) Basics of text retrieval systems, 2) Glossary ontologies, 3) Specifying the meaning of words, 4) Text classification, 5) Language database based reasoning and natural language processing tools.

*Multimodal Data Fusion (521161S)*

After completing the course, students are expected to understand the problem of combining different types of data from different sources (such as images and audio). They should be able to implement basic solutions for the task that requires integration and aggregation of the data provided.

Course Content: This course provides a comprehensive introduction to the concepts and approaches to fusion of many data generated by sensors. The course introduces several real world examples of different types of applications. The content is constructed so that no prior knowledge of data fusion is required. However, a basic understanding of related issues, such as image and signal processing, is useful.

*On the Way to Data Mining (21156S)*

After completing the course, the student will be able to identify what kind of data he or she is going to study and what kind of pre-processing it will require. The specific learning outcomes of the course include: ability to design and implement data collection, ability to combine data from various sources, normalize and transform data, and process missing or incorrect data, ability to generalize results.

Content of the course: 1) Data mining process in general, data collection and different types of data, 2) Data quality and reliability, 3) Data preparation including missing values, outliers and privacy processing, 4) Use of signals from multiple sources 5) Utilization of databases in the data mining process as well as data normalization, transformation and interdependence and distribution of observations; 6) Principles independent of modeling techniques ensuring generalization and data delivery of results, such as train-test-validate, cross-validation and leave-one-out methods.

*Processing and Application of Big Data (521283S)*

After completing the course, the student will be able to explain the mass data (big data), its challenges and opportunities. They will also be able to explain the requirements and general principles for designing and implementing data intensive systems, as well as evaluate the benefits, risks and limitations of possible solutions. In addition, the student knows the principles of mass data management and processing technologies and can apply them at the basic level.

Content of the course: 1) Basics of mass data, 2) Data storage, 3) Single or continuous data processing, 4) Data analysis, 5) Privacy and security, 6) Mass data usage examples.

*Introduction to Social Network Analysis (521157A)*

After completing the course, students are expected to understand the social aspects of the web, gather, clean and present social media data, identify important features of social media, discover and analyze (online) network communities, understand the diffusion process within the social network, and analysis of work tools.

Content of the course: The course describes the basics of social network analysis: 1) Creates the ability to understand the structure and development of the network, 2) Enables the use of appropriate tools and techniques to draw conclusions from the network and to find hidden patterns. Designed for students with a background in information technology, mathematics and social sciences, the course provides a basis for multidisciplinary research.

*Summary*

The training provided will give the practitioner sufficient in-depth knowledge in the field of artificial intelligence and information technology, as well as prepare them for doctoral studies in the field. In addition, courses are offered for training mathematically-oriented data analysts and basic education for all faculties at the University. Such degree programs are also hoped to increase the interest of high school students in pursuing university studies in technology and computer science. Unfortunately, there is no shortcut to happiness: everything relies on mathematical thinking.

## L1.4 References

Russell S & Norvig P (2010) Artificial Intelligence: A Modern Approach, 3rd Edition. Prentice Hall, 1152 p.

Pietikäinen M, Silvén O & Pirttikangas S (2017) Tekoälyn opetusta on lisättävä ja syvennettävä (We should have more and deeper AI teaching in Finland). Helsingin Sanomat 20.4.2017.

Web-Bernard: The 10 Best Free Online Artificial Intelligence and Machine Learning Courses for 2020, Forbes, 16.3.2020

Web-Coursera: Machine Learning, Stanford University

Web-Helsinki: Elements of AI, University of Helsinki

Web-Oulu: CSE-MSc (Study guide for Computer Science and Engineering students). University of Oulu

Wiki-Ng: Andrew Ng

# Table of Figures

**Challenges of Artificial Intelligence – From Machine Learning and Computer Vision to Emotional Intelligence**

Artificial intelligence or machine intelligence is a technology capable for actions considered to be intelligent. We encounter artificial intelligence in computer games, video services, product recommendations, speech recognition capabilities of mobile phones, face-to-face payment points and, for example, lane guards of cars. As the development expertise and implementations of such solutions become more common, the concept of artificial intelligence has shifted to increasingly sophisticated capabilities.

If the recent expectations on artificial intelligence come true, everyone's everyday life and working life, as well as the future of humanity, will be revolutionized. This expert book is a realistic answer to the over marketing of artificial intelligence.

So far, the development of artificial intelligence has been a series of enthusiasm periods, over-sized promises, major efforts, disappointments, and winters of AI research. However, technological advances have been significant at every step.

This book written for wide audience describes how artificial intelligence has evolved since the 1950s based on the most promising research findings. The most important advances in methodology and their impacts are considered. The hand-to-hand approach is to look "under the hood " of the basic principles, implementation challenges, and limitations of what is perceived to be intelligent.

In the future, in addition to current applications, the authors of the book will see artificial intelligence transforming into an interpreter of human actions and emotions. Better anticipation of human procedures will facilitate the use of machinery and equipment and improve safety at home, at work and in traffic.